\documentclass{article}

\usepackage[english]{babel}

\usepackage[letterpaper,top=2cm,bottom=2cm,left=3cm,right=3cm,marginparwidth=1.75cm]{geometry}
\usepackage{setspace,color}
\usepackage{multirow}
\usepackage{latexsym,amssymb,enumerate,amsmath,epsfig,amsthm,authblk,dsfont,fullpage}
\usepackage{algorithm,algorithmic}
\usepackage{graphicx}
\usepackage{graphbox}
\usepackage[colorlinks=true, allcolors=blue]{hyperref}
\allowdisplaybreaks
\usepackage{dsfont}
\newcommand{\ba}{\mathbf{a}}
\newcommand{\bb}{\mathbf{b}}
\newcommand{\bx}{\mathbf{x}}

\newcommand{\by}{\mathbf{y}}

\newcommand{\bs}{\mathbf{s}}
\newcommand{\bw}{\mathbf{w}}

\newcommand{\bg}{\mathbf{g}}
\newcommand{\bq}{\mathbf{q}}

\newcommand{\bp}{\mathbf{p}}
\newcommand{\bn}{\mathbf{n}}

\newcommand{\RR}{\mathbb{R}}

\newcommand{\diver}{\text{div}}
\newcommand{\Real}{\text{Real}}

\newcommand{\cS}{\mathcal{S}}
\newcommand{\cI}{\mathcal{I}}

\newcommand{\cL}{L}
\newcommand{\cH}{H}
\newcommand{\cF}{\mathcal{F}}
\newcommand{\cP}{\mathcal{P}}

\newcommand{\cW}{W}

\newcommand{\bmu}{\boldsymbol\mu}
\newcommand{\bnu}{\boldsymbol\nu}
\newcommand{\blambda}{\boldsymbol\lambda}

\DeclareMathOperator*{\argmin}{arg\,min}


\theoremstyle{definition}

\newtheorem{prop}{Proposition}[section]

\newtheorem{remark}{Remark}

\title{Euler's Elastica Based Cartoon-Smooth-Texture Image Decomposition}
\author{Roy Y. He\footnote{Department of Mathematics, City University of Hong Kong, Tat Chee Avenue, Kowloon, Hong Kong. Email: royhe2@cityu.edu.hk.}, Hao Liu\footnote{Department of Mathematics, Hong Kong Baptist University, Kowloon Tong, Kowloon, Hong Kong. Email: haoliu@hkbu.edu.hk.}}
\date{}
\begin{document}
\maketitle

\begin{abstract}
We propose a novel model for decomposing grayscale images into three distinct components: the structural part, representing sharp boundaries and regions with strong light-to-dark transitions; the smooth part, capturing soft shadows and shades; and the oscillatory part, characterizing textures and noise. To capture the homogeneous structures,  we introduce a combination of $L^0$-gradient and curvature regularization on level lines. This new regularization term enforces strong sparsity on the image gradient while reducing the undesirable staircase effects as well as preserving the geometry of contours.   For the smoothly varying component, we utilize the $L^2$-norm of the Laplacian that favors isotropic smoothness.  To capture the oscillation, we use the inverse Sobolev  seminorm. To solve the associated minimization problem, we design an efficient operator-splitting algorithm. Our algorithm effectively addresses the challenging non-convex non-smooth problem by separating it into sub-problems. Each sub-problem can be solved either directly using closed-form solutions or efficiently using the Fast Fourier Transform (FFT). We provide systematic experiments, including ablation and comparison studies, to analyze our model's behaviors and demonstrate its effectiveness as well as efficiency.
\end{abstract}

\section{Introduction}
\label{sec.intro}

\begin{figure}[t!]
	\centering
	\begin{tabular}{c@{\hspace{2pt}}c@{\hspace{2pt}}c@{\hspace{2pt}}c@{\hspace{2pt}}c@{\hspace{2pt}}c}
		(a)&(b)&(c)&(d)&(e)&(f)\\
		\includegraphics[width=0.15\textwidth]{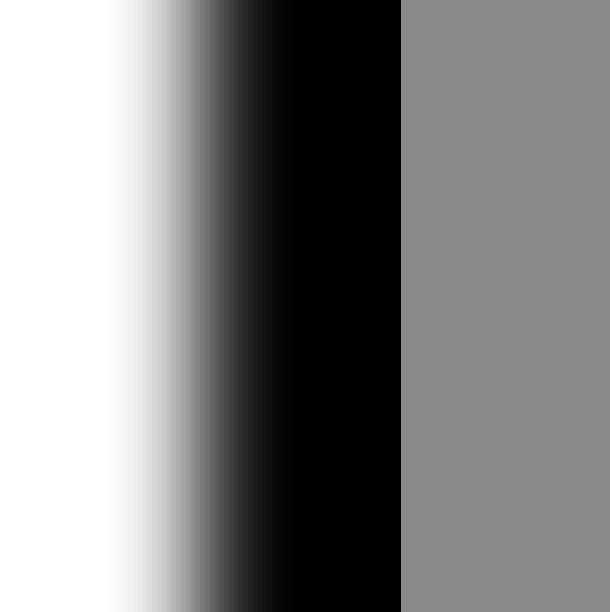}&
		\includegraphics[width=0.15\textwidth]{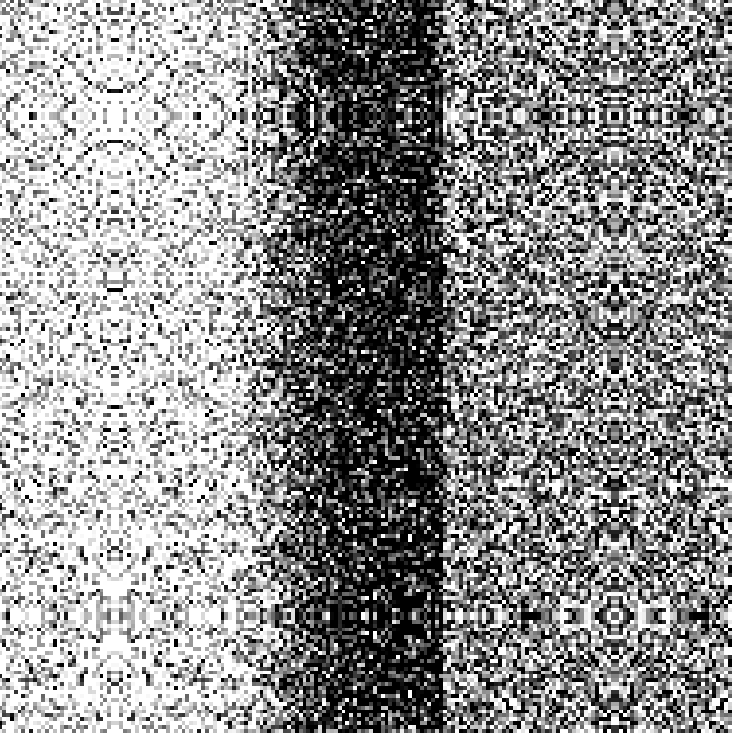}&
		\includegraphics[width=0.15\textwidth]{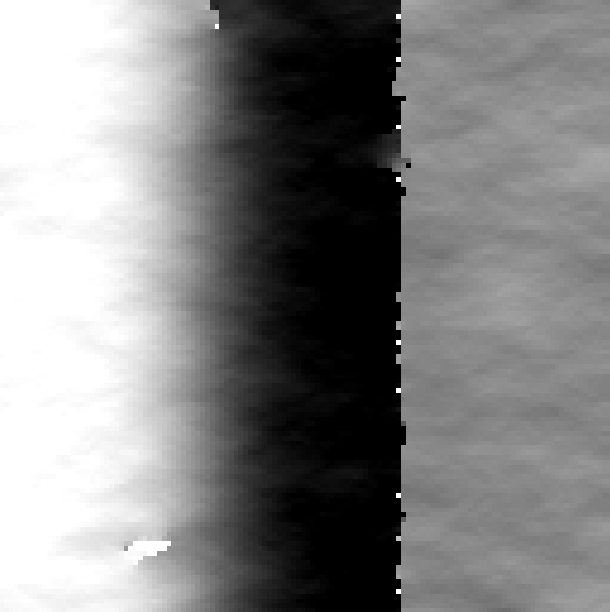}&
		\includegraphics[width=0.15\textwidth]{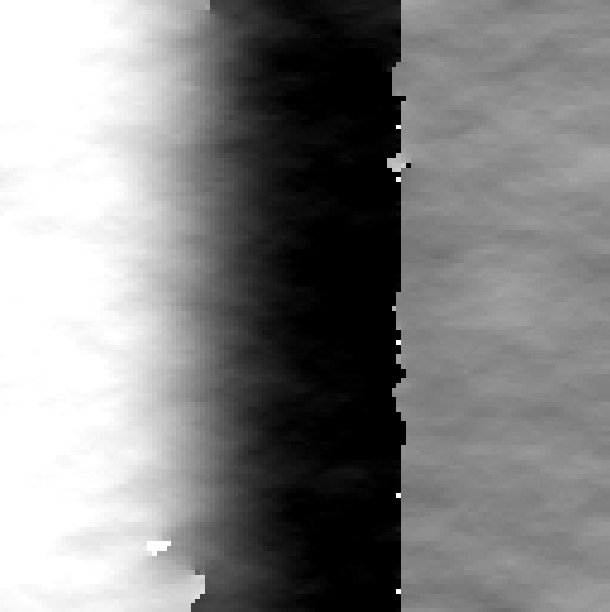}&
		\includegraphics[width=0.15\textwidth]{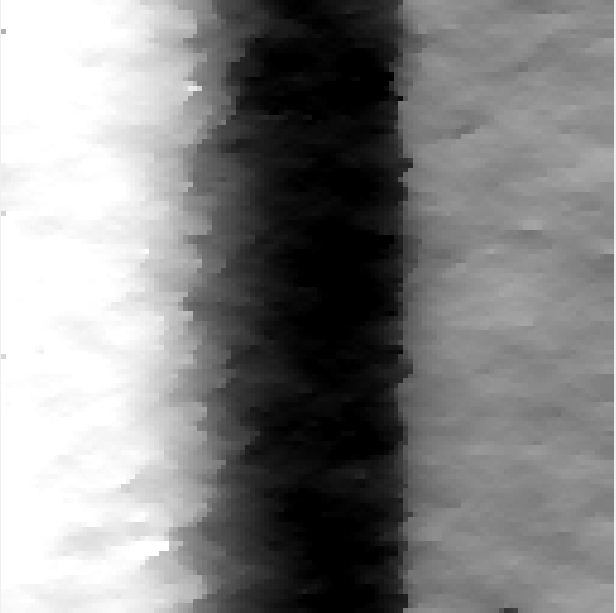}
		&
		\includegraphics[width=0.15\textwidth]{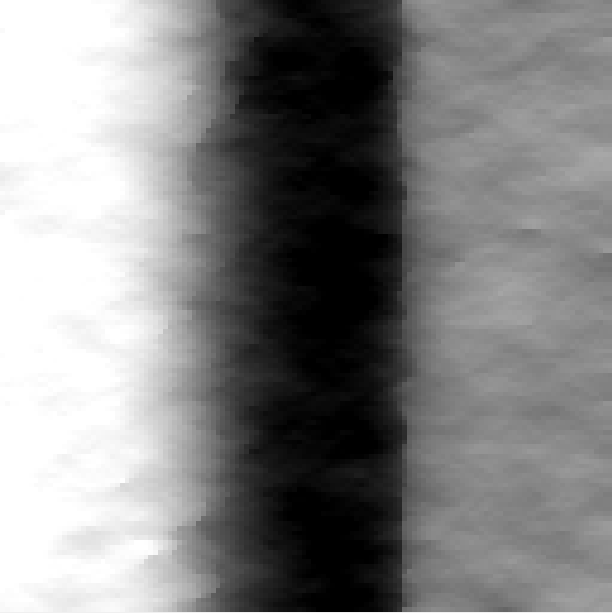}
	\end{tabular}
	\caption{Motivation for curvature regularuzation on level lines. (a) Clean image. (b) Noisy image ($\sigma=120/255$). (c) Denoised image by $L^0$-gradient. (d) Denoised image by the MC penalty function~\cite{zhang2010nearly}. (e) Denoised image by TV. (f) Denoised image by Euler's elastica~\cite{mumford1994elastica}. The noise is captured by $H^{-1}$-norm approximating the Meyer's $G$-norm. Observe that $L^0$-gradient and MC penalty function produce irregular boundaries, and TV-norm creates staircase effects. These artifacts are reduced in (f) when incorporating the curvature regularization on the level lines.}\label{fig_staircase}
\end{figure}

In many tasks of image processing, such as denoising,  segmentation, compression, and pattern analysis,  a given image $f:\Omega\to\mathbb{R}$ defined on  a compact domain $\Omega\subset\mathbb{R}^2$ can be written as a sum of several components, each of which represents a distinctive feature of $f$.
For example, the well known Rudin-Osher-Fatemi (ROF) model~\cite{rudin1992nonlinear} separates a noisy image $f=u+n$ into its noise-free part $u$ and the noisy remainder $n$ by considering
\begin{align}
	\min_{(u,v)\in\text{BV}(\Omega)\times \cL^2(\Omega)/ f=u+n} \left[ \frac{1}{2\lambda}\|n\|^2_{L^2(\Omega)}+\int_\Omega|\nabla u|\right].\label{eq_ROF}
\end{align}
Here $\int_\Omega|\nabla u|$ is the total variation (TV)-norm of a bounded variation function $u\in \text{BV}(\Omega)=\{u\in \cL^1(\Omega): \int_\Omega|\nabla u|<+\infty\}$, and $\lambda>0$ is a weight parameter. 
Various extensions of the ROF model are proposed in the literature mainly from two perspectives. The first branch of works is motivated by the fact that $L^2$-norm in~\eqref{eq_ROF} can fail to capture noise~\cite{meyer2001oscillating,aubert2005modeling}, thus leaving visible oscillations in $u$. Another class of works is based on the observation that TV-norm leads to various artifacts including staircase effects, and loss of geometry as well as contrast~\cite{chan2005recent,ambrosio2003direct,lysaker2003noise,zhu2012image}. See Figure~\ref{fig_staircase} (e) for an illustration.

In his seminal work~\cite{meyer2001oscillating}, Meyer suggested to characterize the  oscillating patterns of an image including textures and noise by the dual space of the closure of the Schwartz class in $\text{BV}(\mathbb{R}^2)$. Later,  Aubert and Aujol~\cite{aubert2005modeling} refined the construction by restricting to the case of open and bounded domain $\Omega$. They proposed the $G(\Omega)$-space as a subspace of the dual of the Sobolev space $\cW^{1,1}_0(\Omega)$, and formulated the Meyer's model as 
\begin{align}
	\min_{(u,v)\in\text{BV}(\Omega)\times G(\Omega)/ f=u+n} \left[ \frac{1}{\lambda}\|v\|_{G(\Omega)}+\int_\Omega|\nabla u|\right],\label{eq_meyer}
\end{align}
where $\|\cdot\|_{G(\Omega)}$ stands for the $G$-norm. A key property of $G$-norm is that, oscillating functions converging to $0$ in $G(\Omega)$ does not have to converge strongly in $\cL^2(\Omega)$~\cite{aubert2005modeling}, which justifies the benefit of using $G$-norm in place of $L^2$-norm in the ROF model for noise. As $G$-norm is related to $L^\infty$-norm, it is computationally challenging to solve~\eqref{eq_meyer}. Vese and Osher~\cite{vese2003modeling} proposed the first practical solution by considering
\begin{align}
	\min_{(u,\bg)\in\text{BV}(\Omega)\times \left(\cL^p(\Omega)\right)^2} \left[ \int_\Omega|\nabla u| + \frac{1}{2\lambda}\int_\Omega|f-u-\partial_xg_1-\partial_yg_2|^2\,d\bx+\mu\|\bg\|_{L^p(\Omega)}\right],\label{eq_veseosher}
\end{align}
where  $d\bx$ stands for the Lebesgue measure in $\mathbb{R}^2$, $\lambda>0$ and $\mu>0$ are weight parameters,  $p\geq1$ is an integer index, and $\|\bg\|_{L^p(\Omega)}=\|\sqrt{g_1^2+g_2^2}\|_{L^p(\Omega)}$ for $\bg=(g_1,g_2)$. They showed that as $\lambda\to0$ and $p\to\infty$, $\partial_xg_1+\partial_yg_2$ belongs to $G(\Omega)$. They derived the Euler-Lagrange equation related to~\eqref{eq_veseosher} and solved the problem after discretization in the case of $p=1$.  In~\cite{osher2003image}, a simplified version of~\eqref{eq_veseosher} when $p=2$ was studied, which leads to using $H^{-1}(\Omega)$ seminorm in place of Meyer's $G$-norm, where $H^{-1}(\Omega)$ is the dual of $H_0^1(\Omega)$. Approaching from a different perspective, Aujol et al.~\cite{aujol2005image} proposed to tackle Meyer's problem by restricting the function space $G(\Omega)$ to a convex subset $G_\mu(\Omega)=\{v\in G(\Omega)~|~\|v\|_{G(\Omega)}\leq \mu\}$ for some $\mu>0$. Analogous to Chambolle's  method~\cite{chambolle2004algorithm} for~\eqref{eq_ROF}, a projection algorithm was proposed in~\cite{aujol2005image} to solve their model. In~\cite{wen2019primal}, Wen et al. considered the Legendre-Fenchel duality of the Meyer's $G$-norm and proposed a primal-dual algorithm for the cartoon-texture decomposition. In the literature, there are many other types of regularizations for capturing texture and noise. For instance, the Bounded Mean Oscillation (BMO) space was considered in~\cite{le2005image,garnett2011modeling}, the negative Hilbert-Sobolev space $\cH^{-s}$ was applied in~\cite{lieu2008image}, and a general family of Hilbert spaces were studied in~\cite{aujol2006structure}. With a more refined decomposition, Aujol and Chambolle~\cite{aujol2005dual} suggested to separate a digital image $f$ as a sum of three components: the geometric part  $u\in\text{BV}(\Omega)$, the texture part $v\in G_{\mu}(\Omega)$, and the noise component $w$ in the Besov space~\cite{meyer2001oscillating} consisting of functions whose wavelet coefficients are in $\ell^\infty$. A fast filter method as well as a nice summary of variational models for cartoon$+$texture decomposition is available in~\cite{buades2010fast}. These variational decomposition frameworks have also been extended and applied to color images~\cite{aujol2006color,le2003color,duval2009projected,wen2022cartoon}, RGB-d images~\cite{chen2013simple}, images on manifold~\cite{wu2013variational}, segmented image decomposition \cite{shen2005piecewise}, and image registration~\cite{chen2024three}.

To address the unsatisfying artifacts of TV-norm~\cite{chan2005recent,zhu2012image,tai2011fast} in the ROF model, much efforts are devoted to constructing new regularizers that incorporate higher-order features or non-convex terms.  The Euler's elastica energy~\cite{mumford1994elastica,masnou1998level, shen2003euler, ambrosio2003direct}
\begin{align}
	\mathcal{R}_{\text{Euler}}(v;a,b)=\int_{\Omega}\left(a+b\left(\nabla\cdot\frac{\nabla v}{|\nabla v|}\right)^2\right)\left|\nabla v\right|\,d\bx\label{eq_ee}
\end{align}
has been used for disocclusion~\cite{mumford1994elastica}, inpainting~\cite{shen2003euler}, and denoising~\cite{ambrosio2003direct,yashtini2015alternating}, where $a,b>0$ are weight parameters. See Figure~\ref{fig_staircase} (f) for the reduction of staircase effects by the Euler's elastica. In~\cite{lysaker2003noise}, Lysaker et al. proposed a regularizer based on the Frobenius norm of Hessian  
\begin{align}
	\mathcal{R}_{H}(v)=\|\partial^2 v\|_{L^1(\Omega)}=\int_\Omega\sqrt{v_{xx}^2+v_{xy}^2+v_{yx}^2+v_{yy}^2}\,d\bx.\nonumber
\end{align}
Regarding the image as a manifold in the Euclidean space with higher dimension, Zhu et al.~\cite{zhu2012image} proposed to use the $L^1$-norm of the mean curvature as the regularizer
\begin{align}
	\mathcal{R}_{\text{mean}}(v)=\int_\Omega\left|\nabla\cdot\left(\frac{\nabla v}{\sqrt{1+|\nabla v|^2}}\right)\right|\,d\bx.\nonumber
\end{align}
From the same geometric point of view, Liu et al.~\cite{liu2022operator} considered the Gaussian curvature   
\begin{align}
	\mathcal{R}_{\text{Gauss}}(v)=\int_\Omega\frac{\text{det}\begin{bmatrix}
			v_{xx}&v_{xy}\\
			v_{yx}&v_{yy}\end{bmatrix}}{(1+|\nabla v|^2)^{3/2}}\,d\bx
	\nonumber\end{align}
to reduce image noise while preserving geometric features. Other  regularizers substituting the TV-norm include the Total Generalized Variation (TGV)~\cite{zhang2018nonconvex,xu2014image}, total variation with adaptive exponent~\cite{blomgren1997total},  non-local TV~\cite{gilboa2009nonlocal}, locally modified TV~\cite{zhu2019first}, and Higher Degree TV (HDTV)~\cite{hu2012higher}.

The aforementioned two streams of research focus on improving different parts of the ROF model~\eqref{eq_ROF}, and there are only a few works combining them. In~\cite{chan2007image}, Chan et al. extended~\cite{aujol2005dual} by considering the inf-convolution of TV-norm and the squared $L^2$-norm of the Laplacian and arrived at the following model
\begin{align}
	&\min_{(v,w,n_1,n_2)\in \text{BV}(\Omega)\times \cH^2(\Omega)\times G_\mu(\Omega)\times B_{1,1}^1}  \bigg[\int_\Omega |\nabla v|+\frac{\alpha}{2}\|\Delta w\|_{L^2(\Omega)}^2\nonumber\\
	&\hspace{5cm} +J^*\left(\frac{n_1}{\mu}\right)+B^*\left(\frac{n_2}{\delta}\right)+\frac{1}{2\lambda}\int_\Omega|f-v-w-n_1-n_2|^2\,d\bx\bigg]\;.\label{eq_chan}
\end{align}
Here $\alpha>0$, $J^*\left(\frac{n_1}{\mu}\right)$ and $B^*\left(\frac{n_2}{\delta}\right)$ are the Legendre-Fenchel conjugate of the $G$-norm and the $B_{-1,\infty}^\infty$-norm, which enforce boundedness $\|n_1\|_{G(\Omega)}\leq \mu$ and $\|n_2\|_{B_{-1,\infty}^{\infty}}\leq\delta$, respectively, and $B_{-1,\infty}^\infty$ is the dual to the homogeneous Besov space $B_{1,1}^1$~\cite{meyer2001oscillating}. In~\eqref{eq_chan},  $v$ corresponds to the image's piecewise constant part, and $w$ can be associated with the smoothly varying part such as soft lighting and blurry shadows. The $n_1$ component corresponds to the textures, and the $n_2$ component captures the noise.  Recently, Huska et al.~\cite{huska2021variational} proposed another three-component decomposition model, where a  discrete image $f:\Omega\cap \mathbb{N}^2\to\mathbb{R}$ is modeled as a sum of a cartoon part $v$ that captures the piecewise constant structures of the image, a smooth part $w$ that models the uneven lighting conditions in the scene, and an oscillatory component $n$ that includes textures and noise. Specifically, their model is
\begin{align}
	\min_{v,w,n\in\mathbb{R}^{N\times M}} \left[\beta_1\|\nabla v\|_0+\beta_2\|\partial^2 w\|_{2}^2+\beta_3\|n\|^4_{H^{-1}(\Omega)}+\frac{1}{2\lambda}\|f-v-w-n\|_2^2\right],\label{eq_huska}
\end{align}
where $\|\nabla v\|_0$ is the  discrete $L^0$-gradient energy~\cite{xu2011image} that counts the pixels with non-zero gradient, $\partial^2w(\bx)\in\mathbb{R}^{2\times 2}$ for each pixel $\bx$ is the  Hessian matrix, $\|\cdot\|_{H^{-1}(\Omega)}$ is the seminorm of the negative Sobolev space $H^{-1}(\Omega)$ as an approximation of the $G$-norm, $\|\partial^2 w\|_{L^2(\Omega)}^2=\left(\sum_{\bx\in \Omega\cap \mathbb{N}^2}w_{xx}^2(\bx)+2w^2_{xy}(\bx)+w^2_{yy}(\bx)\right)^2$, and $\beta_i$, $i=1,2,3$ are positive weight parameters. For numerical computation, they replaced $\|\nabla v\|_0$ by a modified minimax concave (MC) penalty function~\cite{zhang2010nearly}. Note that the regularization on $n$ is the fourth power of the $H^{-1}(\Omega)$ seminorm, which was heuristically found to be useful in distinguishing $w$ and $n$~\cite{huska2021variational}. One of the major drawbacks of solely using the $L^0$-gradient regularization is visualized in Figure~\ref{fig_staircase}. Since $L^0$-gradient regularization is global and heights of boundary jumps do not affect its value, it can lead to strong staircase effects and loss of geometry~\cite{chan2005recent}.    
\begin{figure}[t!]
	\centering
	\includegraphics[width=0.8\textwidth]{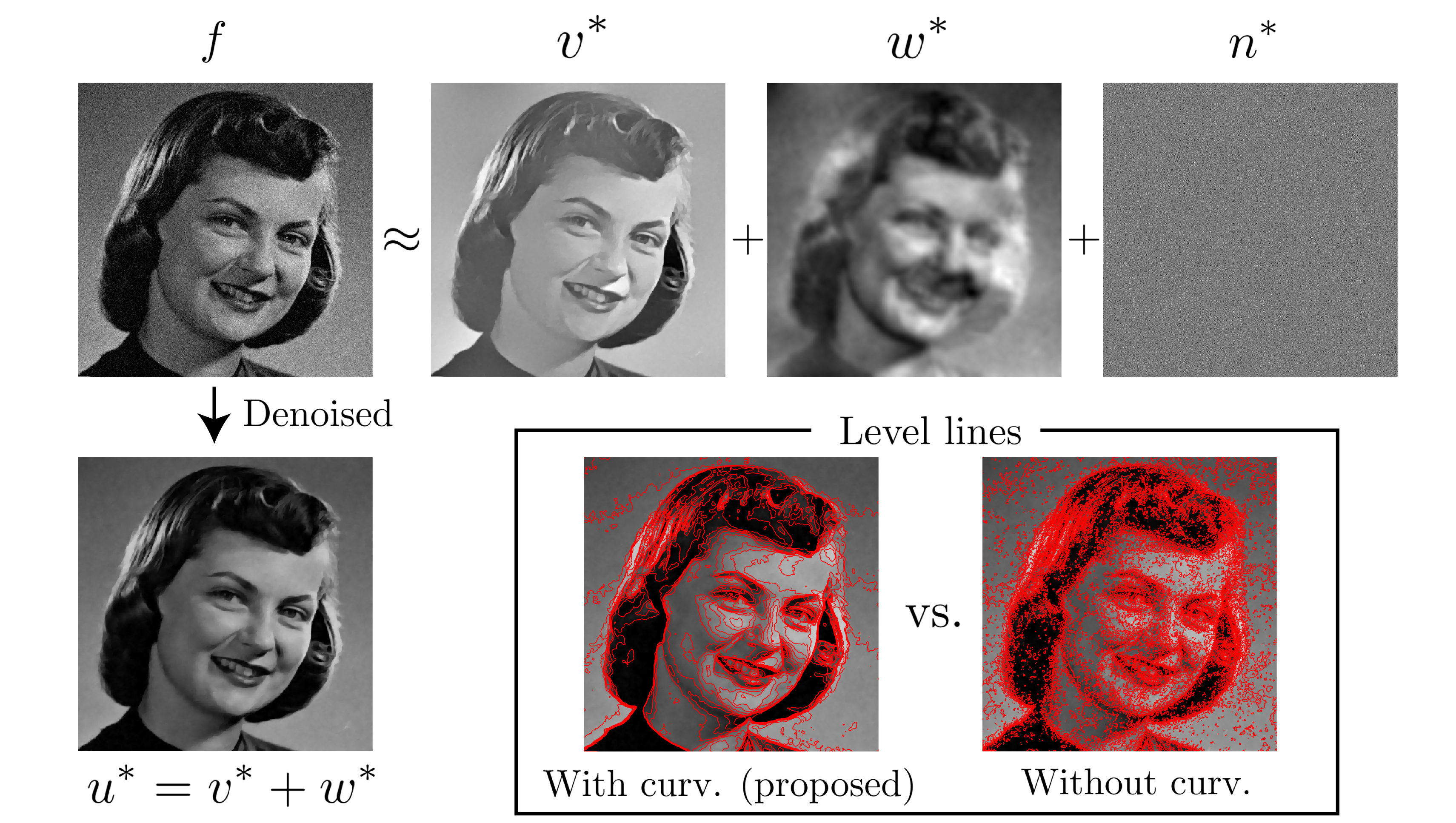}
	\caption{Three-component image decomposition by the proposed method. A grayscale image $f$ is decomposed into a structural part $v^*$, a smooth part $w^*$, and an oscillatory part $n^*$. The sum of $v^*$ and $w^*$, denoted by $u^*$, can be regarded as the denoised version of $f$ if the clean image does not have rich textures. Our effectively reduces the staircase effects.}\label{fig_demo}
\end{figure}

In this paper, we propose a new three-component decomposition model to address the issues discussed above. We decompose any grayscale image into its structural part,  smooth part, and oscillatory part. For the structural part,   our model enforces strong sparsity on the gradient while reducing the presence of staircase effects and encouraging shape-preserving boundary smoothing. We achieve this by proposing a generalized $L^0$-gradient energy defined for images over continuous domains and combining it with the Euler's elastica~\eqref{eq_ee} which controls the magnitude of level-line curvature. For the smooth part, instead of using the Frobenius norm of the Hessian as in~\eqref{eq_huska}, we substitute it with squared $L^2$-norm of the Laplacian, $\|\Delta w\|_{L^2(\Omega)}^2$,  to encourage isotropic smoothness. It can be shown that $\|\Delta w\|^2_{L^2(\Omega)}=\|\partial^2 w\|^2_{L^2(\Omega)}$  for compactly supported smooth functions~\cite{chan2007image}; we prove that they have the same critical points under certain boundary conditions (see Appendix~\ref{append_regularizer}).  For the oscillatory part, we use the squared  $H^{-1}(\Omega)$ semi-norm as a  practical substitute for the Meyer's $G$-norm.  It yields not only effective extraction of the oscillations similarly to the 4-th power in~\eqref{eq_huska} but also significant improvement on the computational efficiency.

Figure~\ref{fig_demo} illustrates the general performance of our model. The structural part, $v^*$, contains definite contours and shades with sharp transition; the smooth part, $w^*$,  captures soft intensity variation on the cheeks and chin; and the oscillatory part, $n^*$, includes noise and textures such as those of the hair. In case the input image does not contain fine-scale textures, the composition $u^*=v^*+w^*$ gives a denoised version of $f$ where the noisy perturbation is removed and sharp boundaries are preserved. The level lines of $u^*$ by our model are more regular compared to those without the curvature regularization. This shows that the proposed model effectively reduces the staircase effects caused by the $L^0$-gradient regularization. 

Our model involves non-convex non-smooth optimization, thus it is challenging to solve. For the discrete $L^0$-gradient minimization, various algorithms and approximations are employed in the literature, such as hard-thresholding~\cite{xu2011image,ding2018image}, region-fusion~\cite{nguyen2015fast}, and  penalty approximation~\cite{huska2021variational}. In~\cite{ono2017l_}, the $L^0$-gradient projection was considered and solved by using a mixed $\ell^{0,1}$ pseudo-norm. Since we introduce a generalized $L^0$-gradient energy, a different strategy is taken. Another nontrivial part of our proposed model is the presence of the nonlinear high-order term for the curvatures of level lines. A classical approach is to derive the first variation and then discretize the associated Euler-Lagrange equation~\cite{shen2003euler}. With higher efficiency, various Operator-Splitting Methods (OSMs) and Augmented Lagrangian Methods (ALMs)~\cite{glowinski1989augmented,glowinski2016some,tai2011fast,liu2022operator,he2020curvature,huska2021variational,liu2023elastica,deng2019new,yashtini2015alternating} have been proposed to handle variational models which contain curvature regularization. OSMs are commonly derived from associating an initial value problem with the formal optimality condition, and ALMs are usually developed via the augmented Lagrangian. These approaches are deeply related and sometimes equivalent~\cite{glowinski1989augmented,glowinski2016some,esser2009applications}. The main advantage of these paradigms come with appropriately splitting the original problem into several sub-problems, each of which can be efficiently handled. Recently, connections between OSMs and neural networks are investigated in \cite{lan2023dosnet,tai2024pottsmgnet,liu2023double}.

To efficiently complete the decomposition, we propose a fast operator-splitting scheme that leverages the Fast Fourier Transform (FFT) and explicit formulas for the involved sub-problems. For an image of size $N\times M$, each iteration of the proposed algorithm has a computational complexity of $\mathcal{O}\left(MN(\log M +\log N)\right)$.  In comparison, the computational complexity in~\cite{huska2021variational} is $\mathcal{O}((MN)^3)$. We present various examples to validate our method. Specifically, we conduct ablation studies to verify the combination of regularization terms in our method. We design experiments to investigate the impact of regularization parameters. We also demonstrate the effectiveness and efficiency of our approach by comparing it to state-of-the-art models for three-component-decomposition.

Our main contributions in this work are
\begin{itemize}
	\item We propose a new  model  for decomposing a grayscale image defined on a continuous domain into its structural part, smooth part, and oscillatory part. To achieve satisfying results,  we incorporate curvature regularization on level lines along with a  sparsity-inducing $L^0$-gradient.  Our model effectively reduces the undesirable staircase effects and irregular boundary curves caused by the geometry-agnostic $L^0$-gradient, yielding more visually appealing components.
	\item We develop a new operator-splitting scheme  to tackle the  non-convex non-smooth optimization problem associated with our model. The proposed algorithm is computationally efficient. Each sub-problem is directly solved or numerically addressed using FFT techniques. Remarkably, when compared to several state-of-the-art methods, our algorithm achieves a speedup of over 10 times. 
	\item We present systematic ablation and comparison studies to justify the  effectiveness and efficiency of our proposed method. We also numerically investigate the effects of model parameters and discuss the problem of parameter selection. 
\end{itemize}
The paper is organized as follows. In Section~\ref{sec_model}, we propose our model. In particular,  we introduce our model accompanied by mathematical remarks in Section~\ref{sec_formal}; and reformulate it in Section~\ref{sec_reform} for the subsequent algorithmic developments. In Section~\ref{sec_algo}, we  propose our operator-splitting scheme to address the non-convex non-smooth functional minimization problem associated with the reformulated model. In Section~\ref{eq_num_discret}, we describe the discretization and implementation details. In Section~\ref{sec_experiment}, we present various experiments to justify the proposed model and compare it with other methods from different aspects. We conclude the paper in Section~\ref{sec_conclude}. 

\section{Proposed Cartoon-Smooth-Texture Decomposition Model}\label{sec_model}
In this section, we propose a new model for cartoon-smooth-texture image decomposition integrating Euler's elastica. We present a formal version of our model in Section~\ref{sec_formal} where each component is characterized by a functional. Considering the difficulty of the problem,  we introduce a reformulation of our model in Section~\ref{sec_formal} and derive a formal optimality condition, which paves our way towards developing an efficient operator splitting scheme in Section~\ref{sec_algo}. 

\subsection{Proposed model}\label{sec_formal}
Suppose $\Omega\subset\mathbb{R}^2$ is a compact domain with a Lipschitz boundary. We assume that a grayscale image $f:\Omega\to\mathbb{R}$ is a sum of three components with distinctive features.  In particular, we consider $f$ to be composed by a \textbf{structural part} $v^*:\Omega\to\mathbb{R}$ that defines objects with sharp boundaries or regions outlined by strong transitions between the light and the dark; a \textbf{smooth part} $w^*:\Omega\to\mathbb{R}$ that characterizes the soft intensity variation, such as shadows and uneven lighting; and an \textbf{oscillatory part} $n^*:\Omega\to\mathbb{R}$ that captures fine-scale textures and noise.
Our objective is to extract these three components from $f$, and we propose to achieve this by considering the following problem
\begin{align}
	&\{v^*,w^*,n^*\}\in \nonumber\\
	&\argmin_{\substack{v,w,n\\v+w+n=f}}\left[\alpha_0{L^0_G(v)}+\alpha_{\text{curv}}\int_{\Omega}{\kappa^2(v)}\left|\nabla v\right|\,d\bx + \alpha_w\|\Delta w\|_{L^2(\Omega)}^2+\alpha_n\|n\|^2_{H^{-1}(\Omega)}\right].\label{eq:proposed_model}
\end{align}
{In our proposed model~\eqref{eq:proposed_model}, each component of the given image $f$ is characterized by different functionals, and we state their mathematical definitions as follows.
	\begin{itemize}
		\item \textbf{Structure part $v$}. Define the function space
		\begin{align}
			\mathcal{V}(\Omega) = \{&v\in H^1(\Omega)~|~\bnu_v \in H(\text{div},\Omega)\;,\nonumber\\
			&\text{where for}~\bx\in\Omega,~\bnu_v(\bx)=\frac{\nabla v(\bx)}{|\nabla v(\bx)|}~\text{if}~|\nabla v(\bx)|\neq 0~\text{and}~\mathbf{0}~\text{otherwise}\}\label{eq_space_v}
		\end{align}
		and $H(\text{div},\Omega)$ denotes the Hilbert space of square-integrable vector fields on $\Omega$ with square-integrable weak divergences; see for example~\cite{arnold2000multigrid}. In this work, we define the $L^0$-gradient energy of $v\in\mathcal{V}(\Omega)$  as
		\begin{equation}
			L^0_G(v)=\int_{\Omega}\mathbb{H}(|\nabla v|)\,d\bx\;,~\label{eq_Dv}
		\end{equation}
		where $\mathbb{H}(\cdot)$ denotes the Heaviside step function, which outputs a value of 1 for positive inputs and a value of 0 otherwise.
		
		Moreover, since $v\in\mathcal{V}(\Omega)$, we can interpret the Euler's elastica part as 
		\begin{equation}
			\int_{\Omega}{\kappa^2(v)}\left|\nabla v\right|\,d\bx=\int_{\Omega}|\nabla\cdot\bnu_v|^2|\nabla v|\,d\bx\;,\label{eq_euler_elastica_functional}
		\end{equation}
		where $\bnu_v$ is defined as in~\eqref{eq_space_v}, and $\nabla\cdot\bnu_v$ denotes its weak divergence. By the definition of $\mathcal{V}(\Omega)$,~\eqref{eq_euler_elastica_functional} is finite and well-defined.

		\item \textbf{Smooth part $w$}. We take $w\in H^2(\Omega)$ and define
		\begin{equation}
			\|\Delta w\|_{L^2(\Omega)}^2=\int_{\Omega}\left(\partial^2_{x} w(\bx) +\partial^2_{y}w(\bx)\right)^2\,d\bx\;.
		\end{equation}
		Here the partial derivatives $\partial^2_x$ and $\partial_y^2$ are understood in the weak sense. Its effect is similar to that of the squared $L^2$-norm of the second-order derivatives, i.e., $\|\partial^2 w\|_{L^2(\Omega)}^2$ used in~\eqref{eq_huska}. Under certain boundary conditions, we show that they are equivalent in the sense of having the same critical points (see Appendix~\ref{append_regularizer}), which generalizes the observation in~\cite{chan2007image}. Using this term, we gain significant computational efficiency thanks to the simpler first variation (see Section~\ref{sec_algo}).
		\item \textbf{Oscillatory part $w$}.   As for the oscillatory part, we take $n\in H^{-1}(\Omega)$ and use the squared inverse Sobolev norm 
		\begin{equation}
			\|n\|_{H^{-1}(\Omega)}^2=\|\nabla\Delta^{-1}n\|_{L^2(\Omega)}^2
		\end{equation}
		as a practical approximation for the Meyer's $G$-norm. 
	\end{itemize}
}

We design functionals in~\eqref{eq:proposed_model} for better visual effects and computational efficiency.
The sparsity-inducing $L^0$-gradient energy~\eqref{eq_Dv}  punishes non-zero gradients independent of their magnitudes, yielding results with higher contrast; this also implies  stronger stair-casing effects. To reduce such artifacts in $v$, our second term regularizes the TV-weighted squared curvatures of its level-lines. It combines high-order derivative information of the image, which has been shown to be an effective strategy for improving the visual effects of the reconstruction~\cite{chan2007image}. The squared $L^2$-norm of the Laplacian encourages isotropic smoothness of $w$, modeling the gradual changes of illumination over the scene.  The inverse Sobolev norm on the oscillatory component was discussed in~\cite{meyer2001oscillating,osher2003image,huska2021variational}.

\begin{remark}    We note that~\eqref{eq_Dv} is a continuous analogue of the discrete $L^0$-gradient energy~\cite{xu2011image,huska2021variational}. In particular,  for a discrete image $v_D:\Omega\times \mathbb{N}^2\to\mathbb{R}$, we can obtain a continuous extension $v:\Omega\to\mathbb{R}$ by bilinear interpolation, then $v\in \mathcal{V}(\Omega)$, and
	$\|\nabla v_D\|_{0}$ equals $L_G^0(v)$ multiplied by a constant factor. It is possible to consider more general function spaces for the structure part, e.g., the space of bounded variation functions $\text{BV}(\Omega)$ or de Giorgi and Ambrosio's $\text{SBV}(\Omega)$~\cite{de1989new}, and the $L^0$-gradient energy may be defined via mollification.  An alternative is to consider a sufficiently smooth approximation such as the minimax concave penalty function~\cite{zhang2010nearly,huska2021variational}. For the Euler's elastica part, the major technical problem is that level-lines of BV functions generally lack the sufficient regularity to define curvature in the classical sense.  One strategy is to restrict $v$ to $C^2(\Omega)$~\cite{masnou2006variational}; another approach is to define the weak curvature for BV functions after mollification~\cite{shen2003euler}; a different paradigm involves considering the solution of the $p$-Laplacian equation as $p\to 1^+$; yet another method relies on the pairing theory~\cite{anzellotti1983pairings} for distributions~\cite{pimenta2021existence}. In our setting, although $v$ needs to be continuous in theory, it does not cause practical problems by our splitting algorithm (see Section~\ref{sec_algo}). We shall leave the challenging study on  generalizations in the future works.
\end{remark}

\begin{remark} No matter which mathematical definitions are used for interpretation, our model~\eqref{eq:proposed_model} is challenging to analyze for the existence of minimizers. The main difficulty is caused by the $L^0$-gradient energy and the Euler's elastica energy.  If we use the TV-norm for regularizing $v$ instead, the resulting model becomes convex, and the existence property can be derived similarly to~\cite{chan2007image}. With an additional restriction on the mean value of $w$, then the uniqueness can be established. If we replace the $L^0$-gradient energy by the TV-norm, we recover the ordinary Euler's elastica~\cite{masnou1998level,masnou2006variational,shen2003euler} for the $v$ component, and the related analysis on semi-continuity in the context of $\Gamma$-convergence can be found in~\cite{bellettini1997variational}. 
\end{remark}

\begin{remark}
	In our proposed model~\eqref{eq:proposed_model}, both of the regularization on the curvature of level-lines of $v^*$ and the regularization on the Laplacian of $w^*$ encourage smoothness. However, they emphasize different types of smoothing. The curvature regularization on  $v^*$ admits a geometric prior that prefers straight level lines, whereas the regularization on $w^*$ has no preferences in diffusion directions, i.e., it is isotropic.

	There exist functions $w$ defined on $\Omega\subset\mathbb{R}^2$ whose curvatures of level lines are constantly zero but not harmonic. For example, consider $w(x_1,x_2) = x_1^2$ defined over $(x_1,x_2)\in \Omega =[-1,1]\times [-1,1]$. There also exist harmonic functions  whose level lines have non-zero curvatures, such as $v(x_1,x_2)=x_1^2-x_2^2$. 
	For general images,  the amount of smoothness in a given image can be transferred between $v^*$ and $w^*$ depending on the ratio $\alpha_{\text{curv}}/\alpha_{w}$. We investigate this special property of our model in Section~\ref{sec_experiment} using real images.
	
	We emphasize that the $v^*$ component is also regularized by~\eqref{eq_Dv}, which is crucial in distinguishing the structural features in $v^*$ from the soft shading and blurry  shapes represented by $w^*$. Consider a Gaussian blurred smooth function
	$f_{\varepsilon}(x_1,x_2)=G_\varepsilon*h(x_1,x_2)$ defined on $[0,1]\times[0,1]$ where $h(x_1,x_2)=1$ if $0\leq x_1\leq 0.5, 0\leq x_2\leq 1$, and $h(x_1,x_2) = 0$ otherwise,  $G_{\varepsilon}$ is the Gaussian kernel with standard deviation $\varepsilon>0$, and $*$ is the convolution. The level lines of $f_{\varepsilon}$ remain straight and $\|\Delta f\|_{L^2(\Omega)}^2=\mathcal{O}(\varepsilon^{-4})>0$ which decreases in $\varepsilon$. If $v^*$ is only penalized by the curvature of its level line and we ignore the oscillatory component for simplicity, then the optimal decomposition is $v^*=f_{\varepsilon}$ and $w^*=0$ for any 
	choice of $\varepsilon$. This is contradictory to the visual perception of homogeneous structures with clear boundaries.  To consider the $L^0$-gradient, we first note that the Gaussian kernel has a compact support in the discrete sense. In particular, for a fixed threshold $\gamma>0$,  we note that
	$$\frac{1}{\varepsilon\sqrt{2\pi}}\exp\left(-\frac{x_1^2}{2\varepsilon^2}\right)>\gamma\;$$
	only if
	$$-\sqrt{-2\varepsilon^2\ln(\gamma\varepsilon\sqrt{2\pi})}\leq x_1\leq \sqrt{-2\varepsilon^2\ln(\gamma\varepsilon\sqrt{2\pi})}\;.$$
	Hence, $ L_G^0(f_{\varepsilon})\approx 2\sqrt{-2\varepsilon^2\ln(\gamma\varepsilon\sqrt{2\pi})}$, which increases when $\varepsilon$ is bigger, i.e., $f_{\varepsilon}$ is blurrier. With appropriate choices of the parameter $\alpha_0$ and $\alpha_{n}$, the optimal decomposition will be $v^*=0$ and $w^*=f_{\varepsilon}$ if $\varepsilon$ is sufficiently large, and the optimal decomposition becomes $v^* = f_{\varepsilon}$ and $w^*=0$ if $\varepsilon$ is sufficiently small. By considering the $L^0$-gradient of $v^*$ component, the decomposition is compatible with visual perception.
\end{remark}
\subsection{Model reformulation and formal optimality condition}\label{sec_reform}
As discussed above, the Model~\eqref{eq:proposed_model} is non-trivial to analyze and solve. To arrive at a practical form that can be numerically addressed, we relax~\eqref{eq:proposed_model} by rewriting the constraint $f=v+w+n$ as a quadratic penalization. Our modified model is  
\begin{align}
	\min_{\substack{v\in \mathcal{V}(\Omega),\\
			w\in \cH^2(\Omega),n\in \cH^{-1}(\Omega)}}\bigg[&\alpha_0 \int_{\Omega}\mathbb{H}(|\nabla v|)\,d\bx +\alpha_{\text{curv}}\int_{\Omega}|\nabla\cdot\bnu_v|^2\left|\nabla v\right|\,d\bx\nonumber\\
	&+ \alpha_w\|\Delta w\|_{L^2(\Omega)}^2
	+\alpha_n\|n\|^2_{H^{-1}(\Omega)} +\frac{1}{2}\|f-(v+w+n)\|_{L^2(\Omega)}^2\bigg],\label{eq:model1}
\end{align}
where $v,w$ and $n$ stand for the same components as in~\eqref{eq:proposed_model}, and $\mathcal{V}(\Omega)$ is defined in~\eqref{eq_space_v}. We then resolve the high-order and non-linear term associated with the curvature by introducing auxiliary variables
$\bq(\bx) = 
\nabla v(\bx)$ and $
\bmu(\bx) = \bq(\bx)/|\bq(\bx)|$
which satisfy the obvious constraints $\bq(\bx)\cdot \bmu(\bx) = |\bq(\bx)|$ and $|\bmu(\bx)|\leq 1$ for any $\bx\in\Omega$ as in~\cite{deng2019new}. Moreover, if we let $\bg=(g_1,g_2)=\nabla \Delta^{-1}v\in L^2(\Omega)\times L^2(\Omega)$ as in~\cite{osher2003image}, then~\eqref{eq:model1} is rewritten as
\begin{equation}
	\begin{cases}
		\displaystyle \min_{\substack{v\in \cH^1(\Omega)\\\bq\in (\cL^2(\Omega))^2,\bmu\in (\cH^1(\Omega))^2,\\ w\in \cH^2(\Omega),\bg\in (\cH^1(\Omega))^2}}\bigg[\alpha_0\int_{\Omega}\mathbb{H}(|\bq|)\,d\bx +\alpha_{\text{curv}}\int_{\Omega}\left(\nabla\cdot\bmu\right)^2\left|\bq\right|\,d\bx\\
		\hspace{2cm}+ \alpha_w\|\Delta w\|_{L^2(\Omega)}^2+\alpha_n\|\bg\|^2_{L^{2}(\Omega)}+\frac{1}{2}\|f-(v+w+\nabla\cdot \bg)\|_{L^2(\Omega)}^2\bigg]\\
		\bq =\nabla v,\\
		\bmu=\frac{\bq}{|\bq|},\\
		|\bmu|\leq 1.
	\end{cases}\label{eq:model_splitted}
\end{equation}

We note that Model~\eqref{eq:model_splitted} can be further simplified by observing the following characterization of minimizers of Model~\eqref{eq:model1}, whose proof is in Appendix~\ref{sec_proof_prop}.
\begin{prop}\label{prop:minimizer}
	Let $(v^*,w^*,n^*)$ be a minimizer of~\eqref{eq:model1}, then we have
	\begin{align}
		\int_{\Omega} v^*+w^*+n^*\,d\bx = \int_{\Omega} f \,d\bx.\label{eq_prop}
	\end{align}
\end{prop}
Proposition \ref{prop:minimizer} is a consequence of $L^2$ minimization, which has been used in the literature, see \cite{deng2019new} for example.
This implies that if $(v^*,\bq^*,\bmu^*,w^*,\bg^*)$ is a minimizer of Model~\eqref{eq:model_splitted}, $v^*$ solves the following Poisson equation
\begin{align}
	\begin{cases}
		\Delta v=\nabla\cdot\bq^* &\mbox{ in } \Omega,\\
		(\nabla v-\bq^*)\cdot \bn=0 &\mbox{ on } \partial\Omega,\\
		\int_\Omega v\,d\bx=\int_\Omega f-w^*-\nabla\cdot\bg^*\,d\bx.
	\end{cases}
	\label{eq.vq}
\end{align}
where $\bn$ denotes the outward normal direction along $\partial\Omega$. Since the condition $\int_{\Omega}\nabla\cdot\bq^*\,d\bx=\int_{\partial\Omega} \bq^*\cdot\bn\,d\sigma(\bx)$ follows from the divergence theorem, where $\sigma(\cdot)$ is the surface measure,~\eqref{eq.vq} admits a unique solution~\cite{evans2022partial}. This fact allows us to express $v$ via $\bq,w$, and $\bg$, thus  reducing the number of optimizing variables by one.

It is difficult to directly optimize the $L^0$-gradient energy since the integrand inside the integral is discontinuous. To bypass this difficulty during the derivation of our algorithm, we consider a smoothed version
\begin{equation}
	\mathcal{R}_a(\bp)=\int_{\Omega} \frac{a|\bp|}{1+a|\bp|} d\bx, \quad a>0\;.\label{eq_Ra}
\end{equation}
When $a$ is finite, $\mathcal{R}_a(\cdot)$ is locally Lipschitzian; moreover, we have 
$$\lim_{a\rightarrow +\infty} \mathcal{R}_a(\bq)=\int_{\Omega}\mathbb{H}(|\bq|)\,d\bx\;.$$

Utilizing~\eqref{eq.vq} and~\eqref{eq_Ra}, we consider an approximation of~\eqref{eq:model_splitted} as 
\begin{align}
	\min_{\substack{\bq\in (\cL^2(\Omega))^2,\bmu\in (\cH^1(\Omega))^2,\\ w\in \cH^2(\Omega),\bg\in (\cH^1(\Omega))^2}}\bigg[&\alpha_0\mathcal{R}_a(\bq)+\alpha_{\rm curv}\int_\Omega \left|\nabla\cdot\bmu\right|^2\left|\bq\right|\,d\bx+\alpha_{w}\int_{\Omega}|\Delta w|^2d\bx +\alpha_{n}\int_{\Omega}|\bg|^2 d\bx \nonumber\\
	&+\frac{1}{2}\int_{\Omega}|f-(v_{\bq,w,\bg}+w+\nabla \cdot \bg)|^2d \bx+I_{\Sigma_f}(\bq,w ,\bg)+I_S(\bq,\bmu)\bigg],
	\label{eq:reform_model}
\end{align}
where $v_{\bq,w,\bg}$ is uniquely determined by $\bq,w,\bg$ via~\eqref{eq.vq}; and the indicator functions
\begin{align*}
	I_{\Sigma_f}(\bq,w ,\bg)=\begin{cases}
		0 & \mbox{ if } (\bq,w ,\bg)\in \Sigma_f,\\
		+\infty & \mbox{ otherwise},
	\end{cases} \quad
	I_S(\bq,\bmu)=\begin{cases}
		0 & \mbox{ if } (\bq,\bmu)\in S,\\
		+\infty & \mbox{ otherwise}.
	\end{cases}
\end{align*}
with
\begin{align*}
	&\Sigma_f = \bigg\{(\bq,w,\bg)\in(L^2(\Omega))^2\times L^2(\Omega)\times (H^1(\Omega))^2~|~\nonumber\\
	&\hspace{3cm} \exists v\in H^1(\Omega)~\text{such that}~\bq=\nabla v~\text{and}~\int_\Omega(v+w+\nabla\cdot \bg-f)\,d\bx=0\bigg\},\\
	&S =\left\{(\bq,\bmu)\in(L^2(\Omega))^2\times(L^2(\Omega))^2~|~\bq\cdot\bmu=|\bq|,|\bmu|\leq 1\right\}.
\end{align*}

Assuming that~\eqref{eq:reform_model} has a minimizer $\{\bq^*,\bmu^*,w^*,\bg^*\}$ (See Remark~\ref{remark_assumption}),  we have $v^*=v_{\bq^*,w^*,\bg^*}$ for the structural part, $w^*$ for the smooth part, and $n^*=\nabla\cdot\bg^*$ for  the oscillatory part. Moreover, we denote
\begin{align*}
	&J_0(\bq,w,\bg)=\frac{1}{2}\int_{\Omega}|f-(v_{\bq,w,\bg}+w+\nabla\cdot \bg)|^2d\bx,\\
	&J_1(\bq) = \alpha_0\mathcal{R}_a(\bq),\\
	&J_2(\bq,\bmu) = \alpha_{\rm curv}\int_\Omega \left|\nabla\cdot\bmu\right|^2\left|\bq\right|\,d\bx,\\
	&J_3(w)=\alpha_{w}\int_{\Omega} |\Delta w|^2d\bx,\\
	&J_4(\bg) = \alpha_n\int_{\Omega}|\bg|^2 d\bx.
\end{align*}
and the minimizer $\{\bq^*,\bmu^*,w^*,\bg^*\}$ satisfies the formal optimality condition
\begin{align}
	\begin{cases}
		\partial_{\bq}J_1(\bq^*)+\partial_{\bq}J_2(\bq^*,\bmu^*)+\partial_{\bq}J_0(\bq^*,w^*,\bg^*)+\partial_{\bq}I_{\Sigma_f}(\bq^*,w^*,\bg^*)+\partial_{\bq}I_S(\bq^*,\bmu^*)\ni 0,\\
		\partial_{\bmu} J_2(\bq^*,\bmu^*)+\partial_{\bmu}I_{S}(\bq^*,\bmu^*)\ni 0,\\
		\partial_wJ_3(w^*)+\partial_wJ_0(\bq^*,w^*,\bg^*)+\partial_{w}I_{\Sigma_f}(\bq^*,w^*,\bg^*)\ni 0,\\
		\partial_{\bg} J_4(\bg^*)+\partial_{\bg} J_0(\bq^*,w^*,\bg^*)+\partial_{\bg}I_{\Sigma_f}(\bq^*,w^*,\bg^*)\ni 0.
	\end{cases}\label{eq:optimality}
\end{align}
where $\partial$ is the generalized gradient \cite[Section 1.2, Equation (2)]{clarke1990optimization} extended to functionals.

\begin{remark}\label{remark_assumption}
	The existence of minimizer for~\eqref{eq:reform_model} can ensure the  validity of~\eqref{eq:optimality}. Indeed,~(\ref{eq:optimality}) involves locally Lipscthiz functions and indicator functionals. According to \cite[Proposition 2.3.3]{clarke1990optimization}, the generalized gradient of the sum of functions is a subset of the sum of the generalized gradient of each function, which can be also extended to  indicator functionals. In this paper, we focus on the algorithmic development,  leaving the question of existence to future research.
	
\end{remark}

\section{Operator-Splitting Algorithm for the Proposed Model}\label{sec_algo}
In this section, we propose an efficient algorithm for Model~\eqref{eq:reform_model} based on an operator-splitting framework. In Section~\ref{sec_reform}, we associate the model with a gradient flow based on its formal optimality condition. In Section~\ref{sec_operator_split}, we present a novel operator-splitting scheme to solve the reformulated model. This non-convex non-smooth minimization problem is decomposed into several subproblems, whose solutions are described in Section~\ref{sec_sub_l0}-\ref{sec_sub_end}. In Section~\ref{sec_period}, we adapt the proposed algorithm to periodic boundary condition. We summarize the proposed scheme in Algorithm~\ref{alg_DOSH}.

\subsection{Proposed  scheme}\label{sec_operator_split}

We associate~\eqref{eq:optimality} with the following initial value problem in analogous to the gradient flow
\begin{align}
	\begin{cases}
		\frac{\partial\bp}{\partial t}+\partial_{\bq}J_1(\bp)+\partial_{\bq}J_2(\bp,\blambda)+\partial_{\bq}J_0(\bp,r,\bs)+\partial_{\bq}I_{\Sigma_f}(\bp)+\partial_{\bq}I_S(\bp,\blambda)\ni 0,\\
		\gamma_1\frac{\partial\blambda}{\partial t}+\partial_{\bmu} J_2(\bp,\blambda)+\partial_{\bmu}I_{S}(\bp,\blambda)\ni 0,\\
		\gamma_2\frac{\partial r}{\partial t}+\partial_wJ_3(r)+\partial_wJ_0(\bp,r,\bs)= 0,\\
		\gamma_3\frac{\partial \bs}{\partial t}+\partial_{\bg} J_4(\bs)+\partial_{\bg} J_0(\bp,w,\bs)= 0,\\
		\{\bp(0),\blambda(0),r(0),\bs(0)\}=\{\bp_0,\blambda_0,r_0,\bs_0\},
	\end{cases}\label{eq:ivp}
\end{align}
where $\{\bp_0,\blambda_0,r_0,\bs_0\}$ is a given initial condition, and $\gamma_1,\gamma_2,\gamma_3>0$ are parameters that control the evolution speed of $\bp,\blambda, r,\bs$, respectively. We note that the steady state solution of~\eqref{eq:ivp} satisfies the optimality condition~\eqref{eq:optimality} for Model~\eqref{eq:reform_model}. To avoid confusion, we distinguish the time-dependent variables in~\eqref{eq:ivp} from the optimization variables in~\eqref{eq:reform_model}, and Table~\ref{tab.variable} shows the correspondences.  

\begin{table}[t!]
	\centering
	\begin{tabular}{c|c|c|c|c}
		\hline
		&\multicolumn{4}{c}{Correspondence}\\
		\hline
		General variables& $\bq$ & $\bmu$ & $w$ & $\bg$  \\
		\hline
		PDE solutions& $\bp$ & $\blambda$ & $r$ & $\bs$\\
		\hline
	\end{tabular}
	\caption{Notation of general variables and PDE solutions and their correspondences.}
	\label{tab.variable}
\end{table}

We find the steady state solution of (\ref{eq:ivp}) by employing the operator-splitting framework~\cite{glowinski2016some,glowinski2017splitting}. Our proposed scheme proceeds as follows:\\[2pt]

\noindent\underline{\textit{Initialization}}
\begin{align}
	\text{Initialize}~\{\bp^0,\blambda^0,r^0,\bs^0\}=\{\bp_0,\blambda_0,r_0,\bs_0\}.\nonumber
\end{align}

Fix some $\tau>0$ as the time step. For $k=0,1,2,\dots$,  denote $t^k=k\tau$ for $k=0,1,2\dots$; and
we update $\{\bp^k,\blambda^k,r^k,\bs^k\} \rightarrow \{\bp^{k+1},\blambda^{k+1},r^{k+1},\bs^{k+1}\}$ via four fractional steps:\\[2pt]

\noindent\underline{\textit{Fractional step 1}}: Solve
\begin{align}
	\begin{cases}
		\begin{cases}
			\frac{\partial\bp}{\partial t}+\partial_{\bq}J_1(\bp)\ni 0,\\
			\gamma_1\frac{\partial\blambda}{\partial t}= 0,\\
			\gamma_2\frac{\partial r}{\partial t}= 0,\\
			\gamma_3\frac{\partial \bs}{\partial t} = 0,
		\end{cases} \mbox{ in } \Omega\times (t^k,t^{k+1}),\\
		\{\bp(t^k),\blambda(t^k),r(t^k),\bs(t^k)\}=\{\bp^{k},\blambda^{k},r^{k},\bs^{k}\}  ,
	\end{cases}\nonumber
\end{align}
and set
\begin{align*}
	\{\bp^{k+1/4},\blambda^{k+1/4},r^{k+1/4},\bs^{k+1/4}\}=\{\bp(t^{k+1}),\blambda(t^{k+1}),r(t^{k+1}),\bs(t^{k+1})\}.
\end{align*}

\noindent\underline{\textit{Fractional step 2}}: Solve 
\begin{align}
	\begin{cases}
		\begin{cases}
			\frac{\partial\bp}{\partial t}+\partial_{\bq}J_2(\bp,\blambda)= 0,\\
			\gamma_1\frac{\partial\blambda}{\partial t}+\partial_{\bmu} J_2(\bp,\blambda)\ni 0,\\
			\gamma_2\frac{\partial r}{\partial t}= 0,\\
			\gamma_3\frac{\partial \bs}{\partial t} =0,
		\end{cases} \mbox{ in } \Omega\times (t^k,t^{k+1}),\\
		\{\bp(t^k),\blambda(t^k),r(t^k),\bs(t^k)\}=\{\bp^{k+1/4},\blambda^{k+1/4},r^{k+1/4},\bs^{k+1/4}\} , 
	\end{cases}\nonumber
\end{align}
and set
\begin{align*}
	\{\bp^{k+2/4},\blambda^{k+2/4},r^{k+2/4},\bs^{k+2/4}\}=\{\bp(t^{k+1}),\blambda(t^{k+1}),r(t^{k+1}),\bs(t^{k+1})\}.
\end{align*}
\noindent\underline{\textit{Fractional step 3}}: Solve
\begin{align}
	\begin{cases}
		\begin{cases}
			\frac{\partial\bp}{\partial t}+\partial_{\bq}I_S(\bp,\blambda)\ni 0,\\
			\gamma_1\frac{\partial\blambda}{\partial t}+\partial_{\bmu}I_{S}(\bp,\blambda)\ni 0,\\
			\gamma_2\frac{\partial r}{\partial t}= 0,\\
			\gamma_3\frac{\partial \bs}{\partial t}= 0,
		\end{cases} \mbox{ in } \Omega \times (t^k,t^{k+1}),\\
		\{\bp(t^k),\blambda(t^k),r(t^k),\bs(t^k)\}=\{\bp^{k+2/4},\blambda^{k+2/4},r^{k+2/4},\bs^{k+2/4}\}  ,
	\end{cases}\nonumber
\end{align}
and set
\begin{align*}
	\{\bp^{k+3/4},\blambda^{k+3/4},r^{k+3/4},\bs^{k+3/4}\}=\{\bp(t^{k+1}),\blambda(t^{k+1}),r(t^{k+1}),\bs(t^{k+1})\}.
\end{align*}

\noindent\underline{\textit{Fractional step 4}}: Solve
\begin{align}
	\begin{cases}
		\begin{cases}
			\frac{\partial\bp}{\partial t}+\partial_{\bq}J_0(\bp,r,\bs) + \partial_{\bq}I_{\Sigma_f}(\bp,r,\bs)\ni 0,\\
			\gamma_1\frac{\partial\blambda}{\partial t}= 0,\\
			\gamma_2\frac{\partial r}{\partial t}+\partial_{w}J_0(\bp,r,\bs) + \partial_w J_3(r) + \partial_{w}I_{\Sigma_f}(\bp,r,\bs)\ni 0,\\
			\gamma_3\frac{\partial \bs}{\partial t}+\partial_{\bg}J_0(\bp,r,\bs) + \partial_{\bg} J_4(\bs)+ \partial_{\bg}I_{\Sigma_f}(\bp,r,\bs)\ni 0,
		\end{cases} \mbox{ in } \Omega \times (t^k,t^{k+1}),\\
		\{\bp(t^k),\blambda(t^k),r(t^k),\bs(t^k)\}=\{\bp^{k+3/4},\blambda^{k+3/4},r^{k+3/4},\bs^{k+3/4}\},
	\end{cases}\nonumber
\end{align}
and set
\begin{align*}
	\{\bp^{k+1},\blambda^{k+1},r^{k+1},\bs^{k+1}\}=\{\bp(t^{k+1}),\blambda(t^{k+1}),r(t^{k+1}),\bs(t^{k+1})\}.
\end{align*}

We time discretize the systems above by the Marchuk-Yanenko scheme and get 
\begin{align}\displaystyle
	&\begin{cases}
		\frac{\bp^{k+1/4}-\bp^{k}}{\tau}+ \partial_{\bq}J_1(\bp^{k+1/4})\ni 0,\\
		\gamma_1\frac{\blambda^{k+1/4}-\blambda^{k}}{\tau}= 0,\\
		\gamma_2\frac{r^{k+1/4}-r^{k}}{\tau}=0,\\
		\gamma_3\frac{\bs^{k+1/4}-\bs^{k}}{\tau}=0,
	\end{cases}
	\label{eq.split.time.3}\\
	&\begin{cases}
		\frac{\bp^{k+2/4}-\bp^{k+1/4}}{\tau}+\partial_{\bq}J_2(\bp^{k+2/4},\blambda^{k+1/4})\ni 0,\\
		\gamma_1\frac{\blambda^{k+2/4}-\blambda^{k+1/4}}{\tau}+ \partial_{\bmu}J_2(\bp^{k+2/4},\blambda^{k+2/4})\ni 0,\\
		\gamma_2\frac{r^{k+2/4}-r^{k+1/4}}{\tau}=0,\\
		\gamma_3\frac{\bs^{k+2/4}-\bs^{k+1/4}}{\tau}=0,
	\end{cases}
	\label{eq.split.time.1}\\
	&\begin{cases}
		\frac{\bp^{k+3/4}-\bp^{k+2/4}}{\tau}+\partial_{\bq}I_S(\bp^{k+3/4},\blambda^{k+3/4})\ni 0,\\
		\gamma_1\frac{\blambda^{k+3/4}-\blambda^{k+2/4}}{\tau}+ \partial_{\bmu}I_S(\bp^{k+3/4},\blambda^{k+3/4})\ni 0,\\
		\gamma_2\frac{r^{k+3/4}-r^{k+2/4}}{\tau}=0,\\
		\gamma_3\frac{\bs^{k+3/4}-\bs^{k+2/4}}{\tau}=0,
	\end{cases}
	\label{eq.split.time.2}\\
	&\begin{cases}
		\frac{\bp^{k+1}-\bp^{k+3/4}}{\tau}+\partial_{\bq}J_0(\bp^{k+1},r^{k+1},\bs^{k+1}) + \partial_{\bq}I_{\Sigma_f}(\bp^{k+1},r^{k+1}, \bs^{k+1})\ni 0,\\
		\gamma_1\frac{\blambda^{k+1}-\blambda^{k+3/4}}{\tau}= 0,\\
		\gamma_2\frac{r^{k+1}-r^{k+3/4}}{\tau} +\partial_{w}J_0(\bp^{k+1},r^{k+1},\bs^{k+1})+ \partial_w J_3(r^{k+1})+\partial_{w}I_{\Sigma_f}(\bp^{k+1},r^{k+1}, \bs^{k+1})\ni0,\\
		\gamma_3\frac{\bs^{k+1}-\bs^{k+3/4}}{\tau}+ \partial_{\bg}J_0(\bp^{k+1},r^{k+1},\bs^{k+1})+ \partial_{\bg} J_4(\bs^{k+1})+\partial_{\bg}I_{\Sigma_f}(\bp^{k+1},r^{k+1}, \bs^{k+1})\ni0.
	\end{cases}
	\label{eq.split.time.4}
\end{align}
for $k=0,1,2,\dots$. We note that each system above corresponds to a subproblem which can be solved explicitly or approximated efficiently.  In the following sections, we describe the details.

\begin{remark}
	Scheme (\ref{eq.split.time.3})-(\ref{eq.split.time.4}) is an approximation of the gradient flow. Its convergence closely relates to that of the gradient flow together with an approximation error. It is shown that when there is only one variable and the operators in each step are smooth enough, the approximation error is in the order of $O(\tau)$ (see \cite{chorin1978product} and \cite[Chapter 6]{glowinski2003finite}). In this article, due to the complicated structures of $J_1,J_2, I_{\Sigma_f}$ and $I_{S}$, we cannot use existing tools to analyze the convergence rate and need to study it separately.
\end{remark}

\subsection{On the solution to sub-problem~\eqref{eq.split.time.3} and~\eqref{eq.split.time.1}}\label{sec_sub_l0}
Given $\bp^k$ for $k=0,1,2,\dots$, the vector field $\bp^{k+1/4}$  solves 
\begin{align*}
	\bp^{k+1/4}=\argmin_{\bq\in (\cL^2(\Omega))^2} \left[ \frac{1}{2}\int_{\Omega} |\bq-\bp^{k}|^2\,d\bx +  \tau\alpha_0\mathcal{R}_a(\bq)\right]\;.
\end{align*}
It can be checked that, when $\bp^k(\bx)=\mathbf{0}$,  $\bp^{k+1/4}(\bx)=\mathbf{0}$; and when  $\bp^k(\bx)\neq \mathbf{0}$, then either $\bp^{k+1/4}=\mathbf{0}$, or  $\bp^{k+1/4}(\bx)$ is a non-zero vector satisfying
\begin{align}
	\left(|\bp^{k+1/4}(\bx)|+\frac{\tau\alpha_0a}{(1+a|\bp^{k+1/4}(\bx)|)^2}\right)\frac{\bp^{k+1/4}(\bx)}{|\bp^{k+1/4}(\bx)|}=\bp^k(\bx)\;.
	\label{eq.q.split3.a_temp}
\end{align}
Denote $p^k=|\bp^k(\bx)|$ and define
\begin{equation}
	g^k_a(y)= a^2y^3+a(2-ap^k)y^2+(1-2ap^k)y- p^k+a\tau\alpha_0\;,\label{eq_g_a}
\end{equation}
then~\eqref{eq.q.split3.a_temp} implies that $|\bp^{k+1}(\bx)|$ is a positive root of~\eqref{eq_g_a}. Since the differential of $g^k_a$ has roots $y_1=-a^{-1}$ and $y_2 = (3a)^{-1}(2ap^k-1)$ with $y_1<y_2$, $g^k_a(y)$ is strictly increasing for $y\in(-\infty,y_1)\cup(y_2,+\infty)$ and strictly decreasing for $y\in(y_1,y_2)$.  Now we consider two cases:
\begin{itemize}
	\item If $p^k\leq (2a)^{-1}$, then  since $g_a^k(0)= a \tau\alpha_0-p^k$, $g_a^k$ has a positive root when 
	$a\tau\alpha_0<p^k\leq (2a)^{-1}$
	and does not have any positive root when $p^k\leq a\tau\alpha_0$.
	\item If $p^k>(2a)^{-1}$, $g_a^k$ has at least one positive root if and only if $g_a^k(y_2)< 0$, which requires  that
	$$u^k(p^k):=4a^3(p^k)^3+3a^2(4(p^k)^2-9\tau\alpha_0)+12ap^k+4>0\;,$$
	and this holds if  $p^k > p^*:=3(\tau\alpha_0/4a)^{1/3}-a^{-1}$. If we set $\tau$ such that $2\tau\alpha_0<a^{-2}$, then $(2a)^{-1}>p^*$, and $u^k((2a)^{-1})=27(1/2-a^2\tau\alpha_0)>0$; hence~\eqref{eq_g_a} always has a positive root when $2\tau\alpha_0<a^{-2}$.
\end{itemize}
In summary, if $2\tau\alpha_0<a^{-2}$ and $p^k>(\tau\alpha_0/2)^{1/2}$,~\eqref{eq.q.split3.a_temp} has at least one solution. 
Note that~\eqref{eq.q.split3.a_temp} indicates that $\bp^{k+1/4}(\bx)$ and $\bp^k(\bx)$ have the same direction, and $|\bp^k(\bx)|=|\bp^{k+1/4}(\bx)|+o(a^{-1})$; thus when $a$ is sufficiently large, $\bp^{k+1/4}(\bx)$ can be approximated by $\bp^k(\bx)$. Meanwhile,  since $\tau=O(a^{-2})$, when $p^k\leq(\tau\alpha_0/2)^{1/2}$, we can approximate $\bp^{k+1/4}(\bx)$ by a zero vector.

In a compact form, we deduce the updated formula
\begin{align}
	\bp^{k+1/4}(\bx)=\begin{cases}
		\mathbf{0} & \mbox{ if } |\bp^{k}(\bx)|^2\leq \tau\alpha_0/2,\\
		\bp^{k}(\bx) & \mbox{ otherwise}.
	\end{cases}
	\label{eq.q.split3}
\end{align}
We note that, up to a scaling factor in the threshold,  the solution~\eqref{eq.q.split3} is identical to the formula in~\cite{xu2011image}  derived for the discrete $L^0$-gradient minimization.

The first equation in (\ref{eq.split.time.1}) is the Euler-Lagrange equation of the following minimization problem
\begin{align*}
	\bp^{k+2/4}=\argmin_{\bq\in (\cL^2(\Omega))^2} \left[\frac{1}{2}\int_{\Omega} |\bq-\bp^{k+1/4}|^2d\bx +\tau \alpha_{\rm curv}\int_{\Omega} \left| \nabla \cdot \blambda^{k+1/4}\right|^2|\bq|d\bx\right]\;,
\end{align*}
whose closed-form solution is given as
\begin{align}
	\bp^{k+2/4}=\max \left\{ 0,1-\frac{\tau \alpha_{\rm curv}\left| \nabla \cdot \blambda^{k+1/4}\right|^2}{|\bp^{k+1/4}|}\right\}\bp^{k+1/4}.
	\label{eq.q.split1}
\end{align}

For the second equation of~\eqref{eq.split.time.1}, we note that $\blambda^{k+2/4}$ solves 
\begin{align}
	\blambda^{k+2/4}=\argmin_{\bmu\in (\cH^1(\Omega))^2} \left[ \frac{\gamma_1}{2} \int_{\Omega} |\bmu-\blambda^{k+1/4}|^2d\bx +\tau\alpha_{\rm curv} \int_{\Omega} \left| \nabla \cdot \bmu\right|^2|\bp^{k+2/4}|d\bx \right],
	\label{eq.mu.split1.min}
\end{align}
and the corresponding Euler-Lagrange equation  is
\begin{align}
	\begin{cases}
		\gamma_1 \frac{\blambda^{k+2/4}-\blambda^{k+1/4}}{\tau} -2\alpha_{\rm curv}\nabla(|\bp^{k+2/4}|\nabla \cdot \blambda^{k+2/4})=0 \mbox{ in } \Omega, \\
		|\bp^{k+2/4}|\nabla \cdot \blambda^{k+2/4}= 0 \mbox{ on } \partial \Omega.
	\end{cases}
	\label{eq.mu.split1}
\end{align}
\subsection{On the solution to sub-problem (\ref{eq.split.time.2})}
\label{sec.split2}

We update $\bp^{k+3/4}$ and $\blambda^{k+3/4}$ using the method proposed in \cite[Section 3.5]{deng2019new}. 
Note that $\bp^{k+3/4}$ and $\blambda^{k+3/4}$ solve
\begin{align}
	(\bp^{k+3/4},\blambda^{k+3/4})=\argmin_{(\bq,\bmu)\in S} \left[ \frac{1}{2} \int_{\Omega} |\bq-\bp^{k+2/4}|^2d\bx + \frac{\gamma_1}{2} \int_{\Omega} |\bmu-\blambda^{k+2/4}|^2d\bx \right].
	\label{eq.split2.1}
\end{align}
Problem (\ref{eq.split2.1}) can be solved in a pointwise manner. For each given $\bx$, $(\bp^{k+3/4}(\bx),\blambda^{k+3/4}(\bx))$ satisfies
\begin{align}
	(\bp^{k+3/4}(\bx),\blambda^{k+3/4}(\bx))= \argmin_{(\bq(\bx),\bmu(\bx))\in \chi} \left[|\bq(\bx)-\bq^{k+2/4}(\bx)|^2 +\gamma_1 |\bmu(\bx)-\blambda^{k+2/4}(\bx)|^2\right]
	\label{eq.split2.2}
\end{align}
with $\chi=\{(\ba,\bb)\in \RR^2: \ba\cdot \bb=|\ba|,|\bb|\leq 1\}$. Define 
\begin{align*}
	\chi_1=\{(\ba,\bb)\in \RR^2: |\ba|=0,|\bb|\leq 1\},\;\text{and}\;\chi_2=\{(\ba,\bb)\in \RR^2: |\ba|\neq0, \ba\cdot \bb=|\bq|,|\bb|= 1\}.
\end{align*}
We have $\chi=\chi_1\cup \chi_2$. We next discuss the solution of (\ref{eq.split2.2}) on $\chi_1$ and $\chi_2$ separately.

Over $\chi_1$, denote the solution to (\ref{eq.split2.2}) by $(\widehat{\bp}^{k+3/4}(\bx),\widehat{\blambda}^{k+3/4}(\bx))$. Problem (\ref{eq.split2.2}) reduces to
\begin{align*}
	\widehat{\blambda}(\bx)^{k+3/4}=\argmin_{\bmu(\bx)\in \RR^2:|\bmu(\bx)|\leq 1} |\bmu(\bx)-\blambda^{k+2/4}(\bx)|^2\;,
\end{align*}
and $\widehat{\bp}^{k+3/4}(\bx)=\mathbf{0}$.
The closed-form solution for $\widehat{\blambda}^{k+3/4}(\bx)$ is given as
\begin{align}
	\widehat{\blambda}^{k+3/4}(\bx)=\frac{\blambda^{k+2/4}(\bx)}{\max\{1,|\blambda^{k+2/4}(\bx)|\}}.
	\label{eq.split2.mu.1}
\end{align}

Over $\chi_2$, denote the solution to (\ref{eq.split2.2}) by $(\widetilde{\bp}^{k+3/4}(\bx),\widetilde{\blambda}^{k+3/4}(\bx))$, and it can be written as
\begin{align}
	&(\widetilde{\bp}^{k+3/4}(\bx),\widetilde{\blambda}^{k+3/4}(\bx))=  \nonumber\\
	&\hspace{2cm} \argmin_{\substack{(\bq(\bx),\bmu(\bx))\in \RR^2\times \RR^2, \\ \bq(\bx)\neq \mathbf{0},\bq(\bx)\cdot\bmu(\bx)=|\bq(\bx)|, \\ |\bmu(\bx)|=1}} \left[ |\bq(\bx)-\bp^{k+2/4}(\bx)|^2 +\gamma_1 |\bmu(\bx)-\blambda^{k+2/4}(\bx)|^2\right].
	\label{eq.split2.3}
\end{align}
Denote $\theta=|\bq(\bx)|$,  $ \by(\bx)=\bp^{k+2/4}(\bx)$, and $\bw(\bx)=\blambda^{k+2/4}(\bx)$ for each $\bx\in\Omega$. Problem~\eqref{eq.split2.3} is equivalent to solving
\begin{align}
	\min_{(\theta,\bmu(\bx))\in \RR\times \RR^2, \theta>0, |\bmu(\bx)|=1} \left[ \theta^2 -2\theta \bmu(\bx)\cdot \by(\bx) -2\gamma_1 \bmu(\bx)\cdot \bw(\bx) \right]\;.
	\label{eq.split2.4}
\end{align}
For a fixed $\theta$, the solution $\bmu^*(\bx,\theta)$ of (\ref{eq.split2.4}) is given by
\begin{align}
	\bmu^*(\bx,\theta)=\frac{\theta \by(\bx)+\gamma_1\bw(\bx)}{|\theta\by(\bx)+\gamma_1\bw(\bx)|}\;.
	\label{eq.split2.5}
\end{align}
Substituting (\ref{eq.split2.5}) to (\ref{eq.split2.4}) leads to
\begin{align}
	\min_{\theta>0} \left[ \theta^2-2|\theta\by(\bx)+\gamma_1\bw(\bx)| \right].
	\label{eq.split2.6}
\end{align}

We solve (\ref{eq.split2.6}) by a fixed point method. Denote 
\begin{align*}
	E(\bx,\theta)=\theta^2-2|\theta\by(\bx)+\gamma_1\bw(\bx)|\;,
\end{align*}
and thus
\begin{align}
	\frac{d E}{d\theta}(\bx,\theta)=2\theta-\frac{2\by(\bx)\cdot(\theta\by(\bx)+\gamma_1 \bw(\bx))}{|\theta\by(\bx)+\gamma_1 \bw(\bx)|}.
	\label{eq.split2.theta.diff}
\end{align}
We initialize $\theta^0=|\by(\bx)|$. For $m=0,1,2,\dots$, we update $\theta^{m+1}$ from $\theta^m$ via
\begin{align}
	\theta^{m+1}=\max \left\{0, \frac{\by(\bx)\cdot(\theta^m\by(\bx)+\gamma_1 \bw(\bx))}{|\theta^m\by(\bx)+\gamma_1 \bw(\bx)|} \right\}
	\label{eq.split2.theta}
\end{align}
until $|\theta^{m+1}-\theta^m|<\varepsilon$ for some small $\varepsilon>0$. 

Denote the converged $\theta$ by $\theta^*(\bx)$, then we compute 
\begin{align}
	\widetilde{\blambda}^{k+3/4}(\bx)=\frac{\theta^*(\bx)\bp^{k+2/4}(\bx)+\gamma_1\blambda^{k+2/4}(\bx)}{|\theta^*(\bx)\bp^{k+2/4}(\bx)+\gamma_1\blambda^{k+2/4}(\bx)|},\quad \widetilde{\bp}^{k+3/4}(\bx)=\theta^* \widetilde{\blambda}^{k+3/4}(\bx)\;.
	\label{eq.split2.tilde}
\end{align}

Finally, we define
\begin{align*}
	G(\bq,\bmu;\bx)=|\bq(\bx)-\bq^{k+2/4}(\bx)|^2 +\gamma_1 |\bmu(\bx)-\blambda^{k+2/4}(\bx)|^2,
\end{align*}
and compare the minimizers in $\chi_1$ and $\chi_2$ to obtain $(\bp^{k+3/4}(\bx), \blambda^{k+3/4}(\bx))$ as
\begin{align}
	(\bp^{k+3/4}(\bx), \blambda^{k+3/4}(\bx)) =\begin{cases}
		(\widehat{\bp}^{k+3/4}(\bx),\widehat{\blambda}^{k+3/4}(\bx)) &\mbox{ if } G(\widehat{\bp}^{k+3/4},\widehat{\blambda}^{k+3/4};\bx)\\ &\hspace{1cm}\leq G(\widetilde{\bp}^{k+3/4},\widetilde{\blambda}^{k+3/4};\bx),\\
		(\widetilde{\bp}^{k+3/4}(\bx),\widetilde{\blambda}^{k+3/4}(\bx)) &\mbox{ otherwise}.
	\end{cases}
	\label{eq.split2.7}
\end{align}

\begin{remark}
	The operator in (\ref{eq.split2.theta.diff}) is nonlinear, nonsmooth, and may be discontinuous (depending on the angle between $\by(\bx)$ and $\bw(\bx)$). Proving the convergence of the fixed-point iteration (\ref{eq.split2.theta}) requires a careful analysis of properties of $\by$ and $\bw$, and an independent study. 
	Intuitively, with a small time step $\tau$, after each iteration, we expect $(\bp^{k+2/4},\blambda^{k+2/4})$ stays close to $(\bp^{(k-1)+3/4},\blambda^{(k-1)+3/4})$ and does not deviate too far from the set $S$. Iteration (\ref{eq.split2.theta}) is expected to converge. 
	Numerically, (\ref{eq.split2.theta}) converges in our experiments.
\end{remark}

\subsection{On the solution to sub-problem (\ref{eq.split.time.4})}\label{sec_sub_end}
In (\ref{eq.split.time.4}), $(\bp^{k+1},r^{k+1},\bs^{k+1})$ solves
\begin{align}
	\min_{\substack{(\bq,w,\bg)\in \Sigma_f, w\in \cH^2(\Omega),\\ \bg\in (\cH^1(\Omega))^2}} & \bigg[\frac{1}{2} \int_{\Omega} |\bq-\bp^{k+3/4}|^2 d\bx + \frac{\gamma_2}{2} \int_{\Omega} |w-r^{k+3/4}|^2 d\bx + \frac{\gamma_3}{2} \int_{\Omega} |\bg-\bs^{k+3/4}|^2 d\bx \nonumber\\
	&+ \tau\alpha_{w}\int_{\Omega} |\Delta w|^2d\bx + \tau \alpha_{n}\int_{\Omega}|\bg|^2 d\bx +\frac{\tau}{2} \int_{\Omega} |f-(v_{\bq,w,\bg}+w+\nabla \cdot \bg)|^2d\bx \bigg],
	\label{eq.step4}
\end{align}
or equivalently
\begin{align}
	\begin{cases}
		(v^{k+1},r^{k+1},\bs^{k+1})= \displaystyle\argmin_{\substack{v\in \cH^1(\Omega), w\in \cH^2(\Omega),\\ \bg\in (\cH^1(\Omega))^2}} \bigg[\frac{1}{2} \int_{\Omega} |\nabla v-\bp^{k+3/4}|^2 d\bx + \frac{\gamma_2}{2} \int_{\Omega} |w-r^{k+3/4}|^2 d\bx \\
		\hspace{4cm}+ \frac{\gamma_3}{2} \displaystyle\int_{\Omega} |\bg-\bs^{k+3/4}|^2 d\bx
		+\tau \alpha_{w}\int_{\Omega} |\Delta w|^2d\bx +  \tau\alpha_{n}\int_{\Omega}|\bg|^2 d\bx\\
		\hspace{4cm} +\frac{\tau}{2} \displaystyle\int_{\Omega} |f-(v+w+\nabla \cdot \bg)|^2d\bx\bigg],\\
		\bp^{k+1}=\nabla v^{k+1}.
	\end{cases}
	\label{eq.step4.v}
\end{align}
The optimality condition for $(v^{k+1},r^{k+1},\bs^{k+1})$ is
\begin{align}
	\begin{cases}
		\begin{cases}
			\frac{-\Delta v^{k+1}+\nabla \cdot\bp^{k+3/4}}{\tau} + v^{k+1}-(f-r^{k+1}-\nabla\cdot \bs^{k+1})=0 & \mbox{ in } \Omega,\\
			(\nabla v^{k+1}-\bp^{k+3/4})\cdot \bn=0 &\mbox{ on } \partial\Omega,
		\end{cases}\\
		\begin{cases}
			\gamma_2\frac{r^{k+1}-r^{k+3/4}}{\tau}+r^{k+1}-(f-v^{k+1}-\nabla\cdot \bs^{k+1}) + 2\alpha_{w}(\nabla^4 r^{k+1})=0 & \mbox{ in } \Omega,\\
			\Delta r^{k+1}=0,\ \nabla^3 r^{k+1}\cdot \bn=0 & \mbox{ on } \partial \Omega,
		\end{cases}\\  
		\begin{cases}
			\gamma_3\frac{\bs^{k+1}-\bs^{k+3/4}}{\tau}-\nabla(\nabla\cdot\bs^{k+1}) +2 \alpha_n \bs^{k+1}+\nabla(f-v^{k+1}-r^{k+1})=0 & \mbox{ in } \Omega,\\
			f-(v^{k+1}+r^{k+1}+\nabla \cdot \bs^{k+1})=0 &\mbox{ on } \partial \Omega.
		\end{cases}
	\end{cases}
	\label{eq.step4.opti}
\end{align}
The proposed operator splitting scheme is summarized in Algorithm~\ref{alg_DOSH}.

\begin{algorithm}[t!]
	\begin{algorithmic}[1]
		\REQUIRE{Gray image $f$; model parameters: $\alpha_0,\alpha_\text{curv},\alpha_w, \alpha_n$; algorithmic parameters: time step $\tau$, evolution speed factor $\gamma_i$, $i=1,2,3$, threshold $\rho$ for termination, maximal number of iteration $\text{Iter}_{\max}$
		}
		
		\FOR{$k=1,\dots,\text{Iter}_{\max}$}
		\STATE Update $\bp^{k}\to\bp^{k+1/4}$ by~\eqref{eq.q.split3}, and set $\blambda^{k+1/4}=\blambda^{k}$, $r^{k+1/4}=r^{k}$,  $\bs^{k+1/4}=\bs^{k}$.
		\STATE Update $\bp^{n+1/4}\to\bp^{k+2/4}$ by~\eqref{eq.q.split1}, $\blambda^{k+1/4}\to\blambda^{k+2/4}$ by solving \eqref{eq.mu.split1}, and set $r^{k+2/4}=r^{k+1/4}$, $\bs^{k+2/4}=\bs^{k+1/4}$.
		\STATE Update $\bp^{k+2/4}\to\bp^{k+3/4}$ and  $\blambda^{k+2/4}\to\blambda^{k+3/4}$ by~\eqref{eq.split2.7}, and set $r^{k+3/4}=r^{k+2/4}$, $\bs^{k+3/4}=\bs^{k+2/4}$.
		\STATE Compute $v^{k+1}$, $r^{k+1}$, and $\bs^{k+1}$ by solving \eqref{eq.step4.opti}, and set $\bp^{k+1}=\nabla v^{k+1}$, $n^{k+1}=\nabla\cdot\bs^{k+1}$, $\blambda^{k+1}=\blambda^{k+3/4}$.
		\IF{$\max\{\frac{\|r^{k+1}-r^k\|_2}{\|r^k\|_2}, \frac{\|v^{k+1}-v^{n}\|_2}{\|v^k\|_2}\} < \rho$}
		\STATE break.
		\ENDIF
		\ENDFOR
		
		\RETURN the converged homogeneous structure  $v^*$, smooth part $w^*$, and oscillatory part $n^*$, such that $f$ is approximated by $v^*+w^*+n^*$. 
	\end{algorithmic}
	\caption{Proposed operator-splitting algorithm}\label{alg_DOSH}
\end{algorithm}

\subsection{Adapt to periodic boundary conditions for efficiency}\label{sec_period}
We note that while~\eqref{eq.mu.split1} and~\eqref{eq.step4.opti} contain high-order derivatives, they are linear. Therefore, we can leverage the Fast Fourier Transform (FFT) to convert these PDE systems to pointwise algebraic systems, for which we can deduce explicit solutions.  To garnish such computational efficiency,  we  extend our model~\eqref{eq:proposed_model}
to periodic data, and this requires equipping  several equations with the periodic boundary condition. Specifically, in (\ref{eq:reform_model}), (\ref{eq.vq}), (\ref{eq.mu.split1.min}), (\ref{eq.mu.split1}), (\ref{eq.step4}), (\ref{eq.step4.v}) and (\ref{eq.step4.opti}), we equip functions spaces with the periodic boundary condition or replace the boundary conditions by the periodic one. We defer the details to Appendix \ref{append_periodic}.

In the rest of this paper, we always assume these adaptations and consider the periodic boundary condition.

\section{Numerical Discretization}\label{eq_num_discret}
In this section, we provide implementation details for solving the proposed minimization problem.
\subsection{Synopsis}
Let $\Omega$ be a rectangular domain with $M\times N$ pixels. Denote the two spatial directions by $x_1,x_2$ with step size $\Delta x_1=\Delta x_2=h$ for some $h>0$. For an image $f$ defined on $\Omega$, we denote $f(i,j)=f(ih,jh)$. Assume that $f$ satisfies the periodic boundary condition. We define the backward ($-$) and forward ($+$) approximation for $\partial f/\partial x_1$ and $\partial f/\partial x_2$ as
\begin{align*}
	&\partial_1^- f(i,j)=\begin{cases}
		\frac{f(i,j)-f(i-1,j)}{h} &\ 1<i\leq M,\\
		\frac{f(1,j)-f(M,j)}{h} &\ i=1,
	\end{cases}
	& \partial_1^+ f(i,j)=\begin{cases}
		\frac{f(i+1,j)-f(i,j)}{h} &\ 1\leq i< M,\\
		\frac{f(1,j)-f(M,j)}{h} &\ i=M,
	\end{cases}\\
	&\partial_2^- f(i,j)=\begin{cases}
		\frac{f(i,j)-f(i,j-1)}{h} &\ 1<i\leq N,\\
		\frac{f(i,1)-f(i,N)}{h} &\ i=1,
	\end{cases}
	& \partial_2^+ f(i,j)=\begin{cases}
		\frac{f(i,j+1)-f(i,j)}{h} &\ 1\leq i< N,\\
		\frac{f(i,1)-f(i,N)}{h} &\ i=N.
	\end{cases}
\end{align*}
For any scalar valued function $f$ and vector valued function $\bp=(p_1,p_2)$, the backward ($-$) and forward ($+$) approximation for gradient and divergence are defined as
\begin{align*}
	\nabla^{\pm}f(i,j)=(\partial^{\pm}_1 f(i,j), \partial^{\pm}_2 f(i,j)), \quad \diver^{\pm} \bp(i,j)=\partial_1^{\pm}p_1(i,j) + \partial^{\pm}_2 p_2(i,j).
\end{align*}
The discretized Laplacian is defined as
\begin{align*}
	\Delta f(i,j)=\diver^-(\nabla^+ f(i,j))=\partial_1^-\partial_1^+ f(i,j) + \partial_2^-\partial_2^+ f(i,j)\;,
\end{align*}
which recovers the central difference scheme.

Define the shifting and identity operator as
\begin{align*}
	(\cS_1^{\pm}f)(i,j)=f(i\pm 1,j),\quad (\cS_2^{\pm}f)(i,j)=f(i,j\pm 1),\quad (\cI f)(i,j)=f(i,j)\;,
\end{align*}
where the periodic boundary condition is assumed.  Denote the discrete Fourier transform and its inverse by $\cF$ and $\cF^{-1}$, respectively. We have
\begin{align*}
	\cF (\cS_1^{\pm}f)(i,j)=e^{\pm 2\pi\sqrt{-1}(i-1)/M}\cF(f)(i,j)\;,\text{and}~\cF (\cS_2^{\pm}f)(i,j)=e^{\pm 2\pi\sqrt{-1}(j-1)/N}\cF(f)(i,j)\;.
\end{align*}
We use $\Real(\cdot)$ to denote the real part of a complex quantity.
\subsection{Discrete analogue of $\bp^{k+1/4}$, $\bp^{k+2/4}$ and $\blambda^{k+2/4}$}
For $\bp^{k+1/4}$, we discretize (\ref{eq.q.split3}) as
\begin{align}
	\bp^{k+1/4}(i,j)=\begin{cases}
		\mathbf{0} & \mbox{ if } (p_1^{k}(i,j))^2+(p_2^{k}(i,j))^2\leq \alpha_0\tau/2,\\
		\bp^{k}(i,j) & \mbox{ otherwise},
	\end{cases}
	\label{eq.q.split3.dis}
\end{align}
and the computational cost is $\mathcal{O}(MN)$.

According to (\ref{eq.q.split1}), $\bp^{k+2/4}$ is updated pointwisely. We thus have
\begin{align}
	\bp^{k+2/4}(i,j)=\max \left\{ 0,1-\frac{\tau\alpha_{w}\left| \diver^- \blambda^{k+1/4}(i,j)\right|^2}{\sqrt{(p_1^{k+1/4}(i,j))^2+(p_2^{k+2/4}(i,j))^2}}\right\}\bp^{k+1/4}(i,j)
	\;,
	\label{eq.q.split1.dis}
\end{align}
whose computational cost is $\mathcal{O}(MN)$.

To compute $\blambda^{k+2/4}=(\lambda_1^{k+2/4},\lambda_2^{k+2/4})$, the updating formula (\ref{eq.mu.split1})  can be rewritten as
\begin{align}
	\gamma_1 \blambda^{k+2/4} -2\tau\alpha_{w}\nabla(|\bp^{k+2/4}|\nabla \cdot \blambda^{k+2/4})=\gamma_1\blambda^{k+1/4}\;.
	\label{eq.mu.split1.dis.1}
\end{align}

Instead of directly solving (\ref{eq.mu.split1.dis.1}), we use the frozen coefficient method \cite{tai2011fast,he2020curvature,liu2021color} and consider the following approximation
\begin{align}
	\gamma_1 \blambda^{k+2/4} -c\nabla(\nabla\cdot \blambda^{k+2/4}) =\gamma_1\blambda^{k+1/4} -c\nabla(\nabla\cdot \blambda^{k+1/4})+ 2\tau\alpha_{w}\nabla(|\bp^{k+2/4}|\nabla \cdot \blambda^{k+1/4})
	\label{eq.mu.split1.dis.2}
\end{align}
for some small coefficient $c>0$. In our experiment, we set $c=1\times10^{-9}$. We discretize (\ref{eq.mu.split1.dis.2}) as
\begin{align}
	&\gamma_1 \blambda^{k+2/4} -c\nabla^+(\diver^- \blambda^{k+2/4}) =\nonumber\\
	&\hspace{3cm} \gamma_1\blambda^{k+1/4} -c\nabla^+(\diver^- \blambda^{k+1/4})+ 2\tau\alpha_{w}\nabla^+(|\bp^{k+2/4}|(\diver^- \blambda^{k+1/4})).
	\label{eq.mu.split1.dis.3}
\end{align}
Equation (\ref{eq.mu.split1.dis.3}) can be written in a matrix form as
\begin{align}
	\begin{pmatrix}
		\gamma_1-c\partial_1^+\partial_1^- & -c\partial_1^+\partial_2^-\\
		-c\partial_2^+\partial_1^- & \gamma_1-c\partial_2^+\partial_2^-
	\end{pmatrix}
	\begin{pmatrix}
		\lambda^{k+2/4}_1 \\ \lambda^{k+2/4}_2
	\end{pmatrix}=
	\begin{pmatrix}
		b_1\\ b_2
	\end{pmatrix}
	\label{eq.mu.split1.dis.4}
\end{align}
with
\begin{align*}
	\begin{cases}
		&b_1=\gamma_1\lambda_1^{k+1/4} -c\partial_1^+(\diver^- \blambda^{k+1/4})+ 2\tau\alpha_{w}\partial_1^+(|\bp^{k+2/4}|(\diver^- \blambda^{k+1/4})),\\
		&b_2=\gamma_1\lambda_2^{k+1/4} -c\partial_2^+(\diver^- \blambda^{k+1/4})+ 2\tau\alpha_{w}\partial_2^+(|\bp^{k+2/4}|(\diver^- \blambda^{k+1/4})).
	\end{cases}
\end{align*}
System (\ref{eq.mu.split1.dis.4}) is equivalent to 
\begin{align}
	\begin{pmatrix}
		\gamma_1-c(\cS_1^+-\cI)(\cI-\cS_1^-)/h^2  & -c(\cS_1^+-\cI)(\cI-\cS_2^-)/h^2\\
		-c(\cS_2^+-\cI)(\cI-\cS_1^-)/h^2 & \gamma_1-c(\cS_2^+-\cI)(\cI-\cS_2^-)/h^2
	\end{pmatrix}
	\begin{pmatrix}
		\lambda^{k+2/4}_1 \\ \lambda^{k+2/4}_2
	\end{pmatrix}=
	\begin{pmatrix}
		b_1\\ b_2
	\end{pmatrix}.
	\label{eq.mu.split1.dis.5}
\end{align}
Applying discrete Fourier transform on both sides of (\ref{eq.mu.split1.dis.5}) gives
\begin{align}
	\begin{pmatrix}
		a_{11} & a_{12}\\
		a_{21} & a_{22}
	\end{pmatrix}
	\cF\begin{pmatrix}
		\mu^{k+2/4}_1 \\ \mu^{k+2/4}_2
	\end{pmatrix}=\cF
	\begin{pmatrix}
		b_1\\ b_2
	\end{pmatrix},
\end{align}
where
\begin{align*}
	&a_{11}(i,j)=\gamma_1-2c(\cos \zeta_i-1)/h^2,\ a_{22}(i,j)=\gamma_1-2c(\cos \eta_j-1)/h^2,\\
	&a_{12}(i,j)=c(\cos \zeta_i+\sqrt{-1}\sin \zeta_i-1)(\cos \eta_j-\sqrt{-1}\sin \eta_j-1)/h^2,\\
	&a_{21}(i,j)=c(\cos \eta_j+\sqrt{-1}\sin \eta_j-1)(\cos \zeta_i-\sqrt{-1}\sin \zeta_i-1)/h^2,
\end{align*}
with $\zeta_i=2\pi(i-1)/M,\ \eta_j=2\pi(j-1)/N$ for $i=1,...,M$ and $j=1,...,N$. We  thus update $\blambda^{k+2/4}$ via
\begin{align}
	\begin{pmatrix}
		\lambda_{1}^{k+2/4}\\ \lambda_{2}^{k+2/4}
	\end{pmatrix}=\Real\left(\cF^{-1}\left[\frac{1}{a_{11}a_{22}-a_{12}a_{21}}
	\begin{pmatrix}
		a_{11}\cF(b_1)-a_{12}\cF(b_2)\\
		-a_{21}\cF(b_1)+a_{22}\cF(b_2)
	\end{pmatrix} \right]\right)\;.\label{eq_lambda_k+1/4}
\end{align}
Since the linear system is solved explicitly under the forward and inverse FFT, the computational cost is $\mathcal{O}(MN(\log M+ \log N))$.

\subsection{Discrete analogue of $\bp^{k+3/4}$, $\blambda^{k+3/4}$}
According to Section \ref{sec.split2}, $\bp^{k+3/4}$ and $\blambda^{k+3/4}$ can be computed pointwisely. On $\chi_1$, for each $i,j$, we set $\widehat{\bp}^{k+3/4}(i,j)=\mathbf{0}$ and discretize (\ref{eq.split2.mu.1}) as
\begin{align}
	\widehat{\blambda}^{k+3/4}(i,j)=\frac{\blambda^{k+2/4}(i,j)}{\max\left\{1,\sqrt{|\blambda^{k+2/4}(i,j)|^2}\right\}}.
	\label{eq.split2.mu.1.dis}
\end{align}

On $\chi_2$, we discretize (\ref{eq.split2.tilde}) as
\begin{align*}
	\widetilde{\blambda}^{k+3/4}(i,j)=\frac{\theta^*\bp^{k+2/4}(i,j)+\gamma_1\blambda^{k+2/4}(i,j)}{|\theta^*\bp^{k+2/4}(i,j)+\gamma_1\blambda^{k+2/4}(i,j)|},\quad \widetilde{\bp}^{k+3/4}(i,j)=\theta^* \widetilde{\blambda}^{k+3/4}(i,j)\;,
\end{align*}
where $\theta^*$ is computed by (\ref{eq.split2.theta}) with $\by(\bx)=\bp^{k+2/4}(i,j), \bw(\bx)=\blambda^{k+2/4}(i,j)$. Functions $\bp^{k+3/4}$ and $\blambda^{k+3/4}$ are then updated using (\ref{eq.split2.7}).

\subsection{Discrete analogue of $\bp^{k+1},r^{k+1}$ and $\bs^{k+1}$}
\label{sec.step4.dis}

Reorganizing the system (\ref{eq.step4.opti}), we get a linear system in $v^{k+1},r^{k+1}$ and $\bs^{k+1}=(s_1^{k+1},s_2^{k+1})$
\begin{align}
	\begin{cases}
		(-\Delta +\tau I) v^{k+1} + \tau r^{k+1} + \tau \nabla\cdot \bs^{k+1}=-\nabla \cdot \bp^{k+3/4} +\tau f,\\
		\tau v^{k+1} + [(\gamma_2+\tau)I+2\tau \alpha_{w} \nabla^4]r^{k+1} +\tau \nabla\cdot \bs^{k+1}=\gamma_2 r^{k+3/4}+\tau f,\\
		-\tau \nabla v^{k+1}-\tau\nabla r^{k+1}+[(\gamma_3+2\tau \alpha_n) I-\tau \nabla\diver]\bs^{k+1}=\gamma_3\bs^{k+3/4}-\tau\nabla f.
	\end{cases}
	\label{eq.split4}
\end{align}
Upon discretization, we obtain a linear system
\begin{align*}
	\mathbf{A}\begin{bmatrix}
		v^{k+1}\\ r^{k+1} \\ s_1^{k+1} \\ s_2^{k+1}
	\end{bmatrix}=\begin{bmatrix}
		b^k_1\\b^k_2\\b^k_3\\b^k_4
	\end{bmatrix}~\text{with}~\begin{cases}
		&b^k_1=-\diver^-\bp^{k+3/4}+\tau f,\\
		&b^k_2=\gamma_2r^{k+3/4}+\tau f,\\
		&b^k_3=\gamma_3 s_1^{k+3/4}-\tau \partial_1^+ f,\\
		& b^k_4=\gamma_3 s_2^{k+3/4} -\tau \partial_2^+ f,
	\end{cases}
\end{align*}
where
\begin{align*}
	&\mathbf{A}=\nonumber\\
	&\begin{bmatrix}
		\begin{aligned}
			&-\frac{1}{h^2}\cdot\\
			&[(\cI-\cS_1^-)(\cS_1^+-\cI)\\&+ (\cI-\cS_2^-)(\cS_2^+-\cI)]\\
			&+\tau \cI
		\end{aligned}
		& \tau & \tau(\cI-\cS_1^-)/h & \tau (\cI-\cS_2^-)/h\\
		\tau & 
		\begin{aligned}
			&(\gamma_2+\tau)\cI\\&+2\tau\alpha_{w} \cP
		\end{aligned} & \tau(\cI-\cS_1^-)/h & \tau (\cI-\cS_2^-)/h\\
		-\tau(\cS_1^+-\cI)/h & -\tau (\cS_1^+-\cI)/h & \begin{aligned}
			&\gamma_3+2\tau\alpha_n-\\
			&\frac{\tau}{h^2}(\cS_1^+-\cI)(\cI-\cS_1^-)
		\end{aligned}
		& -\frac{\tau}{h^2}(\cS_1^+-\cI)(\cI-\cS_2^-)\\
		-\tau(\cS_2^+-\cI)/h & -\tau (\cS_2^+-\cI)/h 
		& -\frac{\tau}{h^2}(\cS_2^+-\cI)(\cI-\cS_1^-)
		& \begin{aligned}
			&\gamma_3+2\tau\alpha_n-\\
			&\frac{\tau}{h^2}(\cS_2^+-\cI)(\cI-\cS_2^-)
		\end{aligned}
	\end{bmatrix}
\end{align*}
with 
\begin{align*}
	\cP=&\frac{1}{h^4}\Big[((\cI-\cS_1^-)(\cS_1^+-\cI))^2+(\cI-\cS_2^-)(\cS_2^+-\cI)(\cI-\cS_1^-)(\cS_1^+-\cI) \nonumber\\
	&\quad + (\cI-\cS_1^-)(\cS_1^+-\cI)(\cI-\cS_2^-)(\cS_2^+-\cI) + ((\cI-\cS_2^-)(\cS_2^+-\cI))^2\Big].
\end{align*}
Note that $\mathcal{A}$ does not depend on the iteration, thus it can be constructed beforehand and applied repeatedly in each iteration. Applying the discrete Fourier transform $\mathcal{F}$, we get
\begin{align}
	\mathbf{D}\ \cF\left(\begin{bmatrix}
		v^{k+1}\\ r^{k+1} \\ s_1^{k+1} \\ s_2^{k+1}
	\end{bmatrix}\right) =\cF\left(\begin{bmatrix}
		b^k_1\\b^k_2\\b^k_3\\b^k_4
	\end{bmatrix}\right)
	\label{eq.split4.fftM}
\end{align}
with $\mathbf{D}\in\mathbb{R}^{4\times 4}$ whose $(l,m)$-th entries $d_{lm}$, $1\leq l,m\leq 4$  are listed below
\begin{align*}
	&d_{11}(i,j)=(\tau-2(\cos \zeta_i-1)/h^2-2(\cos \eta_j-1)/h^2),\quad d_{12}=\tau,\\
	&d_{13}(i,j)=\tau[1-(\cos \zeta_i-\sqrt{-1}\sin \zeta_i)]/h, \quad d_{14}(i,j)=\tau[1-(\cos \eta_j-\sqrt{-1}\sin \eta_j)]/h,\\
	&d_{21}=\tau, \\
	& d_{22}(i,j)=\gamma_2+\tau+2\tau\alpha_{w} \frac{4}{h^4}\left[(\cos \zeta_i-1)^2+2(\cos \zeta_i-1)(\cos \eta_j-1) +(\cos \eta_j-1)^2\right],\\
	&d_{23}=d_{13}, \quad d_{24}=d_{14},\\
	&d_{31}(i,j)=d_{32}(i,j)=-\tau (\cos \zeta_i+\sqrt{-1}\sin \zeta_i-1)/h,\\
	&d_{33}(i,j)=\gamma_3+2\tau \alpha_n-2\tau(\cos \zeta_i-1)/h^2, \\
	&d_{34}(i,j)=\tau(\cos \zeta_i+\sqrt{-1}\sin \zeta_i-1)(\cos \eta_j-\sqrt{-1}\sin \eta_j-1)/h^2,\\
	&d_{41}(i,j)= d_{42}(i,j)=-\tau (\cos \eta_j+\sqrt{-1}\sin \eta_j-1)/h,\\
	& d_{43}(i,j)=\tau(\cos \eta_j+\sqrt{-1}\sin \eta_j-1)(\cos \zeta_i-\sqrt{-1}\sin \zeta_i-1)/h^2, \\
	&d_{44}(i,j)=\gamma_3+2\tau\alpha_n-2\tau(\cos \eta_j-1)/h^2.
\end{align*}
To stabilize the computation, we add a small constant $\kappa=1\times10^{-9}$ on the diagonal of $\mathbf{D}$. The inverse $\mathbf{D}^{-1}$ can be explicitly computed using the Cramer's rule. 
We can efficiently get $v^{k+1},r^{k+1}$ and $\bs^{k+1}$ by inverse FFT
\begin{align}
	\begin{bmatrix}
		v^{k+1}\\ r^{k+1} \\ s_1^{k+1} \\ s_2^{k+1}
	\end{bmatrix} =\Real\left(\cF^{-1}\left[\mathbf{D}^{-1}\cF\left(\begin{bmatrix}
		b^k_1\\b^k_2\\b^k_3\\b^k_4
	\end{bmatrix}\right)\right]\right)\;.
	\label{eq.split4.fft.sys}
\end{align}
Similarly to~\eqref{eq_lambda_k+1/4}, as we obtain the updated variables by explicit formula, the computational cost is still $\mathcal{O}(MN(\log M+\log N))$. Overall, the computational cost of a single iteration of our proposed algorithm that updates $(\bp^{k},\blambda^k,r^k,\bs^k)$ to $(\bp^{k+1},\blambda^{k+1},r^{k+1},\bs^{k+1})$ is in the order of $\mathcal{O}(MN(\log M+\log N))$.

\section{Numerical Experiments}\label{sec_experiment}

In this section, we present various experiments to show the behaviors of the proposed model, analyze the effects of model parameters, and justify its effectiveness in the task of decomposition and denoising by ablation and comparison studies.   To evaluate the image quality, we employ the PSNR metric
\begin{align*}
	\text{PSNR}(u_{\text{ref}},\widehat{u}) = 10\log_{10}\left(\frac{1}{\|\widehat{u}-u_{\text{ref}}\|_2}\right)\;,
\end{align*}
where $u_{\text{ref}}$ is the reference image, $\widehat{u}$ is the corresponding approximation, and their intensity values are scaled to $[0,1]$. We test our method on various synthetic images as well as grayscale photos. For model parameters $\alpha_0,\alpha_\text{curv}, \alpha_w$ and $\alpha_n$, we discuss their selections in Section~\ref{sec_parameters} and fine-tune them according to the image contents. For algorithmic parameters, we fix $\tau=0.1,\gamma_1=1.0,\gamma_2=0.01$, and $\gamma_3=20$. We set the global constants for numerical stability $c=1\times10^{-9}$ and  $\kappa=1\times10^{-9}$; here $c$ is the frozen coefficient in (\ref{eq.mu.split1.dis.2}), and $\kappa$ is added to the diagonal of $\mathbf{D}$ in~\eqref{eq.split4.fftM} for stability.  We take the maximal number of iterations $\text{Iter}_{\max}=1000$, and fix the threshold for the terminating condition $\rho=1\times10^{-6}$, i.e., the algorithm is terminated if $\max\left\{\frac{\|r^{k+1}-r^k\|_2}{\|r^k\|_2}, \frac{\|v^{k+1}-v^{n}\|_2}{\|v^k\|_2}\right\} < \rho$. For the default initial conditions, we fix $v_0=0.001\times\widetilde{f}+0.999\times0.5$ where $\widetilde{f}$ is the input $f$ convolved with a standard Gaussian kernel, $\bp_0=\nabla^+ v_0$, $r_0=f-v_0$, and $\blambda_0=\bs_0=\mathbf{0}$.  To reduce the boundary artifacts, we pad the input with 30 pixels on each side in a symmetric manner.  The algorithm is implemented with MATLAB 2022a in the environment of Windows 11, and all the experiments are conducted in a laptop equipped with 12th Gen Intel(R) Core(TM) i7-12800H, 2400Mhz, and 14 cores.

\subsection{General performances}
\begin{figure}[t!]
	\centering
	\begin{tabular}{c@{\hspace{2pt}}c@{\hspace{2pt}}c@{\hspace{2pt}}c@{\hspace{2pt}}c@{\hspace{2pt}}c}
		$f_{\text{clean}}$&$f_{\text{noise}}$&$u^*$&$v^*_{\text{scaled}}$&$w^*_{\text{scaled}}$&$n^*_{\text{scaled}}$\\\hline
		\includegraphics[width=0.15\textwidth]{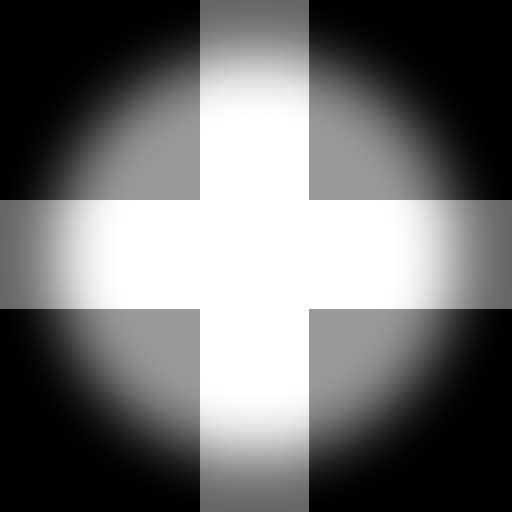}&\includegraphics[width=0.15\textwidth]{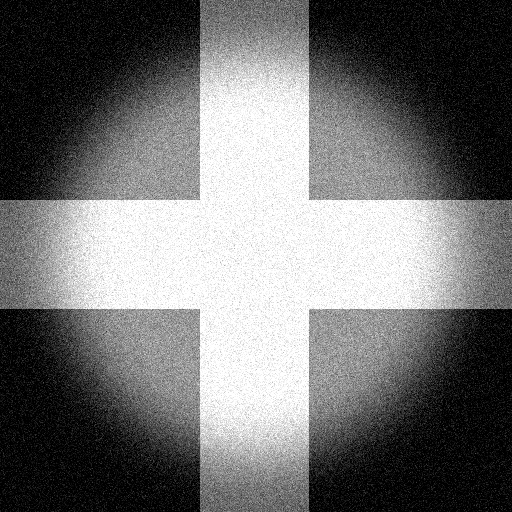}&
		\includegraphics[width=0.15\textwidth]{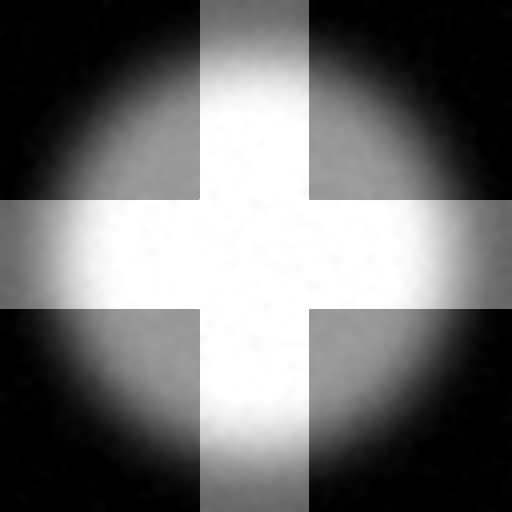}&
		\includegraphics[width=0.15\textwidth]{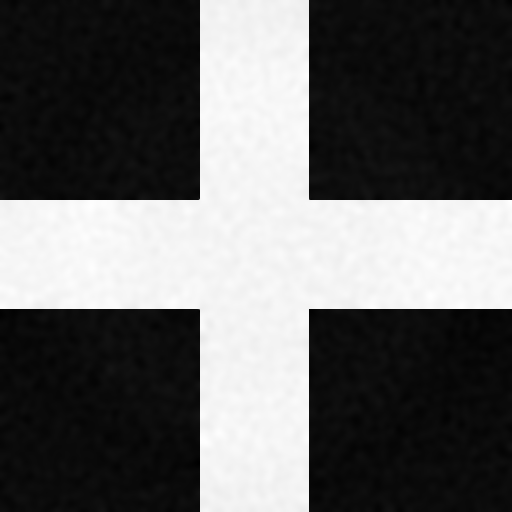}&
		\includegraphics[width=0.15\textwidth]{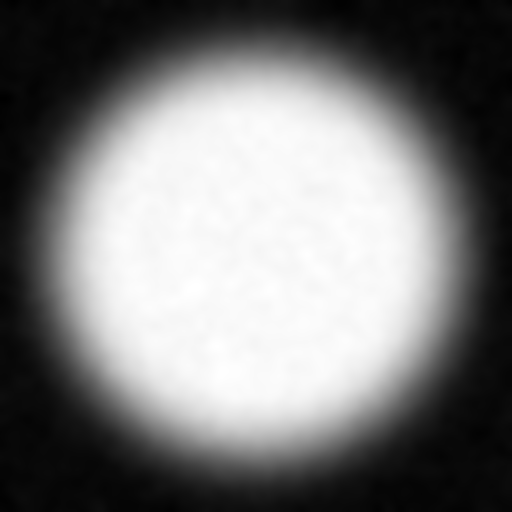}&
		\includegraphics[width=0.15\textwidth]{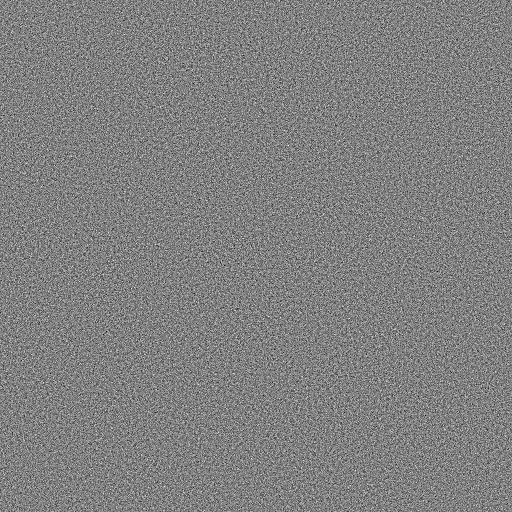}\\
		&PSNR$=22.10$&PSNR$=43.60$&
		&&STD$=19.69/255$\\\hline
		\includegraphics[width=0.15\textwidth]{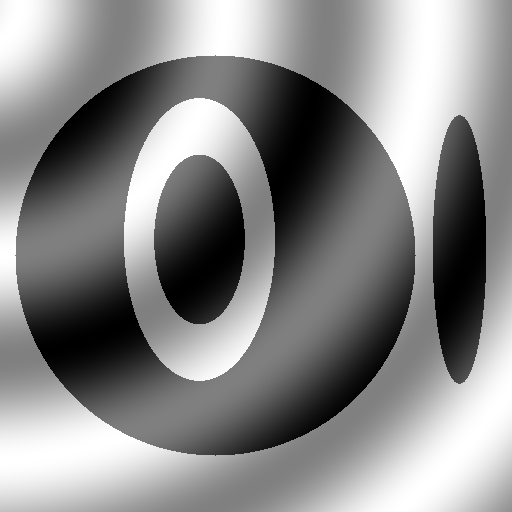}&\includegraphics[width=0.15\textwidth]{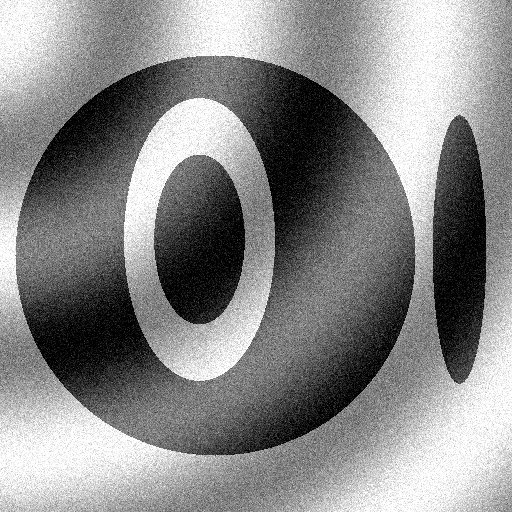}&
		\includegraphics[width=0.15\textwidth]{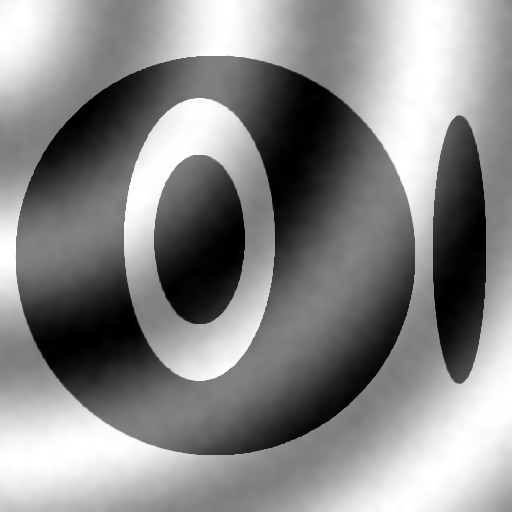}&
		\includegraphics[width=0.15\textwidth]{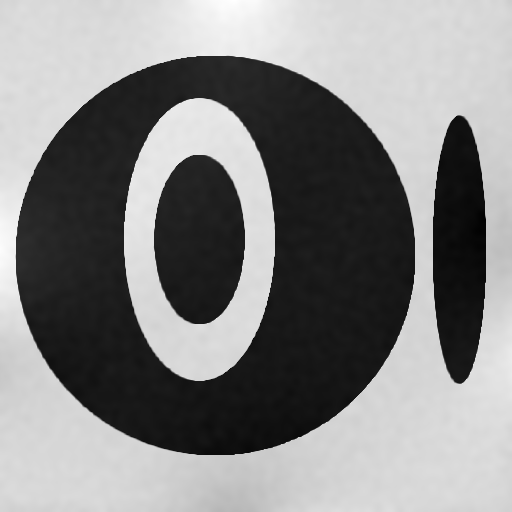}&
		\includegraphics[width=0.15\textwidth]{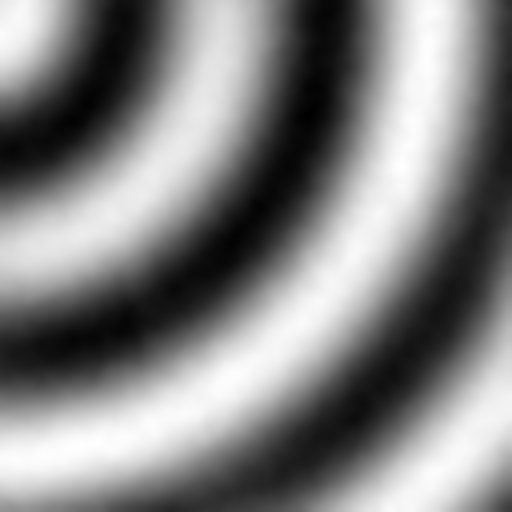}&
		\includegraphics[width=0.15\textwidth]{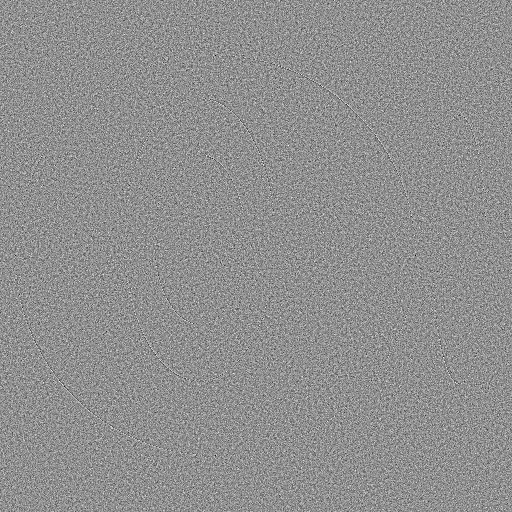}\\
		&PSNR$=22.10$&PSNR$=35.87$&
		&&STD$=19.97/255$\\\hline
		\includegraphics[width=0.15\textwidth]{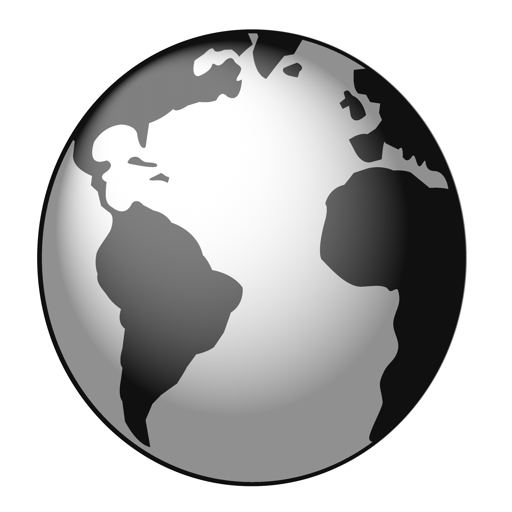}&\includegraphics[width=0.15\textwidth]{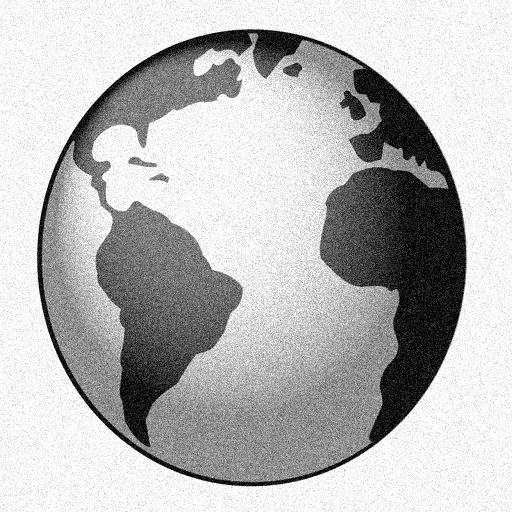}&
		\includegraphics[width=0.15\textwidth]{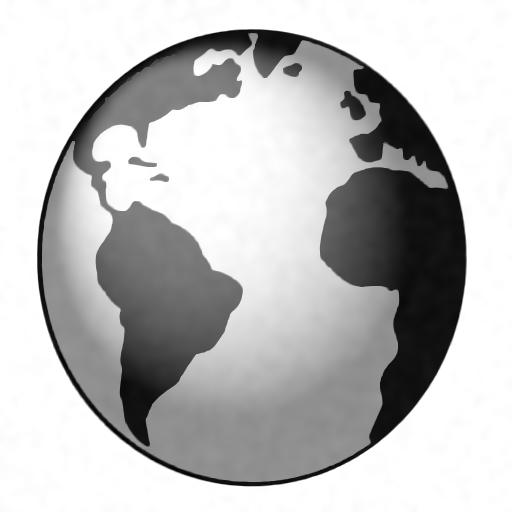}&
		\includegraphics[width=0.15\textwidth]{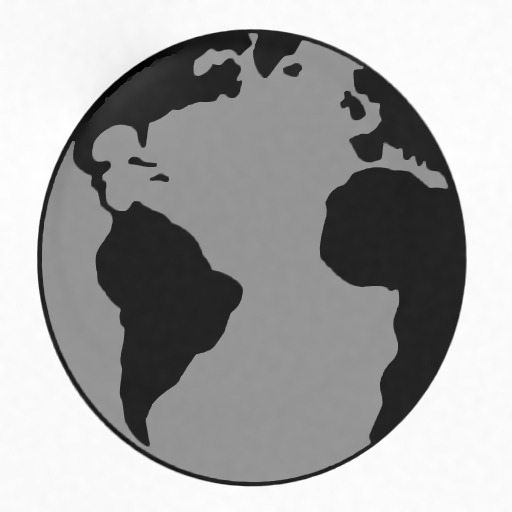}&
		\includegraphics[width=0.15\textwidth]{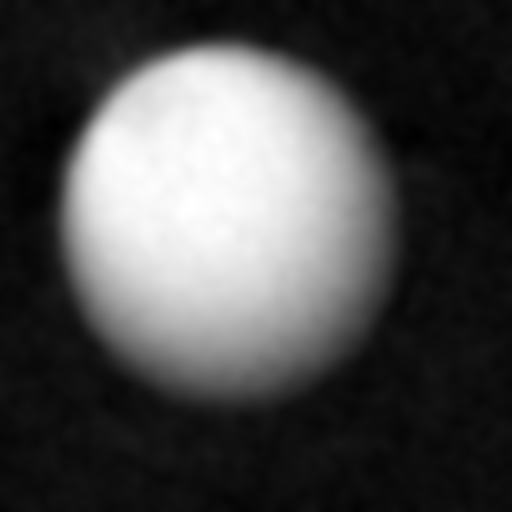}&
		\includegraphics[width=0.15\textwidth]{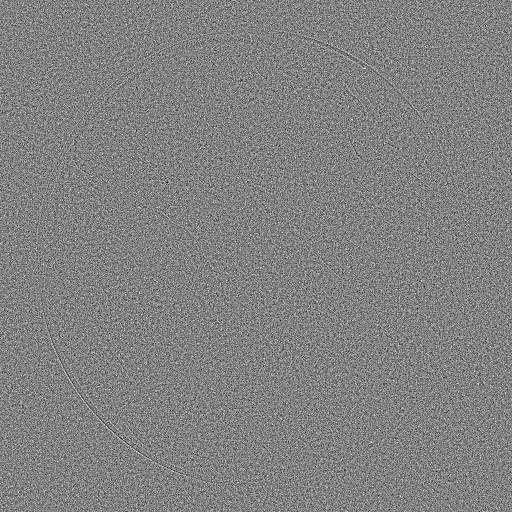}\\
		&PSNR$=22.12$&PSNR$=36.44$&
		&&STD$=19.82/255$\\\hline
		\includegraphics[width=0.15\textwidth]{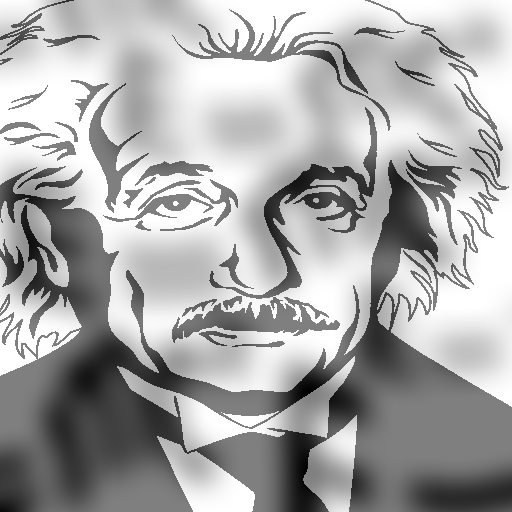}&
		\includegraphics[width=0.15\textwidth]{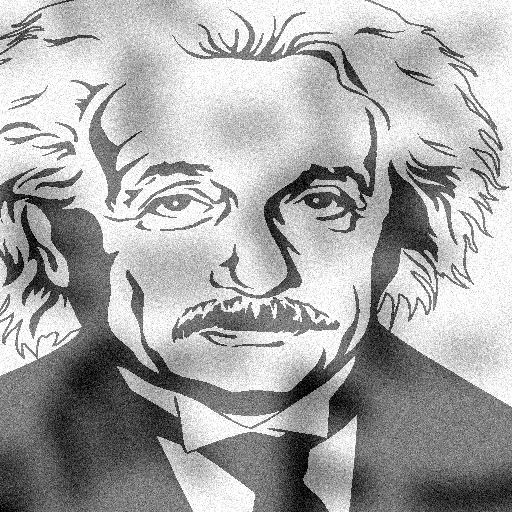}&
		\includegraphics[width=0.15\textwidth]{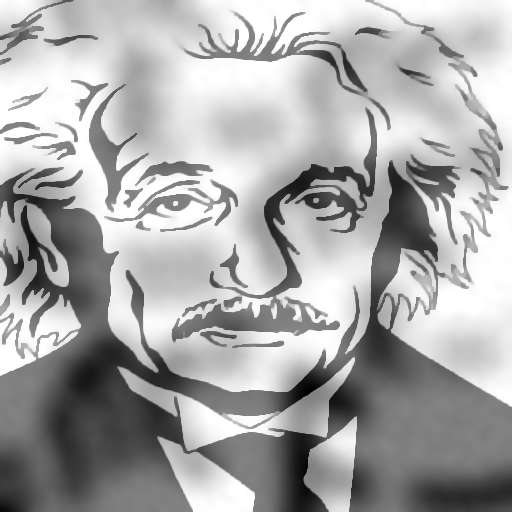}&
		\includegraphics[width=0.15\textwidth]{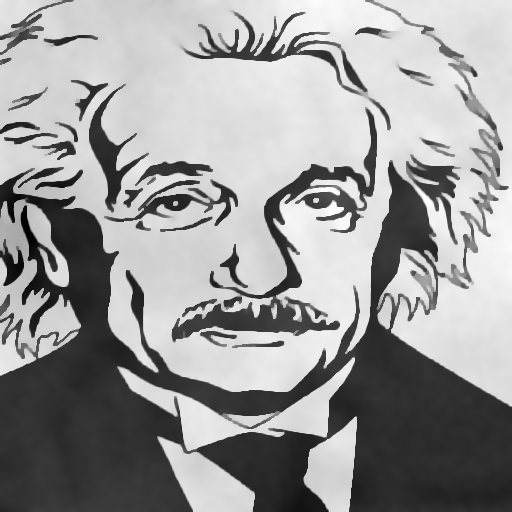}&
		\includegraphics[width=0.15\textwidth]{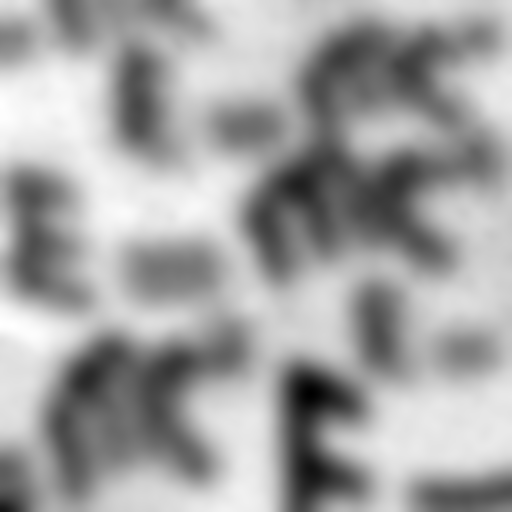}&
		\includegraphics[width=0.15\textwidth]{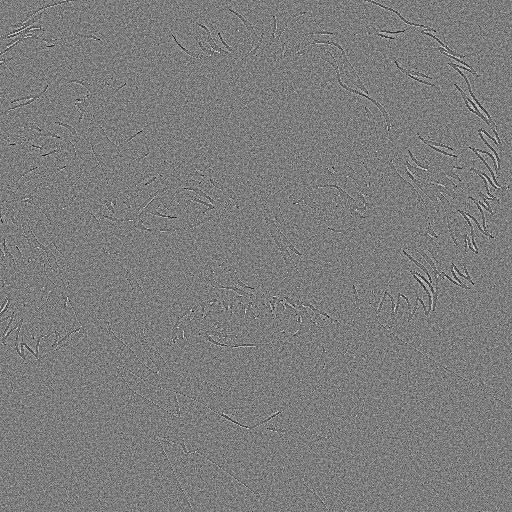}\\
		&PSNR$=22.12$&PSNR$=25.69$&&&STD$=23.19/255$\\\hline
	\end{tabular}
	\caption{Decomposition of synthetic images  with noise ($\sigma=20/255$). The first column ($f_{\text{clean}}$) shows the clean images; the second column ($f_{\text{noise}}$) shows the noisy inputs;   the third column ($u^*$) shows summations of the identified $v^*$ and 
		$w^*$ components; the subsequent columns show  respectively $v^*$, $w^*$ and $n^*$, and for visualization, they are linearly scaled between 0 and 1. Model parameters for these examples are: $\alpha_0=2\times10^{-2}$, $\alpha_\text{curv}=0.1$, $\alpha_w = 80$, and $\alpha_n=1\times10^{-5}$. PSNR values for $f_{\text{noise}}$ and $u^*$ are computed, and the standard deviations (STD) of respective $n^*$ are reported.}\label{fig_syn_imgs}
\end{figure}

\begin{figure}
	\centering
	\begin{tabular}{c@{\hspace{2pt}}c@{\hspace{2pt}}c@{\hspace{2pt}}c}
		(a)&(b)&(c)&(d)\\
		\raisebox{0.1cm}{\includegraphics[width=0.185\textwidth]{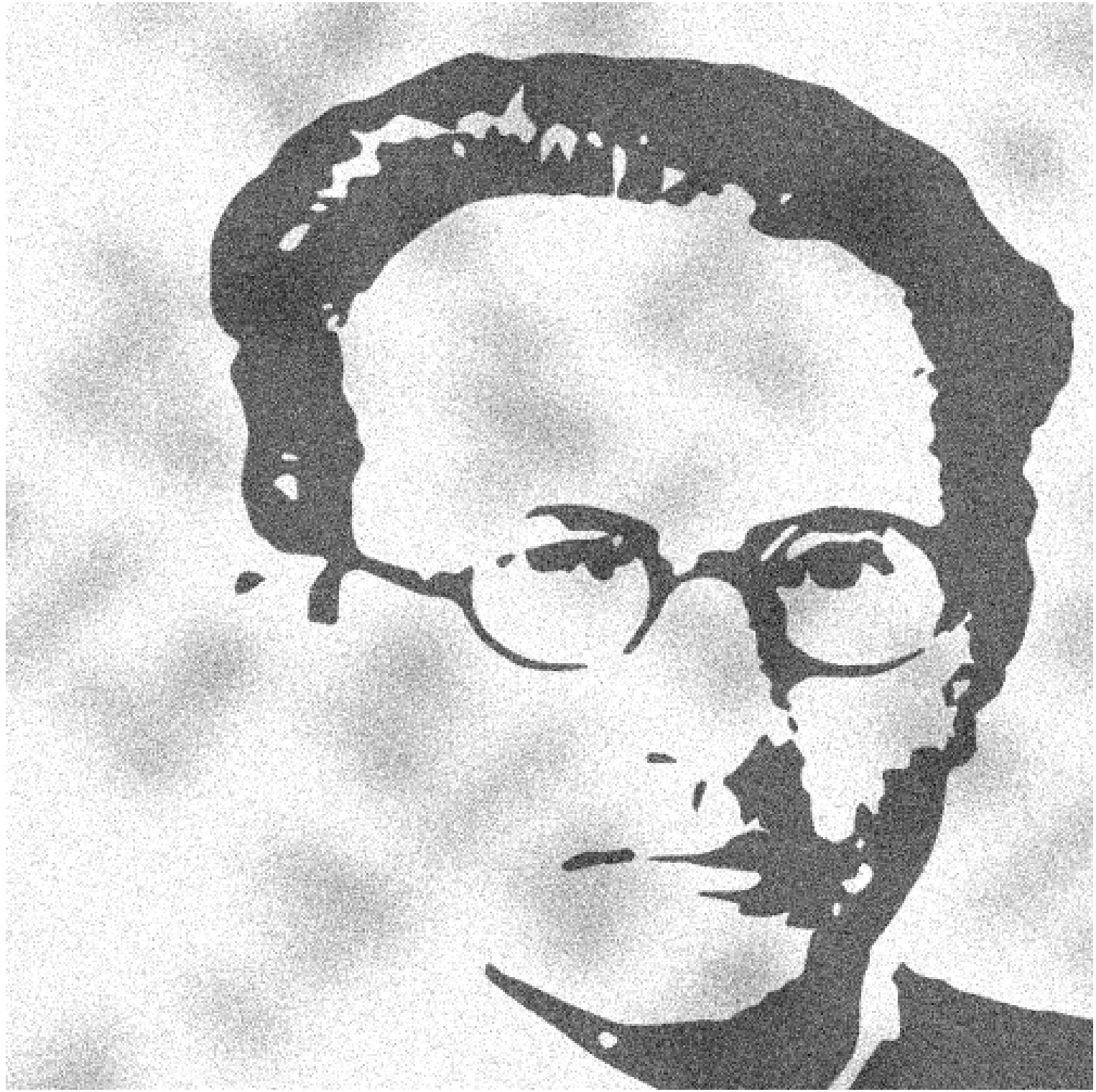}}&
		\includegraphics[width=0.23\textwidth]{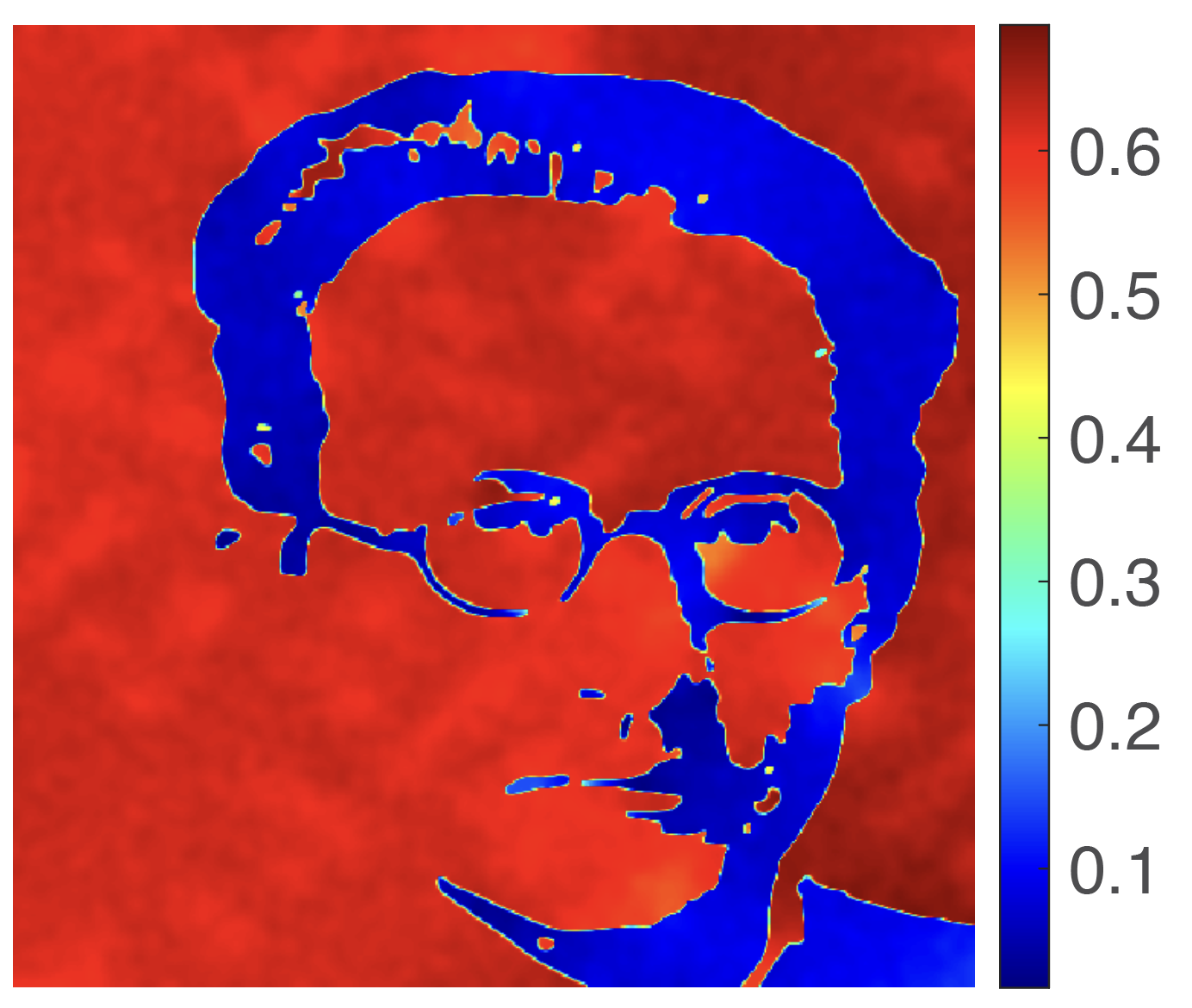}&
		\includegraphics[width=0.23\textwidth]{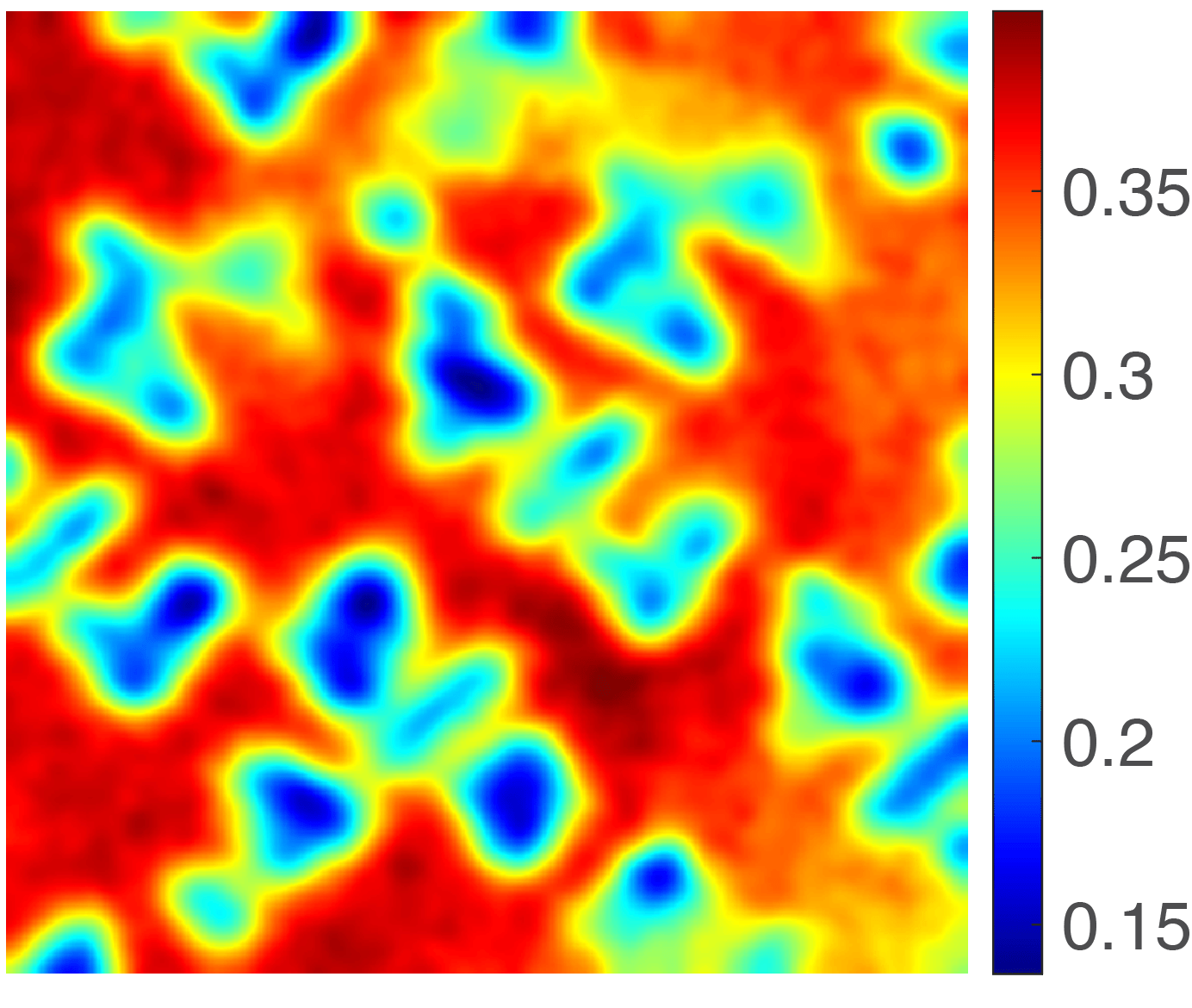}&
		\includegraphics[width=0.23\textwidth]{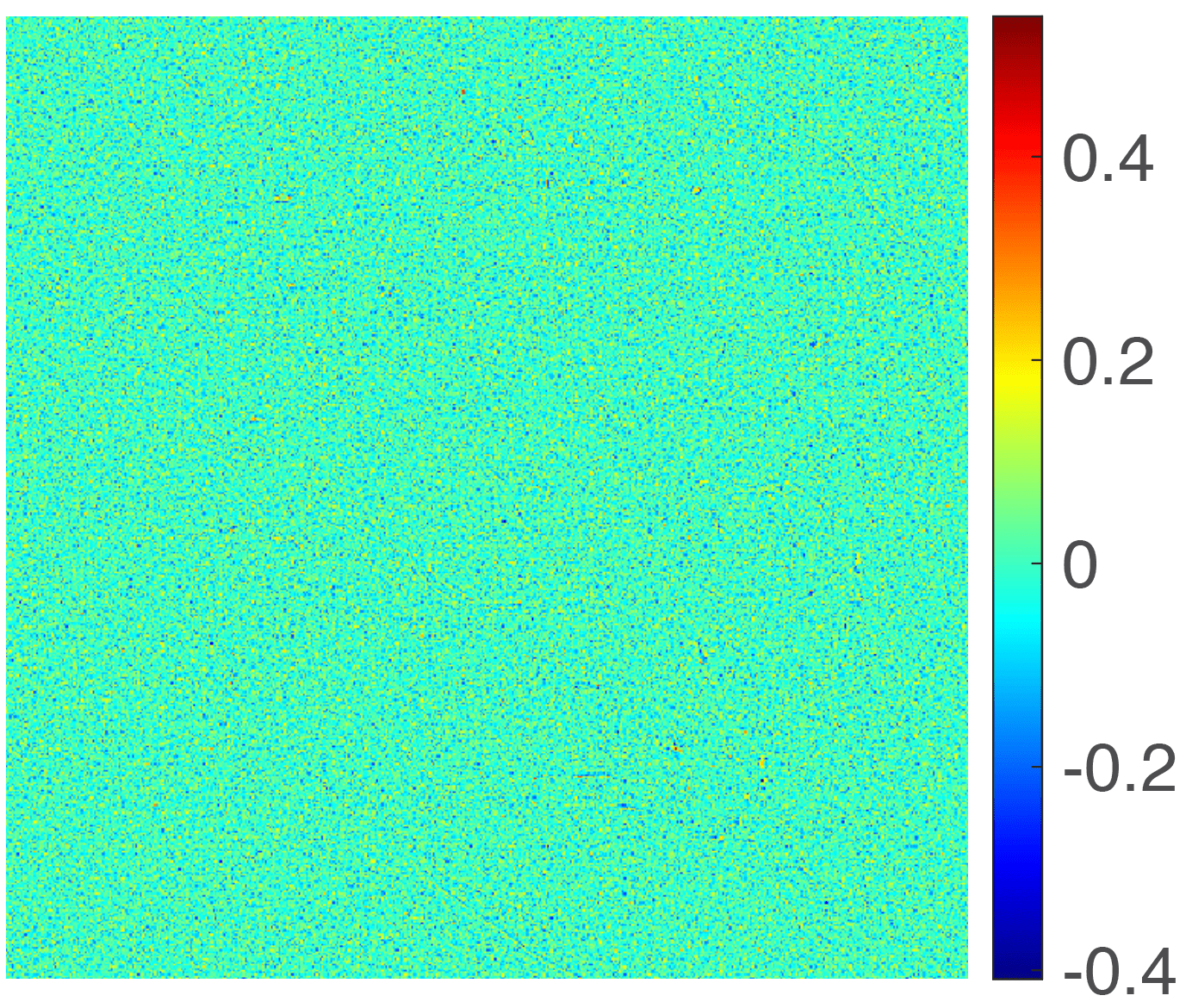}\\
		(f)&(g)&(h)&(i)\\
		\includegraphics[width=0.185\textwidth]{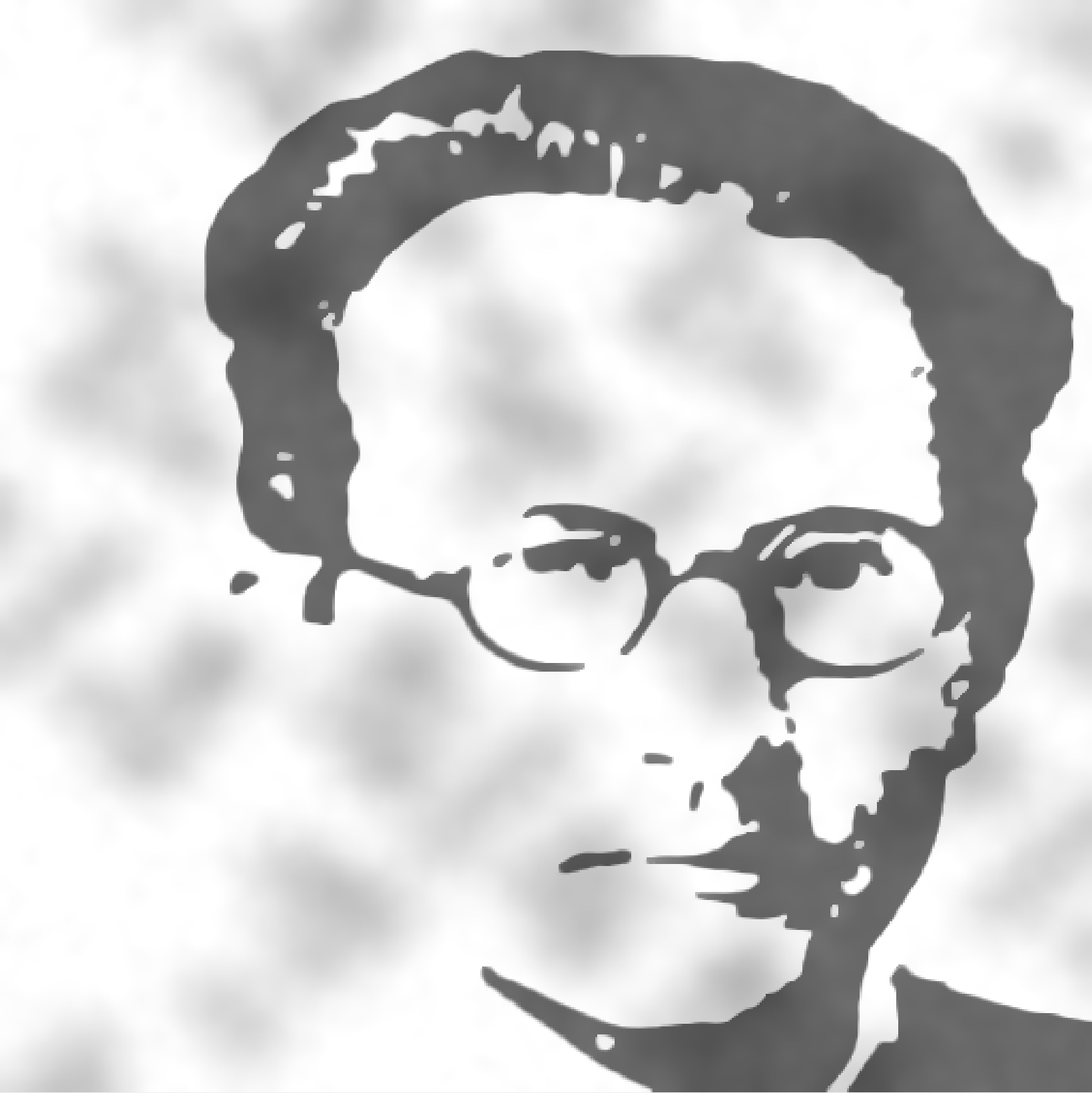}&
		\includegraphics[width=0.185\textwidth]{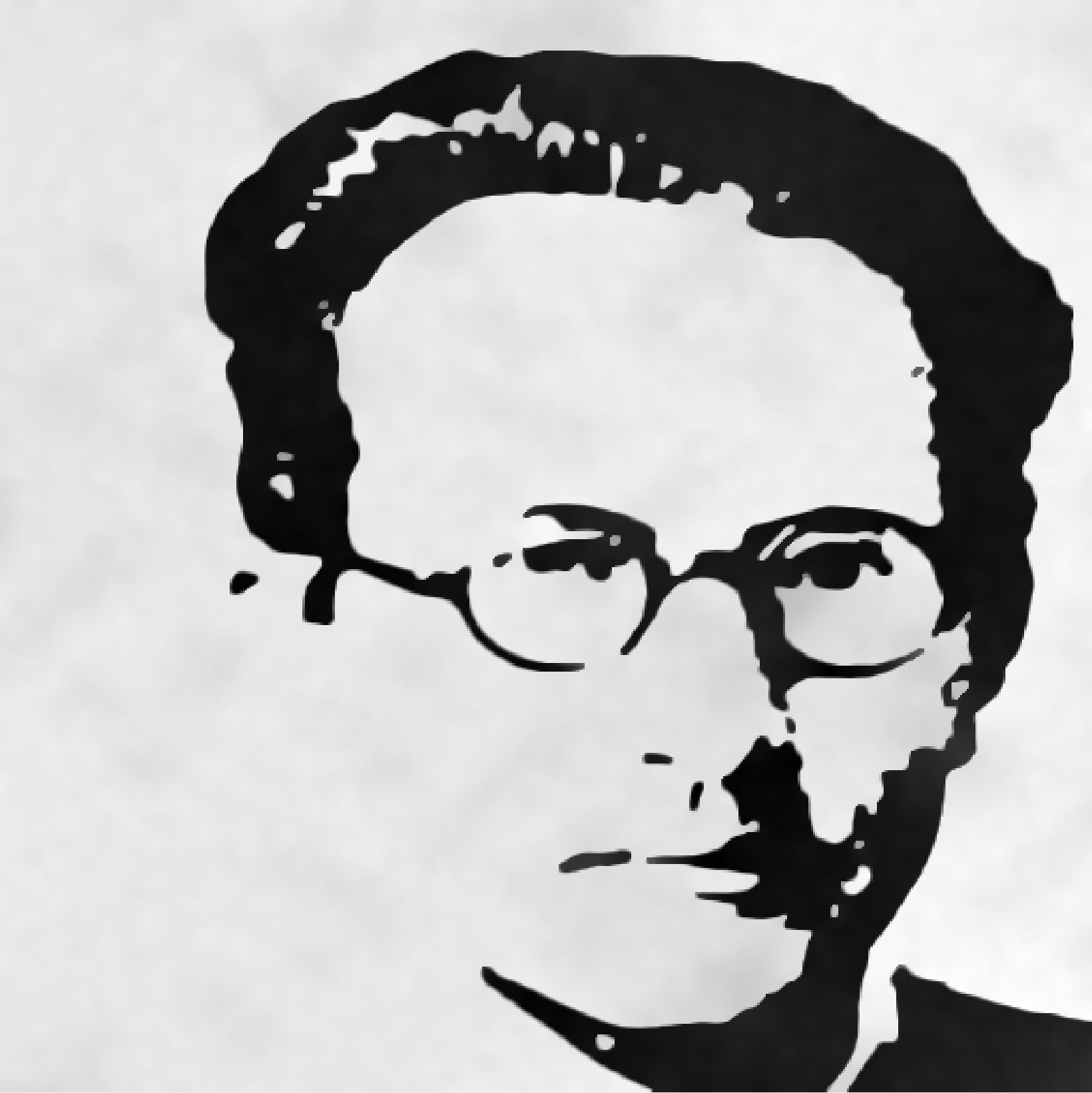}&
		\includegraphics[width=0.185\textwidth]{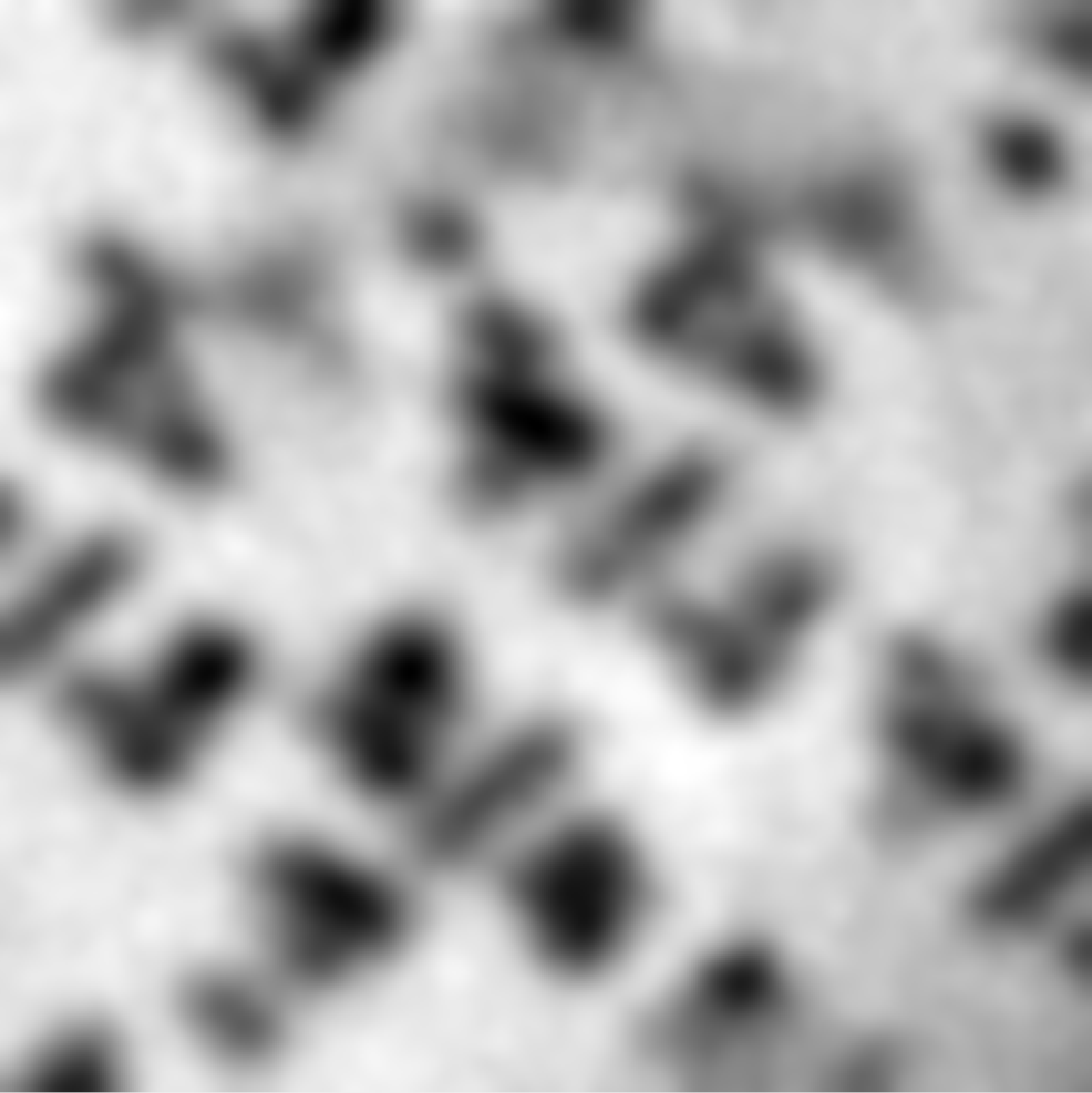}&
		\includegraphics[width=0.185\textwidth]{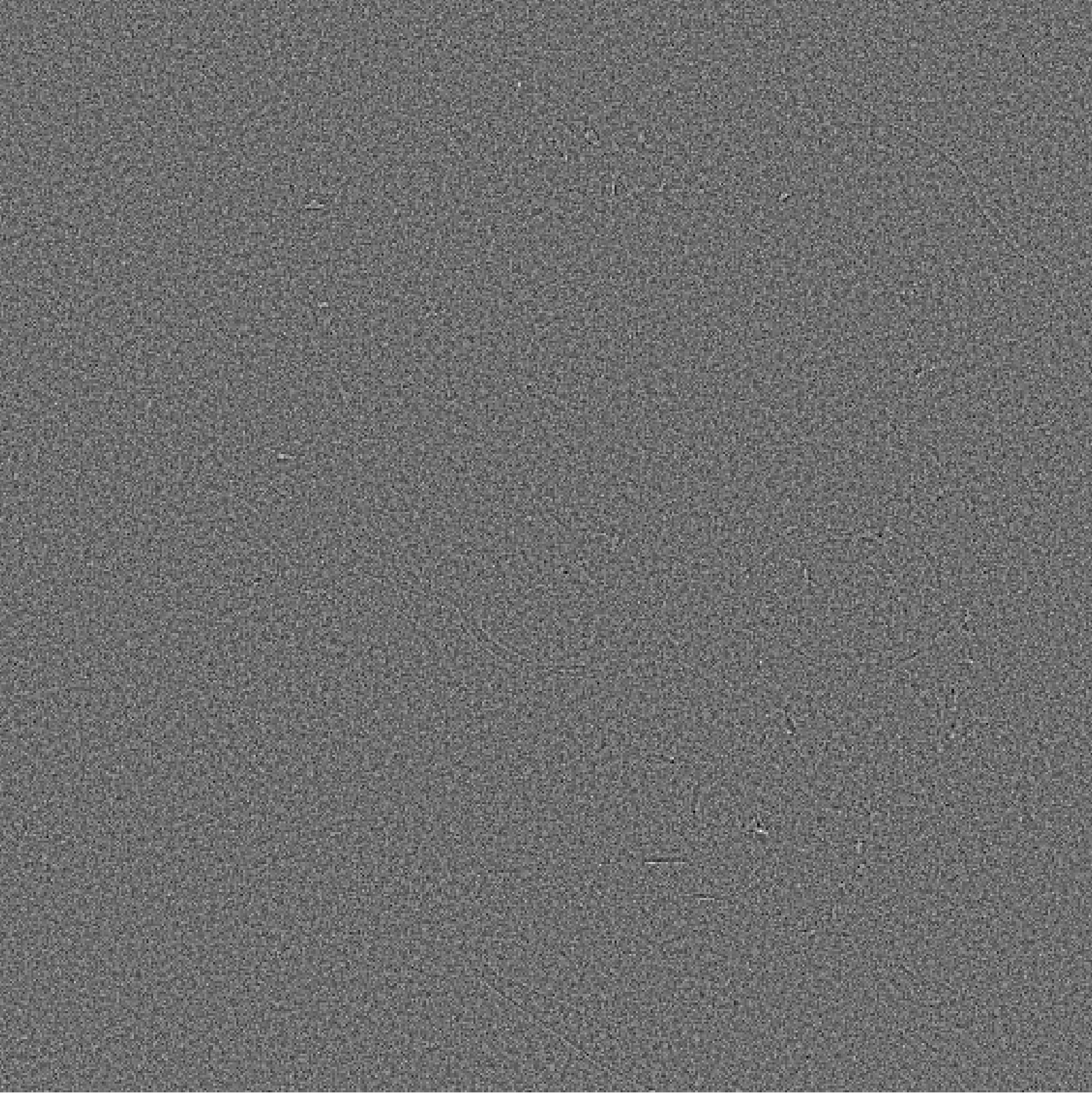}
	\end{tabular}
	\caption{Illustration of the proposed method without scaling the components. (a) Input noisy image ($\sigma=20/255$). (b) Identified structure component $v^*$. (c) Identified smooth component $w^*$. (d) Identified noise component $n^*$. (f) The reconstructed $u^*$ component (PSNR$=36.05$). (g)-(i) are the scaled versions of (b)-(d), respectively. The proposed method successfully separates visually distinctive components. Noticeably, $n^*$ in (d) captures only unstructured oscillations (STD=$19.77/255$).}\label{fig_mesh_exp}
\end{figure}

\begin{figure}[t!]
	\centering
	\begin{tabular}{c@{\hspace{2pt}}c@{\hspace{2pt}}c@{\hspace{2pt}}c@{\hspace{2pt}}c@{\hspace{2pt}}c}
		$f_{\text{clean}}$&$f_{\text{noise}}$&$u^*$&$v^*_{\text{scaled}}$&$w^*_{\text{scaled}}$&$n^*_{\text{scaled}}$\\\hline
		\includegraphics[width=0.15\textwidth]{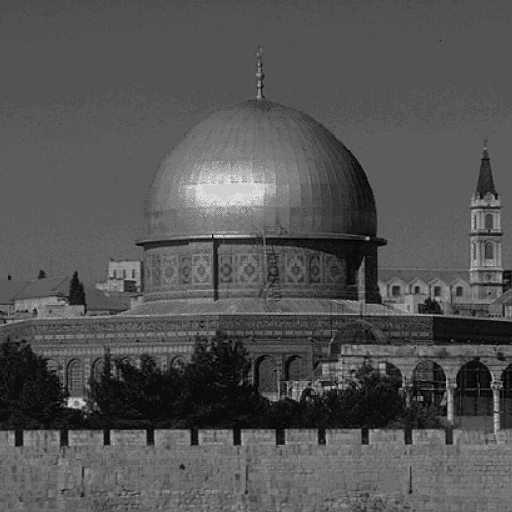}&\includegraphics[width=0.15\textwidth]{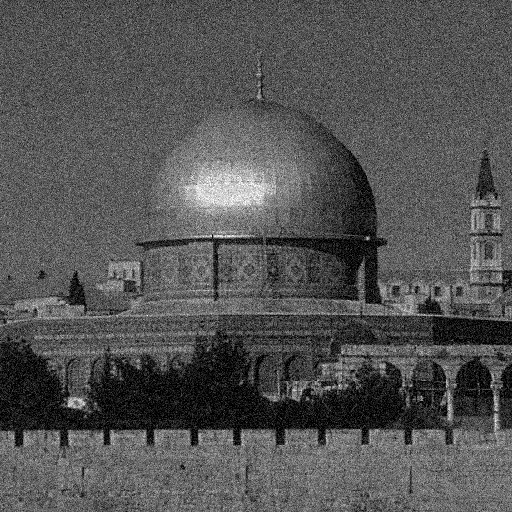}&
		\includegraphics[width=0.15\textwidth]{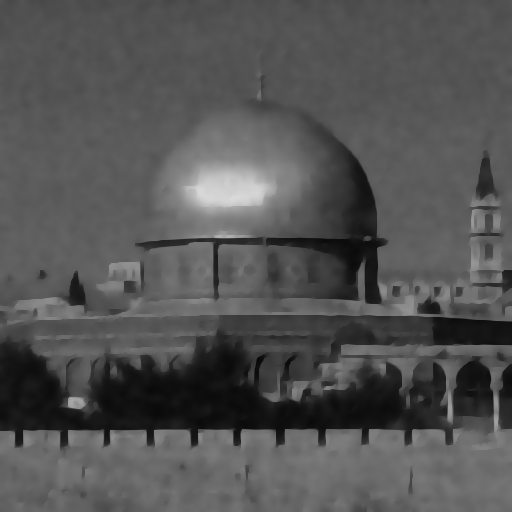}&
		\includegraphics[width=0.15\textwidth]{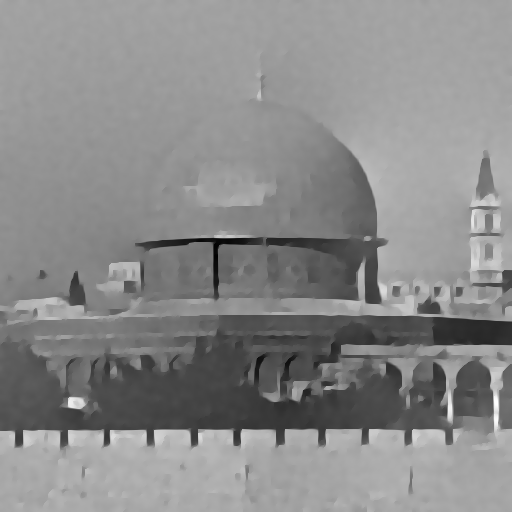}&
		\includegraphics[width=0.15\textwidth]{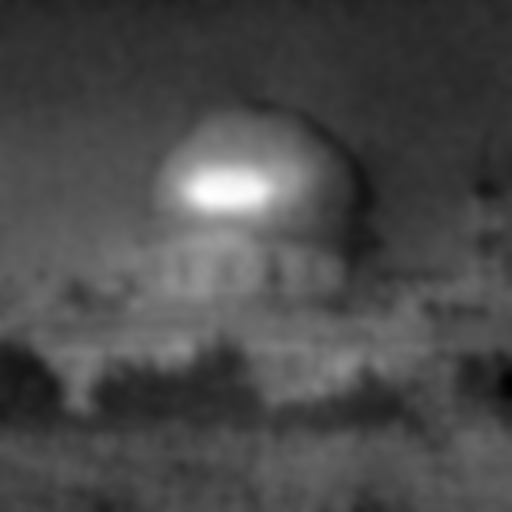}&
		\includegraphics[width=0.15\textwidth]{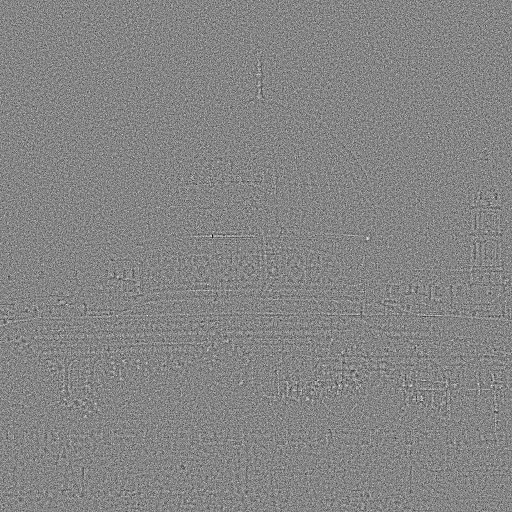}\\
		&PSNR$=22.11$&PSNR$=28.50$&&&STD$=19.87/255$\\\hline
		\includegraphics[width=0.15\textwidth]{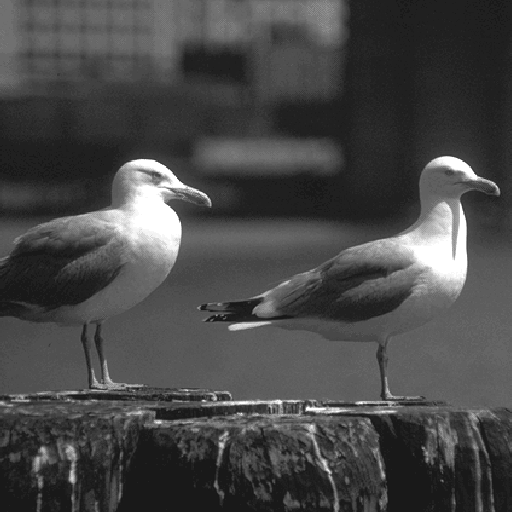}&\includegraphics[width=0.15\textwidth]{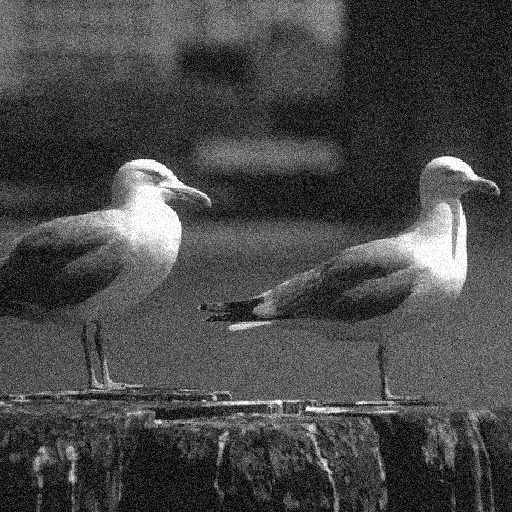}&
		\includegraphics[width=0.15\textwidth]{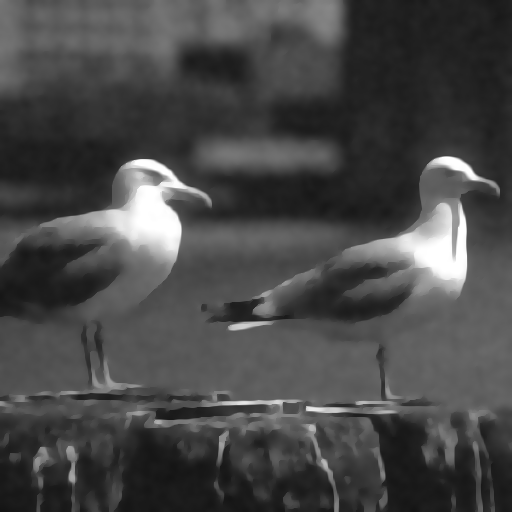}&
		\includegraphics[width=0.15\textwidth]{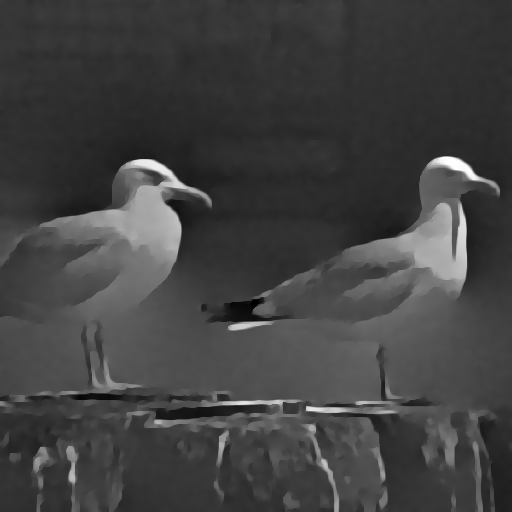}&
		\includegraphics[width=0.15\textwidth]{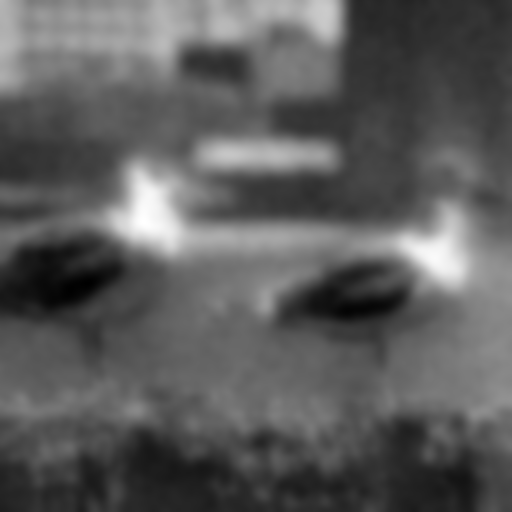}&
		\includegraphics[width=0.15\textwidth]{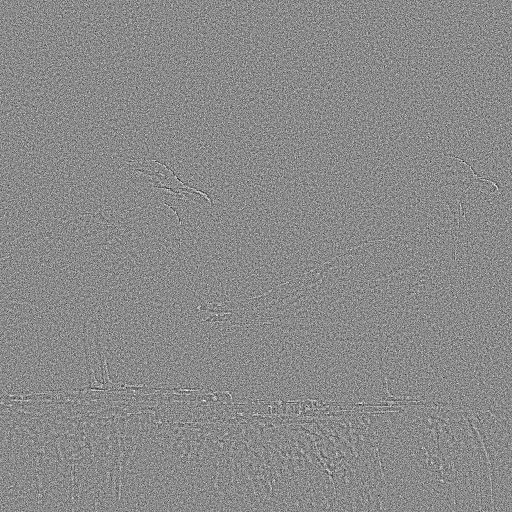}\\
		&PSNR$=22.11$&PSNR$=30.57$&&&STD$=19.24/255$\\\hline
	\end{tabular}
	\caption{Decomposition of photographic images  with noise ($\sigma=20/255$). The first column ($f_{\text{clean}}$) shows the clean images; the second column ($f_{\text{noise}}$) shows the noisy inputs;   the third column ($u^*$) shows summations of the identified $v^*$ and 
		$w^*$ components; the subsequent columns show  respectively $v^*$, $w^*$ and $n^*$, and for visualization, they are linearly scaled between 0 and 1. Model parameters for these examples are: $\alpha_0=2\times10^{-3}$, $\alpha_\text{curv}=0.5$, $\alpha_w = 50$, and $\alpha_n=0.1$. PSNR values for $f_{\text{noise}}$ and $u^*$ are computed, and the standard deviations (STD) of respective $n^*$ are reported.}\label{fig_real}
\end{figure}

We illustrate   general performances of the proposed model using two sets of experiments. In Figure~\ref{fig_syn_imgs}, we show the decomposition results for synthetic images  generated by convex combinations of piecewise constant components and smoothly varying shades.  In Figure~\ref{fig_real}, we show the results for grayscale photos with richer details. For both sets of examples, Gaussian noise with intensity $\sigma=20/255$ is added.

In the first row of Figure~\ref{fig_syn_imgs}, a white cross in the dark is shed by a soft light at the center. The proposed model successfully extracts the cross $v^*$ where the intensity variation due to the lighting is removed, and the important geometric features such as sharp corners as well as straight boundaries are accurately reconstructed. The lighting is captured by the $w^*$ component, and it is well separated from $v^*$. The $n^*$ component corresponds to the noise whose standard deviation is close to the true noise level.  In the second row, we test the proposed model on a combination of silhouettes of ellipses and a circular wave of lighting. Under the non-uniform lighting, the ellipses are correctly identified in $v^*$ and filled with nearly constant intensity. The $w^*$ component shows the correct wave pattern of the lighting. In the third row, we have a shaded globe, where the source of the light is fixed at the top left corner. The proposed method correctly assigns in the $v^*$ component the silhouettes of the globe with sharp boundaries. Noticeably, the proposed model keeps the soft shadow characterizing the volumetric feature of a ball in the $w^*$ component with the correct direction of lighting. In the last row, we have a challenging case where a binarized portrait of \textit{Einstein} is blended with multiple blurry shades of the mass-energy equation. The $w^*$ component is more complicated than those in the previous examples. Our method successfully extracts the soft contents with complex shapes and leaves a relatively clean configuration for the $v^*$ component.

To further examine the identified components by our method, we show in Figure~\ref{fig_mesh_exp} the decomposition results without scaling. In (a), we show the noisy input image ($\sigma=20/255$) where shadows of Schr\"{o}dinger equations are cast over a binary portrait of \textit{Erwin Schr\"{o}dinger}. In (b)-(d), we present the identified structure $v^*$, smooth part $w^*$, and oscillatory part $n^*$ respectively without scaling. In (f), we show the reconstructed $u^*$ of the proposed method; and in (g)-(i), we show the scaled versions of (b)-(d), respectively. The ranges of the values confirm that the identified variations are substantial in (b) and (c); and $n^*$ in (d) captures unstructured oscillations whose standard deviation $19.77/255$ is close to the true one. 

In Figure~\ref{fig_real}, we apply the proposed model to photo-realistic grayscale images. In all examples, we observe that contents with sharp boundaries are captured by $v^*$; soft shadows and illumination variation are modeled by $w^*$; and the $n^*$ keeps fine textures plus noise. Noticeably, in the first row,  the rectangular reflectances on the dome are separated from the soft light and shadow defining the volume of the dome. In the second row, the foreground with crisp boundaries is distinguished from the background with blurry shapes.

The examples in Figure~\ref{fig_syn_imgs}  and \ref{fig_real} justify the effectiveness of the proposed method. Thanks to the combination of the $L^0$-gradient energy~\eqref{eq_Dv} and curvature regularization~\eqref{eq_euler_elastica_functional}, the identified $v^*$ component successfully preserves sharp structure boundaries while avoiding severe staircase effects. The extracted $w^*$ component consistently matches the soft intensity variation caused by lighting and shadows. The extracted $n^*$ component holds the oscillatory residuals including noise and fine-scale textures. We note that when the input images do not contain rich textures,  the $n^*$ components rendered by the proposed model do not contain structural details such as boundaries. It allows us to use the summation of $v^*$ and $w^*$, which is denoted by $u^*$, to reconstruct the clean image. This is justified by the PSNR values  reported in Figure~\ref{fig_syn_imgs}  and \ref{fig_real}. Moreover, for general real images, since smooth transition is not necessarily uniform in the whole domain, and $w^*$ by construction does not allow discontinuous gradients, the structure component $v^*$ are not strictly piecewise constant. Consequently, unnatural staircase effects may appear when applying $L^0$-gradient regularization alone. With the curvature regularization in our proposed model, mild transition is allowed in $v^*$ (See Remark 2.3), thus ameliorating the artifact. We further elaborate this in the next experiments.

\subsection{Ablation study}\label{sec_ablation_study}

To justify the combination of  regularizations in our model~\eqref{eq:proposed_model}, we compare it with the following settings:
\begin{enumerate}
	\item \textbf{Model I. } An image $f$ is decomposed into two parts: cartoon part $u^*$ and oscillatory part $n^*$
	\begin{align}
		\{u^*,n^*\}=\argmin_{u,n} \left[ \alpha_0\|\nabla u\|_0+\alpha_{n}\|n\|^2_{H^{-1}(\Omega)}+\frac{1}{2}\int_\Omega\left|f-(u+n)\right|^2\,d\bx \right]\;.\label{eq_abl_model1}
	\end{align}
	\item \textbf{Model II.} An image $f$ is decomposed into three parts: cartoon part $v^*$, smooth part $w^*$, and oscillatory part $n^*$
	\begin{align}
		\{v^*,w^*,n^*\}=\argmin_{v,w,n} \bigg[\alpha_0\|\nabla v\|_0&+\alpha_{w}\|\Delta w\|_{L^2(\Omega)}^2+\alpha_{n}\|n\|^2_{H^{-1}(\Omega)}\nonumber\\
		&+\frac{1}{2}\int_\Omega\left|f-(v+w+n)\right|^2\,d\bx\bigg]\;.\label{eq_abl_model2}
	\end{align}
	\item \textbf{Model III.} An image $f$ is decomposed into two parts: homogeneous structure $u^*$ and oscillatory part $n^*$
	\begin{align}
		\{u^*,n^*\}=\argmin_{u,n} \bigg[\alpha_0\|\nabla u\|_0&+\alpha_{\text{curv}}\int_\Omega \left(\nabla\cdot\frac{\nabla u}{|\nabla u|}\right)^2\left|\nabla u\right|\,d\bx+\alpha_{n}\|n\|^2_{H^{-1}(\Omega)}\nonumber\\ &+\frac{1}{2}\int_\Omega\left|f-(u+n)\right|^2\,d\bx\bigg]\;.\label{eq_abl_model3}
	\end{align}
\end{enumerate}
Compared to the proposed model, there is no smoothly varying component in Model I, and the regularity of $u^*$ is mainly affected by the $L^0$-gradient. In Model II, the noise-less component $u^*$ is decomposed into a cartoon part $v^*$ and a smooth part $w^*$, the same as in the proposed model; but there is no curvature regularization on $v^*$. In Model III, we impose both $L^0$-gradient and curvature regularization on the noise-free approximation $u^*$, but there is no component that models the smooth shading. 

\begin{figure}
	\centering
	\begin{tabular}{c@{\hspace{2pt}}c@{\hspace{2pt}}c@{\hspace{2pt}}c@{\hspace{2pt}}c@{\hspace{2pt}}c}
		(a)&(b)&(c)&(d)&(e)&(f)\\
		Input &Model I&Model II&Model III&Proposed&True\\\hline
		\includegraphics[width=0.15\textwidth]{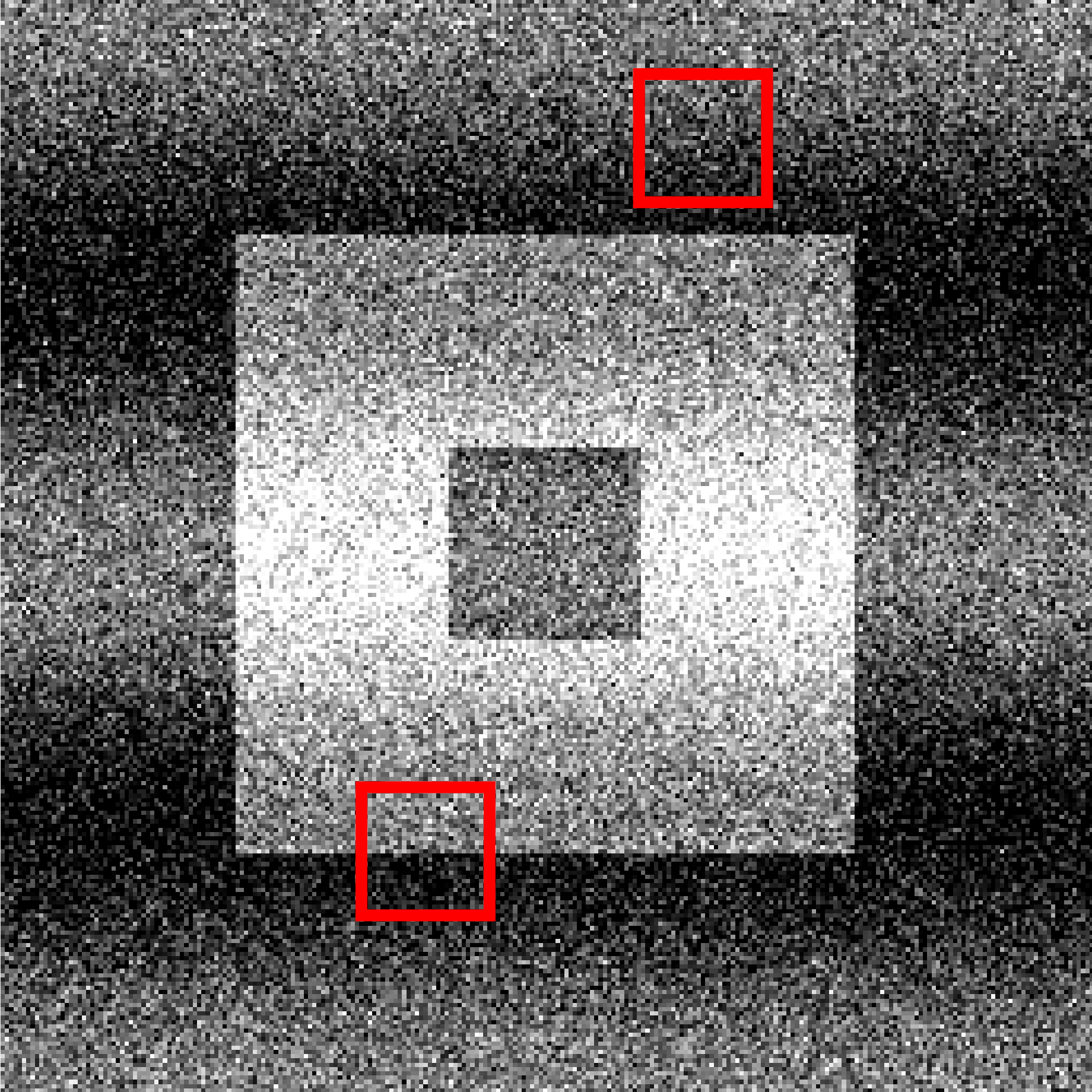}&
		\includegraphics[width=0.15\textwidth]{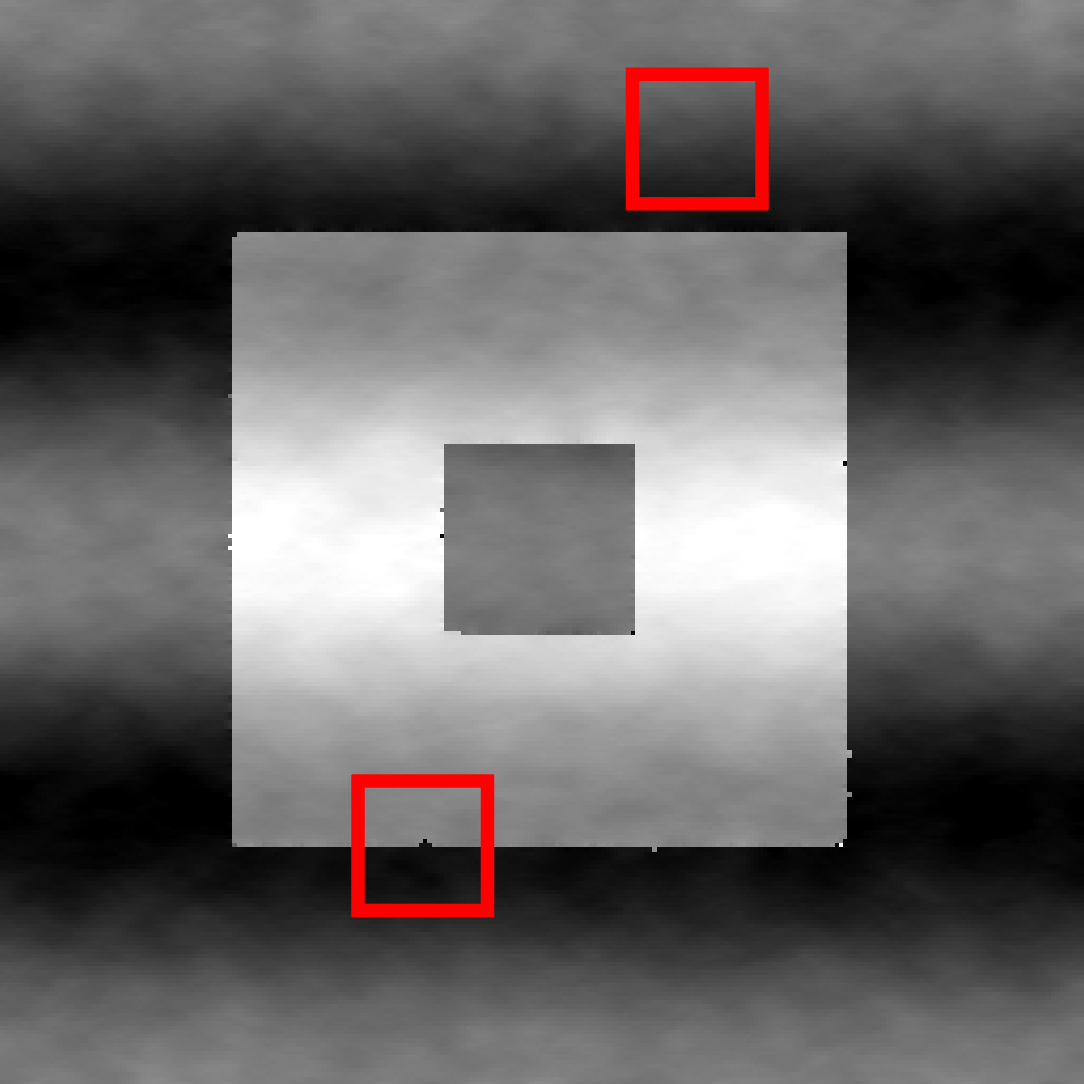}&
		\includegraphics[width=0.15\textwidth]{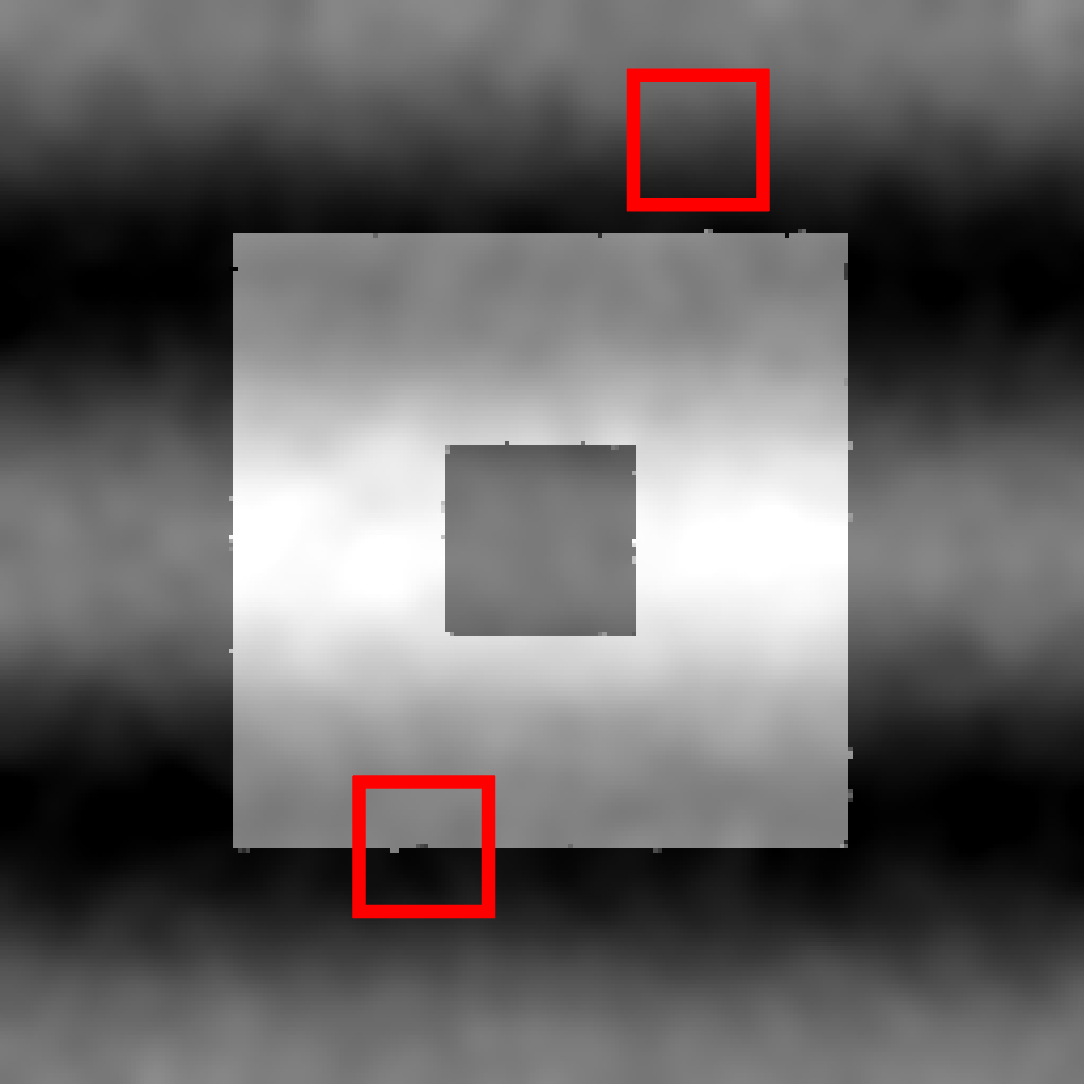}&
		\includegraphics[width=0.15\textwidth]{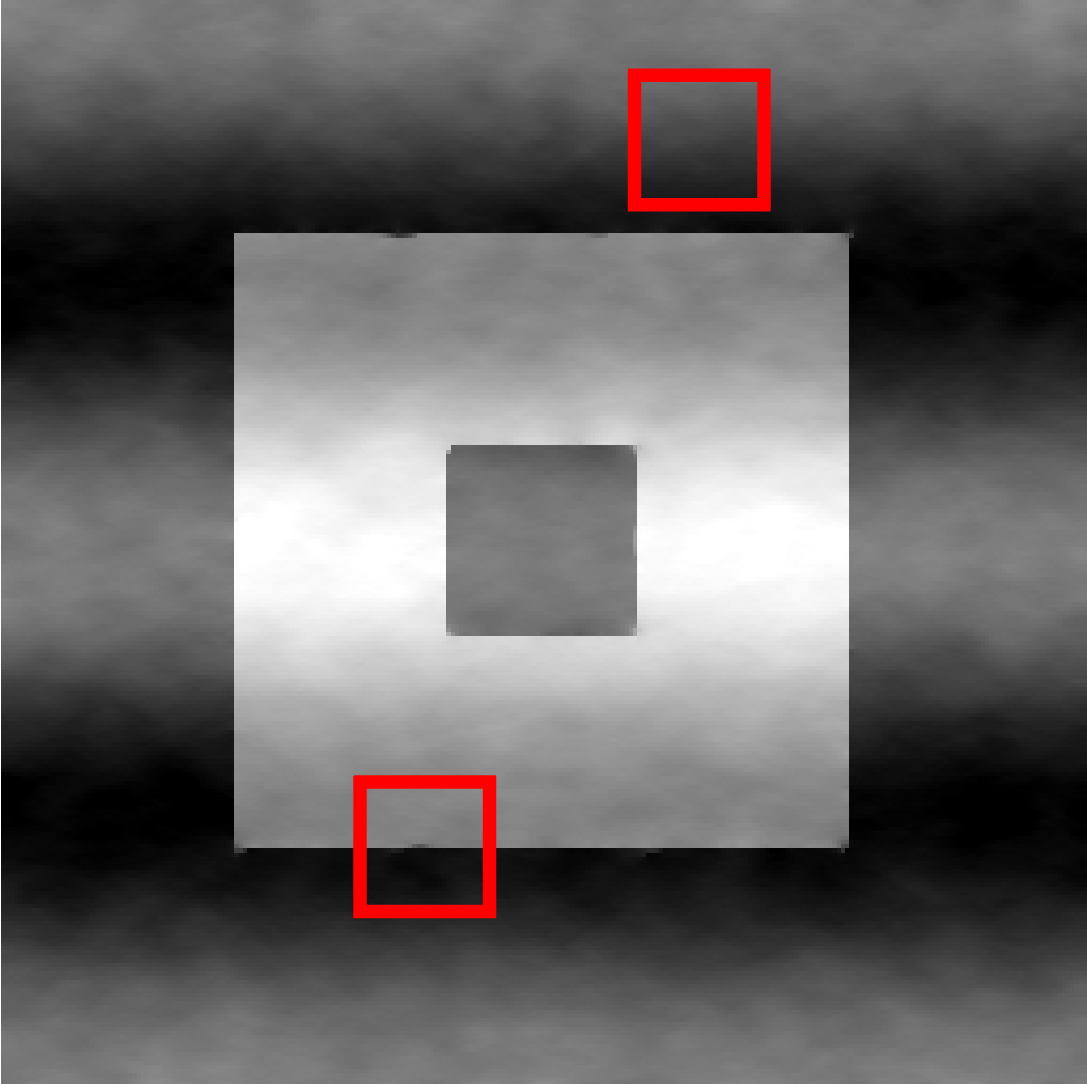}&
		\includegraphics[width=0.15\textwidth]{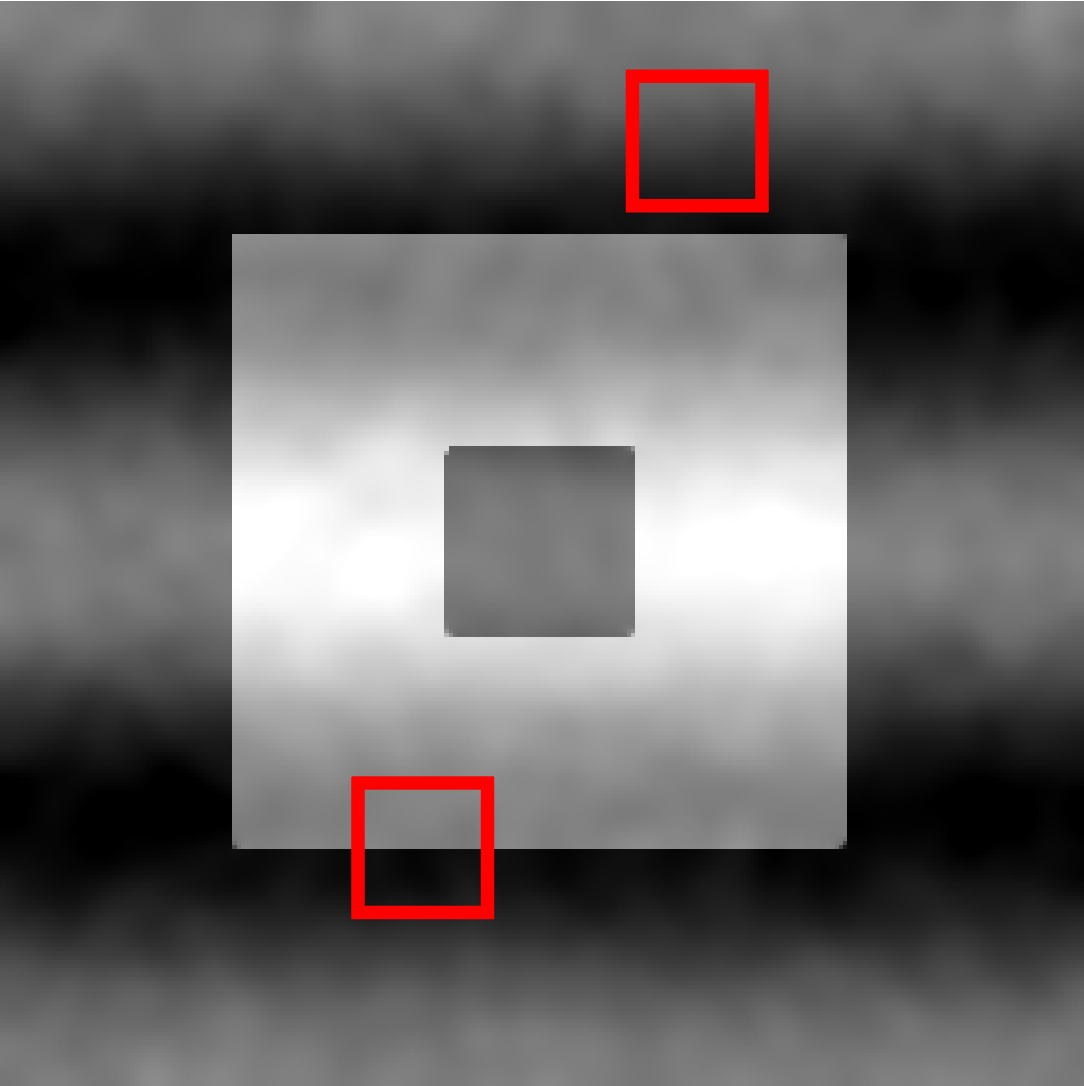}&
		\includegraphics[width=0.15\textwidth]{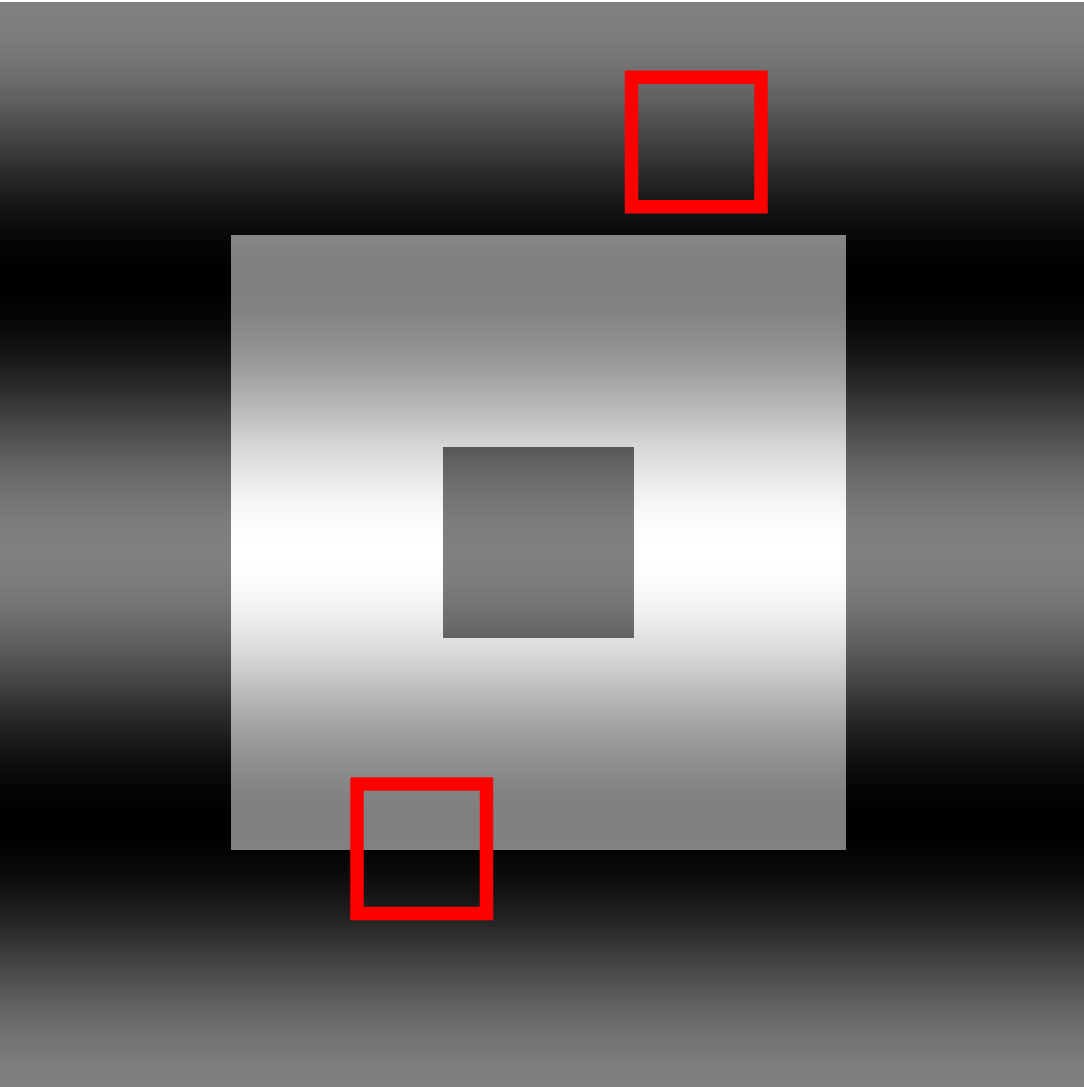}\\
		\includegraphics[width=0.15\textwidth]{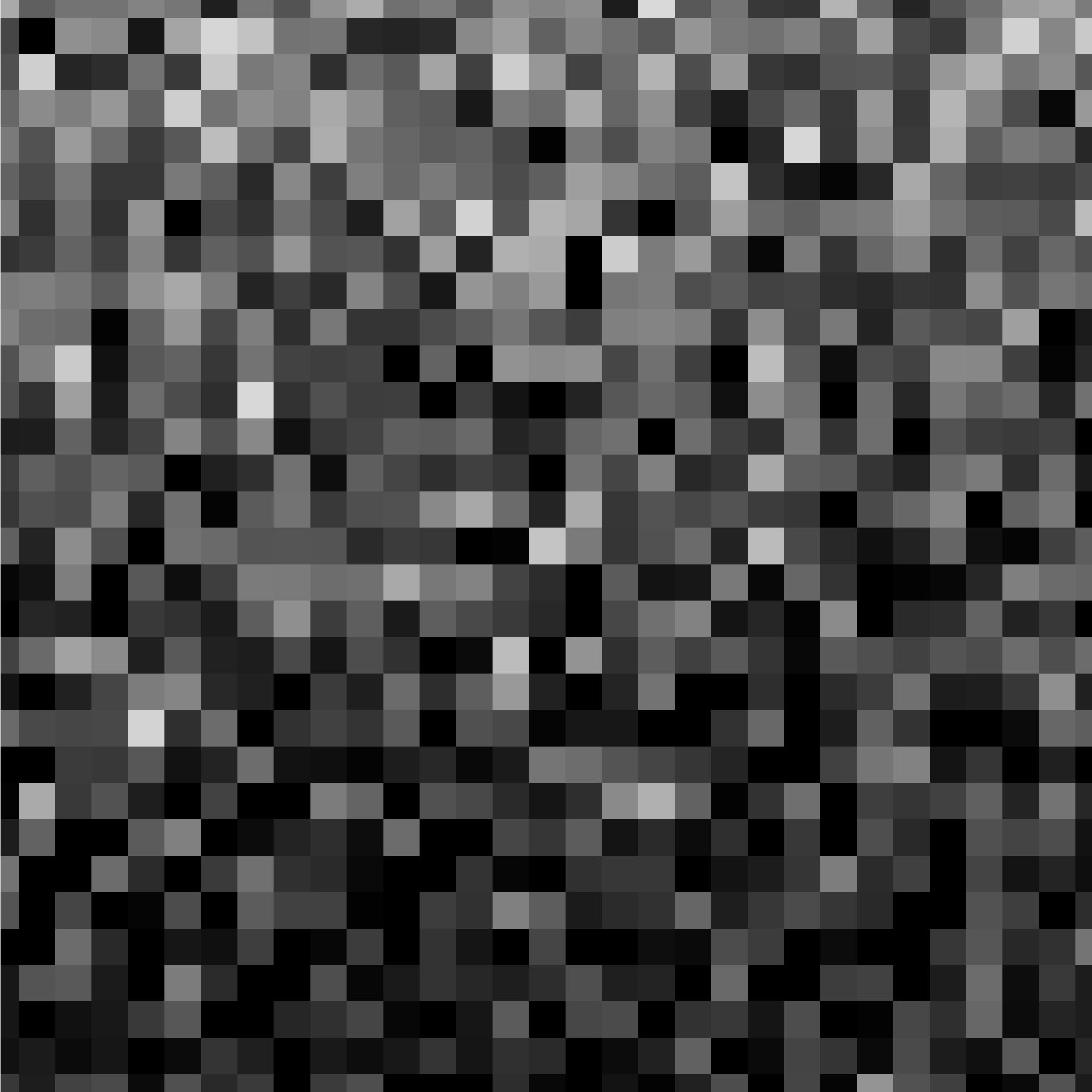}&
		\includegraphics[width=0.15\textwidth]{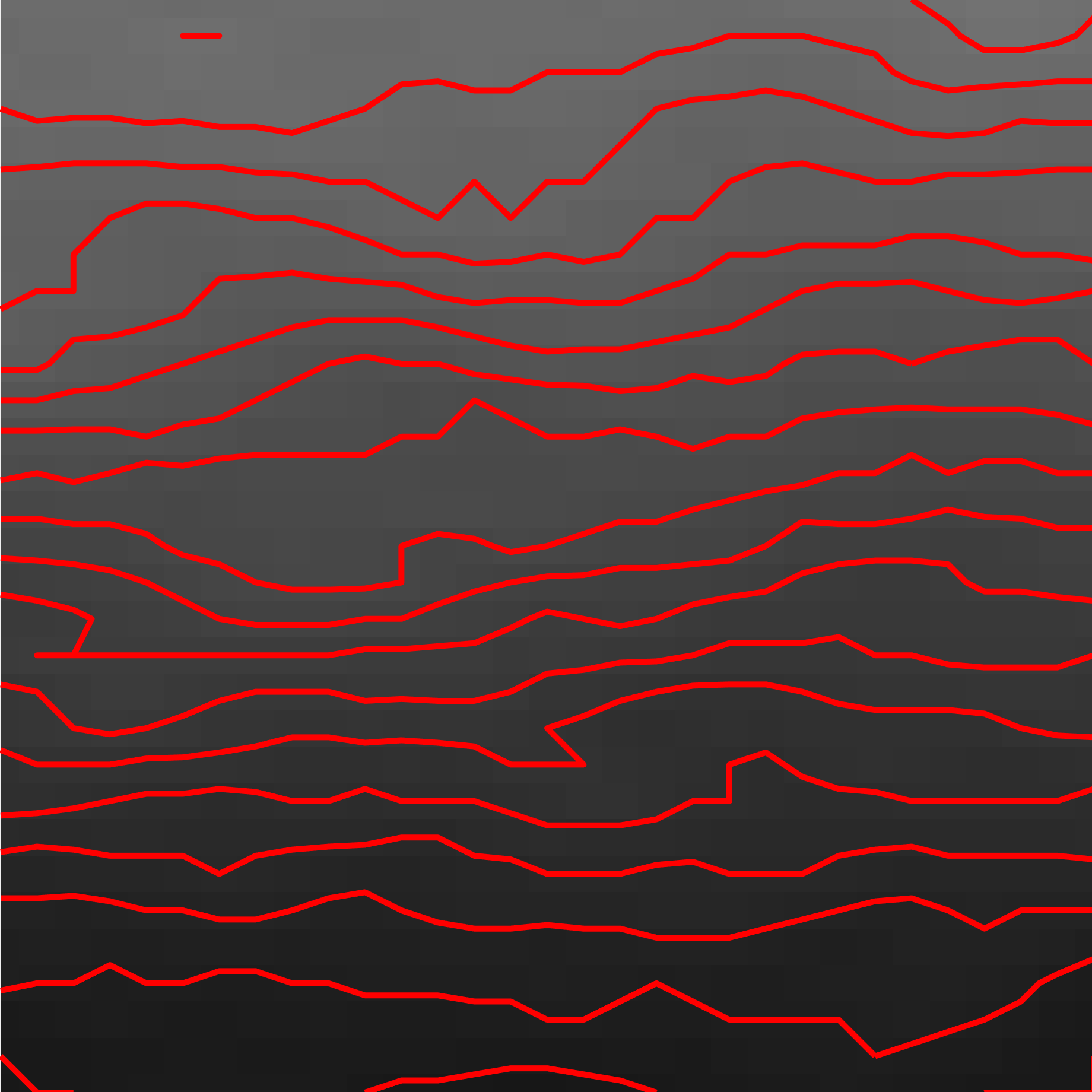}&
		\includegraphics[width=0.15\textwidth]{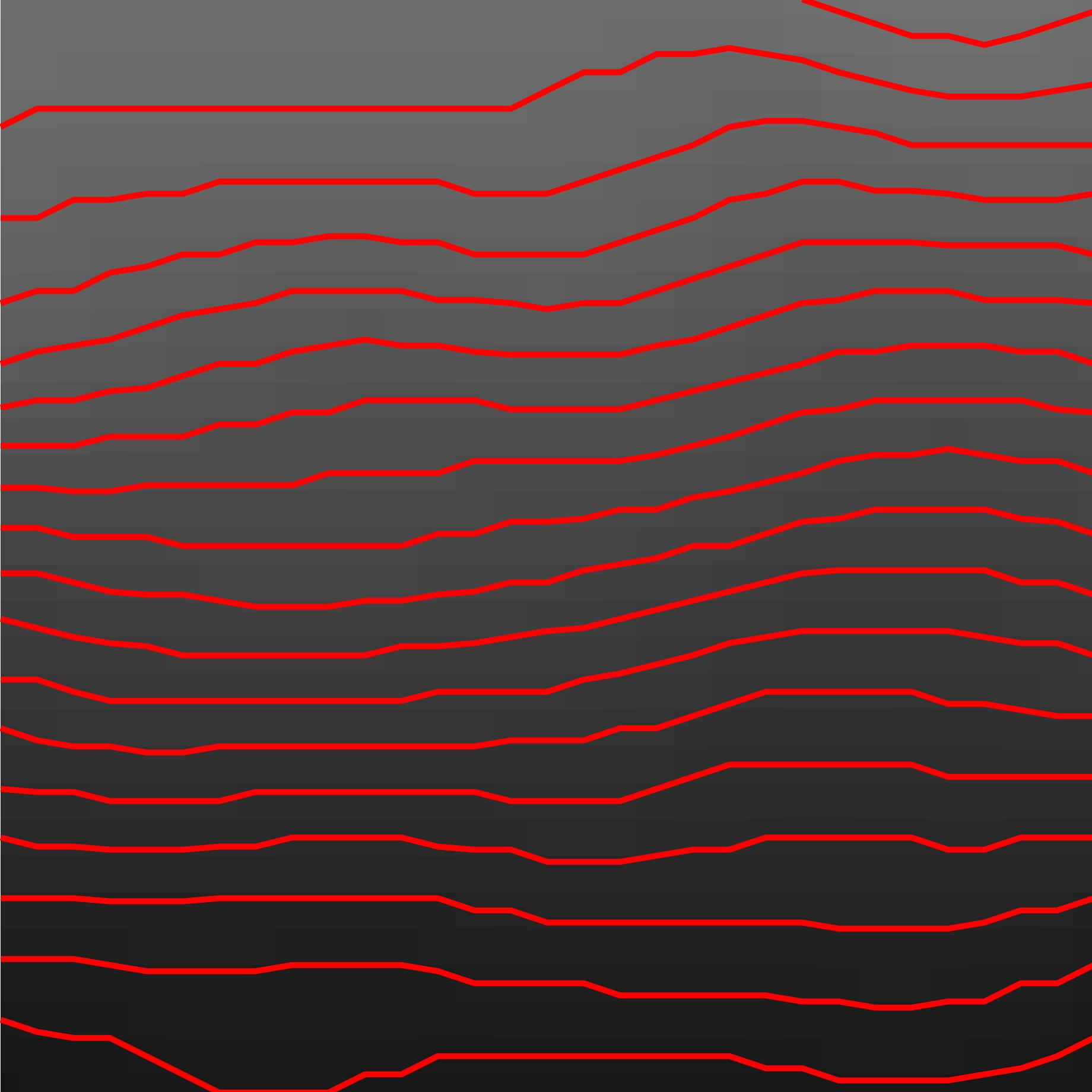}&
		\includegraphics[width=0.15\textwidth]{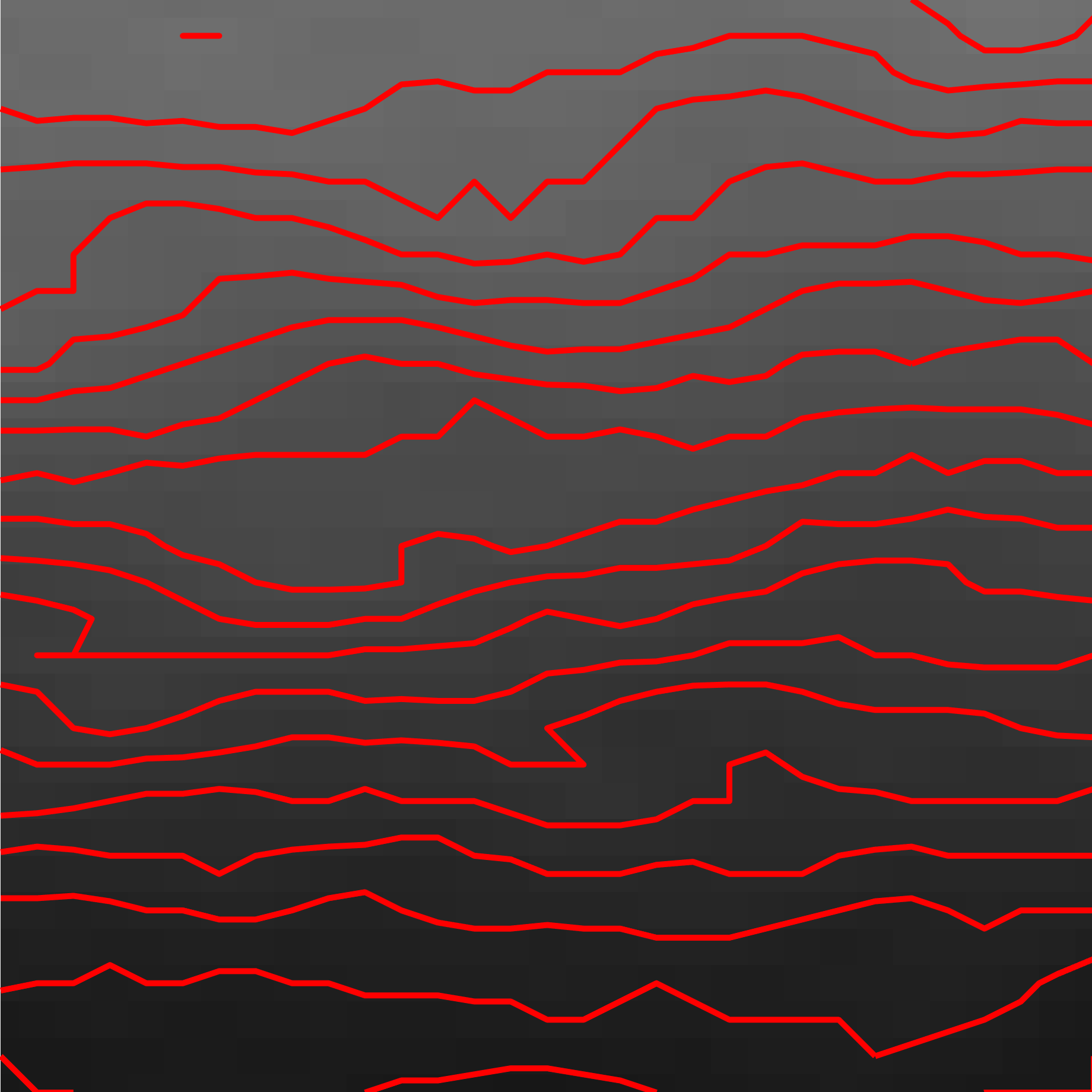}&
		\includegraphics[width=0.15\textwidth]{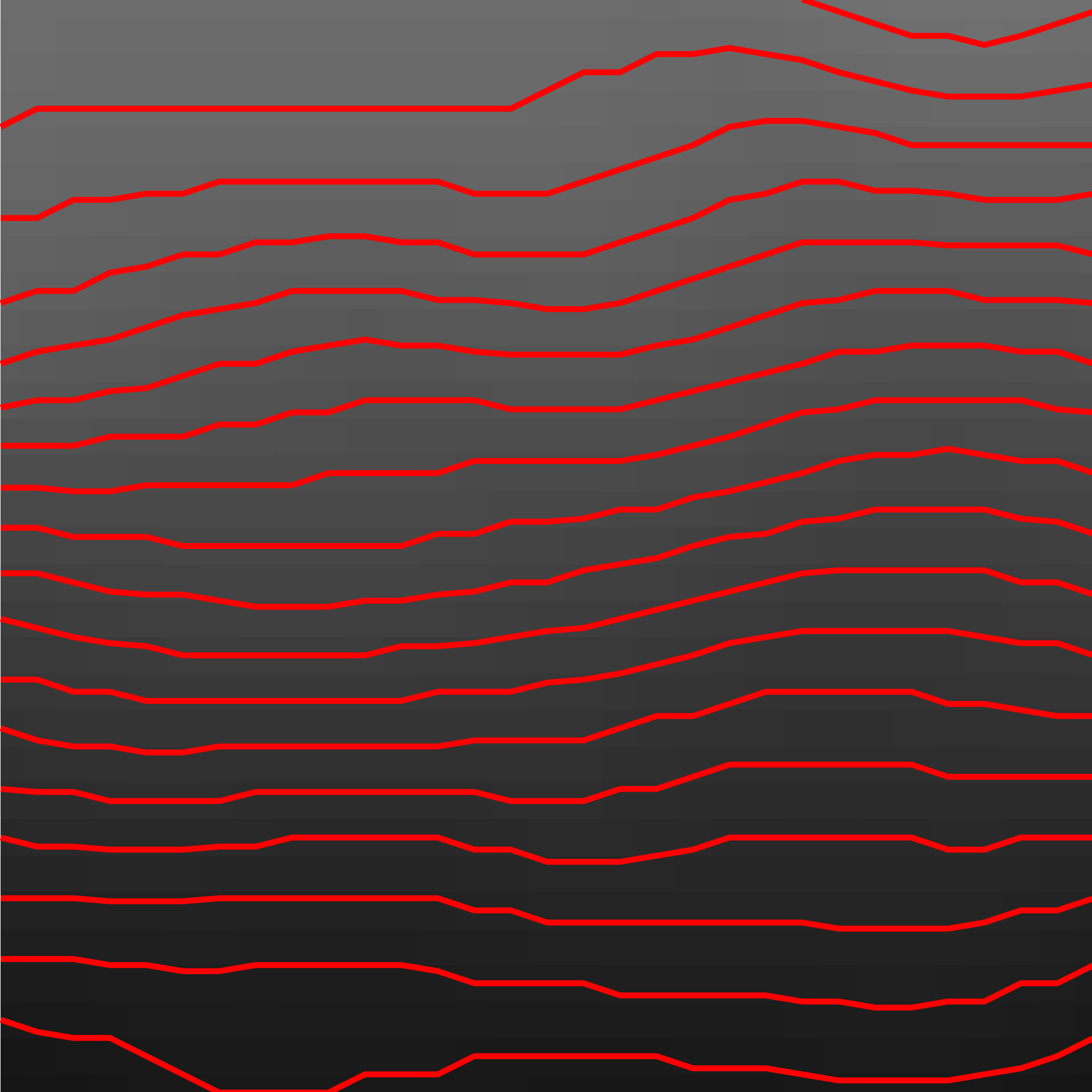}&
		\includegraphics[width=0.15\textwidth]{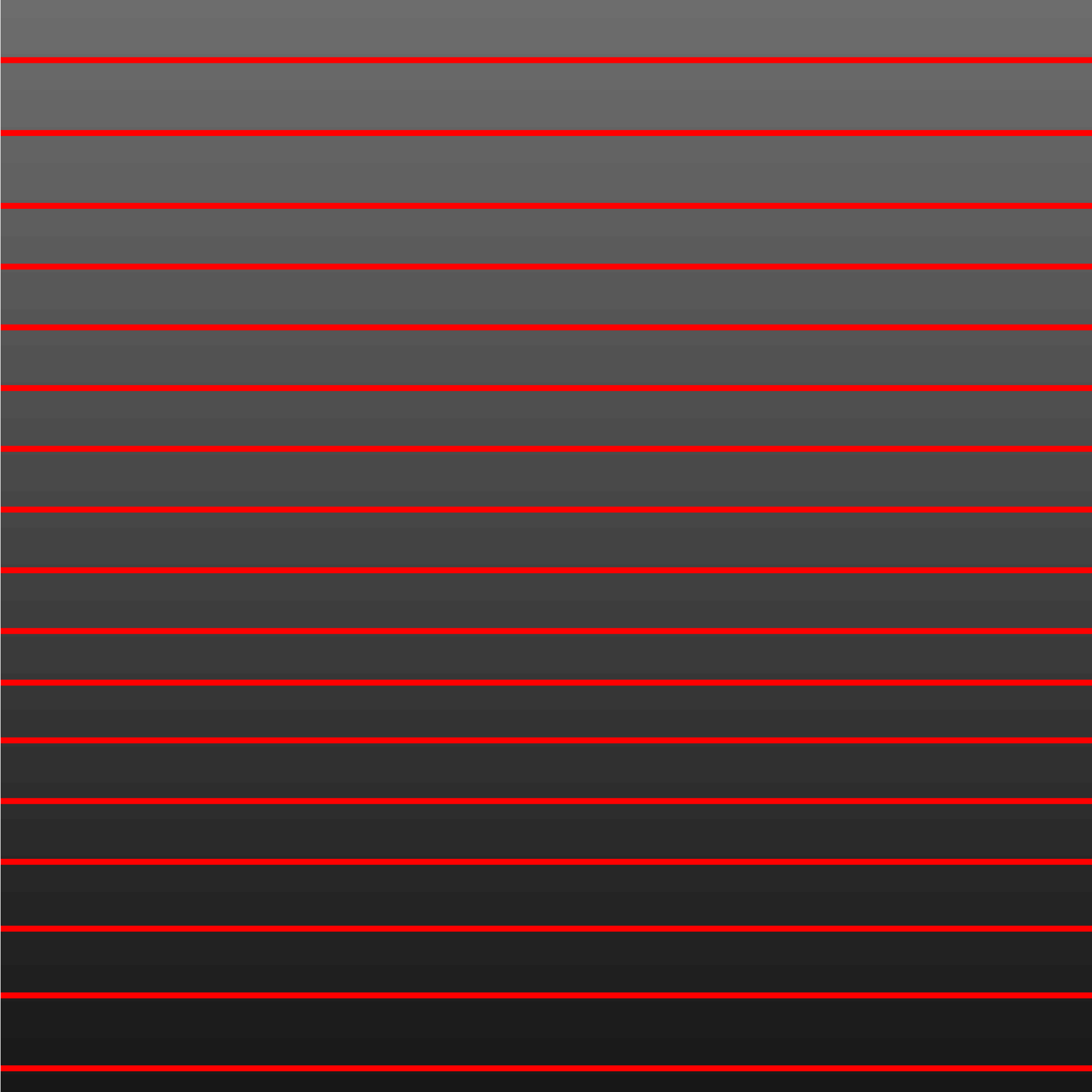}\\
		\includegraphics[width=0.15\textwidth]{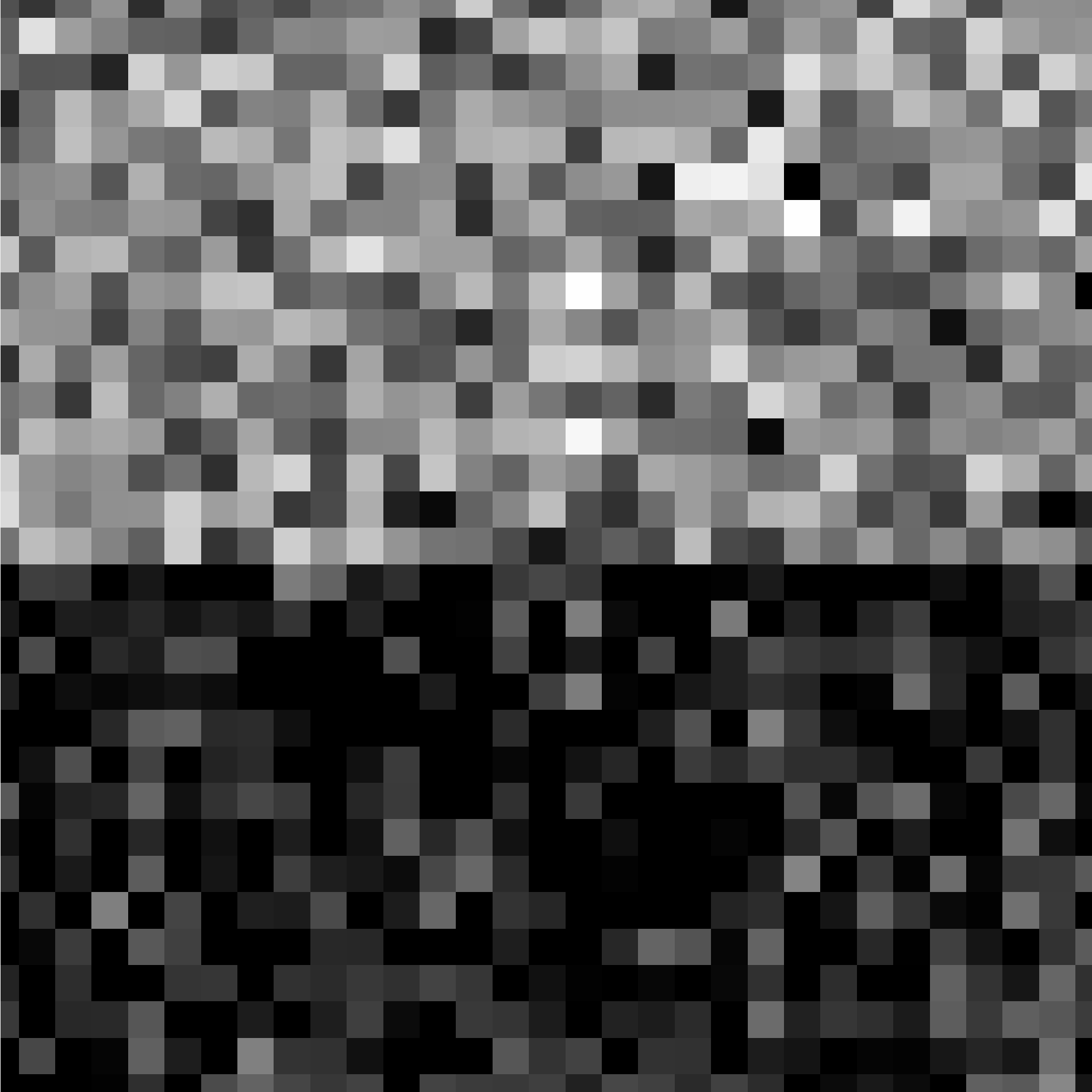}&
		\includegraphics[width=0.15\textwidth]{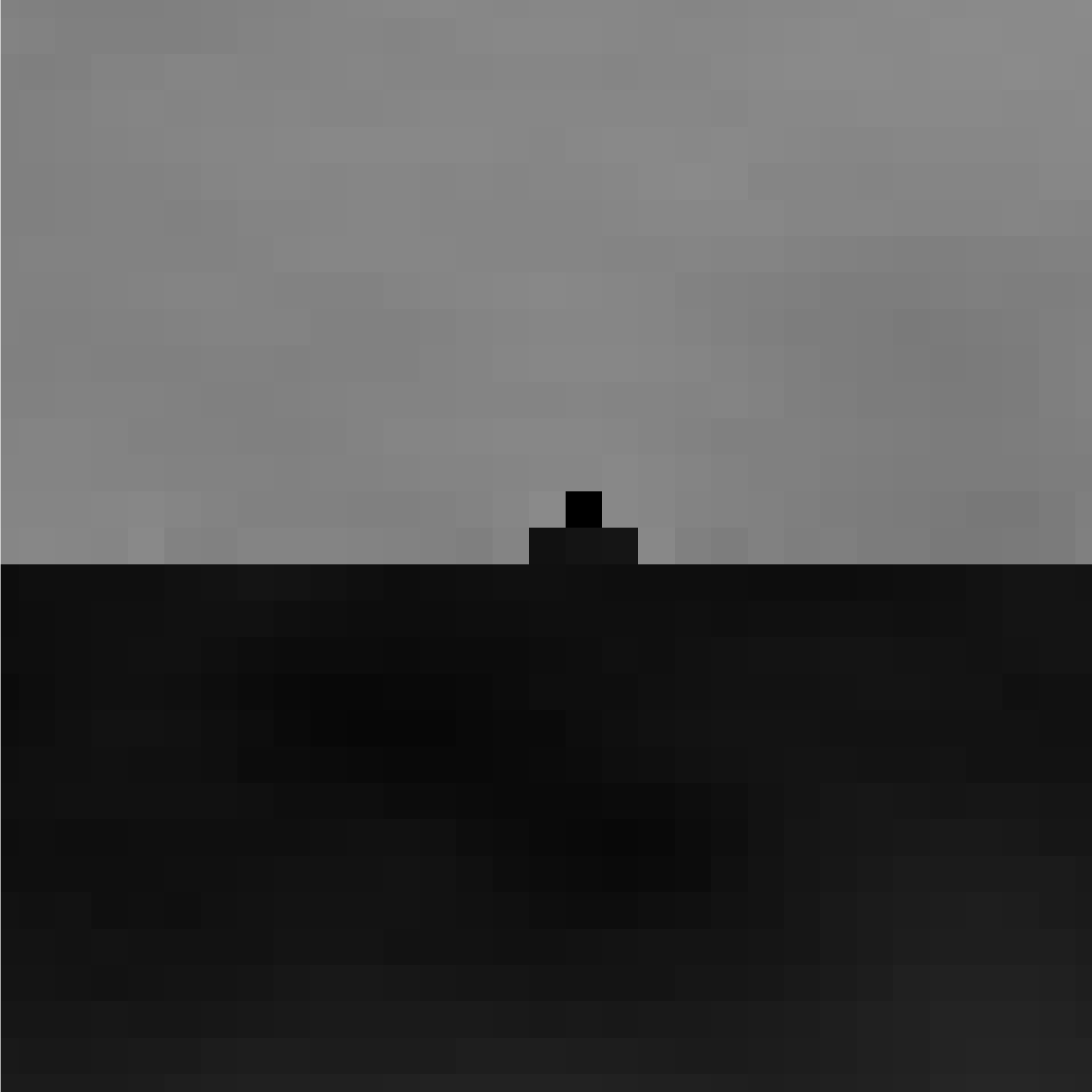}&
		\includegraphics[width=0.15\textwidth]{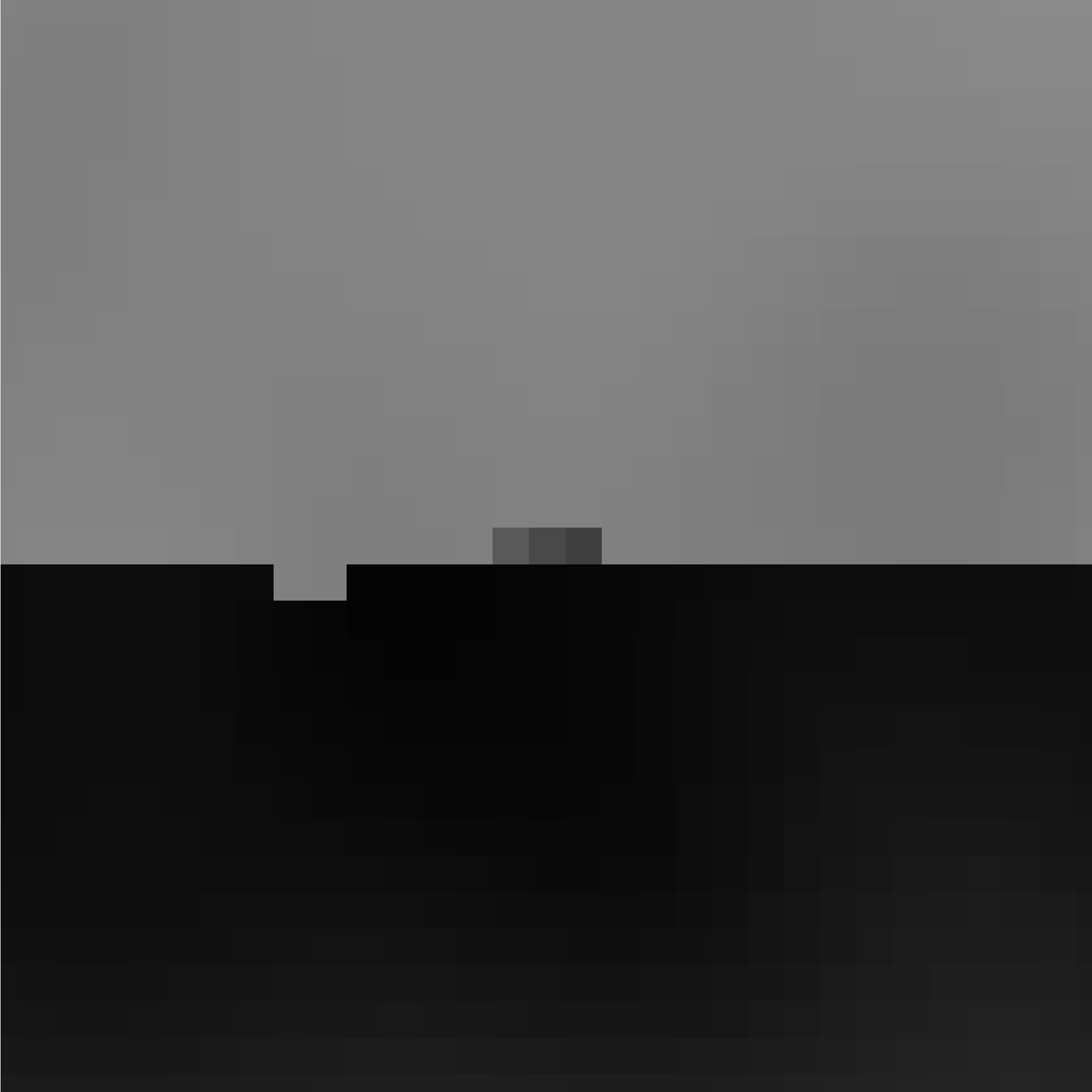}&
		\includegraphics[width=0.15\textwidth]{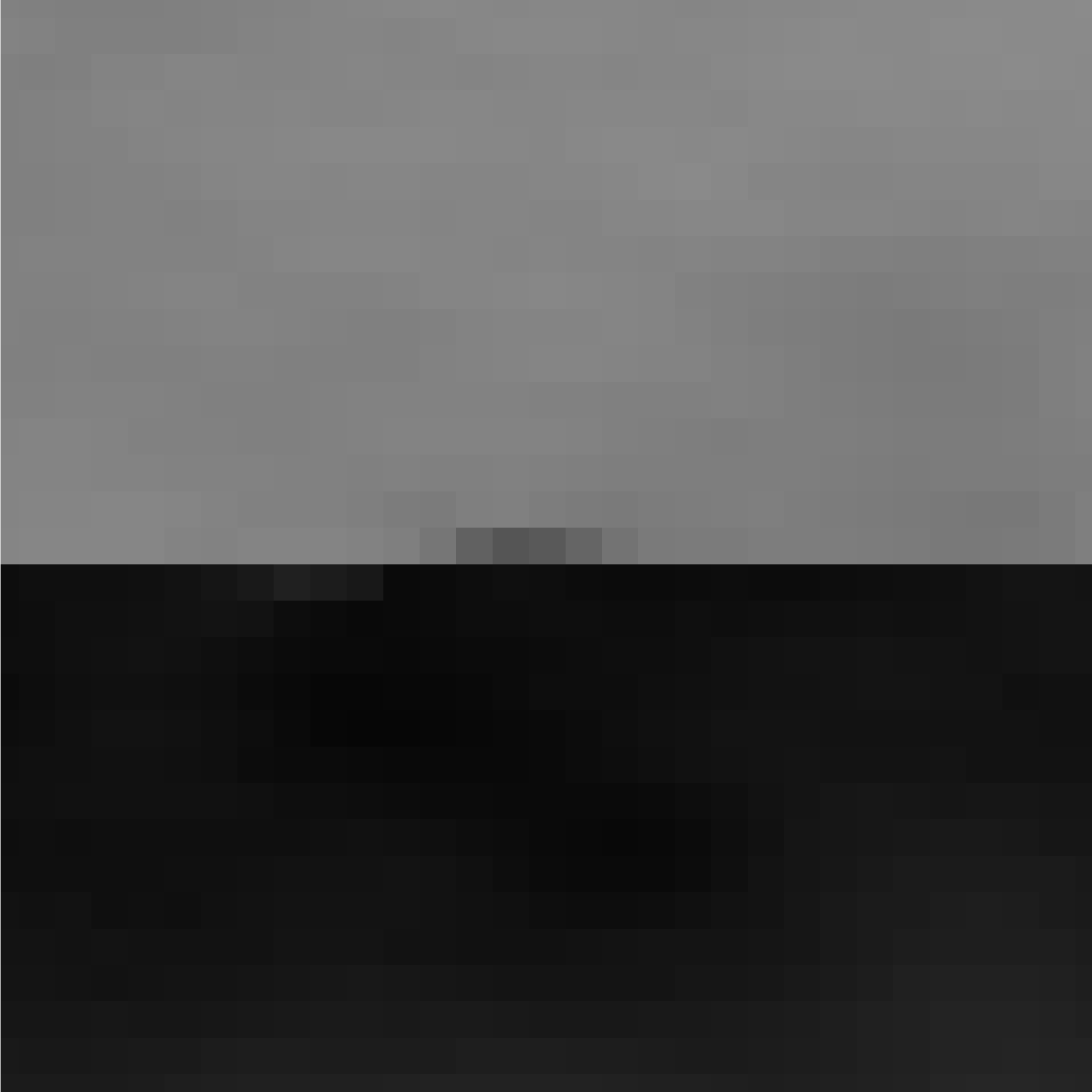}&
		\includegraphics[width=0.15\textwidth]{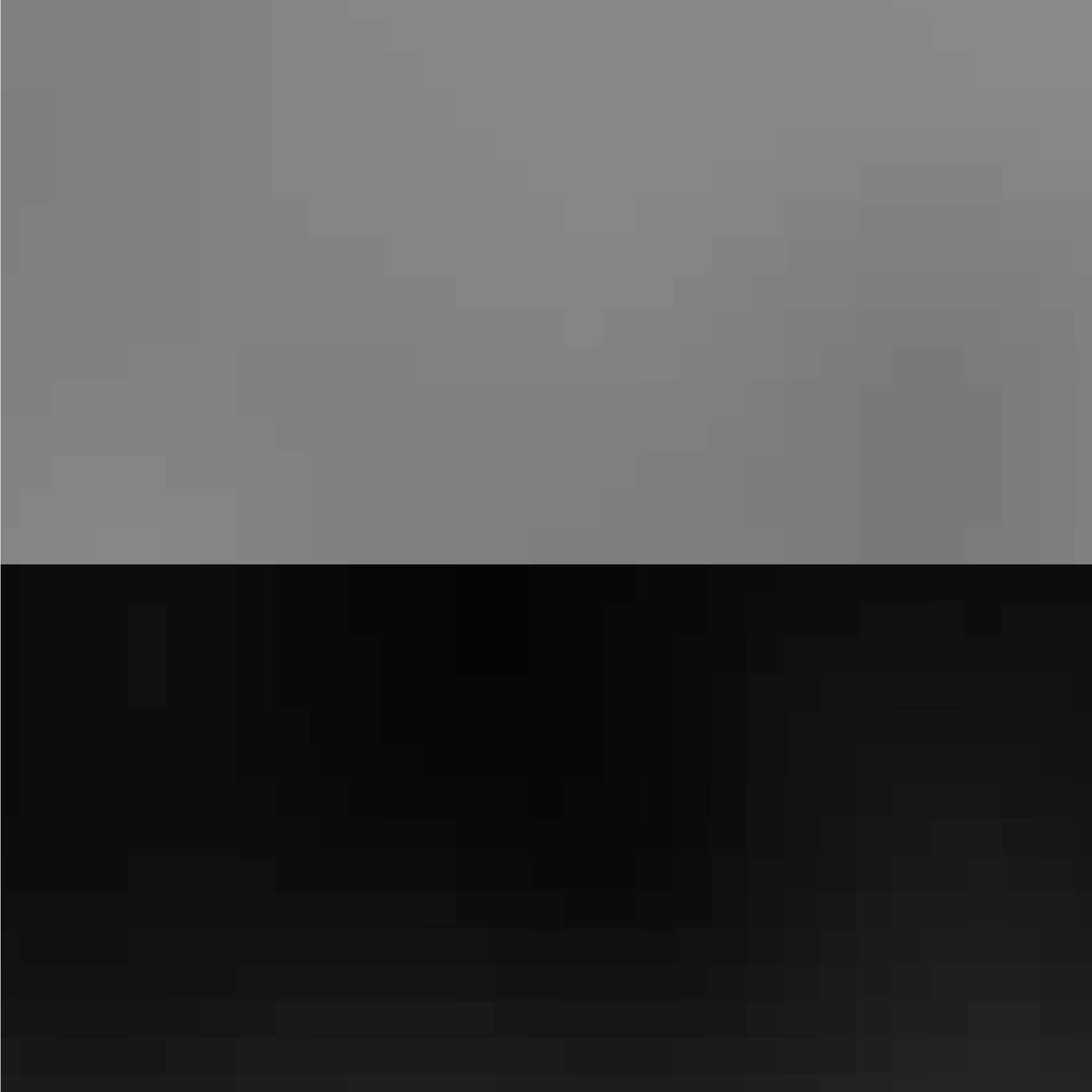}&
		\includegraphics[width=0.15\textwidth]{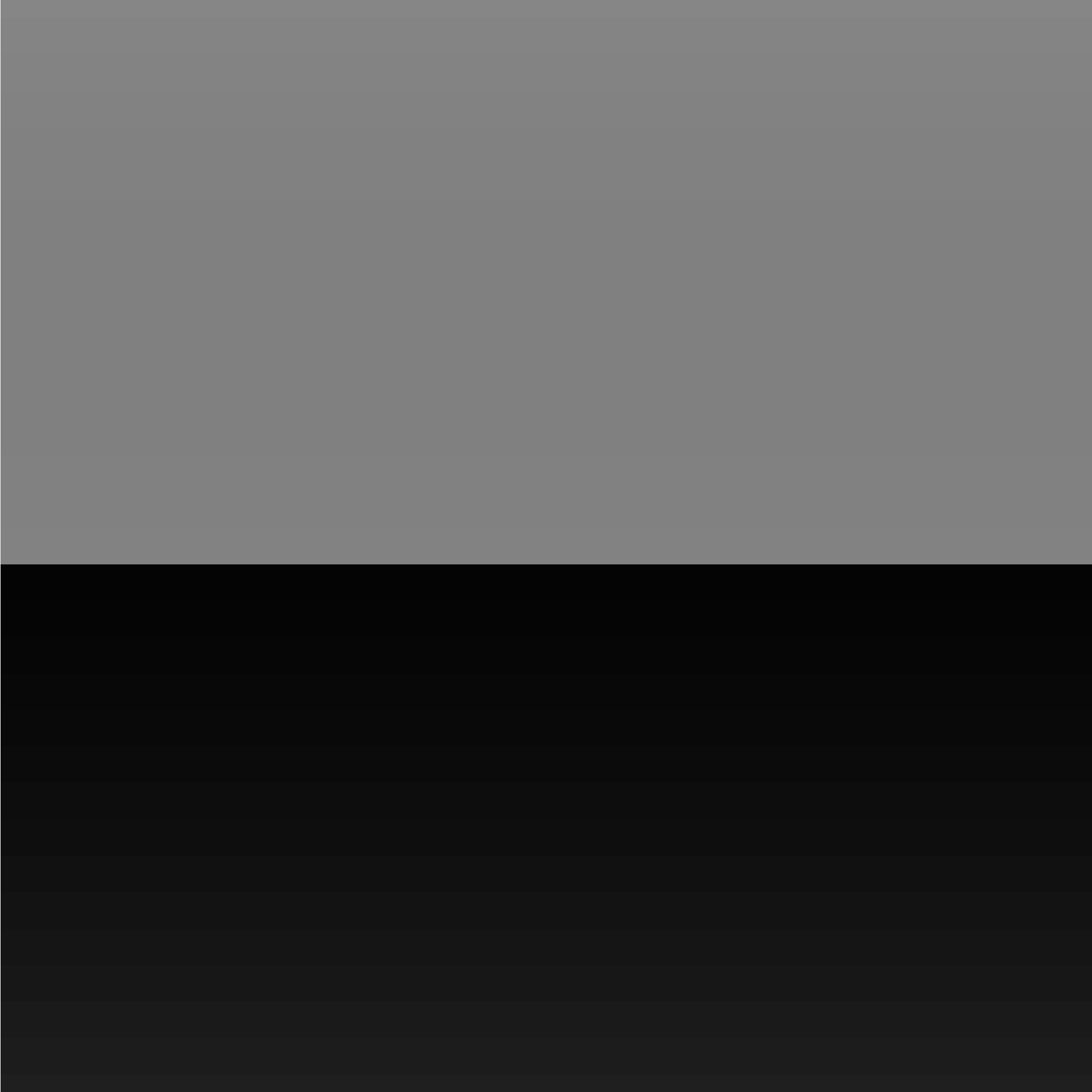}
		\\\hline
		PSNR$(u^*)$&$32.31$&$34.64$&$35.03$&$\mathbf{35.66}$\\\hline
	\end{tabular}
	\caption{Ablation study. (a) Noisy input (size: $256\times256$, $\sigma=50/255$), and regions in red boxes are zoomed-in in the subsequent rows. (b) Noise-free component $u^*$ by Model I~\eqref{eq_abl_model1}, $\alpha_0=1.2\times10^{-1}$. (c) Noise-free component $u^*$ by Model II~\eqref{eq_abl_model2}, $\alpha_0=4\times10^{-2}$, $\alpha_{w}=80$. (d) Noise-free component $u^*$ by Model III~\eqref{eq_abl_model3}, $\alpha_0=8\times10^{-2},\alpha_{\text{curv}}=2$. (e) Noise-free component $u^*$ by the proposed model, $\alpha_0=5\times10^{-2},\alpha_{w}=80,\alpha_\text{curv}=0.5$. (f) True image. For all scenarios, we fixed $\alpha_n=1\times10^{-4}$. The noise-free components are obtained by summing the identified components except for the respective oscillatory parts.}\label{fig_ablation}
\end{figure}

Figure~\ref{fig_ablation} shows the $u^*$ components learned from the various settings above. In column (a), we present a noisy grayscale image of a squared ring, and the regions inside red boxes are zoomed-in and shown in the subsequent rows. We added  Gaussian noise with intensity  $\sigma =50/255$ on the input. In  column (b), we show the $u^*$ component reconstructed by Model I~\eqref{eq_abl_model1}. As was expected, using  the regularization $\|\nabla u\|_0$ enforces strong sparsity on the gradient of reconstructed $u^*$ component and  yields sharp boundaries. However, analogous to the TV-norm regularization,  using  only $L^0$-gradient as regularization creates strong staircase effects. Moreover, $L^0$-gradient is local, yet larger-scale geometric information is necessary to render better results. The boundary in the  third row shows undesirable irregularities. The smooth region is better reconstructed in Model II, where an extra component $w^*$ is introduced. However, due to the lack of geometric information in $L^0$-gradient, the boundary lines still present zig-zags. In Model III, the curvature regularization on $u^*$ is used, which imposes geometric priors on the level lines of $v^*$. The  straight boundary lines are better preserved compared to Model I and Model II. Without the smooth part $w^*$, the intensity transition in the last row is not as smooth as in Model II. Combining the benefits of geometric regularization on the level lines and adding an extra smoothly varying component, our proposed model outperforms the others. Observe that the transition is smooth in the second row indicated by the red contour curves, and the boundary line is straight in the third row. These reconstructed features match the underlying image shown in the last column.

From this set of experiments, we conclude that $L^0$-gradient regularization induces strong sparsity on the gradient which causes significant staircase artifacts and irregular boundaries. These undesirable phenomenons are effectively mitigated by the curvature regularization on the level lines. In order to recover smoothly varying regions while preserving sharp boundaries, our experiments show that adding an extra smooth part in the decomposition is important.

\subsection{Effect of  parameter $\alpha_{w}$ and $\alpha_{\text{curv}}$}\label{sec_effect_alphaw}
In our proposed model, both $v^*$ and $w^*$ favor regularity. Particularly, $v^*$ prefers shapes with definite and regular boundaries; whereas $w^*$ tends to be blurry.  We demonstrate that image content can be distributed and converted between these two components, yielding different decomposition results. 

We first study the effect of the model parameter $\alpha_{w}$ associated with the term $\int_\Omega|\Delta w|^2\,d\bx$ that encourages isotropic smoothness of the $w^*$ component. Intuitively, larger values of $\alpha_{w}$ induce smoother $w^*$ where image details are smeared. If $\alpha_{w}$ is too large, regions with smoothly varying intensity and sufficiently regular level lines can be captured by the $v^*$ component.

\begin{figure}[t!]
	\begin{tabular}{c|c@{\hspace{1pt}}c@{\hspace{1pt}}c@{\hspace{1pt}}c@{\hspace{1pt}}c}
		$\alpha_{w}$&10&50&100&500&1000\\\hline
		$v_{\text{scaled}}^*$&
		\includegraphics[align=c,width=0.15\textwidth]{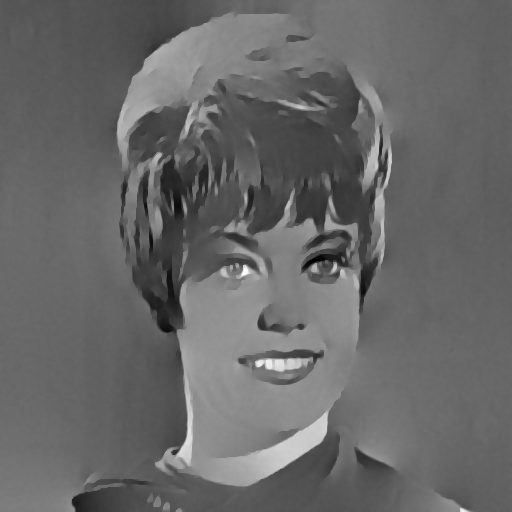}&
		\includegraphics[align=c,width=0.15\textwidth]{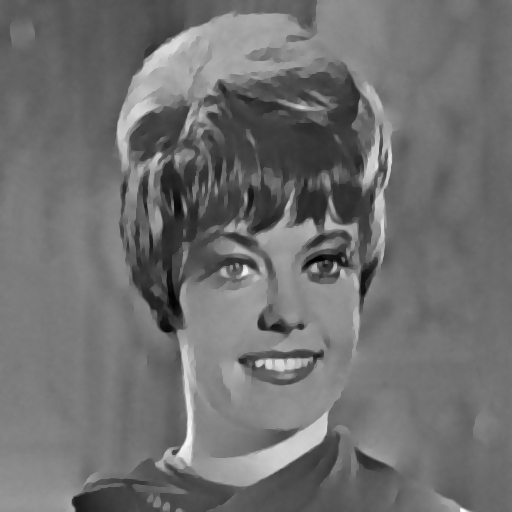}&
		\includegraphics[align=c,width=0.15\textwidth]{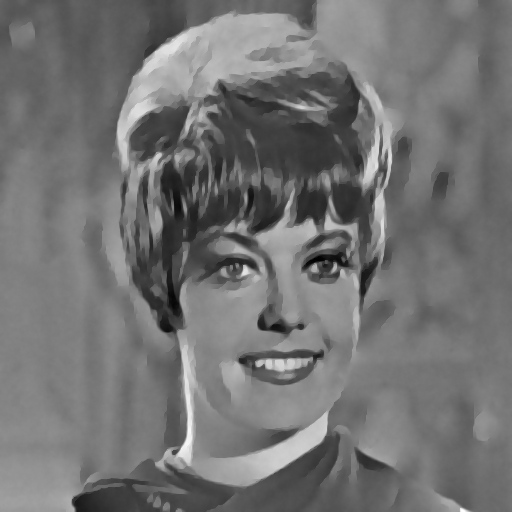}&
		\includegraphics[align=c,width=0.15\textwidth]{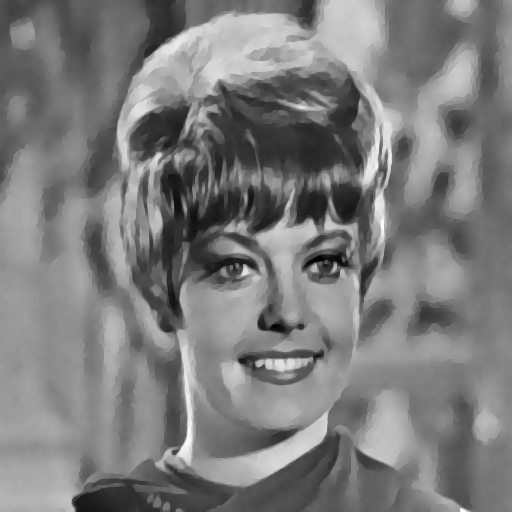}&
		\includegraphics[align=c,width=0.15\textwidth]{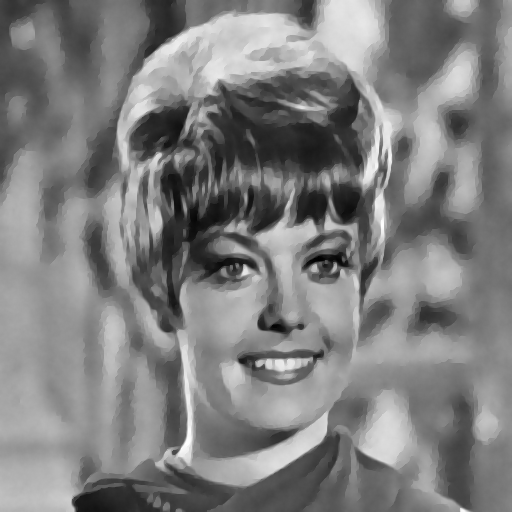}
		\\
		$w_{\text{scaled}}^*$&
		\includegraphics[align=c,width=0.15\textwidth]{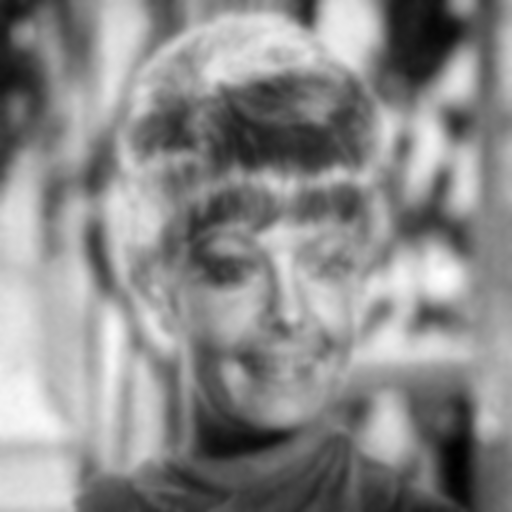}&
		\includegraphics[align=c,width=0.15\textwidth]{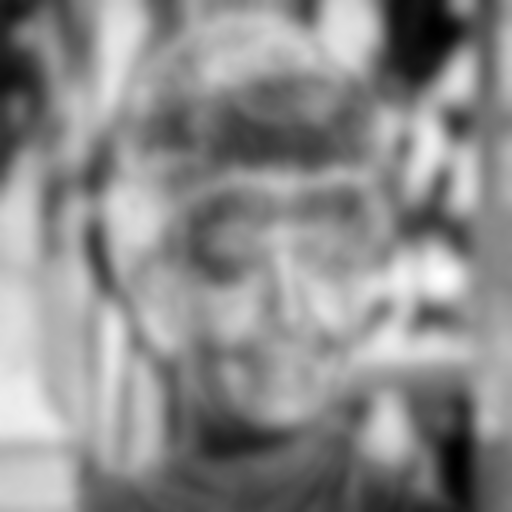}&
		\includegraphics[align=c,width=0.15\textwidth]{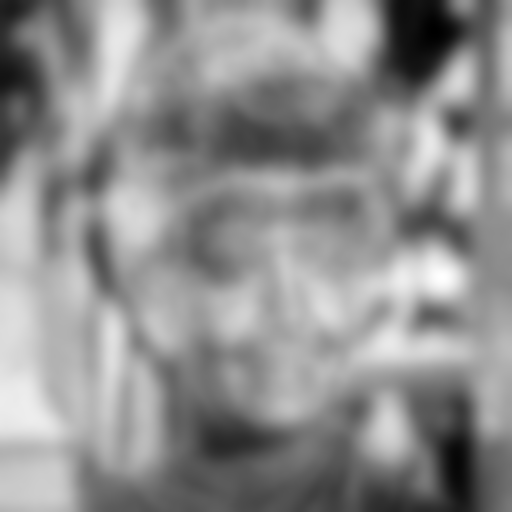}&
		\includegraphics[align=c,width=0.15\textwidth]{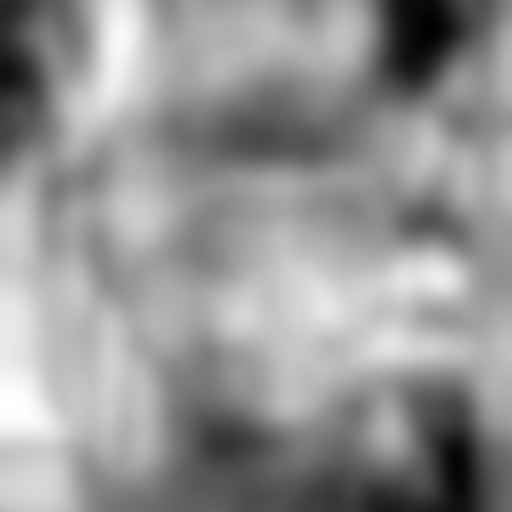}&
		\includegraphics[align=c,width=0.15\textwidth]{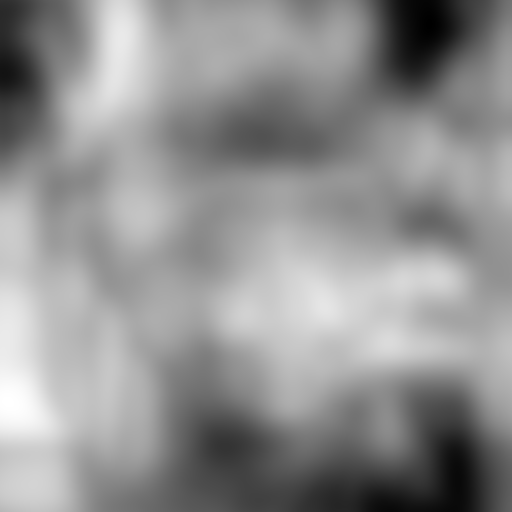}
		\\
		$u^*$&
		\includegraphics[align=c,width=0.15\textwidth]{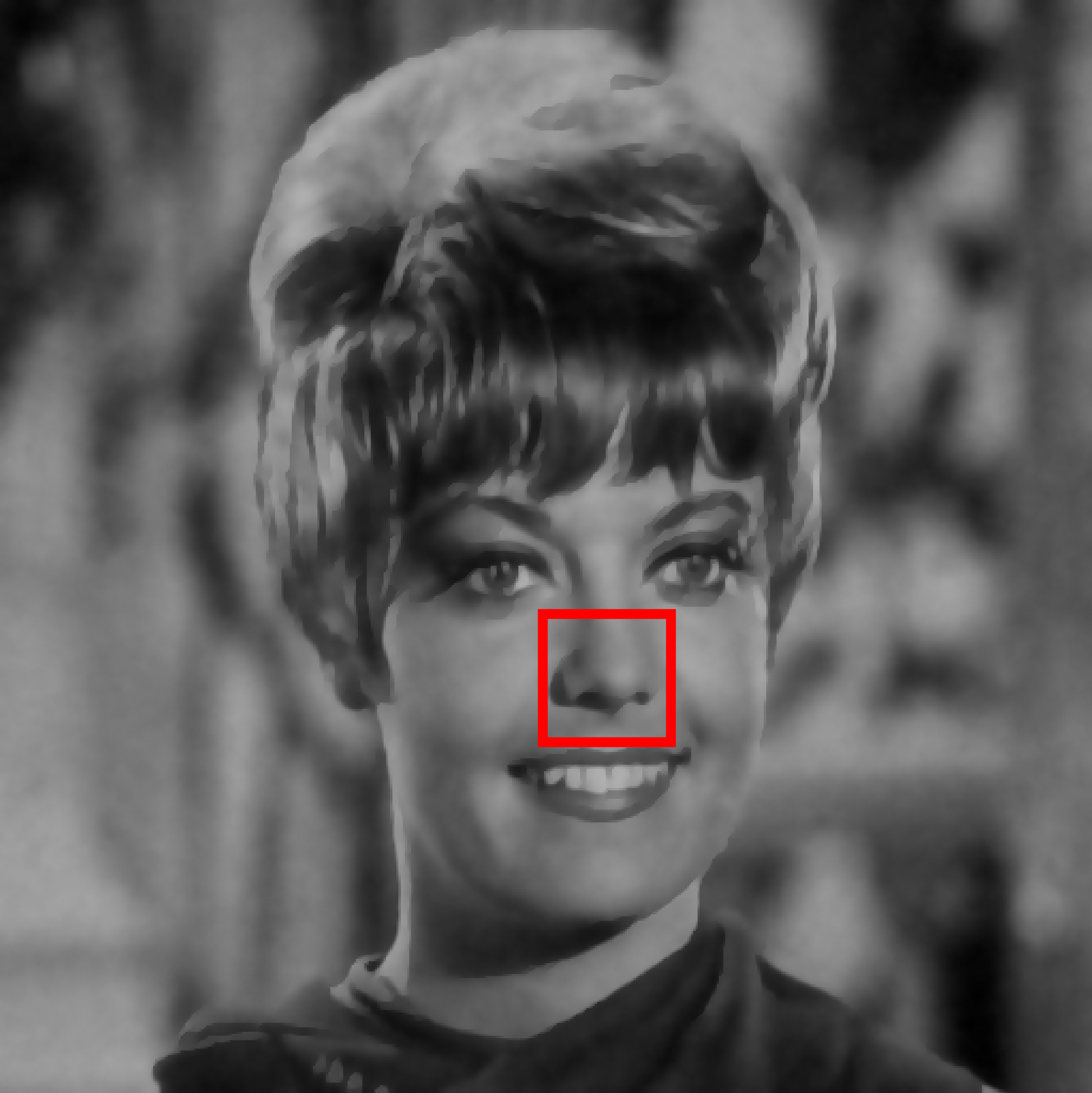}&
		\includegraphics[align=c,width=0.15\textwidth]{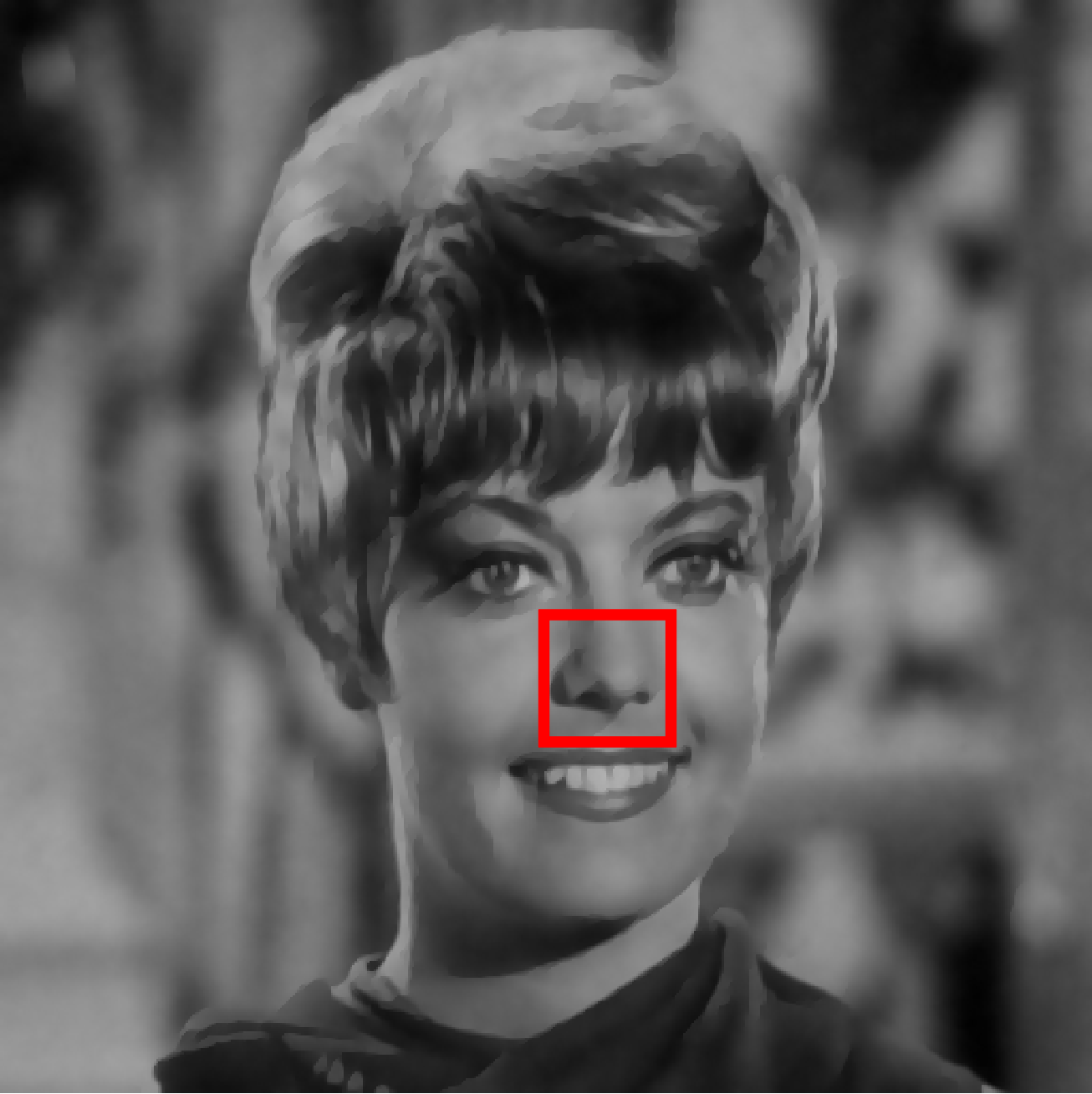}&
		\includegraphics[align=c,width=0.15\textwidth]{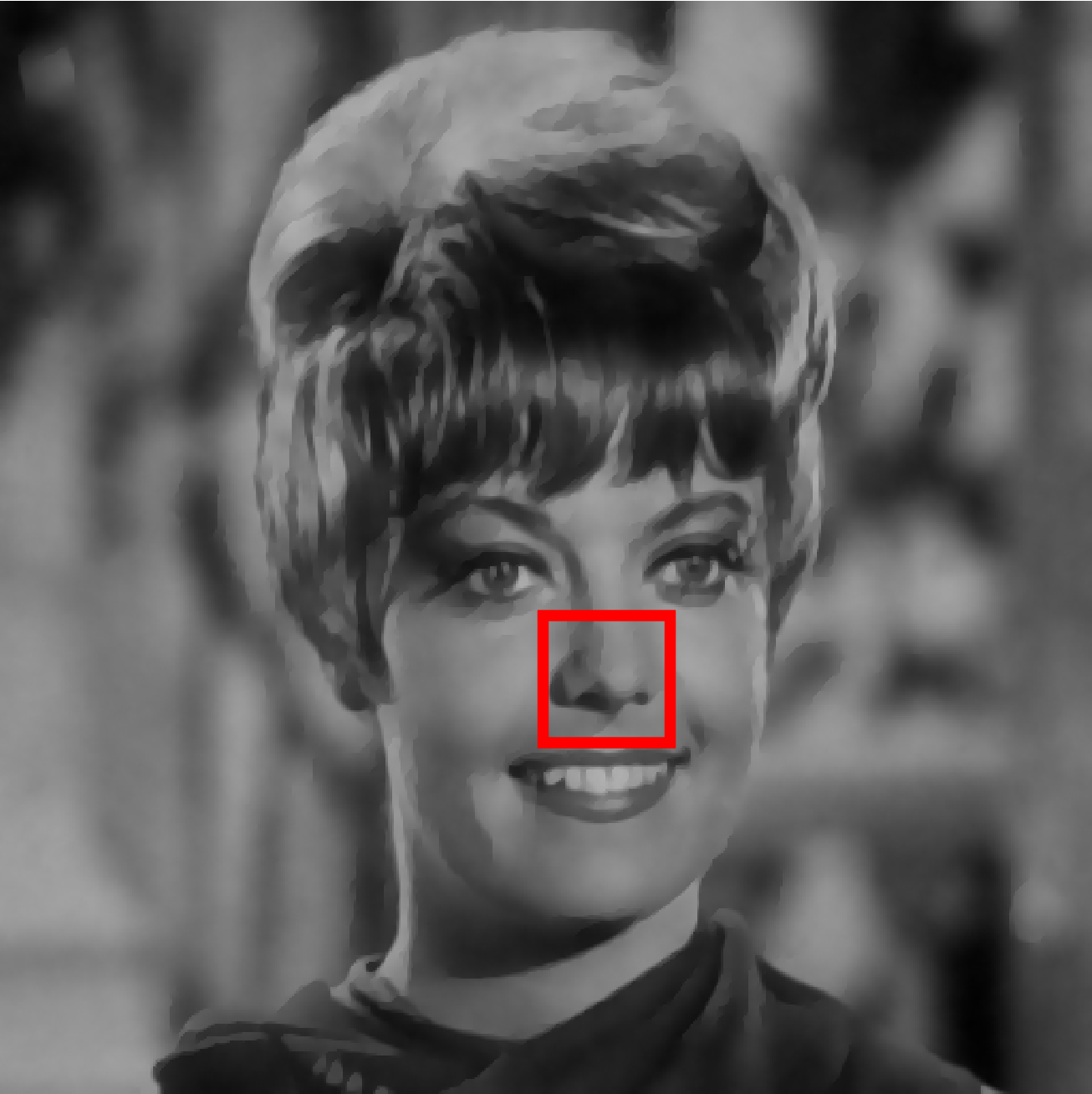}&
		\includegraphics[align=c,width=0.15\textwidth]{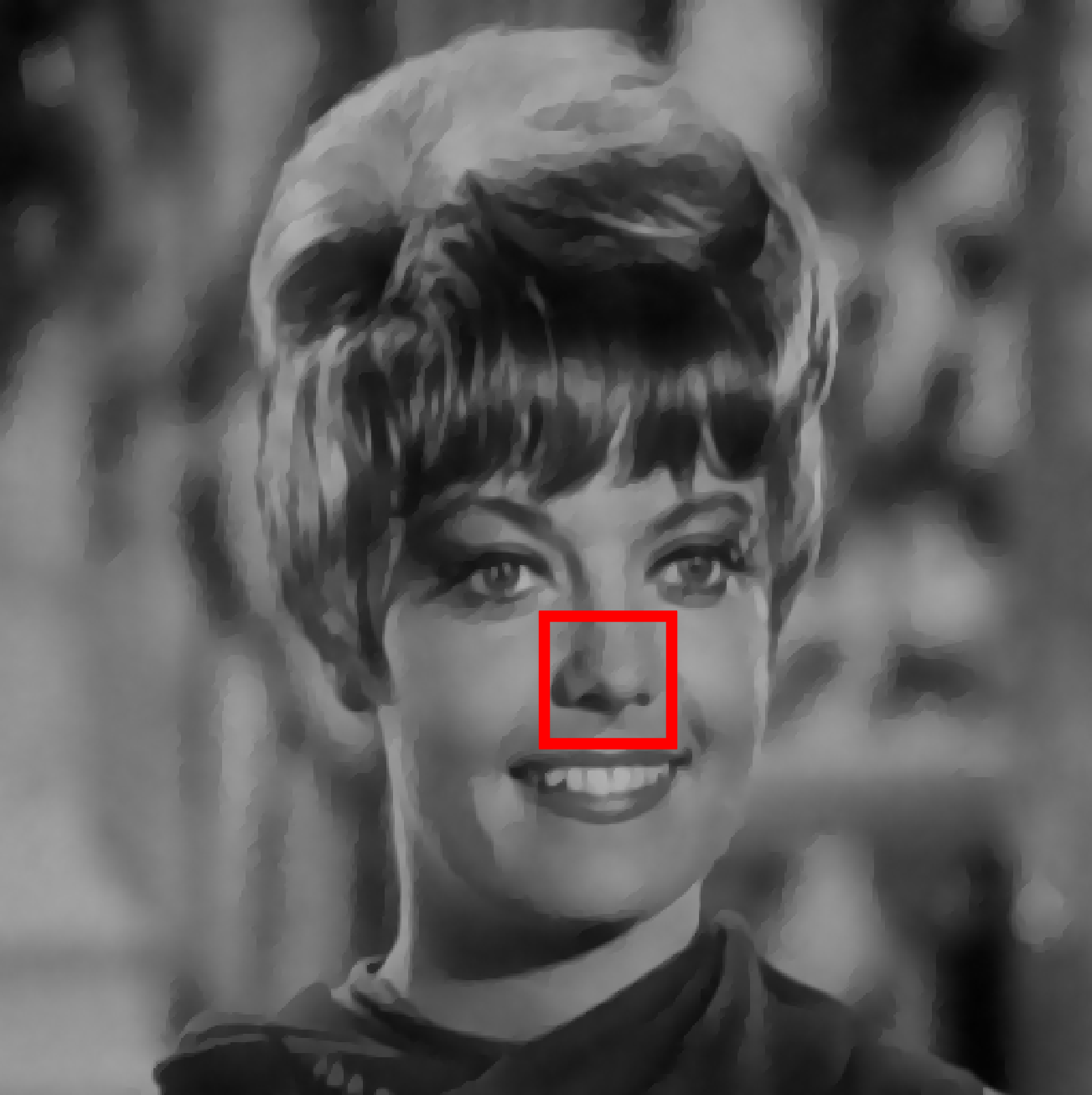}&
		\includegraphics[align=c,width=0.15\textwidth]{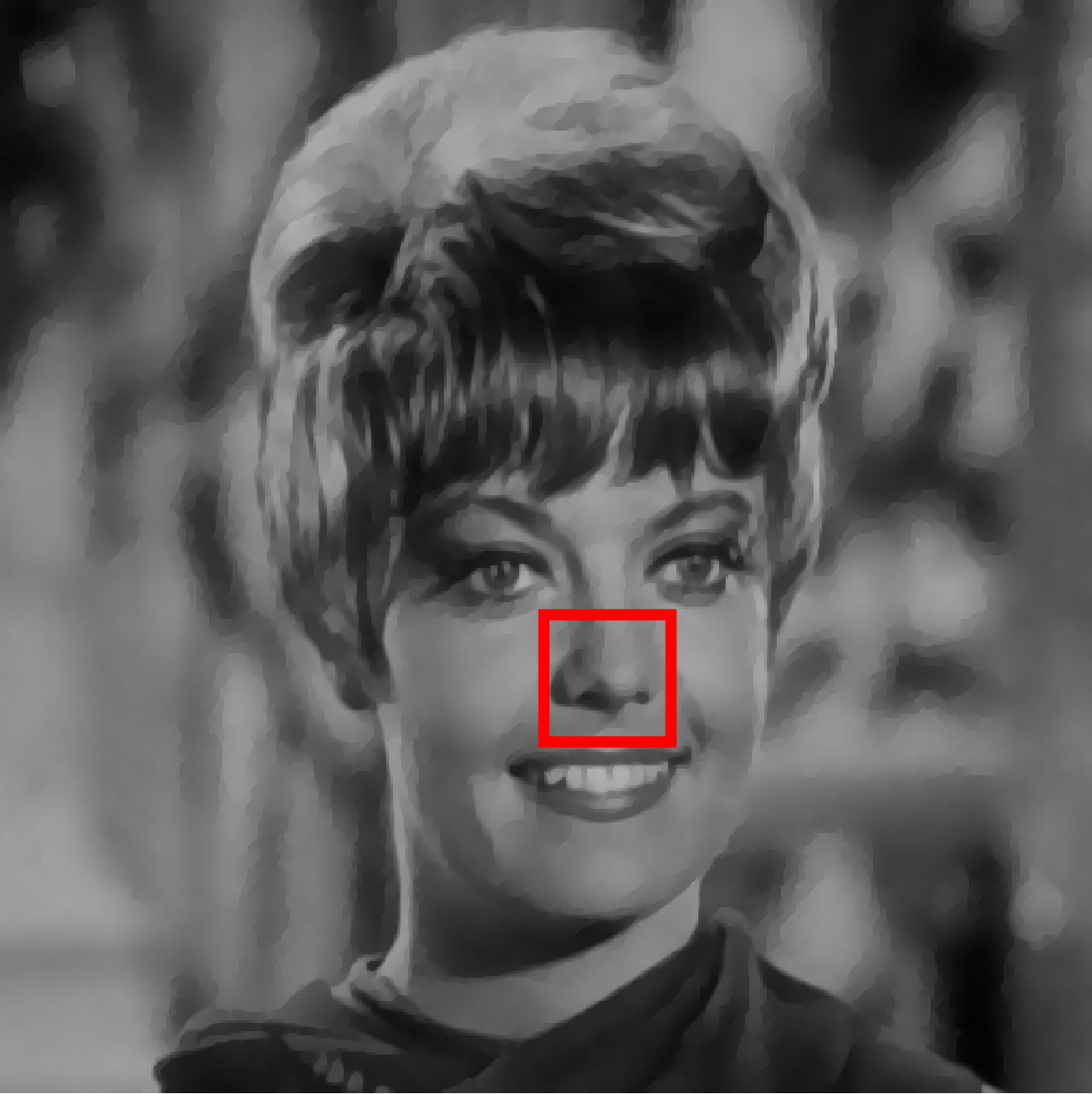}\\
		Zoom-in $u^*$&
		\includegraphics[align=c,width=0.15\textwidth]{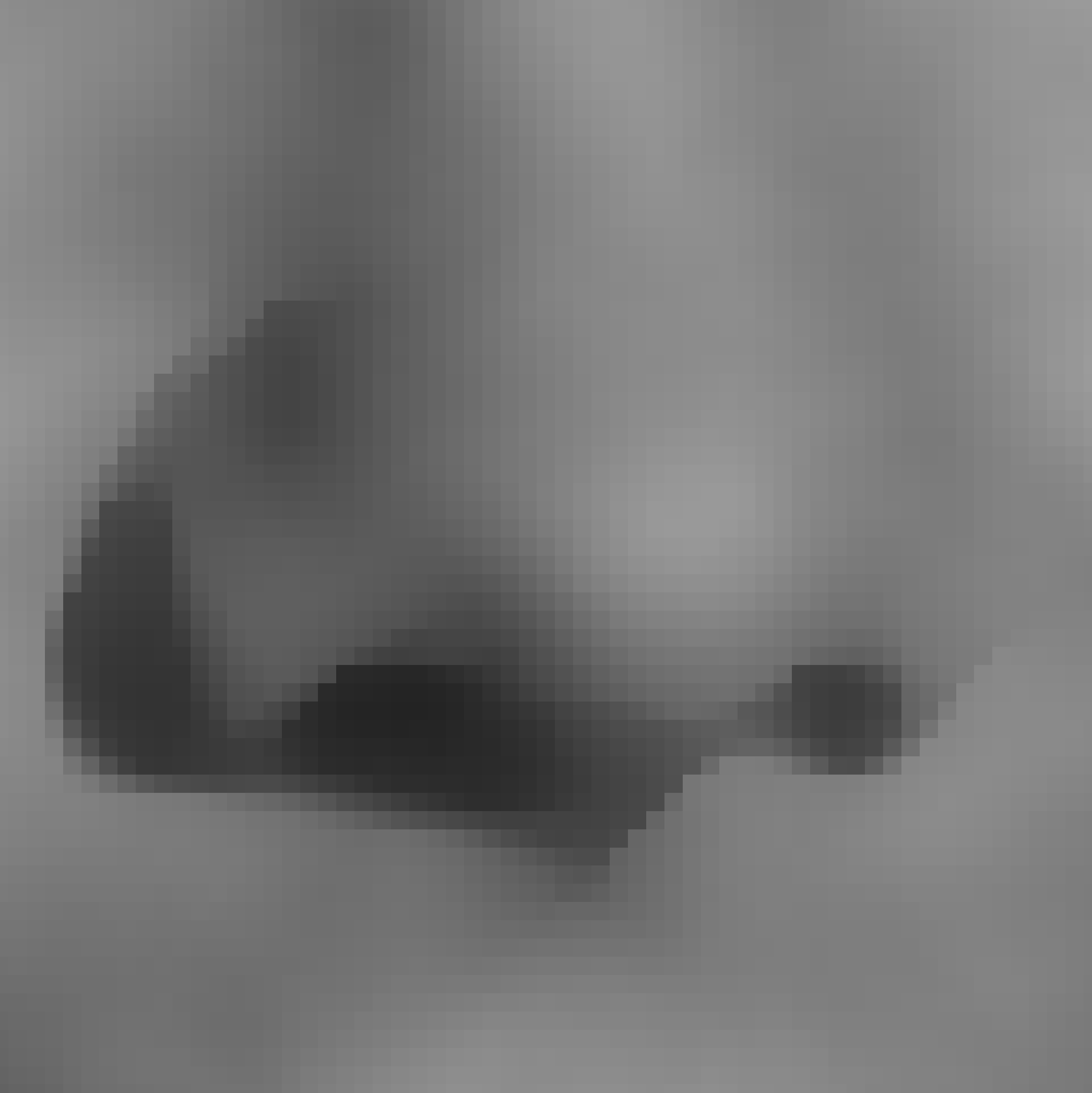}&
		\includegraphics[align=c,width=0.15\textwidth]{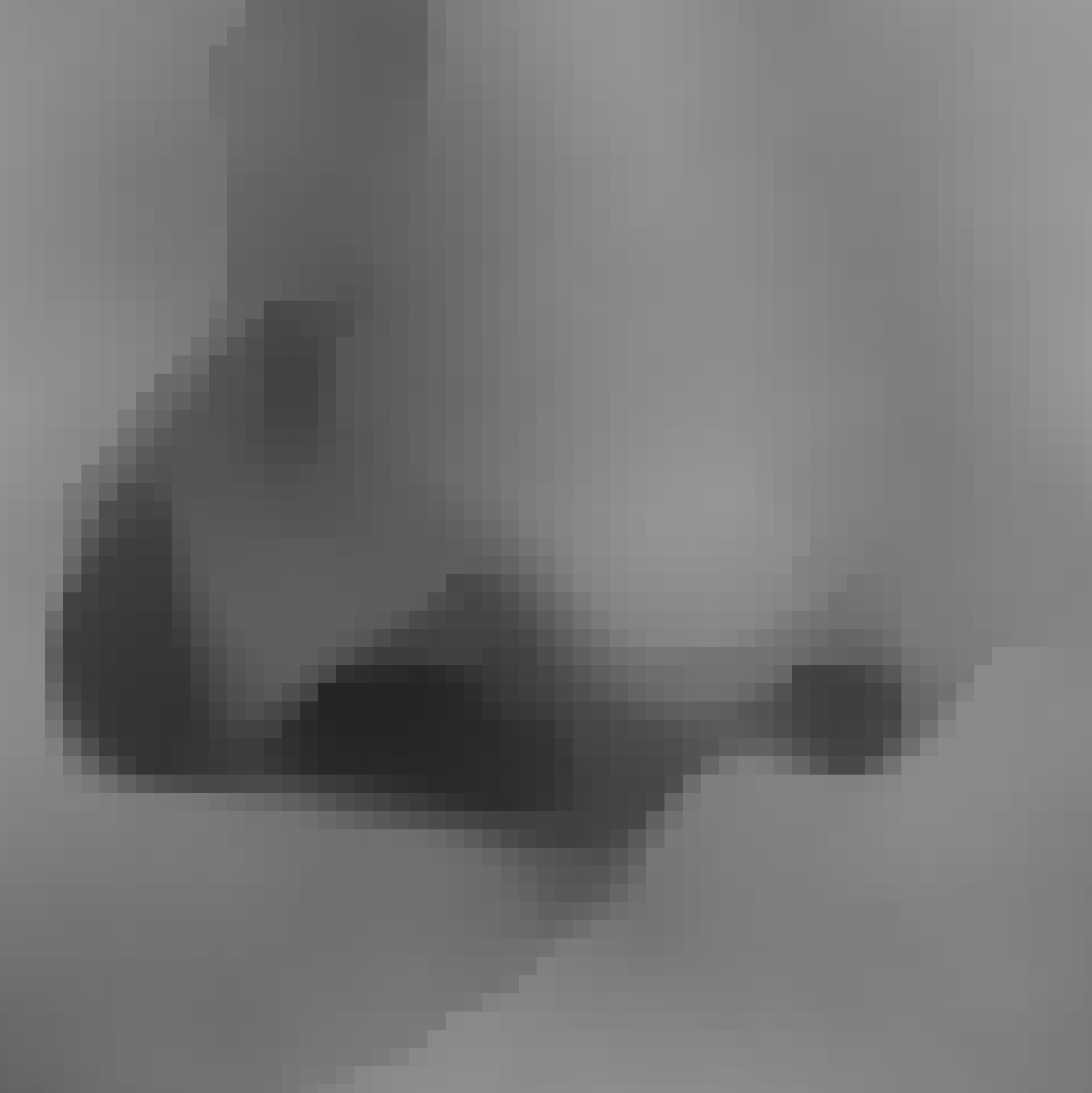}&
		\includegraphics[align=c,width=0.15\textwidth]{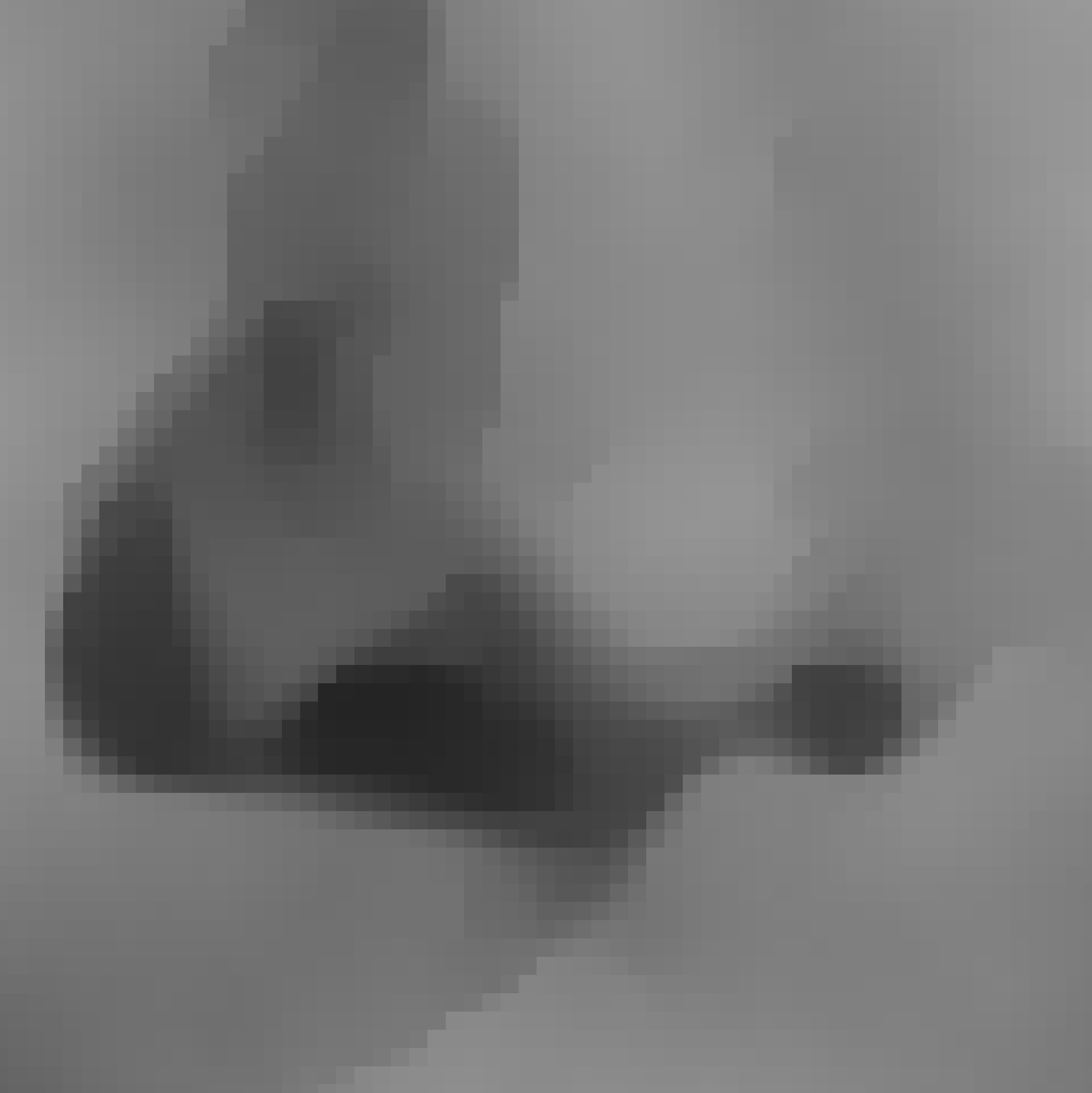}&
		\includegraphics[align=c,width=0.15\textwidth]{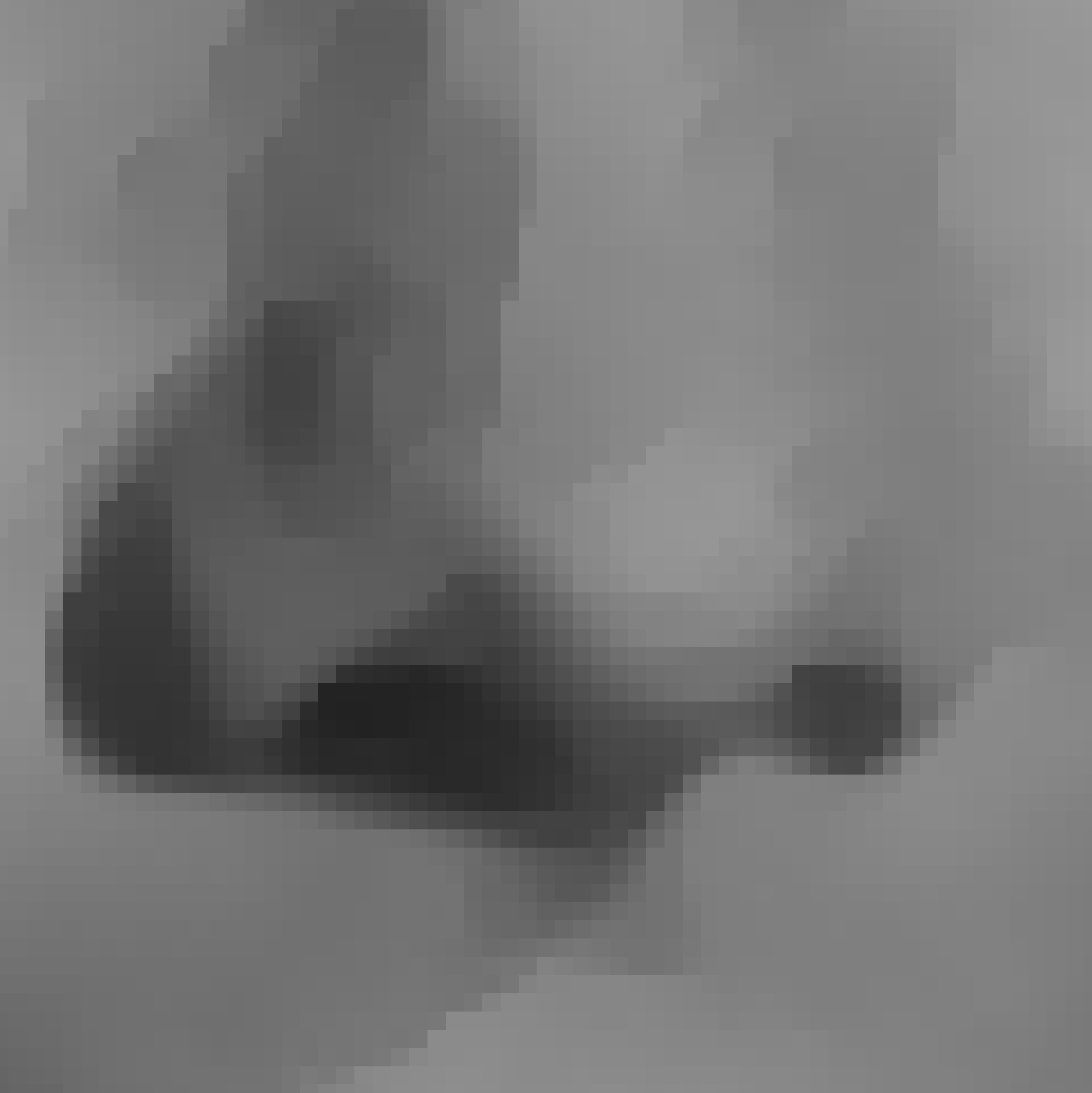}&
		\includegraphics[align=c,width=0.15\textwidth]{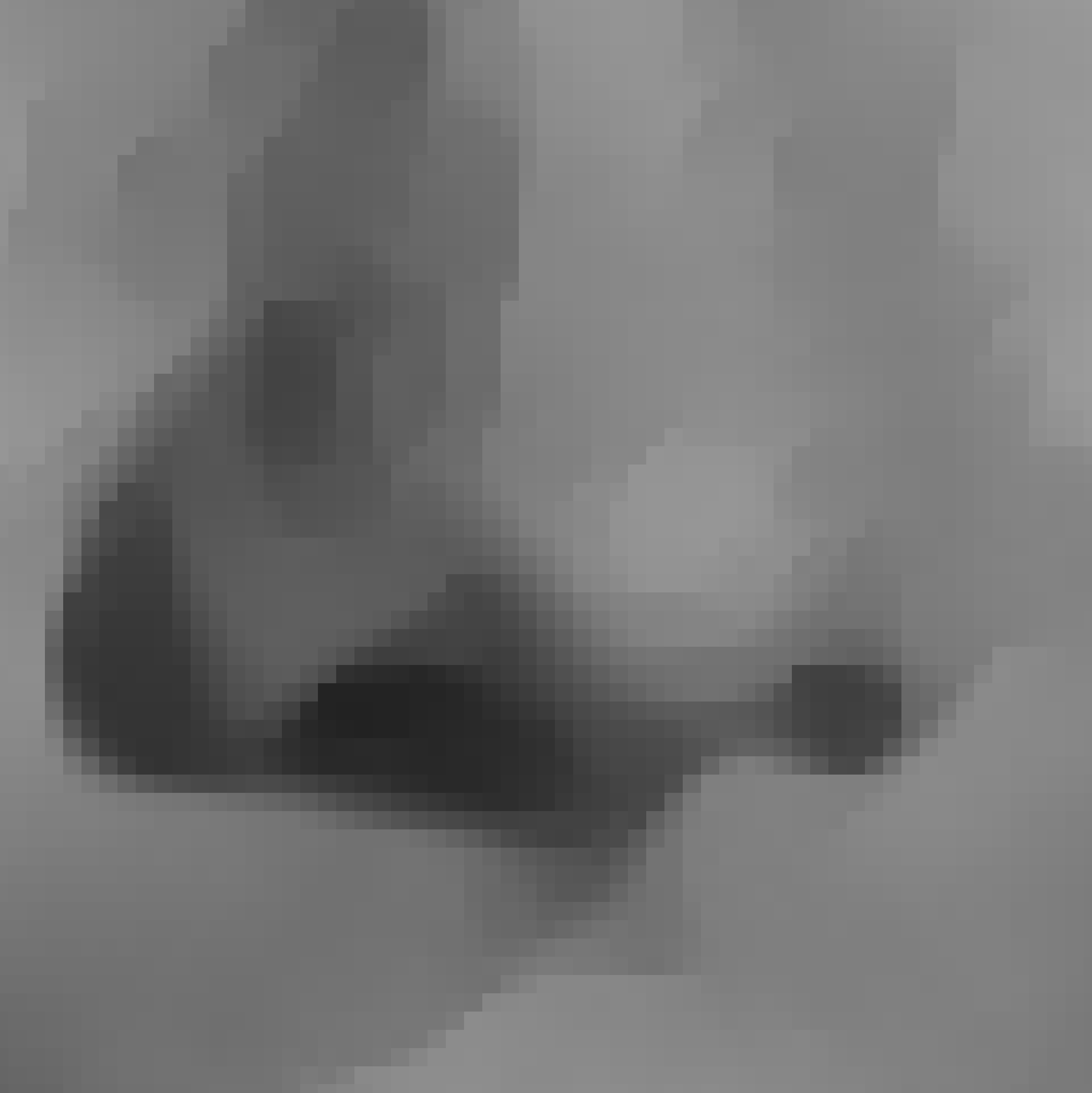}
		\\\hline
		PSNR($u^*$)&$34.58$&$$34.59$$&$34.53$&$34.34$&$34.28$\\\hline
	\end{tabular}
	\caption{Effect of $\alpha_{w}$. When $\alpha_{w}$ increases,  smooth intensity variations in $w^*$ (second row) become structural shapes in $v^*$ (first row): $v^*$ contains richer details, e.g., the shades in the background, and $w^*$ is more blurry. Overall, intensity transitions in $u^*$ (third row) become sharper, which is further illustrated by the zoom-ins of the red box (fourth row). The input image \textit{Zelda} is added with Gaussian noise ($\sigma=10/255$). Other parameters are fixed: $\alpha_0=2\times10^{-3}$, $\alpha_\text{curv}=0.5$, $\alpha_n = 10$. }\label{fig_alpha2}
\end{figure}

We illustrate this conversion between $w^*$ and $v^*$ affected by $\alpha_{w}$ in Figure~\ref{fig_alpha2}. Given a noisy image ($\sigma=10/255$), we apply the proposed model with $\alpha_{w}$ increasing from $10$ to $1000$ while fixing the other parameters. For each value of $\alpha_{w}$, we show the scaled $v^*$ and $w^*$ components in the first and second row, respectively, and in the third row, we show the respective sum $u^*=v^*+w^*$. Observe that with a wide range of values of $\alpha_{w}$, the noise-free reconstruction $u^*$ remains visually similar, and the PSNR values are stable. When $\alpha_{w}$ is small, most of the intensity variation is captured by $w^*$, while $v^*$ only captures objects' large-scale contours. Although hair, face, eyes, and many recognizable shapes are visible in $w^*$, these distinct parts are blended with each other smoothly without sharp boundaries. As $\alpha_{w}$ increases, we see that $w^*$ becomes more blurry, and $v^*$ shows richer intensity variations such as those in the background. Note that these smoothly varying components in $v^*$ have definite boundaries. For example, the zoom-ins of the red boxed region in the fourth row show that intensity transitions become sharper when $\alpha_w$ increases. Other softer smooth parts without distinctive boundaries, such as the shades on the cheeks and chin, are captured by the $w^*$ component.   

Second, we show the effect of the parameter $\alpha_\text{curv}$ which is closely related to the regularity of the level lines in the $v^*$ component. Similar to the previous study on the effect of $\alpha_{w}$, by changing $\alpha_\text{curv}$, smoothly varying components can transfer between $w^*$ and $v^*$.  In particular, if $\alpha_\text{curv}$ is large, certain regions with smooth intensity variation can be pushed into $w^*$ for lower total loss.

\begin{figure}[t!]
	\centering
	\begin{tabular}{c|c@{\hspace{1pt}}c@{\hspace{1pt}}c@{\hspace{1pt}}c@{\hspace{1pt}}c}
		$\alpha_{\text{curv}}$&0.01&0.1&1&5&10\\\hline
		$v_{\text{scaled}}^*$&\includegraphics[align=c,width=0.15\textwidth]{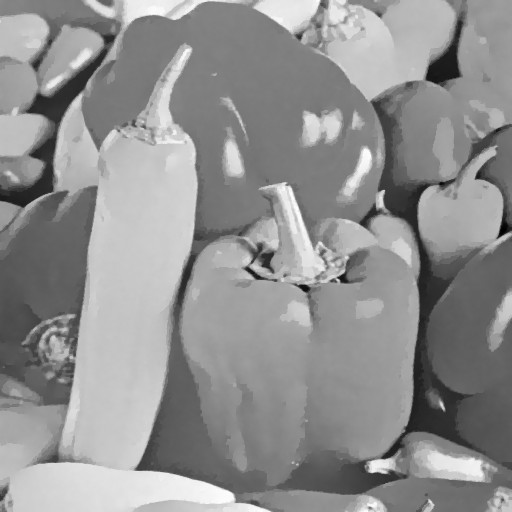}&\includegraphics[align=c,width=0.15\textwidth]{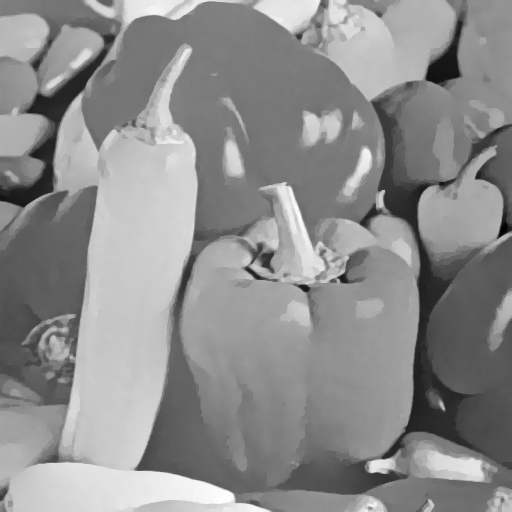}&
		\includegraphics[align=c,width=0.15\textwidth]{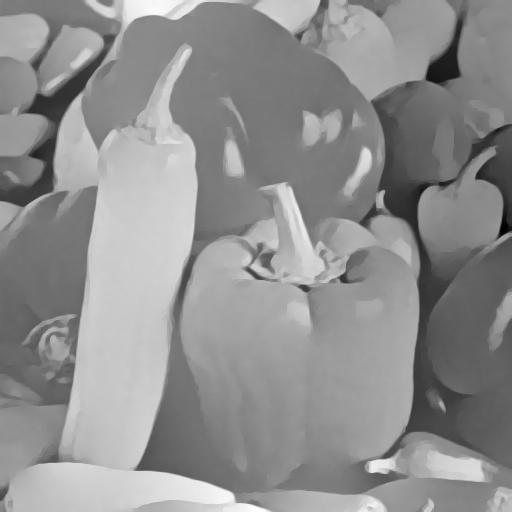}&
		\includegraphics[align=c,width=0.15\textwidth]{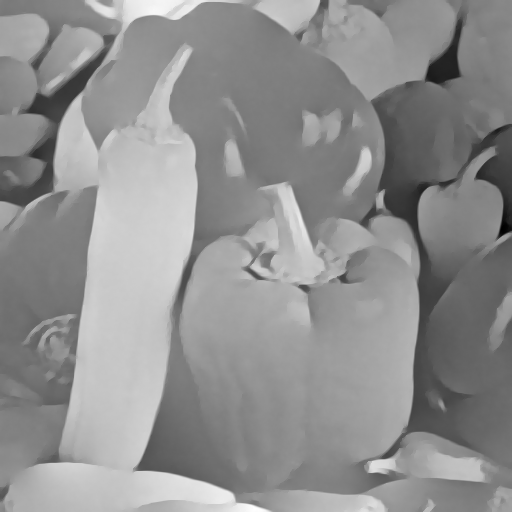}&
		\includegraphics[align=c,width=0.15\textwidth]{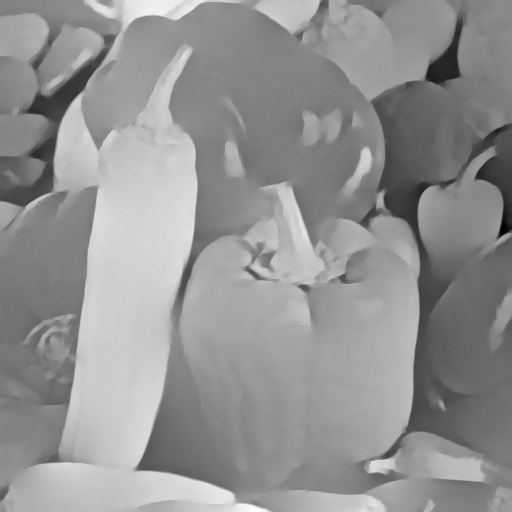}\\
		$w_{\text{scaled}}^*$&\includegraphics[align=c,width=0.15\textwidth]{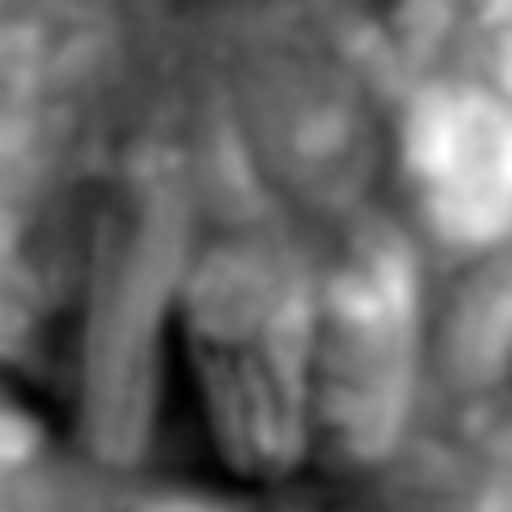}&\includegraphics[align=c,width=0.15\textwidth]{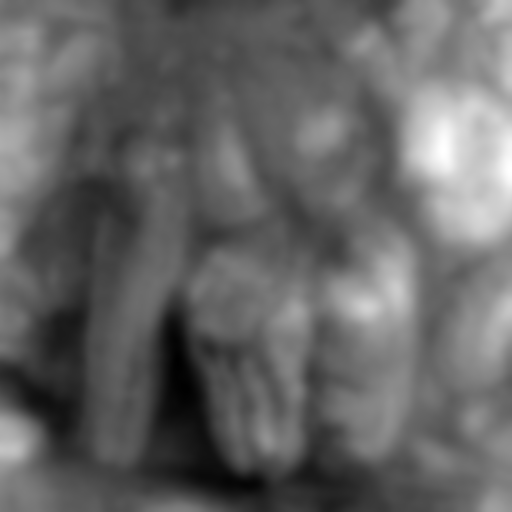}&
		\includegraphics[align=c,width=0.15\textwidth]{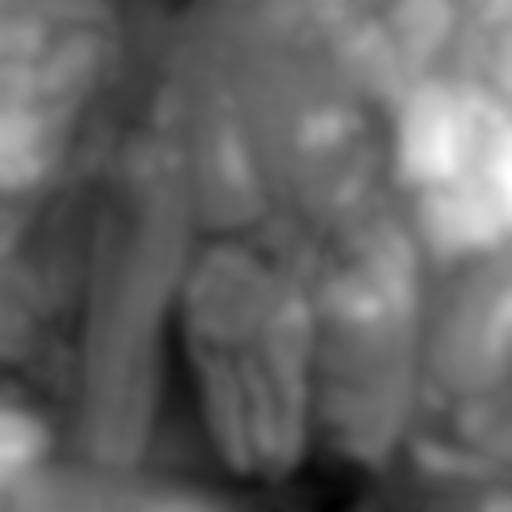}&
		\includegraphics[align=c,width=0.15\textwidth]{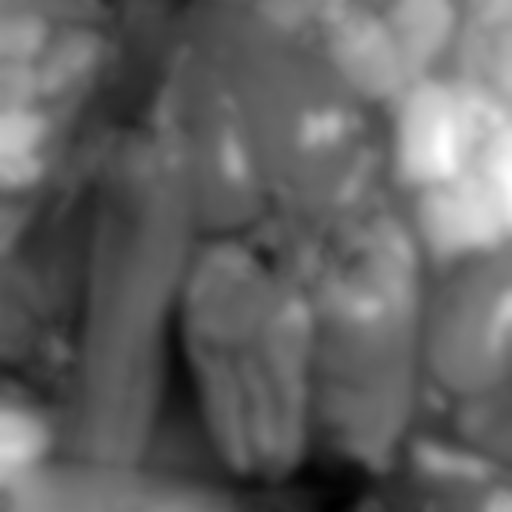}&
		\includegraphics[align=c,width=0.15\textwidth]{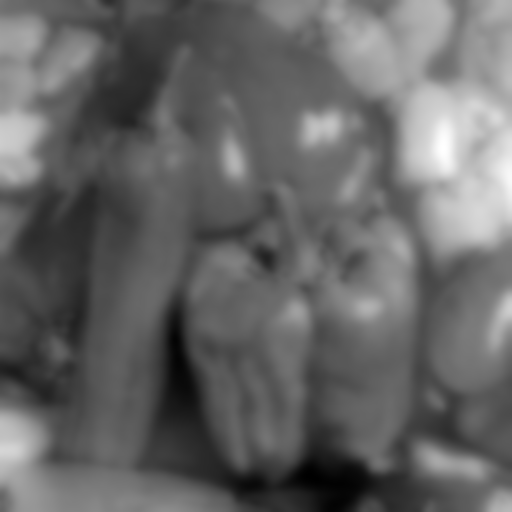}\\
		$u^*$&
		\includegraphics[align=c,width=0.15\textwidth]{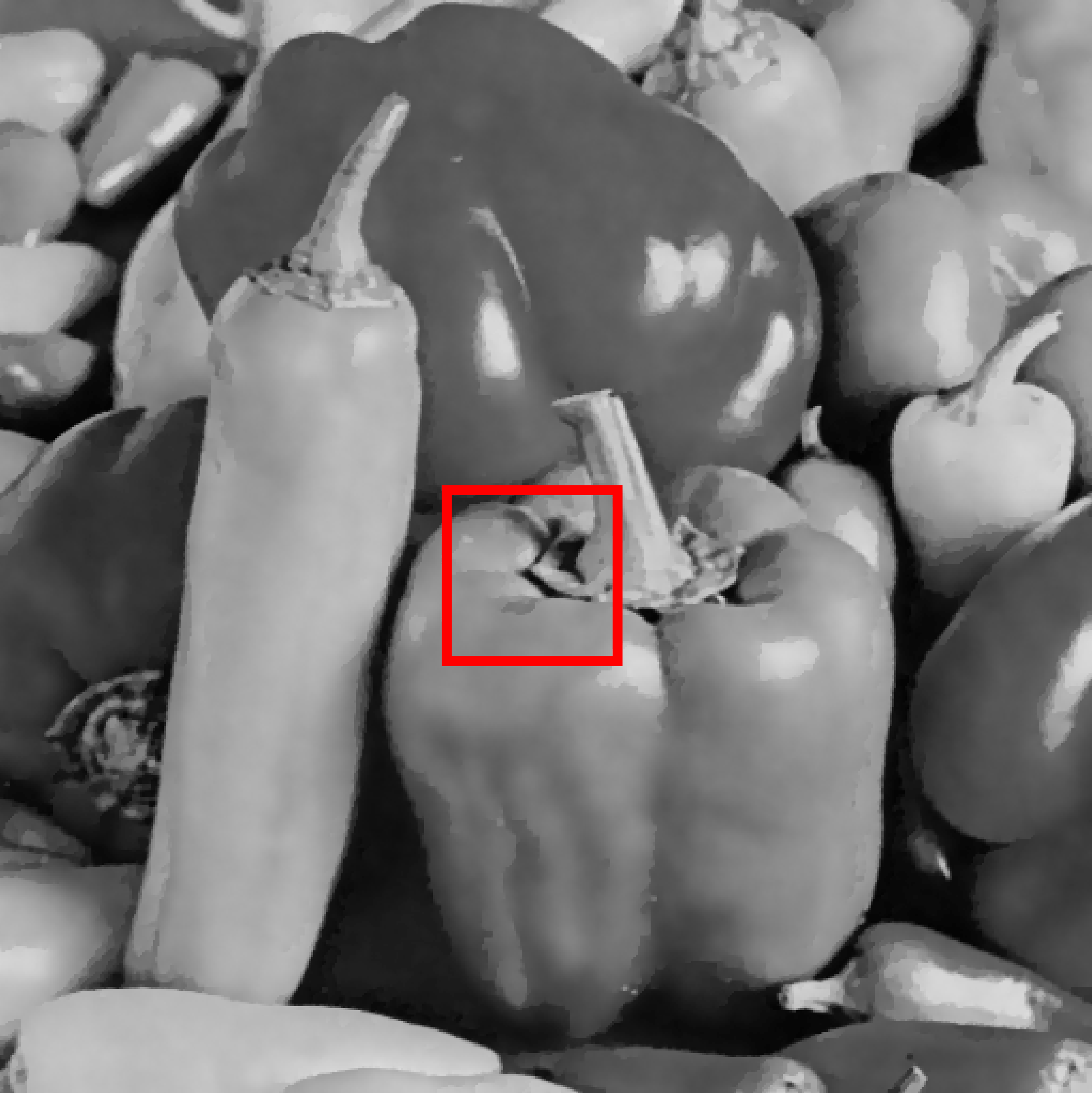}&\includegraphics[align=c,width=0.15\textwidth]{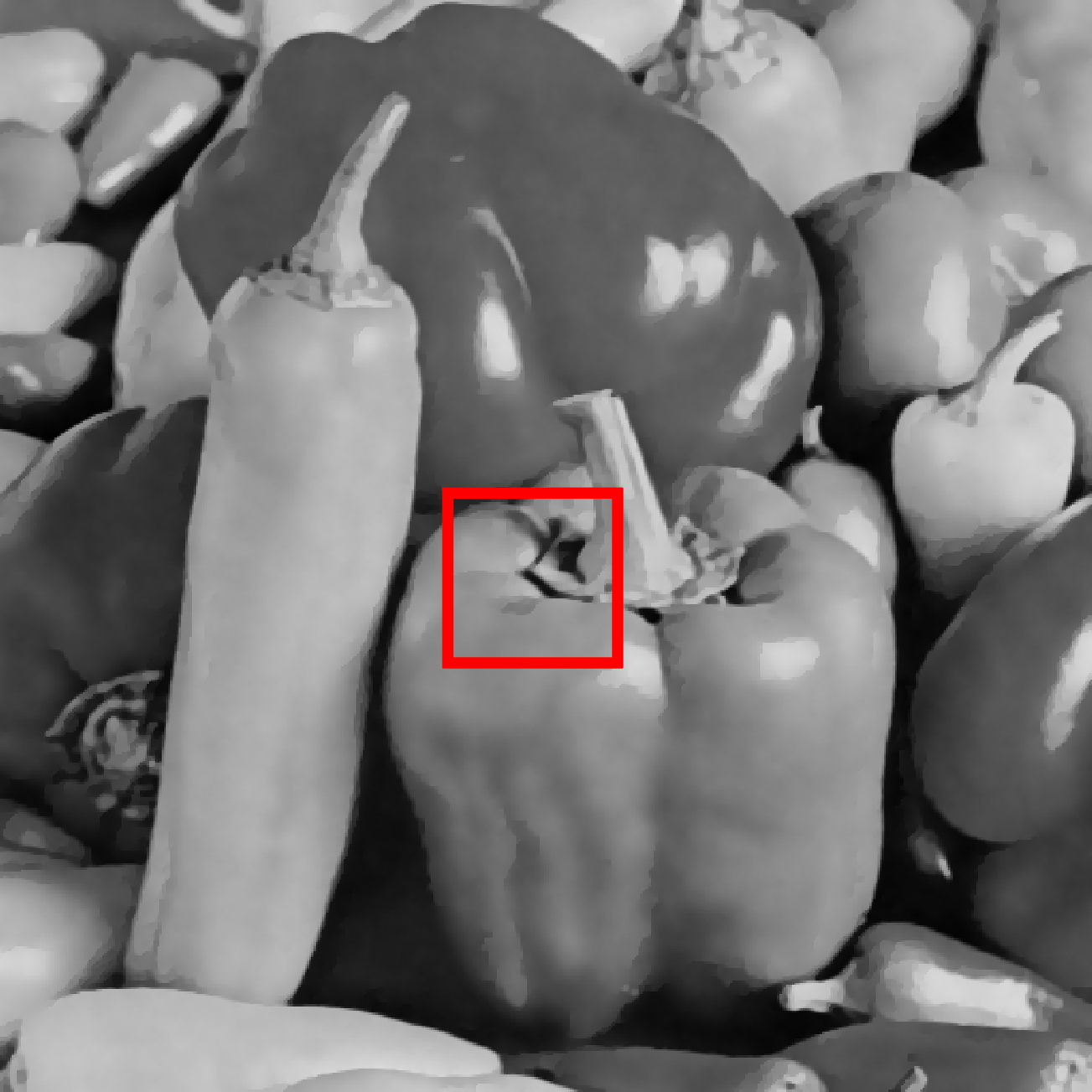}&
		\includegraphics[align=c,width=0.15\textwidth]{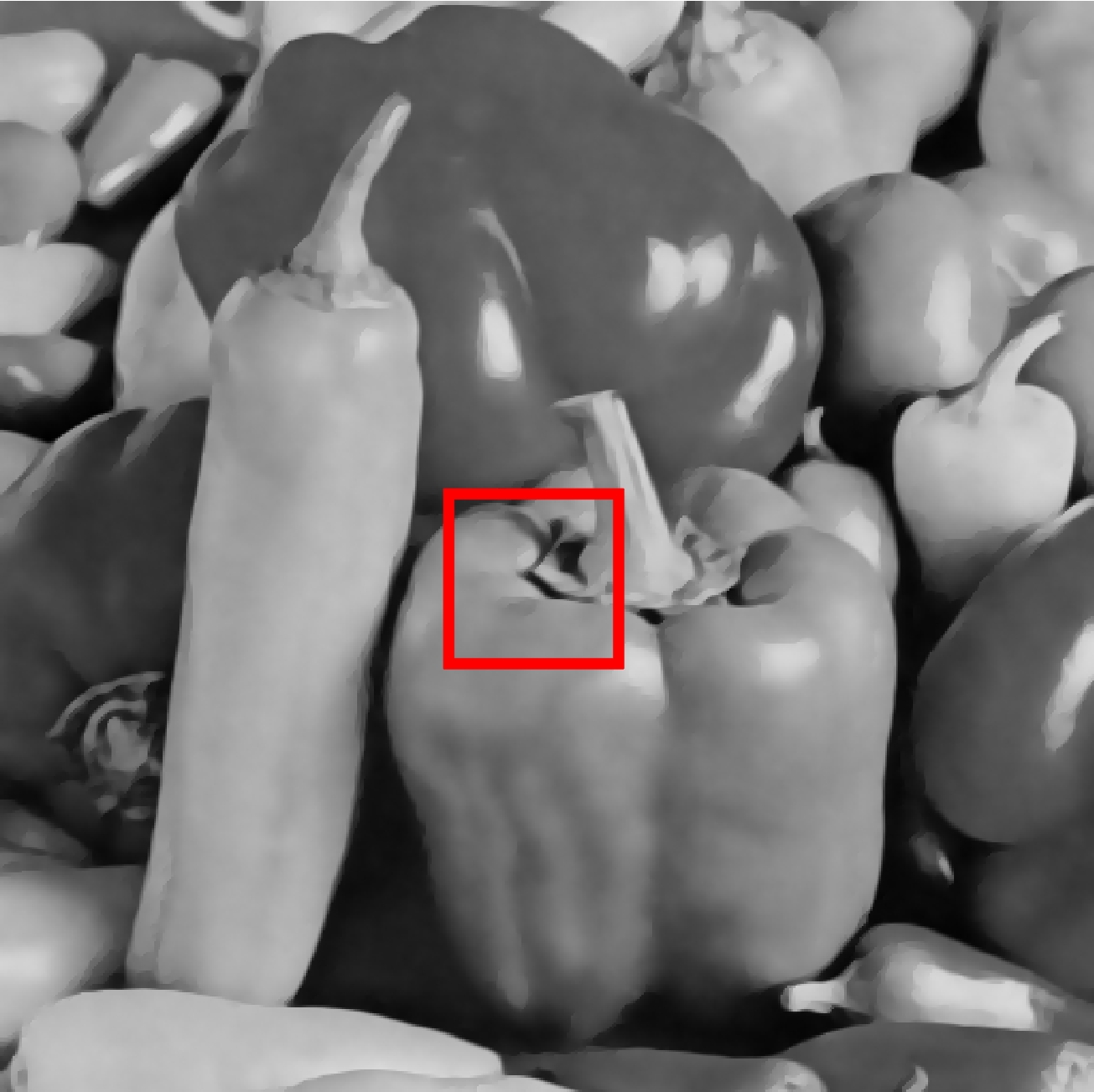}&
		\includegraphics[align=c,width=0.15\textwidth]{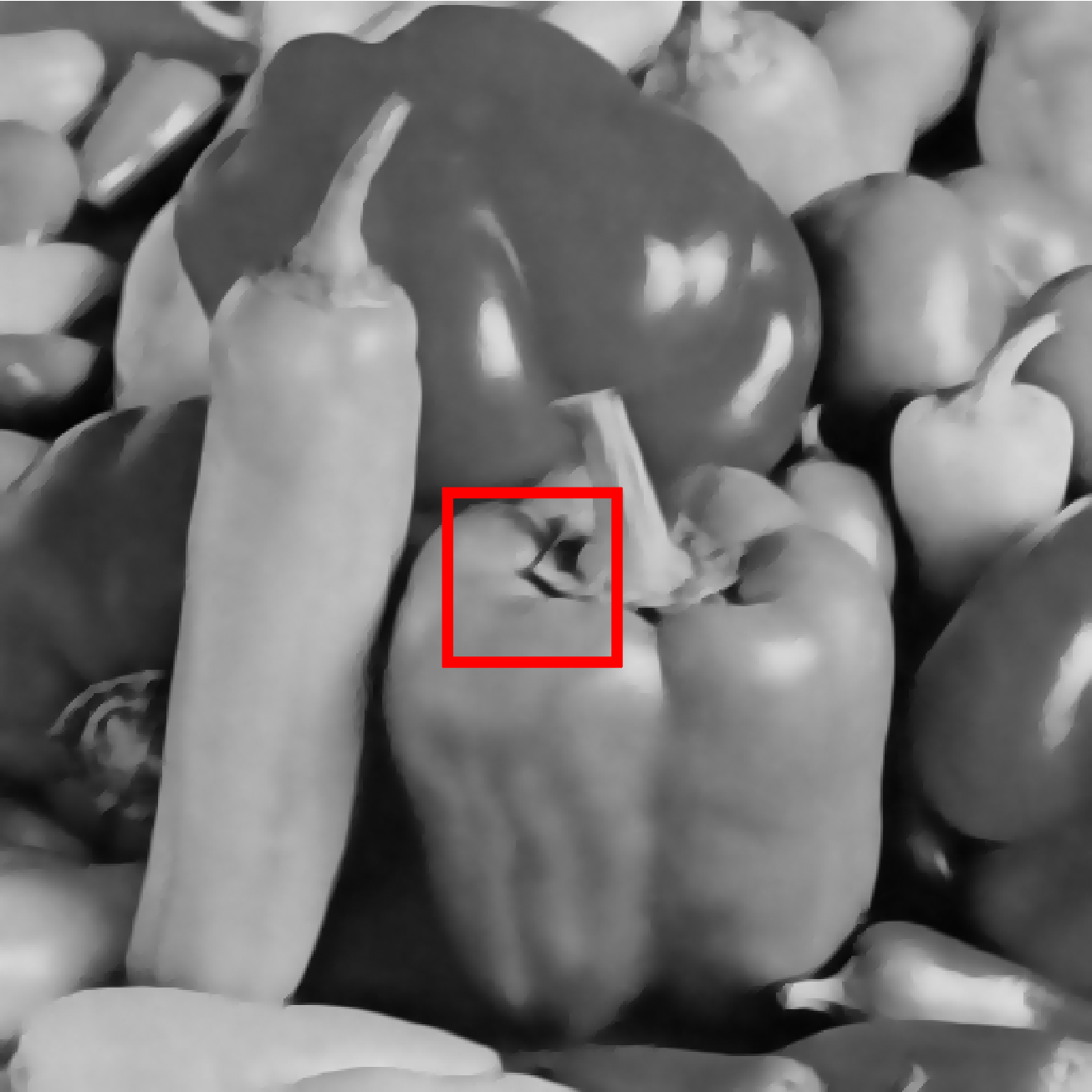}&
		\includegraphics[align=c,width=0.15\textwidth]{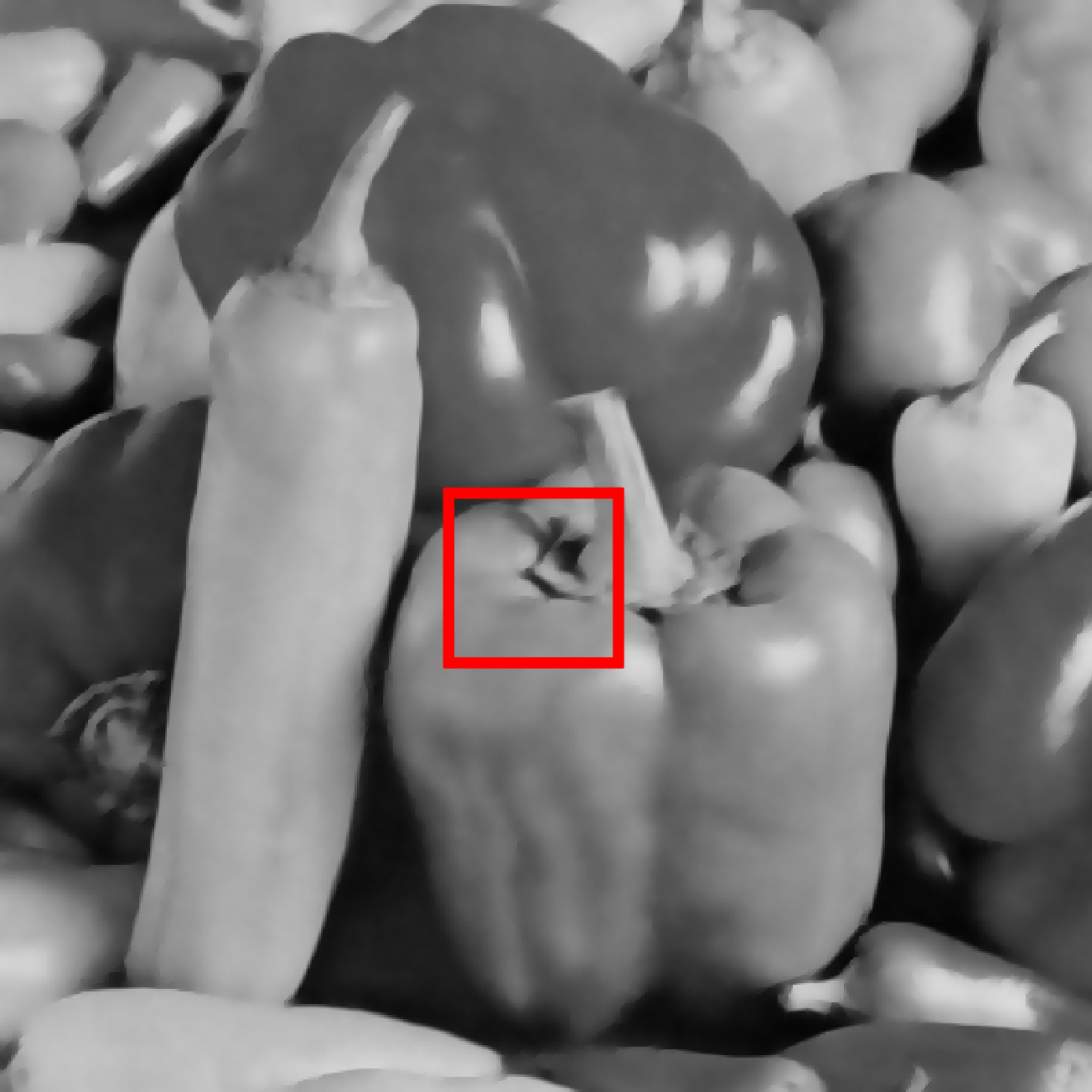}\\
		Zoom-in $u^*$&
		\includegraphics[align=c,width=0.15\textwidth]{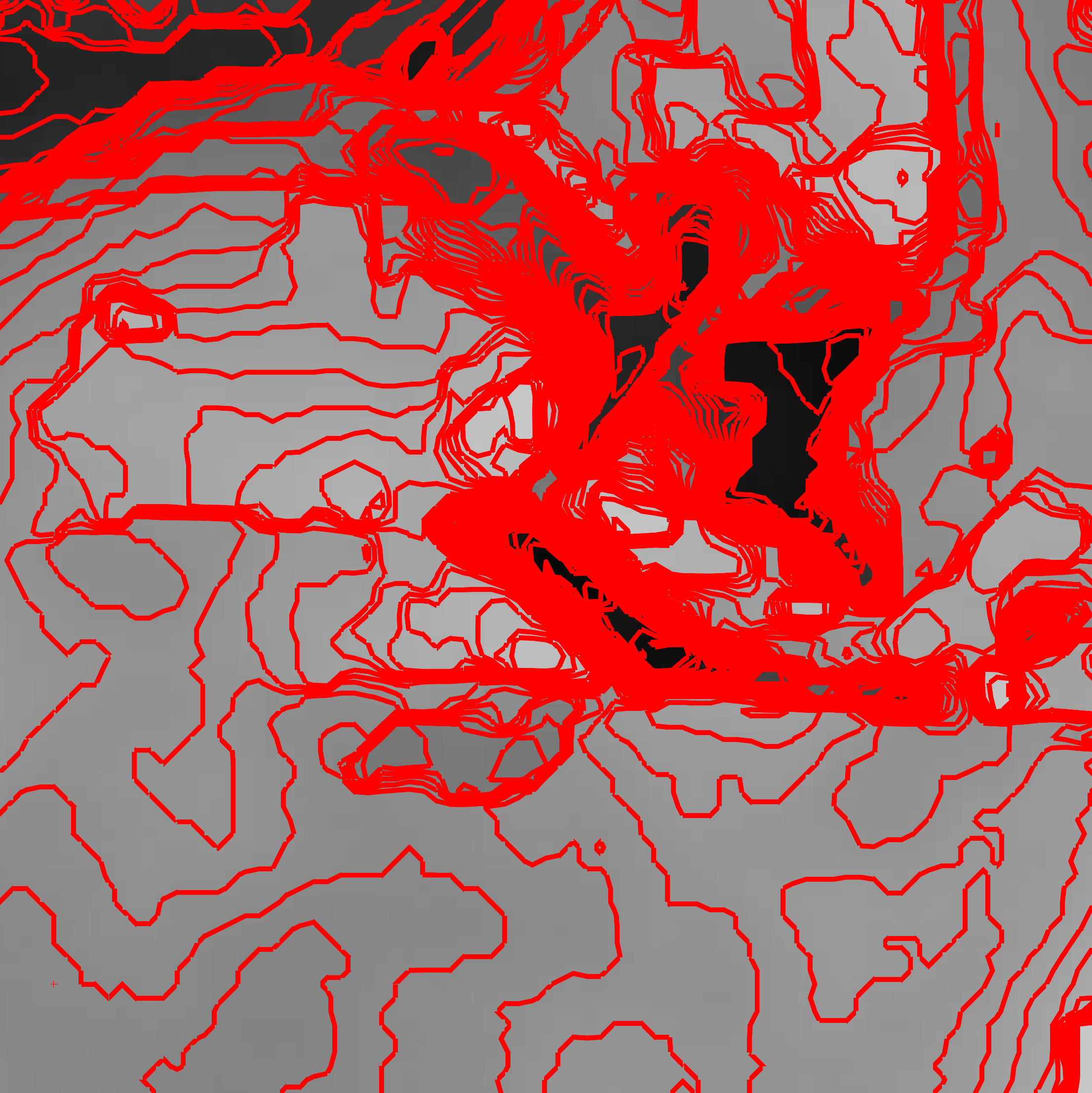}&\includegraphics[align=c,width=0.15\textwidth]{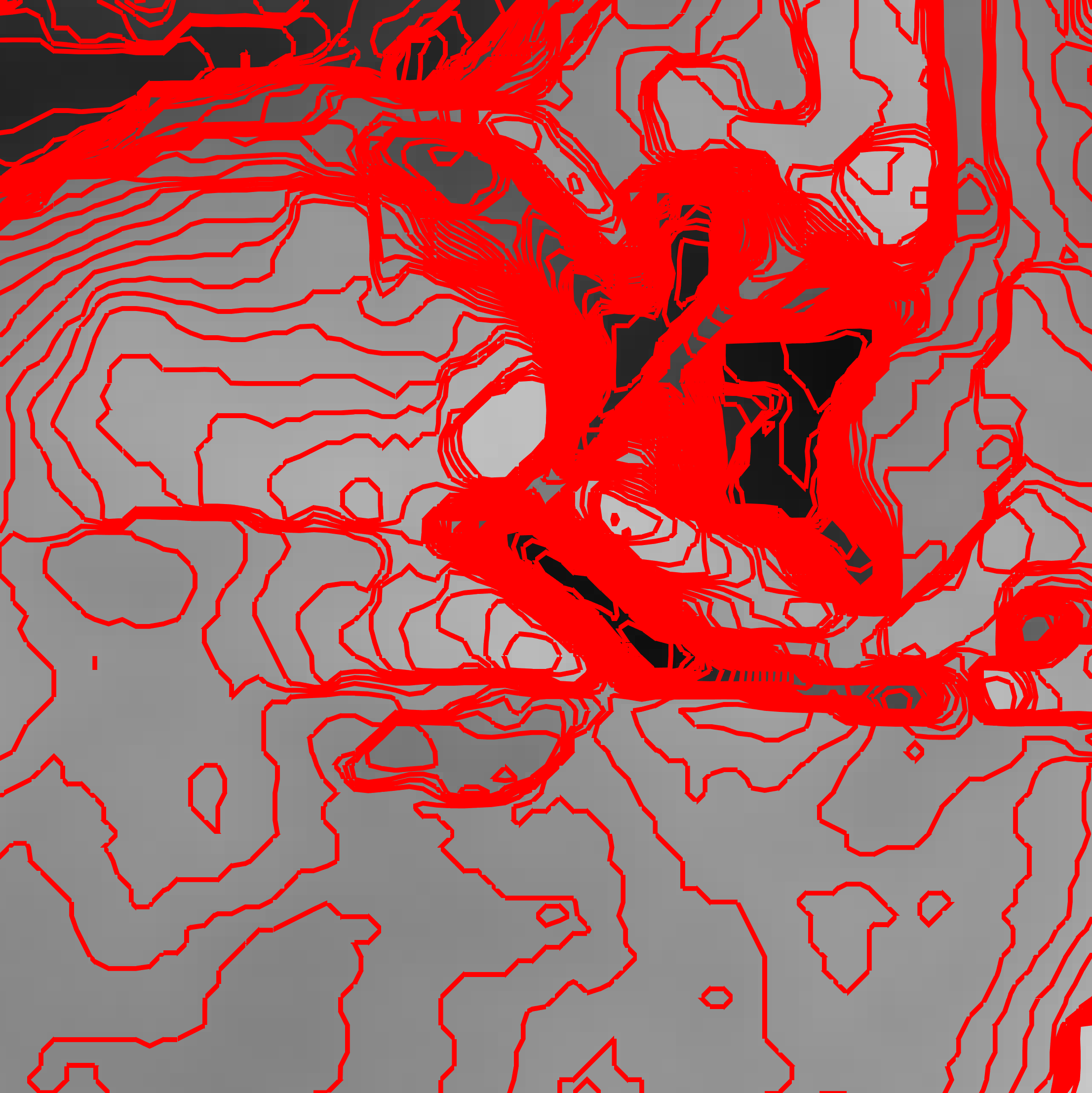}&
		\includegraphics[align=c,width=0.15\textwidth]{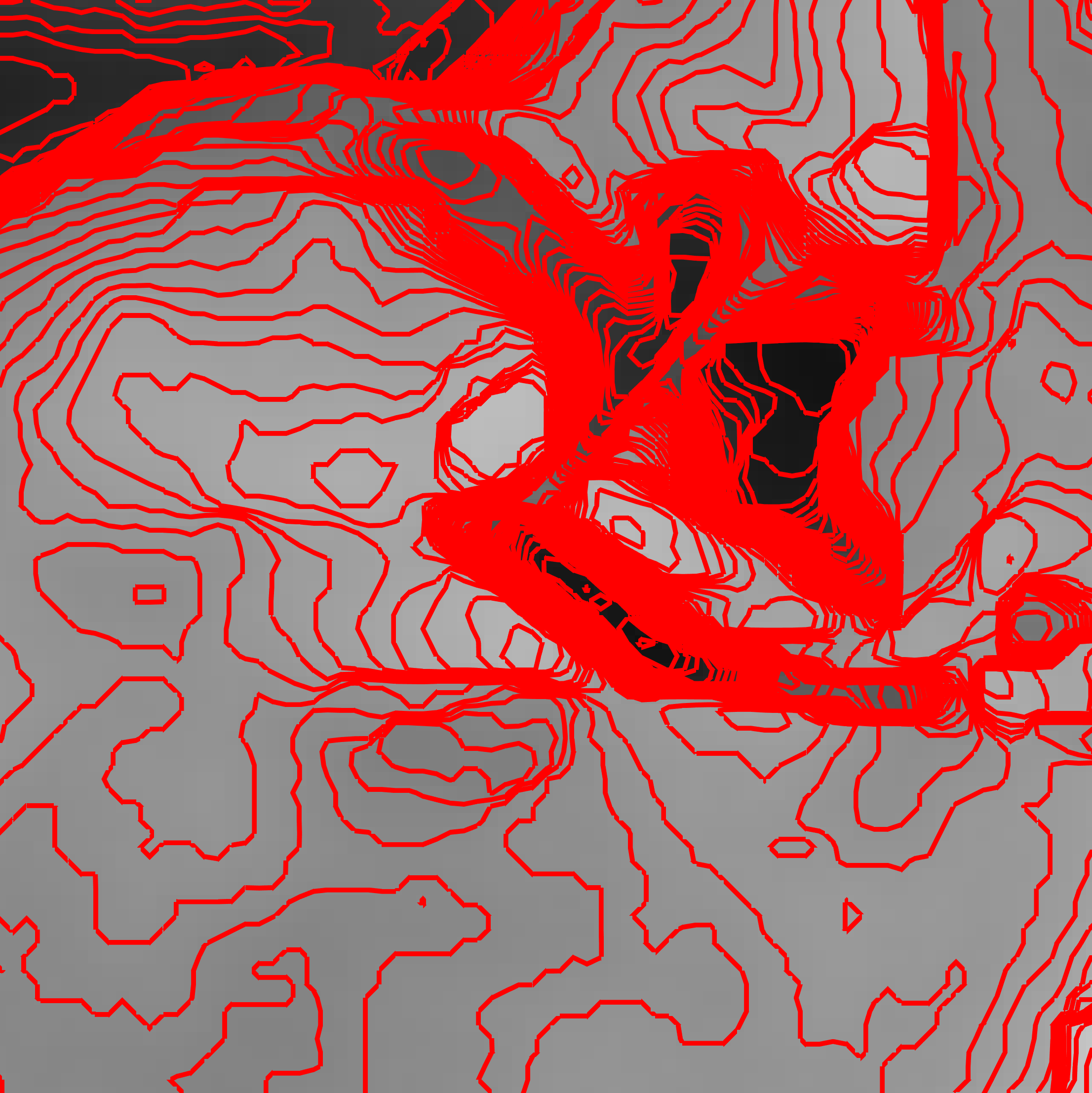}&
		\includegraphics[align=c,width=0.15\textwidth]{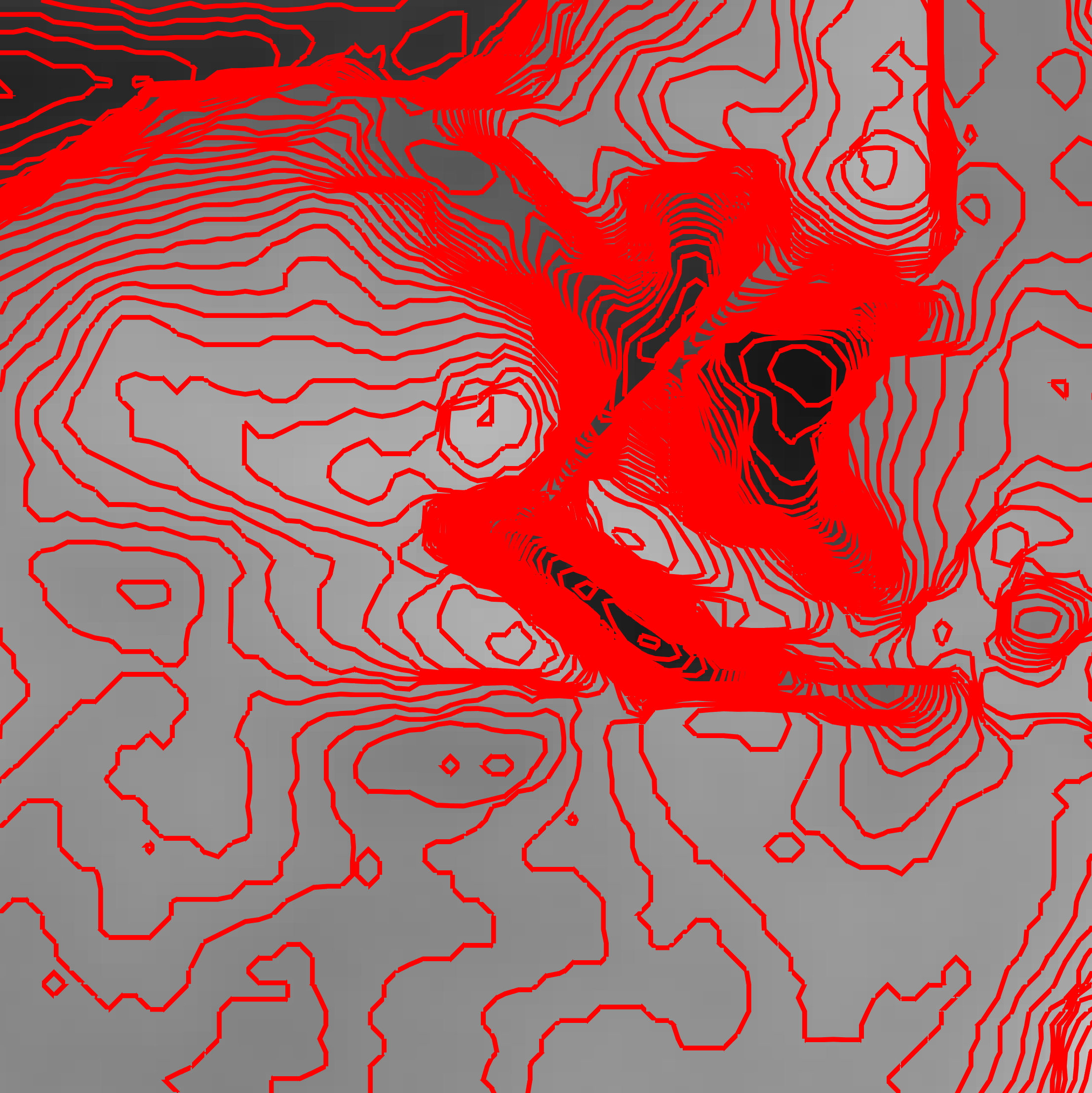}&
		\includegraphics[align=c,width=0.15\textwidth]{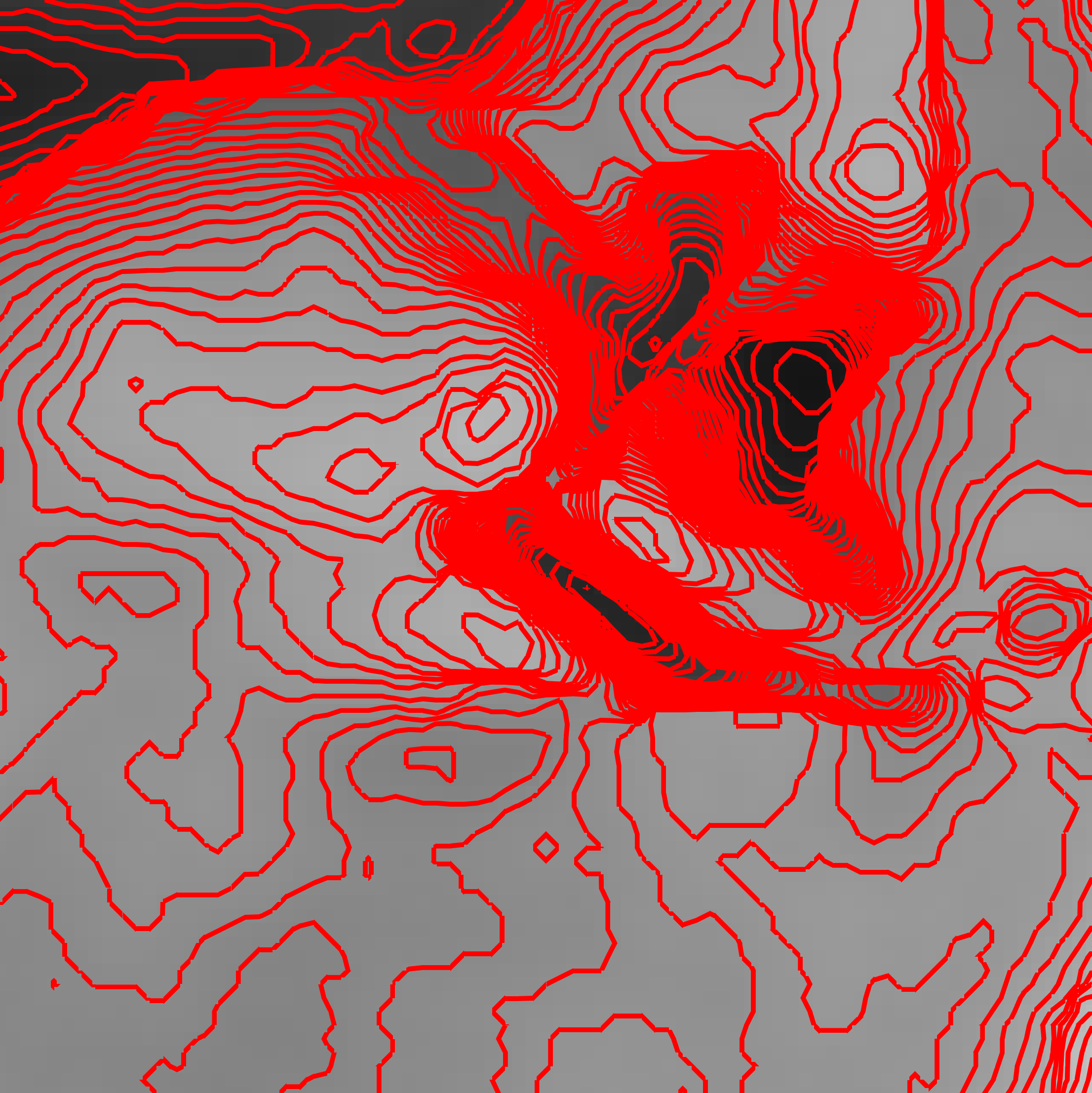}\\\hline
		PSNR($u^*$)&$33.06$&$32.77$&$32.05$&$31.19$&$30.67$\\\hline
	\end{tabular}
	\caption{Effect of $\alpha_{\text{curv}}$ ($\sigma=10/255$). When $\alpha_{\text{curv}}$ increases, $v^*$ (first row) becomes smoother while objects' boundaries are preserved. The $w^*$ component (second row) gains more geometric features such as shapes. Consequently, the sum $u^*$ (third row) is smoother. Thanks to the curvature regularization, stair-casing in the zoomed-in contours (last row) is effectively removed when increasing $\alpha_{\text{curv}}$. Other parameters are fixed: $\alpha_0=2\times10^{-3}$, $\alpha_w=50$, and $\alpha_n=10$.}\label{fig_alpha_curv}
\end{figure}

Figure~\ref{fig_alpha_curv} presents results of the proposed model on a noisy image $(\sigma=10/255)$ with varying values of $\alpha_{\text{curv}}$ when the other regularization parameters are fixed. When $\alpha_\text{curv}$ increases from $0.01$ to $10$, we observe that soft shades on peppers in $v^*$ are suppressed while the boundaries are preserved. Meanwhile, the $w^*$ component gains more variation in the intensity. If we keep increasing $\alpha_{\text{curv}}$ above $0.1$, the intensity variation near the boundaries and the highlights due to reflection slowly fade away in $v^*$. In contrast, shapes of peppers are more apparent in $w^*$. Since the sharp gradients in $v^*$ are reduced and the intensity variation in $w^*$ is intrinsically smooth, the reconstructed $u^*$ component becomes slightly blurry. This transition is also reflected by the decreasing PSNR. In Figure~\ref{fig_alpha_curv}, we also show the level lines of $u^*$ in a zoomed-in region. The effect of $\alpha_{\text{curv}}$ on $u^*$ is obvious. Increasing $\alpha_{\text{curv}}$ yields more regular level lines, and the staircase effect due to $L^0$-gradient regularization is thus ameliorated.

\subsection{Comparison on cartoon+smooth+oscillation decomposition}\label{subsec_decompose}

\begin{figure}[t!]
	\centering
	\begin{tabular}{c|@{\hspace{2pt}}c@{\hspace{2pt}}c@{\hspace{2pt}}c@{\hspace{2pt}}c}
		&\multicolumn{4}{c}{CEP~\cite{chan2007image} (CPU time: 76.16 sec)}\\
		$f_{\text{noise}}$&$u^*$&$v^*_{\text{scaled}}$&$w^*_{\text{scaled}}$&$n^*_{\text{scaled}}$\\
		\multirow{10}{0.18\textwidth}{\includegraphics[width=0.18\textwidth]{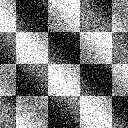}}&
		\includegraphics[width=0.18\textwidth]{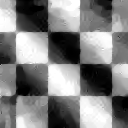}&
		\includegraphics[width=0.18\textwidth]{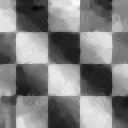}&
		\includegraphics[width=0.18\textwidth]{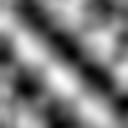}&
		\includegraphics[width=0.18\textwidth]{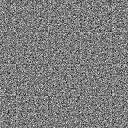}\\
		&PSNR$=29.04$&&&\\\cline{2-5}
		&\multicolumn{4}{c}{HKLM~\cite{huska2021variational} (CPU time: 199.48 sec)}\\
		&$u^*$&$v^*_{\text{scaled}}$&$w^*_{\text{scaled}}$&$n^*_{\text{scaled}}$\\
		&
		\includegraphics[width=0.18\textwidth]{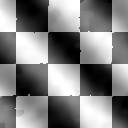}&
		\includegraphics[width=0.18\textwidth]{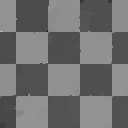}&
		\includegraphics[width=0.18\textwidth]{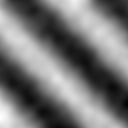}&
		\includegraphics[width=0.18\textwidth]{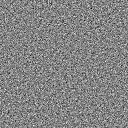}\\
		&PSNR$=25.08$&&&\\\cline{2-5}
		&\multicolumn{4}{c}{Proposed (CPU time: \textbf{14.83} sec)}\\
		&$u^*$&$v^*_{\text{scaled}}$&$w^*_{\text{scaled}}$&$n^*_{\text{scaled}}$\\
		&
		\includegraphics[width=0.18\textwidth]{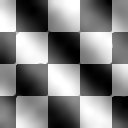}&
		\includegraphics[width=0.18\textwidth]{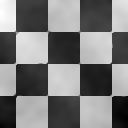}&
		\includegraphics[width=0.18\textwidth]{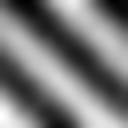}&
		\includegraphics[width=0.18\textwidth]{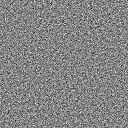}\\
		&PSNR$=\mathbf{32.06}$&&&STD$=49.08$\\\hline
	\end{tabular}
	\caption{Comparison of CEP~\cite{chan2007image}, HKLM~\cite{huska2021variational} and the proposed model for decomposing a noisy synthetic image (size: $128\times128$, noise: $\sigma=50/255$). 
		For CEP (see also~\eqref{eq_chan}), we use $\lambda=0.5,\alpha=10,\mu=2$ and $\delta$ determined by SURE~\cite{donoho1995adapting}. For HKLM (see also~\eqref{eq_huska}), we use $\beta_1=1\times10^{-4}$, $\beta_2=0.5$, $\beta_3=5\times10^{-6}$, and $a=30$ for the MC penalty (an approximation of the $L^0$-norm, see \cite{huska2021variational} for details). For the proposed model, we use $\alpha_0=4\times10^{-2}$, $\alpha_{\text{curv}}=0.15$, $\alpha_w=50$ and $\alpha_n=1\times10^{-3}$. }  \label{fig_compare1}
\end{figure}

\begin{figure}[t!]
	\centering
	\begin{tabular}{c|@{\hspace{2pt}}c@{\hspace{2pt}}c@{\hspace{2pt}}c@{\hspace{2pt}}c}
		&\multicolumn{4}{c}{CEP~\cite{chan2007image} (CPU time: 1655.30 sec)}\\
		$f_{\text{noise}}$&$u^*$&$v^*_{\text{scaled}}$&$w^*_{\text{scaled}}$&$n^*_{\text{scaled}}$\\
		\multirow{10}{0.18\textwidth}{\includegraphics[width=0.18\textwidth]{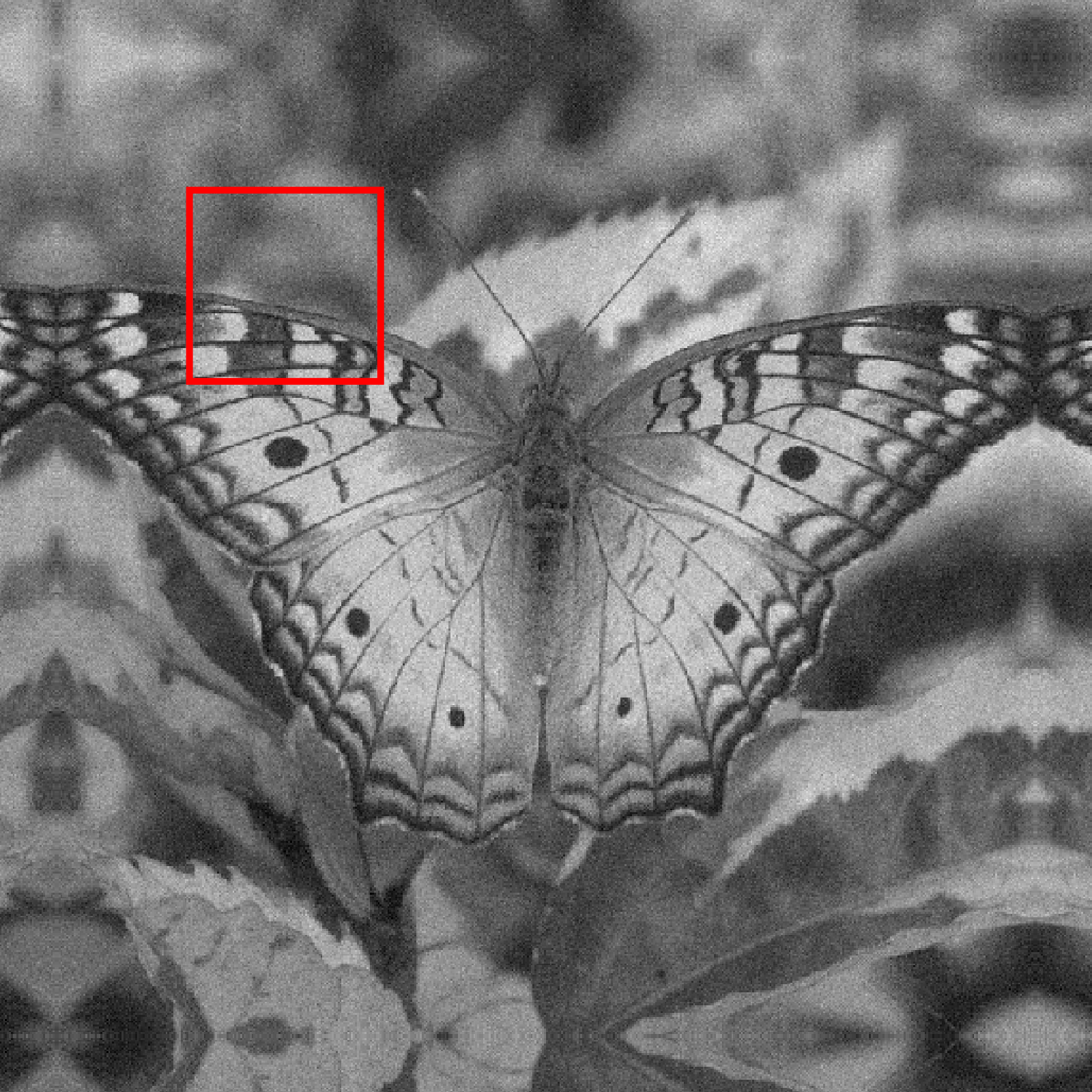}}&
		\includegraphics[width=0.18\textwidth]{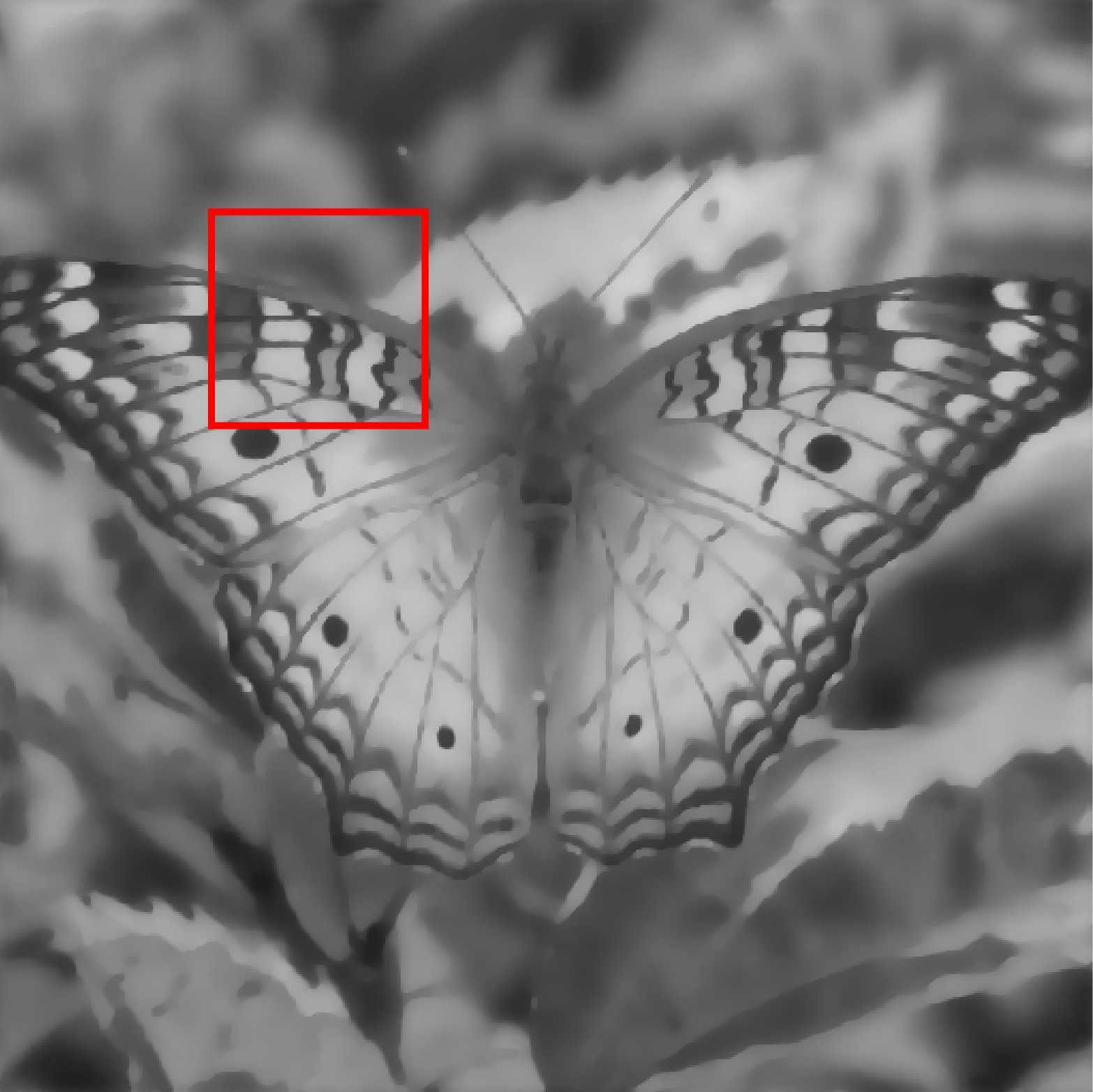}&
		\includegraphics[width=0.18\textwidth]{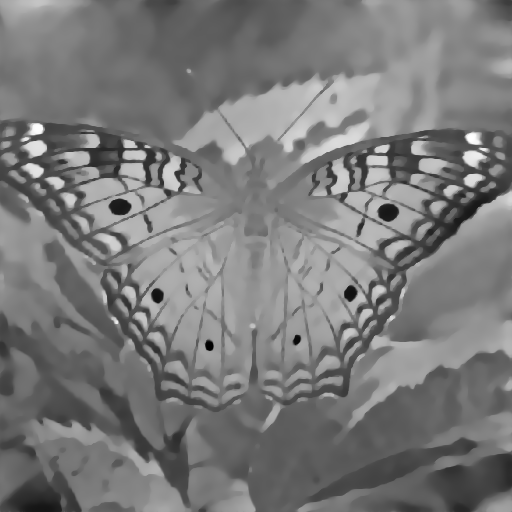}&
		\includegraphics[width=0.18\textwidth]{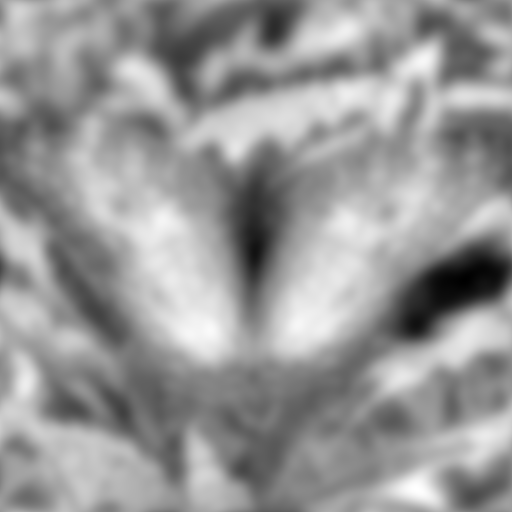}&
		\includegraphics[width=0.18\textwidth]{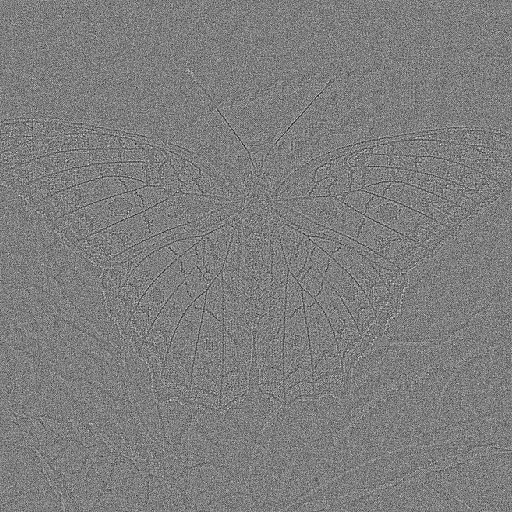}\\
		&PSNR$=29.11$&&&\\\cline{2-5}
		&\multicolumn{4}{c}{HKLM~\cite{huska2021variational} (CPU time: 4201.45 sec)}\\
		&$u^*$&$v^*_{\text{scaled}}$&$w^*_{\text{scaled}}$&$n^*_{\text{scaled}}$\\
		&
		\includegraphics[width=0.18\textwidth]{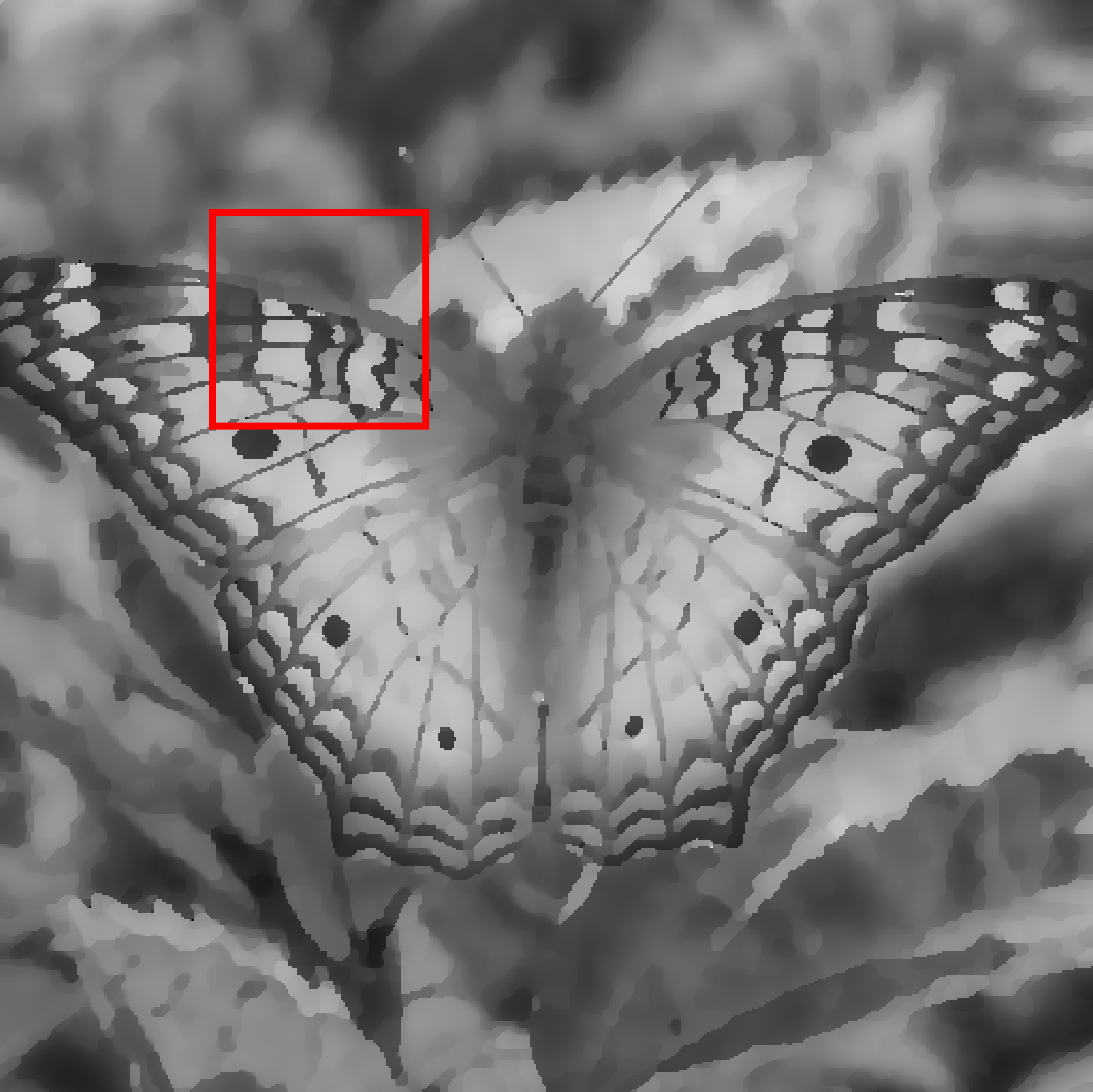}&
		\includegraphics[width=0.18\textwidth]{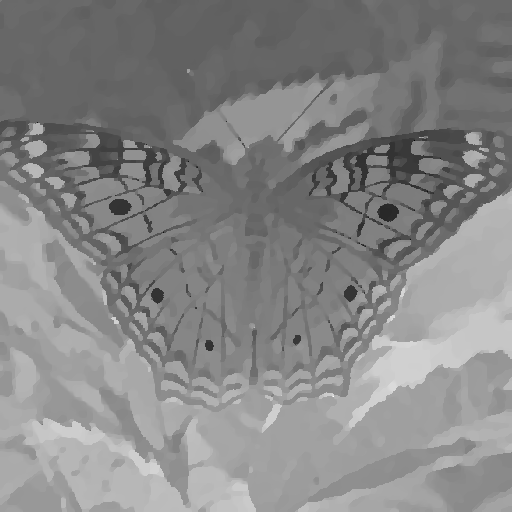}&
		\includegraphics[width=0.18\textwidth]{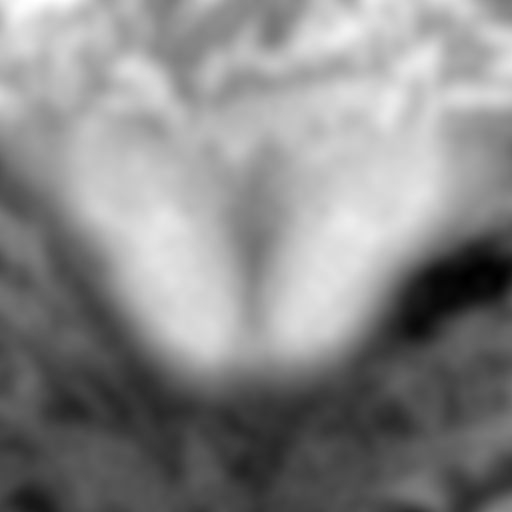}&
		\includegraphics[width=0.18\textwidth]{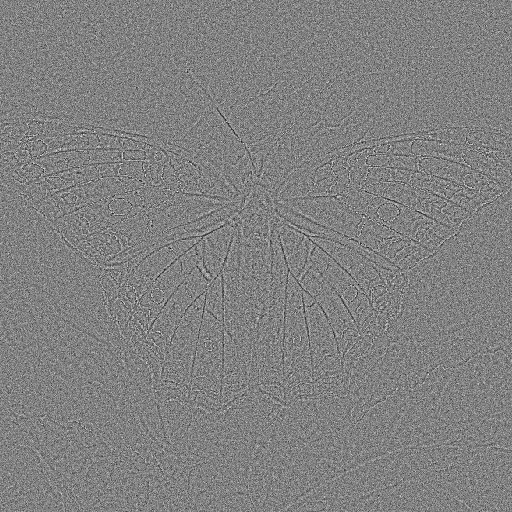}\\
		&PSNR$=27.93$&&&\\\cline{2-5}
		&\multicolumn{4}{c}{Proposed (CPU time: \textbf{179.20} sec)}\\
		&$u^*$&$v^*_{\text{scaled}}$&$w^*_{\text{scaled}}$&$n^*_{\text{scaled}}$\\
		&
		\includegraphics[width=0.18\textwidth]{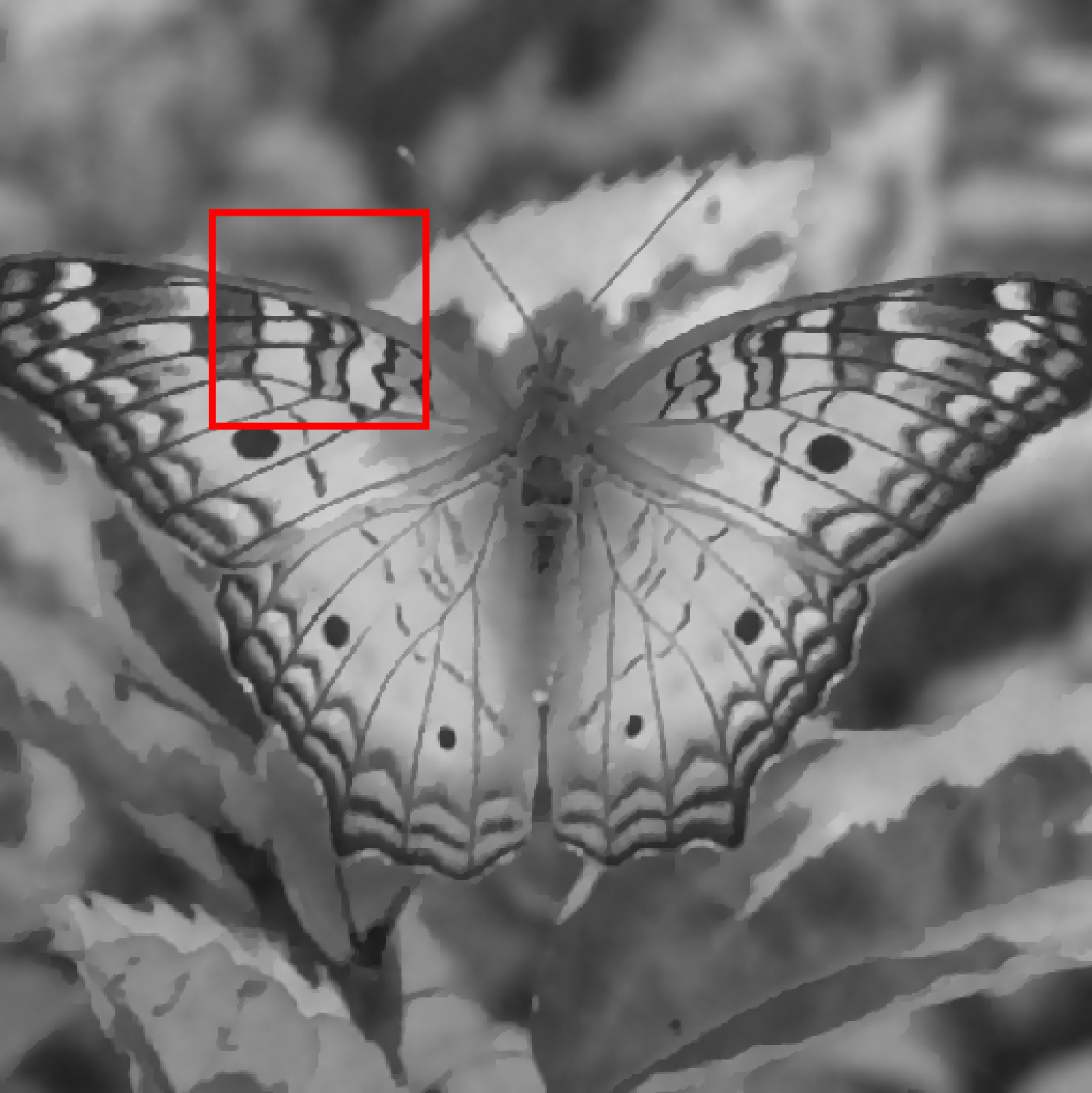}&
		\includegraphics[width=0.18\textwidth]{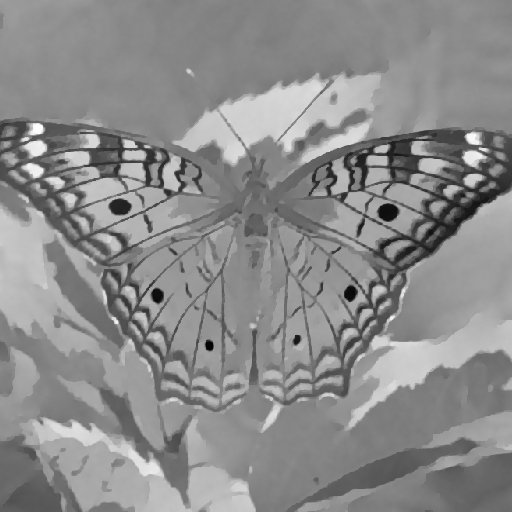}&
		\includegraphics[width=0.18\textwidth]{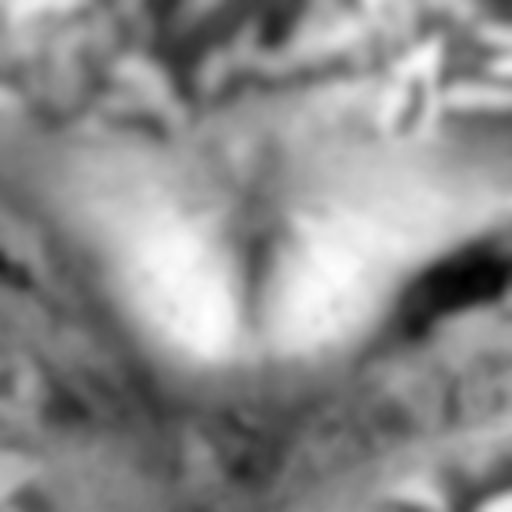}&
		\includegraphics[width=0.18\textwidth]{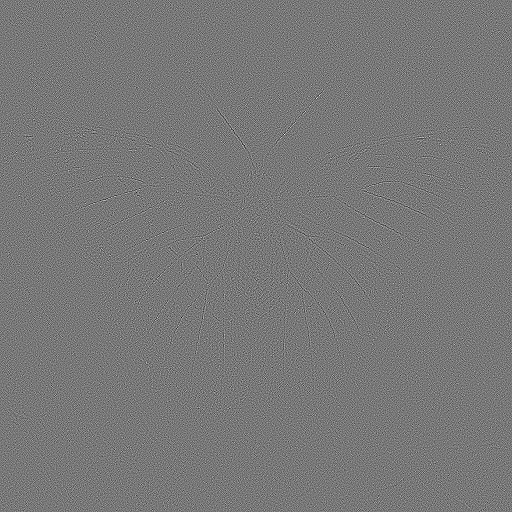}\\
		&PSNR$=\mathbf{31.46}$&&&\\\hline
	\end{tabular}
	\caption{Comparison on cartoon+smooth+oscillation decomposition of CEP~\cite{chan2007image}, HKLM~\cite{huska2021variational} and the proposed model. $f_{\text{noise}}$ is the input \textit{butterfly} (size: $512\times512$, noise: $\sigma=10/255$).  For CEP (see also~\eqref{eq_chan}), we set $\lambda=5,\alpha=10, \mu=2$ and $\delta$ determined from SURE~\cite{donoho1995adapting}. 
		For~\cite{huska2021variational} (see also~\eqref{eq_huska}), we use $\beta_1=1\times10^{-4}$, $\beta_2=3$, $\beta_3=5\times10^{-6}$, and $a=30$ for the MC penalty. For the proposed model, we use $\alpha_0=2\times10^{-3}$, $\alpha_{\text{curv}}=0.05$, $\alpha_w=30$ and $\alpha_n=10$. }  \label{fig_compare_butterfly}
\end{figure}

\begin{figure}[t!]
	\centering
	\begin{tabular}{c@{\hspace{2pt}}c@{\hspace{2pt}}c@{\hspace{2pt}}c@{\hspace{2pt}}c}
		(a)&(b)&(c)&(d)&(e)\\
		\includegraphics[width=0.18\textwidth]{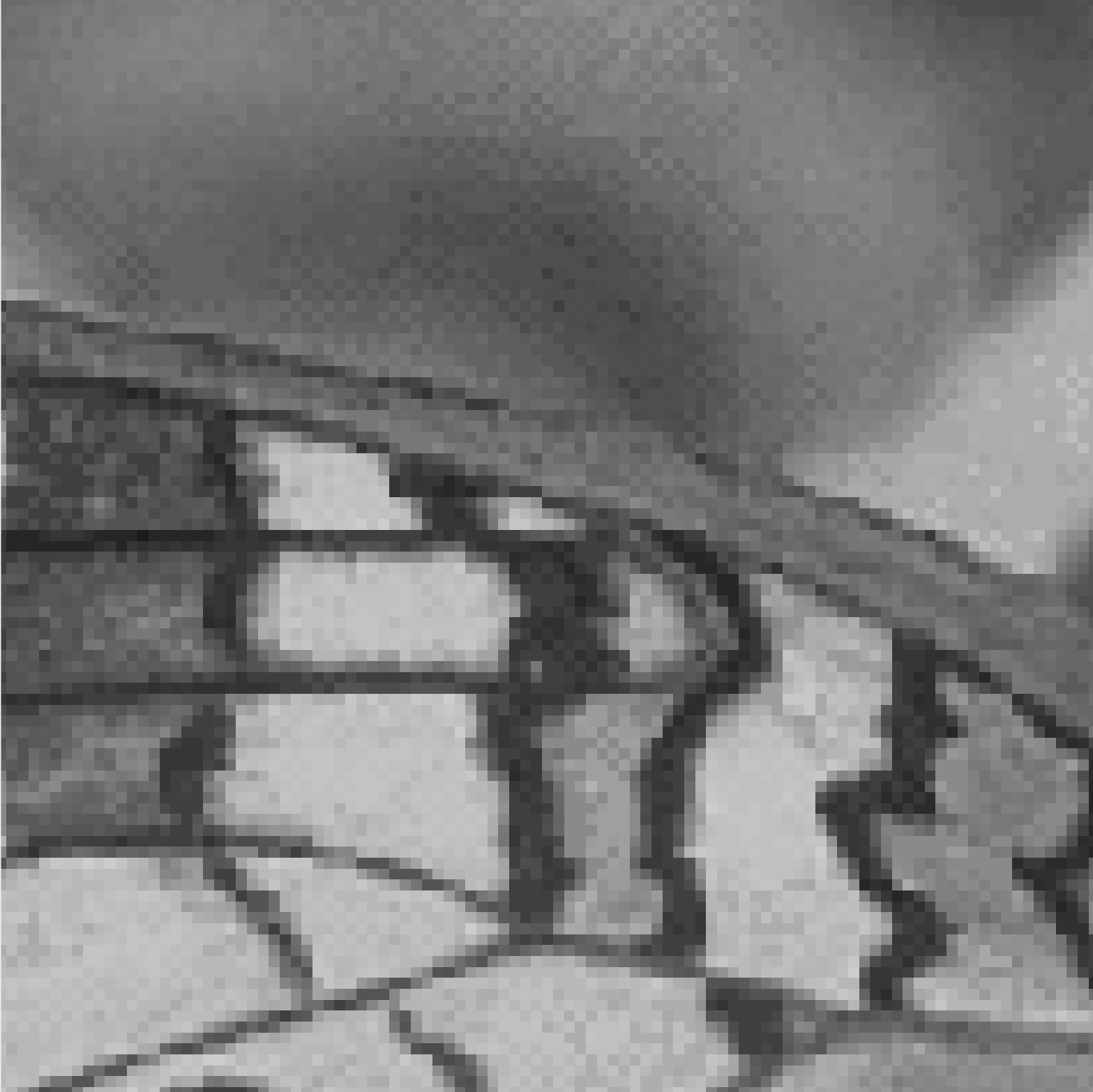}&
		\includegraphics[width=0.18\textwidth]{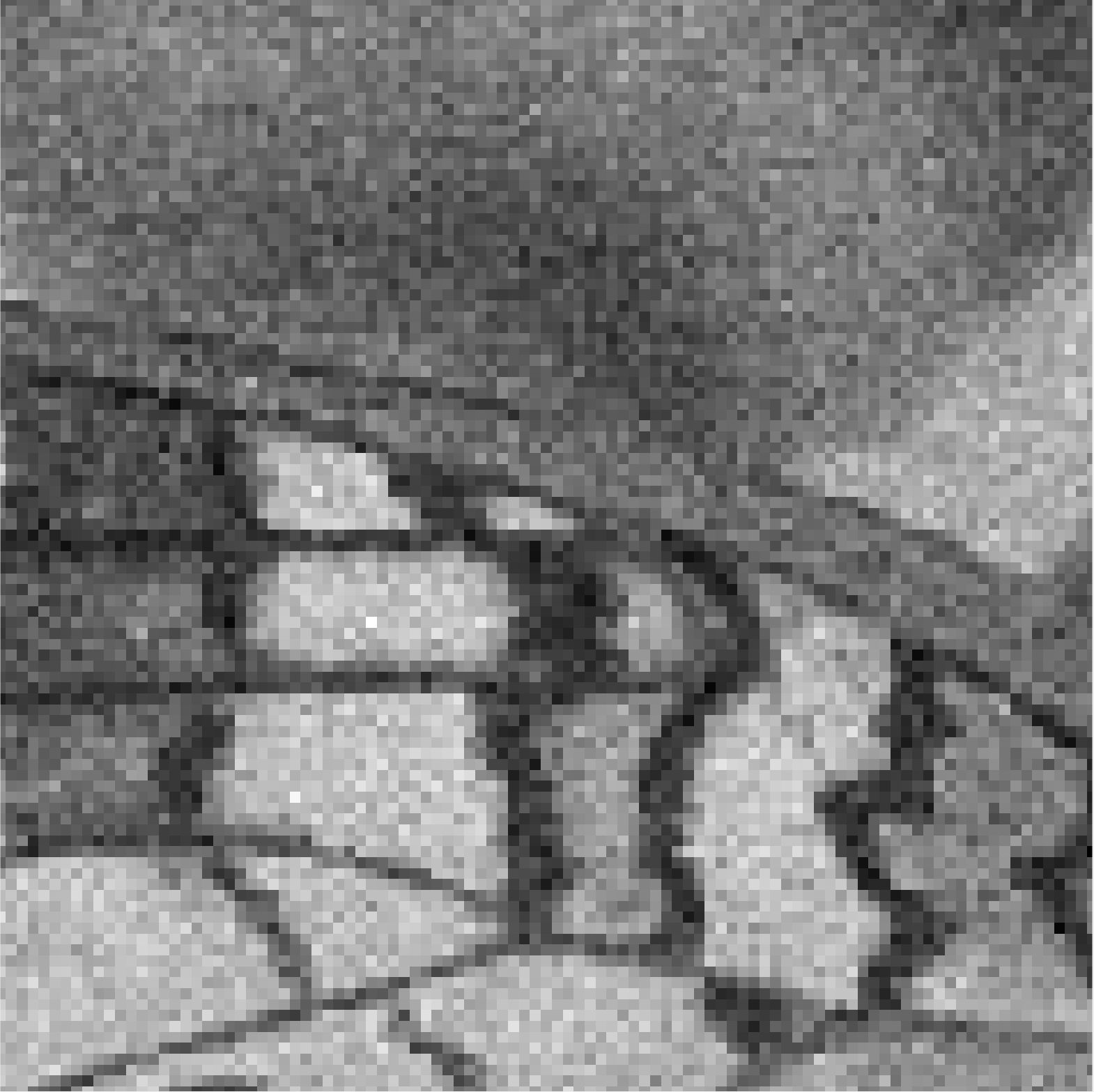}&
		\includegraphics[width=0.18\textwidth]{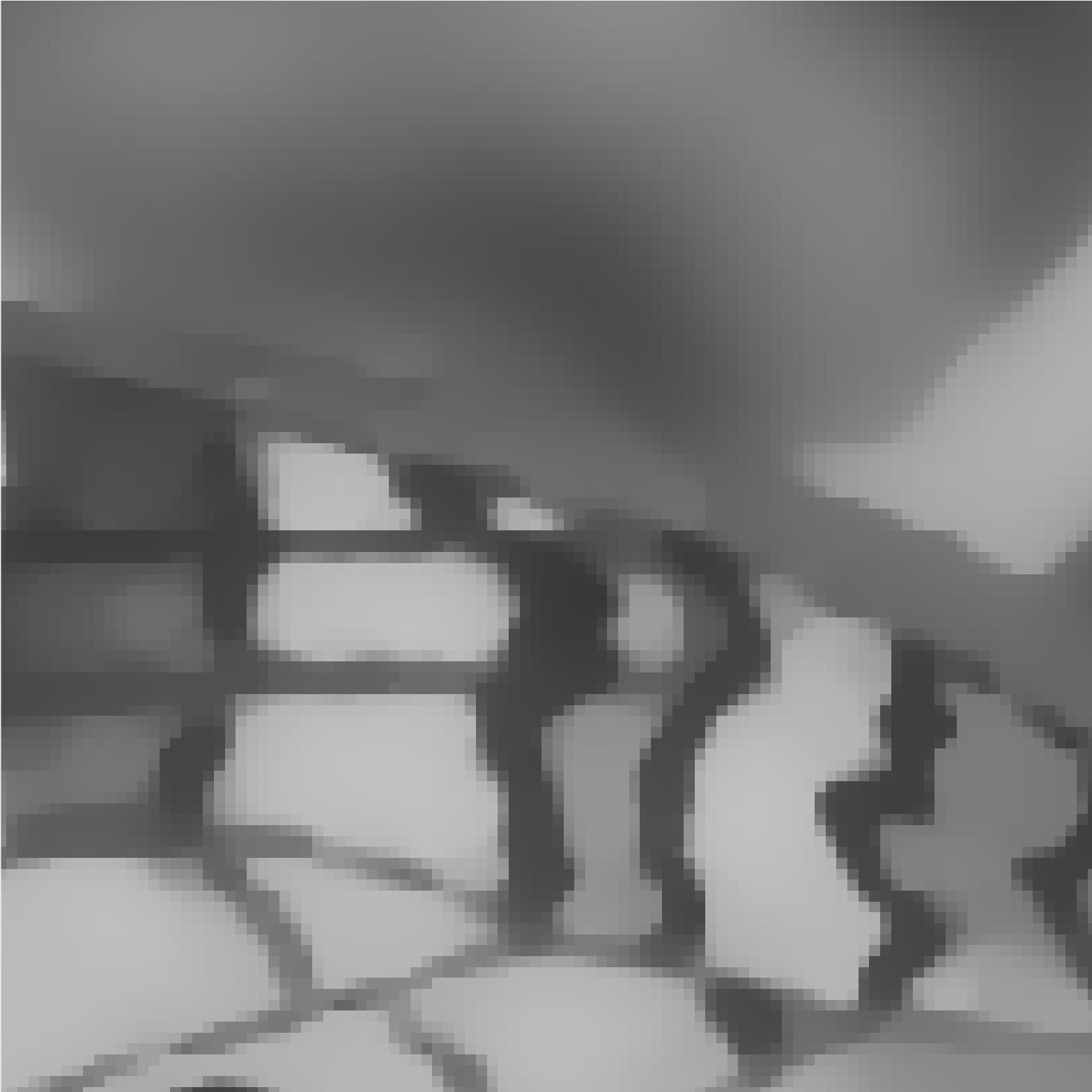}&
		\includegraphics[width=0.18\textwidth]{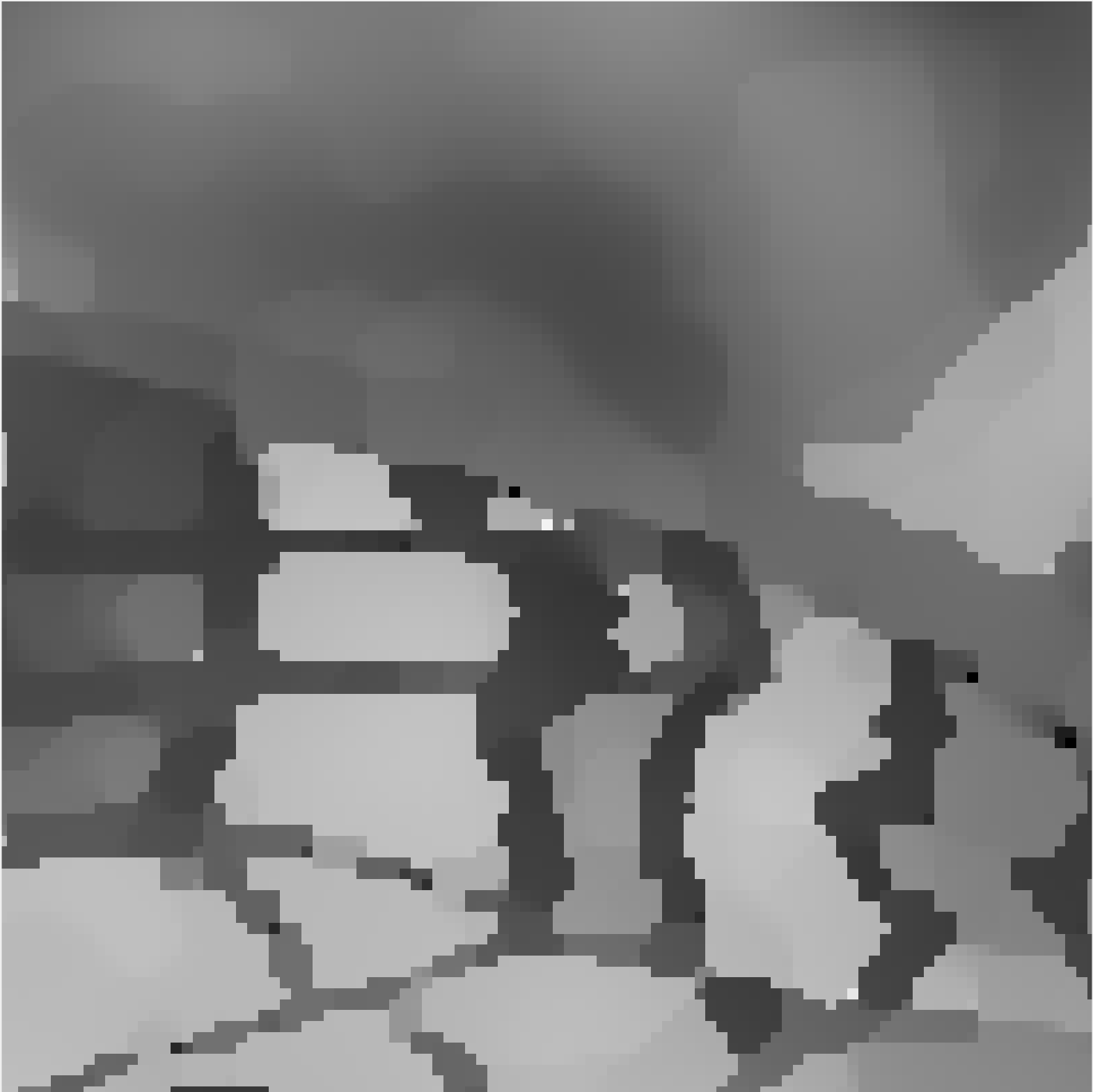}&
		\includegraphics[width=0.18\textwidth]{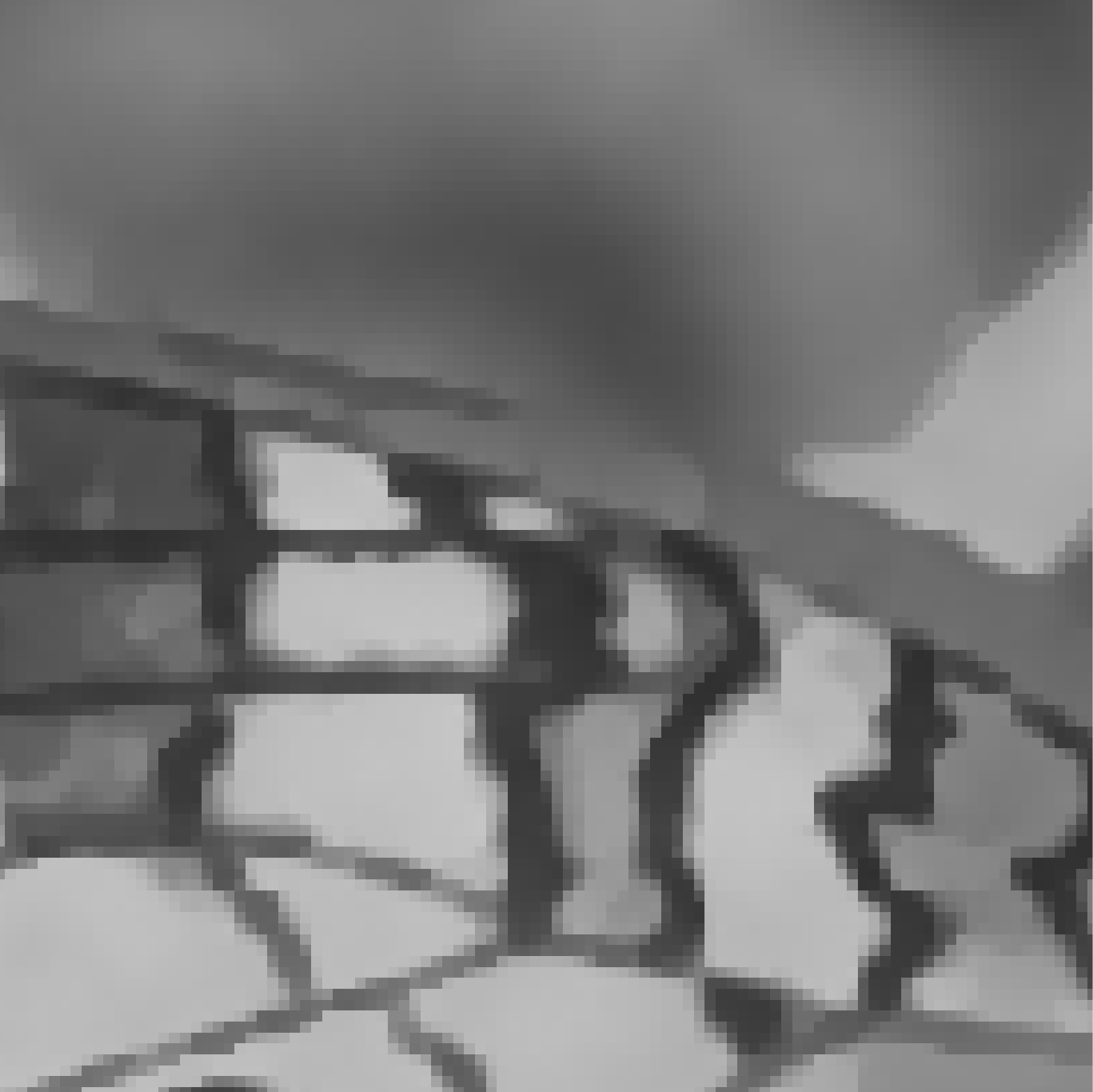}
	\end{tabular}
	\caption{Zoom-ins of the boxed region in Figure~\ref{fig_compare_butterfly}. (a) Clean image. (b) Noisy input. (c)  CEP~\cite{chan2007image}.  (d) HKLM~\cite{huska2021variational}. (e) Proposed.} \label{fig_compare_zoom}
\end{figure}

We compare the proposed model with models proposed by Chan et al.~\cite{chan2007image} (CEP) and Huska et al.~\cite{huska2021variational} (HKLM), which are stated in~\eqref{eq_chan} and~\eqref{eq_huska}, respectively. In both the proposed model and HKLM, an input image is decomposed into three components corresponding to  cartoon-like part $v^*$, smooth part $w^*$, and oscillatory part $n^*$. For the four-component model~\eqref{eq_chan}, in addition to $v^*$ and $w^*$, we take  $n^*=n_1^*+n_2^*$, i.e., the oscillatory part is the summation of the texture and noise.  

In Figure~\ref{fig_compare1}, we show the decomposition results from CEP, HKLM, and DOSH when applied to a noisy synthetic image ($\sigma=50/255$), respectively. For CEP~\cite{chan2007image} (see also~\eqref{eq_chan}), we set $\lambda = 0.5, \alpha=10$, $\mu=2$ and the wavelet threshold $\delta$ is determined using Stein's Unbiased Risk Estimate (SURE)~\cite{donoho1995adapting}. For HKLM~\cite{huska2021variational} (see also~\eqref{eq_huska}),  the authors proposed an automatic scheme for tuning the regularization parameters. However, we find that they fail to yield good decomposition results in general for our test images. To reach good PSNR values in $u^*$, after experimenting with different ranges of parameters, we manually set $\beta_1=1\times10^{-4}$, $\beta_2=0.5$, $\beta_3=5\times10^{-6}$, and the MC penalty (an approximation of the $L^0$-norm, see \cite{huska2021variational} for details) parameter $a=30$.   For the proposed model, we use $\alpha_0=4\times10^{-2}$, $\alpha_{\text{curv}}=0.15$, $\alpha_w=50$ and $\alpha_n=1\times10^{-3}$.   We also present the summations of $v^*$ and $w^*$ as the noise-less approximations for $f_{\text{noise}}$ and evaluate their respective PSNR values. In this example with heavy noise, both HKLM and our proposed model successfully separate the piecewise constant part from $f_{\text{noise}}$ with strong sparsity of gradient thanks to the $L^0$-gradient regularization, and CEP with TV-norm fails to produce clean $v^*$ part. Moreover, we observe that, with a heavy noise, regularization using only $L^0$-gradient as in HKLM yields irregular boundaries. In contrast, with the additional regularization on the curvature of level-set curves of $v^*$, our proposed model renders more satisfying results.

In Figure~\ref{fig_compare_butterfly}, we compare these methods using a grayscale photo of \textit{butterfly} with noise ($\sigma=10/255$) and show the zoom-in comparisons in Figure~\ref{fig_compare_zoom}.  For CEP (see also~\eqref{eq_chan}), we set $\lambda=5,\alpha=10, \mu=2$ and $\delta$ determined from SURE~\cite{donoho1995adapting}. 
For HKLM (see also~\eqref{eq_huska}), we use $\beta_1=1\times10^{-4}$, $\beta_2=3$, $\beta_3=5\times10^{-6}$, and $a=30$ for the MC penalty. For the proposed method, we use $\alpha_0=2\times10^{-3}$, $\alpha_{\text{curv}}=0.05$, $\alpha_w=30$ and $\alpha_n=10$.  All methods separate the structural part including patterns on the wing in the front,  the smooth part including the blurry leaves in the back, as well as the oscillatory component such as noise. Compared to CEP, both HKML and our proposed model produce better separations between the foreground details and background variations. Since HKLM uses only $L^0$-gradient for regularizing the $v^*$ component, we observe that $u^*$ of HKML has the sharpest boundaries compared to CEP and the proposed model in Figure~\ref{fig_compare_zoom}. This also yields irregular curves and strong staircase effects in (d), which are effectively suppressed in CEP in (c) and the proposed model in (e).  Since CEP uses TV-norm  and the proposed model uses the stronger sparsity-inducing $L^0$-gradient, the  $v^*$ component and consequently the $u^*$ component resulted from the proposed model are clearer than those from CEP. Moreover, for this example, the noise-less approximation $u^*$ produced by the proposed model has the highest PSNR value compared to the other two. We conclude that our proposed model produces more visually appealing decomposition results with better quality.

Our proposed model is also more efficient than CEP and HKLM.  For the image of size $512\times512$  in Figure~\ref{fig_compare_butterfly}, CEP takes around 1655 seconds, HKLM takes over 4000 seconds, whereas our model only requires around 180 seconds.  We note that HKLM requires to solve a dense linear system of size $(4NM)\times (4NM)$, which costs around $\mathcal{O}((NM)^3)$. As discussed in Section~\ref{eq_num_discret},   by leveraging FFT, the heaviest computation in our model requires solving a linear system of size $4\times 4$ for each pixel in parallel, and the solution can be explicitly expressed. Overall, the computational cost of each iteration of our method is in the order or $\mathcal{O}(MN(\log M+ \log N))$. This is similar for CEP, which  involves both FFT and the discrete wavelet transform.

We note that the  components identified by either HKLM and the proposed method may  depend on the selected model parameters. In Section~\ref{sec_further_comparison}, we compare the proposed method with HKLM on results reported in the original paper using their fine-tuned parameters. In the next section, we study our model parameters in detail.

\subsection{Selection of model parameters}\label{sec_parameters}
\begin{figure}[t!]
	\centering
	\begin{tabular}{c@{\hspace{2pt}}c@{\hspace{2pt}}c}
		(a)&(b)&(c)\\
		\includegraphics[width=0.2\textwidth]{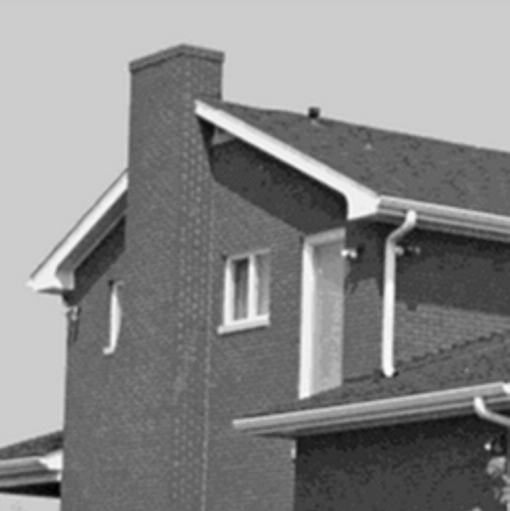}&
		\includegraphics[width=0.2\textwidth]{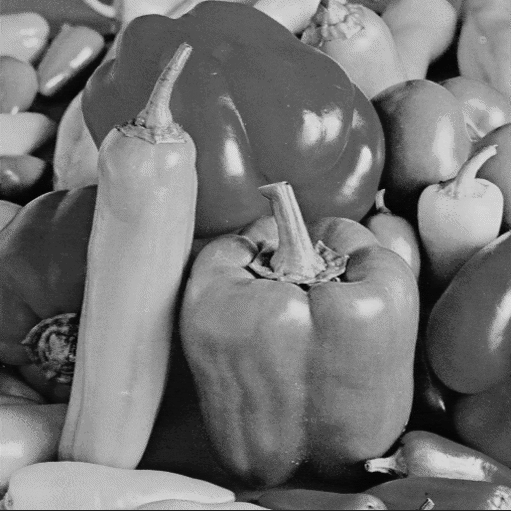}
		&\includegraphics[width=0.2\textwidth]{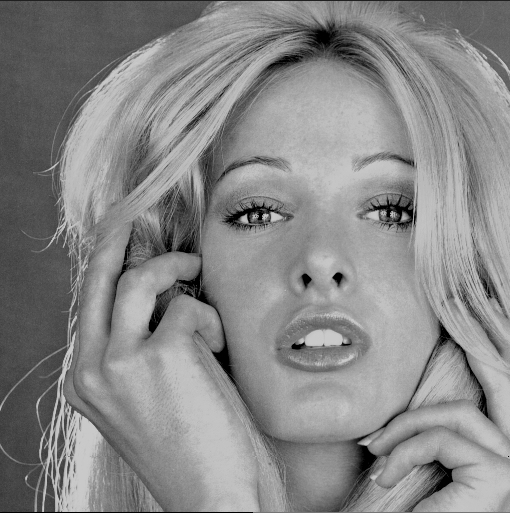}
	\end{tabular}
	\caption{Images used for studying parameters in  subsection~\ref{sec_parameters}. (a) \textit{house}, (b) \textit{pepper}, (c) \textit{blonde}.}\label{fig_test_image}
\end{figure}
\begin{figure}
	\centering
	\begin{tabular}{c@{\hspace{2pt}}c@{\hspace{2pt}}c}
		(a)&(b)&(c)\\
		\includegraphics[width=0.3\textwidth]{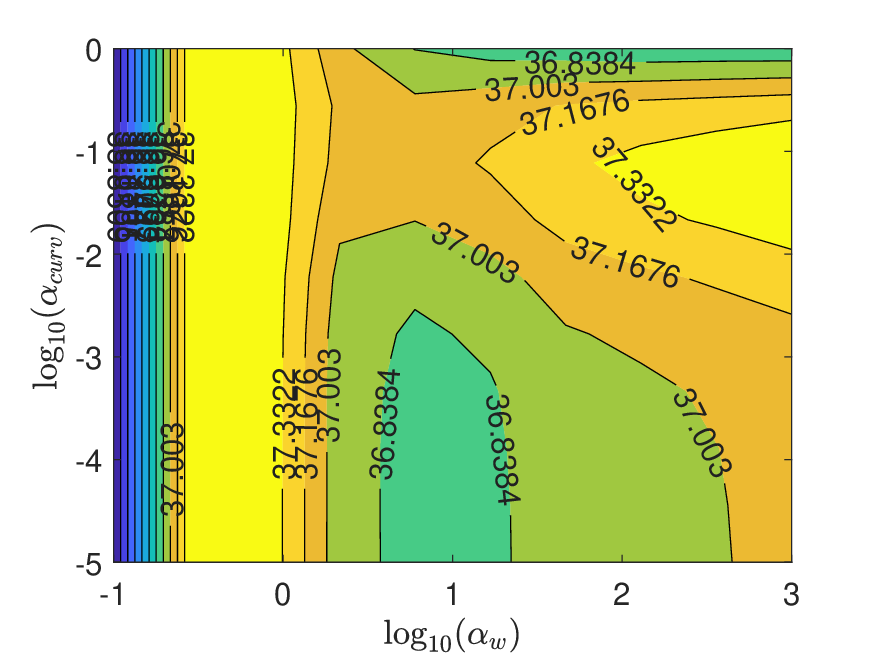}&
		\includegraphics[width=0.3\textwidth]{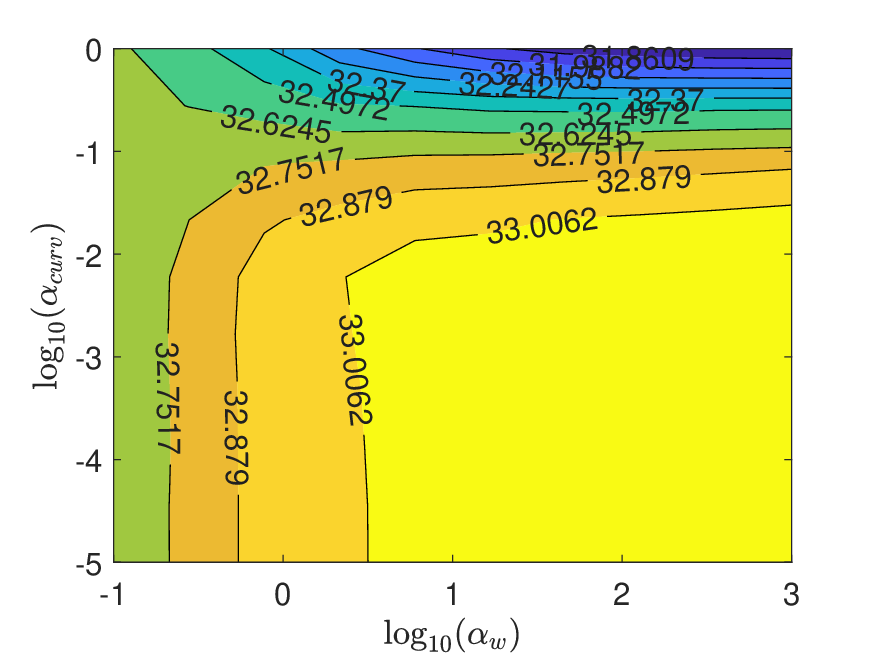}
		&\includegraphics[width=0.3\textwidth]{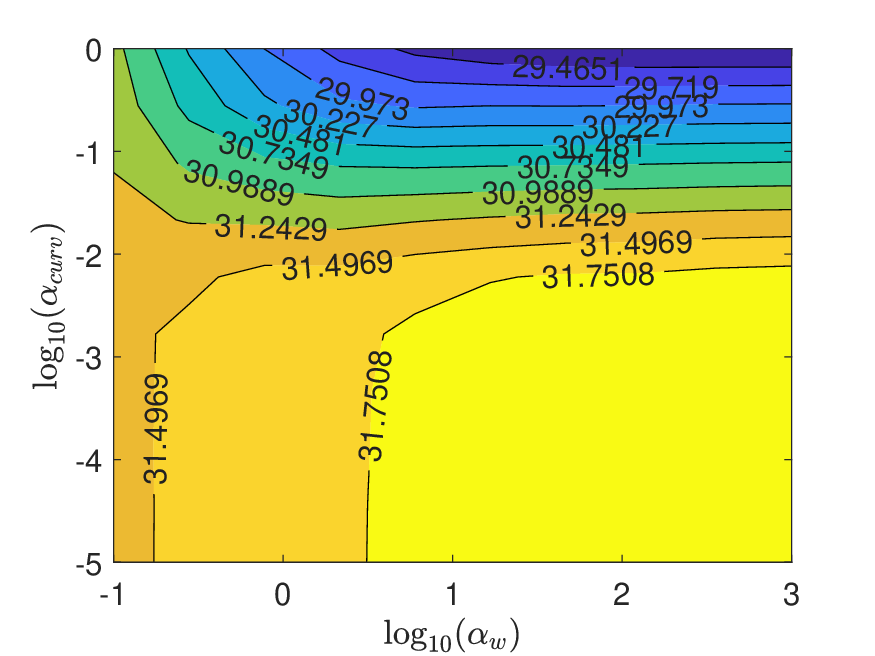}
	\end{tabular}
	\caption{PSNR contour plots of $u^*$ components generated by the proposed model for different combinations of  $\alpha_{\text{curv}}$ and $\alpha_{w}$. The inputs are (a) \textit{house}, (b) \textit{pepper}, and (c) \textit{blonde}, which have increasing levels of details and textures. Images have Gaussian noise ($\sigma=10/255$). Other parameters are fixed: $\alpha_0=2\times10^{-3}$, $\alpha_{n}=10$. The parameter $\alpha_{w}$ is relatively stable when $\alpha_{w}\geq 1$. Choice of $\alpha_\text{curv}$ can depend on image contents. Higher values of $\alpha_{\text{curv}}$ are suitable for images with simple shapes and lower values for those with rich details. }\label{fig_PSNR}
\end{figure}

\begin{figure}[t!]
	\centering
	\begin{tabular}{c@{\hspace{2pt}}c@{\hspace{2pt}}c}
		(a)&(b)&(c)\\
		\includegraphics[width=0.3\textwidth]{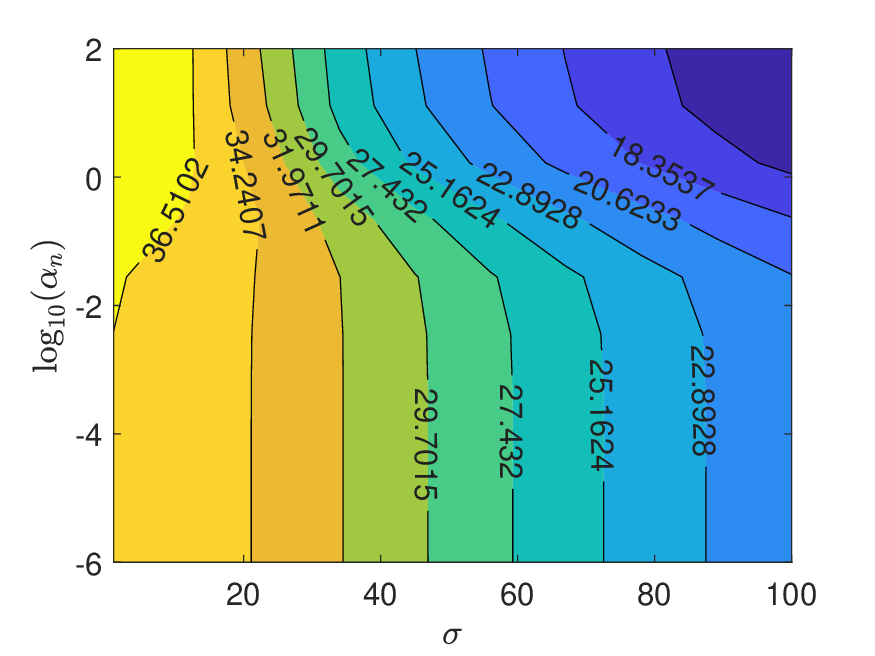}&
		\includegraphics[width=0.3\textwidth]{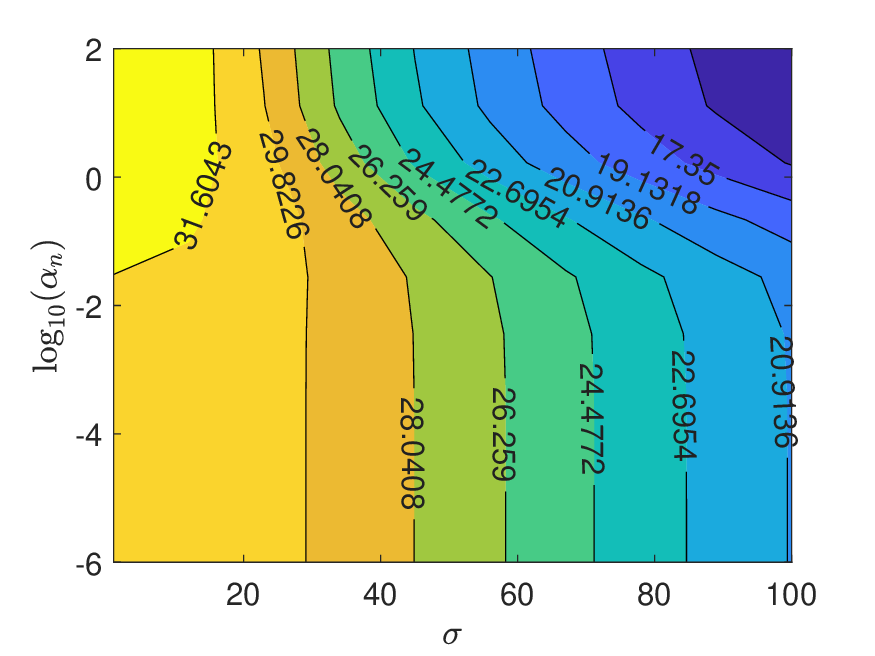}
		&\includegraphics[width=0.3\textwidth]{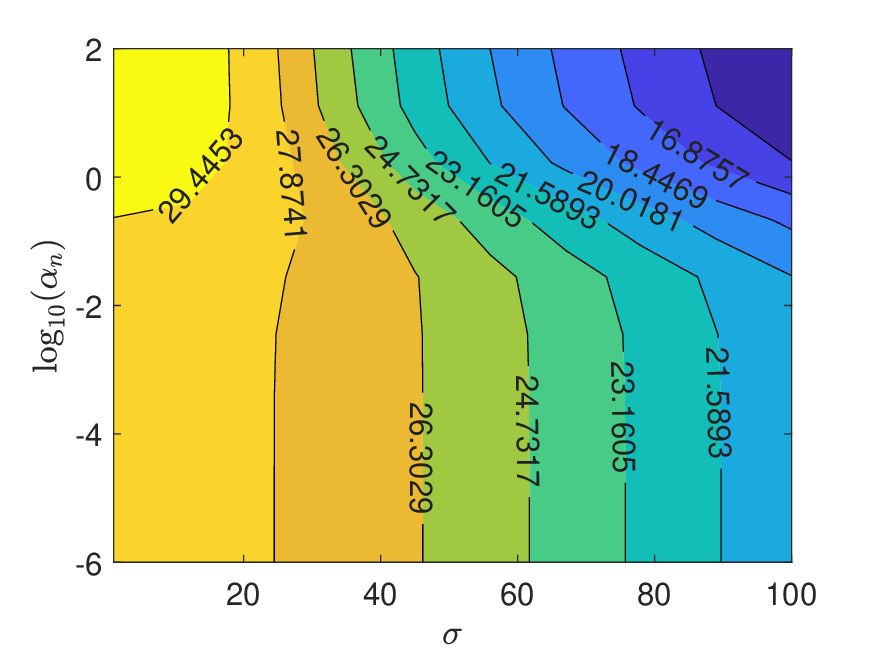}
	\end{tabular}
	\caption{PSNR contour plots of $u^*$ components generated by the proposed model for different values of  $\alpha_{n}$ under various levels of noise $\sigma$. The inputs are (a) \textit{house}, (b) \textit{pepper}, and (c) \textit{blonde}, which have increasing levels of details and textures. Other parameters are fixed: $\alpha_0=2\times10^{-3}$, $\alpha_{\text{curv}}=0.1$, $\alpha_w=50$. For low noise regime $(\sigma< 20)$, $\alpha_{n}\approx 1$ is appropriate; for middle levels of noise $(20\leq \sigma<60)$, we suggest using $\alpha_{n}\approx 10^{-4}$; and for high noise regime, $\alpha_{n}\approx 1\times10^{-6}$ is more suitable.}\label{fig_PSNR2}
\end{figure}

\begin{figure}[t!]
	\centering
	\begin{tabular}{c@{\hspace{2pt}}c@{\hspace{2pt}}c}
		(a)&(b)&(c)\\
		\includegraphics[width=0.32\textwidth]{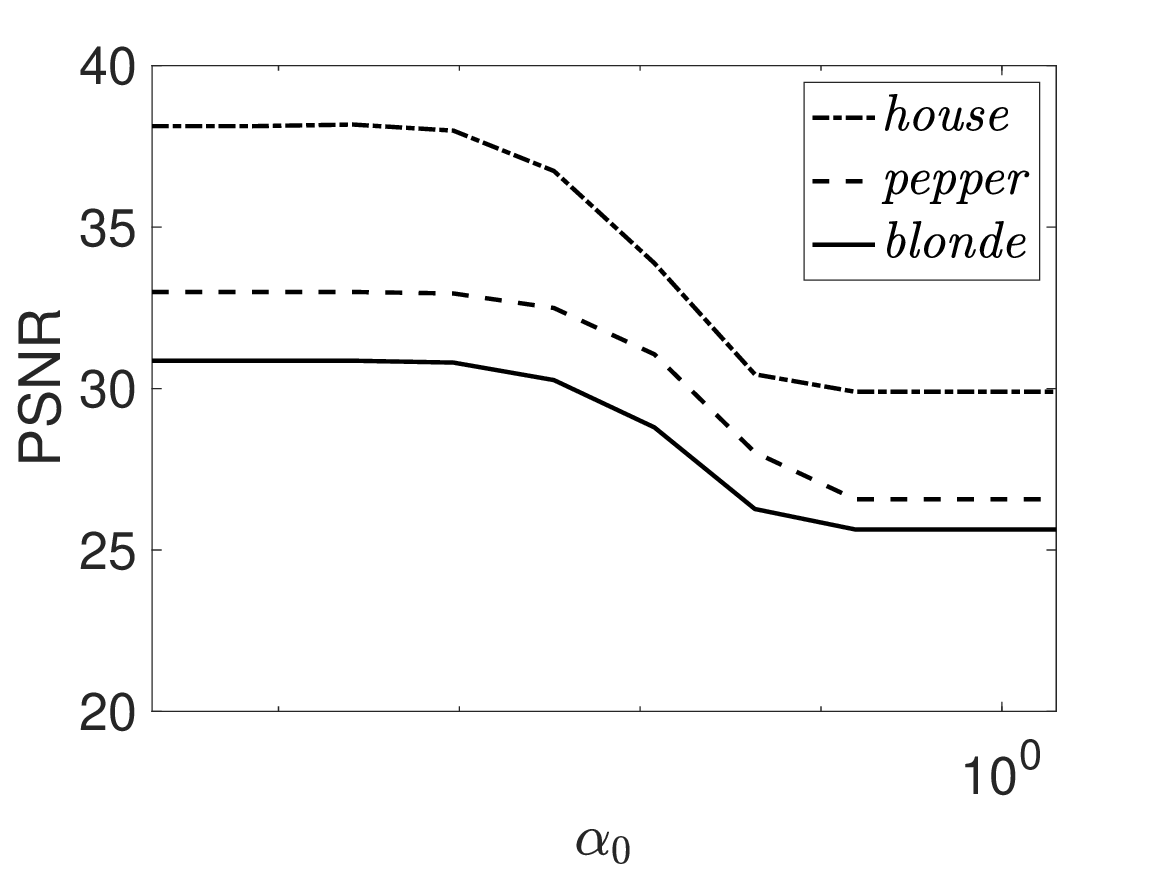}&
		\includegraphics[width=0.32\textwidth]{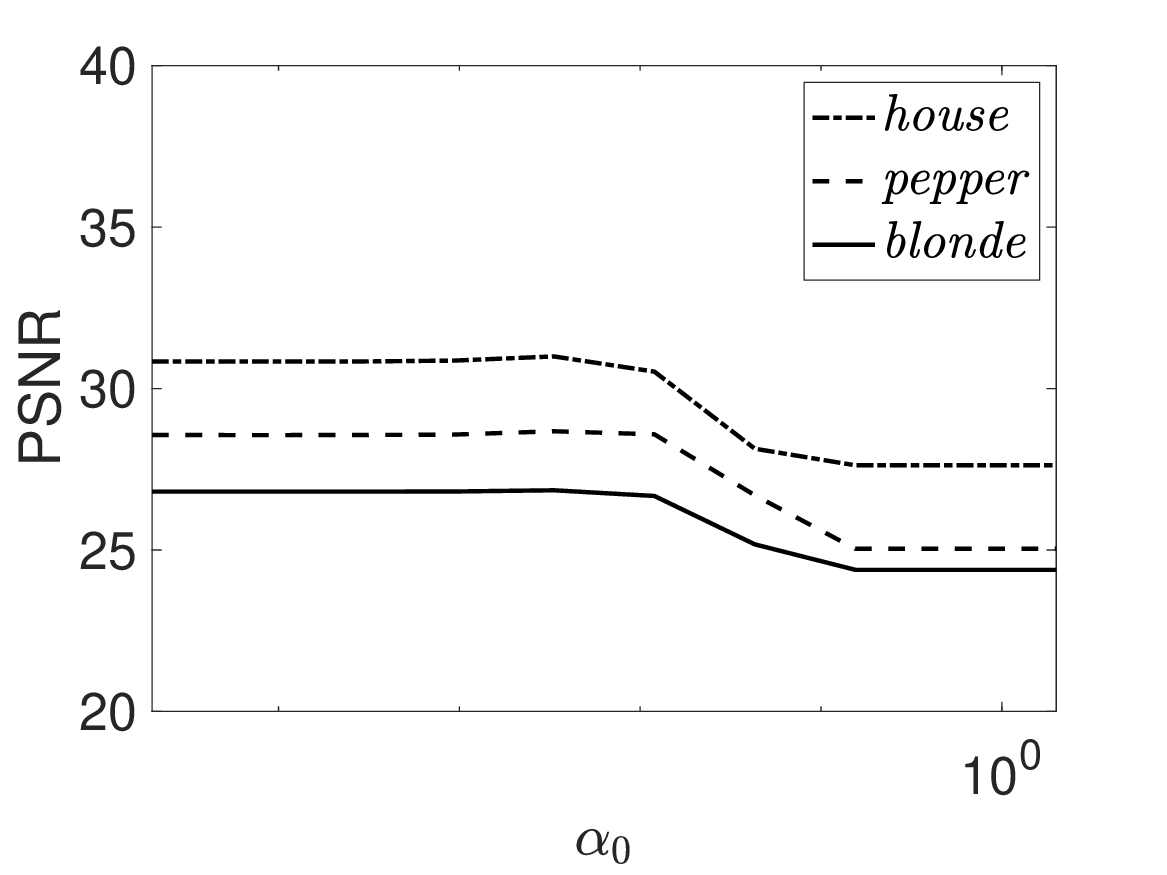}&
		\includegraphics[width=0.32\textwidth]{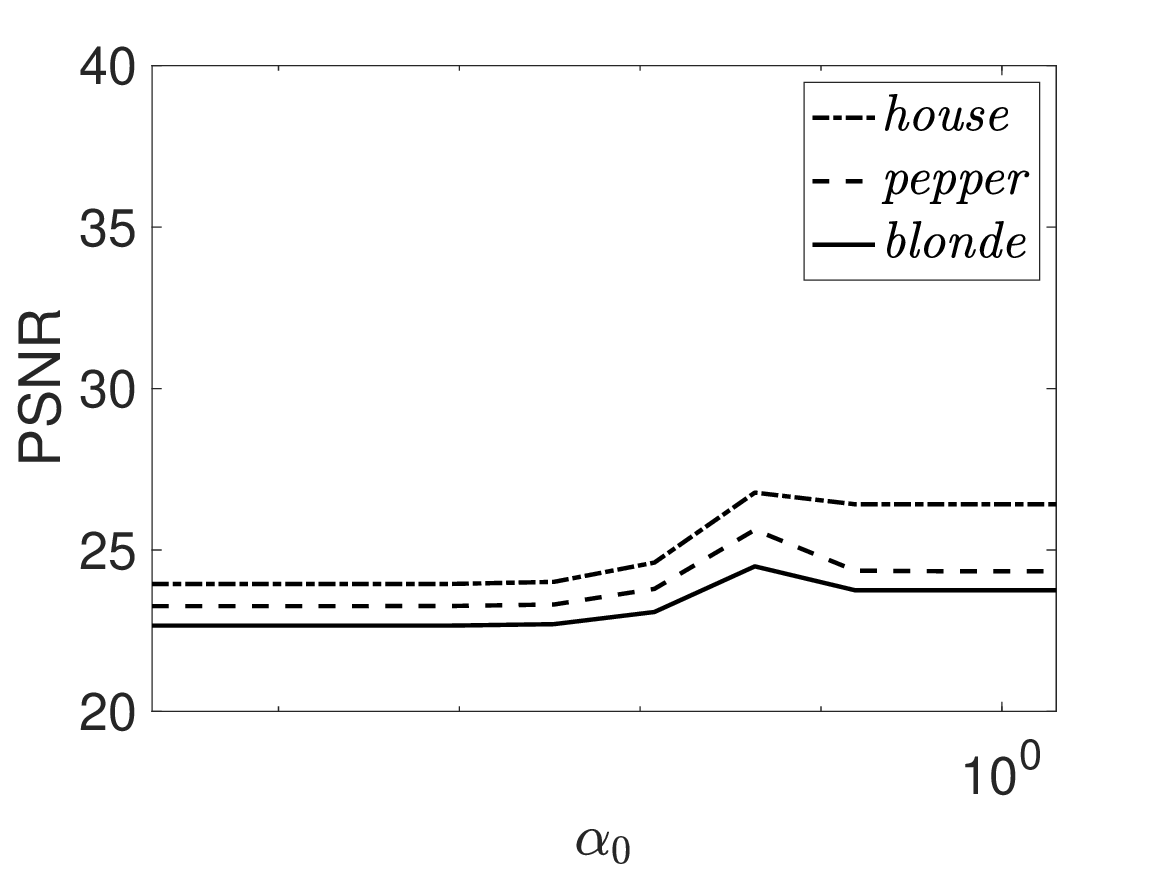}
	\end{tabular}
	\caption{  PSNR values of $u^*$ components for noisy inputs with varying $\alpha_0$. (a) $\sigma=10$, (b) $\sigma=40$, (c) $\sigma=80$. In these examples, $\alpha_\text{curv}=0.1$, $\alpha_w=50$. The parameter $\alpha_n$ is fixed at $10$, $1\times10^{-2}$ and $1\times10^{-4}$, respectively.}\label{fig_PSNR3}
\end{figure}
There are four model parameters in the proposed model: $\alpha_0, \alpha_{\text{curv}}, \alpha_w$ and $\alpha_n$, and it is generally difficult to determine their optimal values a priori. We demonstrate proper ranges for these model parameters that render good quality for $u^*$ in terms of PSNR metric.

We take \textit{house}, \textit{pepper}, and \textit{blonde} in Figure~\ref{fig_test_image}  with increasing levels of details.  In the first set of experiments, we add Gaussian noise ($\sigma=10/255$) and apply the proposed model to them with different combinations of $\alpha_{\text{curv}}$  and $\alpha_w$ while fixing $\alpha_0=2\times10^{-3}$ and $\alpha_n=10$ from heuristics. Figure~\ref{fig_PSNR} shows the PSNR contours of the $u^*$ components learned from the test images. We observe that, for $\alpha_{w}<1$,  the PSNR values are relatively low due to lack of smoothness, and they increase as $\alpha_{w}$ grows. When $1\leq \alpha_{w}\leq10^3$, PSNR values become relatively insensitive to the choice of $\alpha_{w}$. As for $\alpha_{\text{curv}}$, we observe that the optimal values depend on image contents. Specifically, if the images are relatively simple, such as \textit{house}, $\alpha_{\text{curv}}$ around $0.1\sim 1$ yields slightly better quality. As the images contain richer details such as those in \textit{pepper} and \textit{blonde},  the maximal PSNR values are obtained when $\alpha_\text{curv}$ takes smaller values around $10^{-3}\sim10^{-2}$. We suggest fixing $\alpha_{\text{curv}}$ in the order of $0.1$ and $\alpha_{w}$ in the order of $10$. In case fine-tuning is needed, a proper range for $\alpha_{\text{curv}}$ is $10^{-3}\sim 1$, where smaller values are suitable for complex images, and larger values for simple ones. As for $\alpha_w$, keeping it above $1$ is a good practice. Investigations of their effects on the decomposition are collected in Section~\ref{sec_effect_alphaw}. 

We conduct our second set of experiments to search for appropriate $\alpha_n$ while fixing $\alpha_{\text{curv}}=0.1$ and $\alpha_{w}=50$. Since $\alpha_n$ is directly associated with the noise level, we show in Figure~\ref{fig_PSNR2} the PSNR values of $u^*$ components learned from test images by the proposed model, where both $\alpha_n$ and the noise level $\sigma$ vary. We observe that for noise levels below $20$, the optimal choices of $\alpha_n$ stably remain around $0.1\sim 10$. When the noise level grows above $20$, we observe that the optimal choice of $\alpha_n$ has a clear dependence on $\sigma$. In particular, smaller $\alpha_n$ is needed for larger $\sigma$ to capture oscillations due to noises. We could fit $\alpha_n$ in log-scale with $\sigma>20$ using a quadratic polynomial; instead, we suggest a strategy for images with noise $\sigma<100$ with better generalizability. We roughly divide the levels of noise into three regimes: low noise ($\sigma<20$), medium noise ($20\leq \sigma< 60$),  and high noise ($60\leq\sigma< 100$). For images with low noise, use $\alpha_n =10$; for images with medium noise, use $\alpha_n=1\times10^{-2}$; and for those with heavy noise, use $\alpha_n=1\times10^{-4}$.

With $\alpha_{\text{curv}}$, $\alpha_w$, and  $\alpha_n$ determined above, we proceed to tune $\alpha_0$.  Using various values of $\alpha_0$, we record in Figure~\ref{fig_PSNR3} the corresponding PSNR values of the $u^*$ components for the three test images with different levels of noise. Specifically, we add Gaussian noises with $\sigma=10/255$ in (a), $\sigma=40/255$ in (b), and $\sigma=80/255$ in (c). As discussed above, we use $\alpha_n=10$ for (a), $\alpha_n=1\times10^{-2}$ for (b), and $\alpha_n=1\times10^{-4}$ for (c); and we fix $\alpha_\text{curv}=0.1$ and $\alpha_w=50$. For the  tested images, when the noise is mild as in (a) and (b),  we observe two plateaus in PSNR values of $u^*$ components connected by decreasing phases between $2\times10^{-5}<\alpha_0<2$. The transitions from high PSNR to low PSNR regime are smooth, and they correspond to removing rich image details such as textures. However, when the noise level is high as in (c), we observe that as $\alpha_0$ increases, the PSNR values  improve slightly. This shows that $L^0$-gradient can contribute to perturbation reduction in case of heavy noise. For general images, we suggest using $\alpha_0=2\times10^{-3}$ as the default.

\section{Conclusion}\label{sec_conclude}
In this paper, we propose a novel model for decomposing grayscale images into three components: the structural part, the smooth part, and the oscillatory part, each of which has a distinctive visual feature. To take advantage of the strong sparsity induced by the $L^0$-gradient regularization while avoiding the undesirable staircase effect, our model integrates the curvature regularization on the level lines of the structural part. As a result, the reconstructed structural part effectively preserves sharp boundaries enclosing homogeneous regions. To solve the associated energy minimization problem, we develop an efficient numerical algorithm based on operator splitting. The original non-convex non-smooth problem is separated into four sub-problems. Each of them either admits a closed-form solution or enjoys a form suitable for applying FFT. We present various experiments to justify the proposed model. In particular, we show that our model successfully separates underlying patterns from soft shadows even when both of them have complicated geometries with fine details.  We also justify the regularization terms included in the proposed model and illustrate the effects of parameters. We compare our model with CEP~\cite{chan2007image} and HKLM~\cite{huska2021variational} for three-component decomposition and show that our model achieves visually appealing results and outstanding computational efficiency.  To amend the loss of fine-scale details, a natural extension is to model noise and textures separately.   The proposed method can also be extended to color image decomposition when applied channel-wise. To avoid potential artifacts, we leave it as a future work to investigate appropriate regularizers that leverage correlations among color channels.

\section*{Acknowledgments}
We would like to thank Dr. Martin Huska at University of Bologna for kindly sharing the code of \cite{huska2021variational}. We would also like to thank Prof. Sung Ha Kang at Georgia Institute of Technology for inspiring discussions on high-order regularizations.

\bibliographystyle{abbrv}
\bibliography{sample}

\appendix

\section{Proof of Proposition~\ref{prop:minimizer}}\label{sec_proof_prop}
\begin{proof}
	The proof is analogous to~\cite{deng2019new}. Denote 
	\begin{align*}
		& E_1(v,w,n)= \alpha_0 L_G^0(v) + \alpha_{\text{curv}}\int_{\Omega}\kappa^2(v)\left|\nabla v\right|+  \alpha_w\|\Delta w\|_{L^2(\Omega)}^2+ \alpha_n  \|n\|_{H^{-1}(\Omega)}^2,\\ 
		&E_2(v,w,n)=\frac{1}{2}\int_{\Omega}|f-(v+w+n)|_2^2d\bx.
	\end{align*}
	Let $(v^*,w^*,n^*)$ be a minimizer of Model~\eqref{eq:model1}. Consider the tuple $(v^*+c_1,w^*+c_2,n^*)$, where $c_1,c_2\in \RR$ are scalar fields. Notice that
	\begin{align*}
		E_1(v^*,w^*,n^*)=E_1(v^*+c_1,w^*+c_2,n^*)\;,
	\end{align*}
	and
	\begin{align}
		E_2(v^*+c_1,w^*+c_2,n^*)=&\frac{1}{2}\int_{\Omega}|f-(v^*+c_1+w^*+c_2+n^*)|^2d\bx\nonumber\\
		=&E_2(v^*,w^*,n^*)+(c_1+c_2)\int_{\Omega}(v^*+w^*+n^*-f)d\bx + \frac{|\Omega|}{2}(c_1+c_2)^2,
		\label{eq:minimizer:E2}
	\end{align}
	where $|\Omega|$ is the area of $\Omega$. The function on the right-hand-side of (\ref{eq:minimizer:E2}) is quadratic in $(c_1+c_2)$, which takes its minimal value  $E_2(v^*,w^*,n^*)-(2|\Omega|)^{-1}\left(\int_{\Omega}(f-(v^*+w^*+n^*))d\bx\right)^2$ when
	\begin{align}
		c_1+c_2=\frac{\int_{\Omega}(f-(v^*+w^*+n^*))d\bx}{|\Omega|}.
		\label{eq.c1c2}
	\end{align}
	If $\int_{\Omega}(f-(v^*+w^*+n^*))d\bx\neq 0$, for any $c_1$ and $c_2$ that satisfy (A.2), it would imply that 
	\begin{align*}
		&E_1(v^*,w^*,n^*)+E_2(v^*,w^*,n^*)\\
		>&E_1(v^*,w^*,n^*)+E_2(v^*,w^*,n^*)-(2|\Omega|)^{-1}\left(\int_{\Omega}(f-(v^*+w^*+n^*))d\bx\right)^2\\
		=&E_1(u^*+c_1,w^*+c_2,n^*)+E_2(u^*+c_1,w^*+c_2,n^*),
	\end{align*}
	which contradicts to the assumption that $(v^*,w^*,n^*)$ is a minimizer of Model~\eqref{eq:model1}. We thus conclude  that $\int_{\Omega} v^*+w^*+n^*d\bx = \int_{\Omega} f d\bx$.
\end{proof}
\section{Equivalence between Hessian and Laplacian regularizer}
\label{append_regularizer}
We  compare 
\begin{align*}
	E_H(w)=\frac{1}{2}\|\partial^2w\|_{L^2(\Omega)}^2
	\quad \mbox{ with } \quad
	E_L(w)=\frac{1}{2}\|\Delta w\|_{L^2(\Omega)}^2.
\end{align*}
Functional $E_H$ is used in \cite{huska2021variational} and $E_L$ is used in this paper as a regularizer for the $w$ component. It was observed in~\cite{chan2007image} that $E_H=E_L$ for compactly supported smooth functions. We show that more general boundary conditions, $E_H$ and $E_L$ have the same set of critical points.
To prove that, it is sufficient to show that their variations concerning $w$ have the same form.
For simplicity, we assume $\Omega=[0,L_1]\times [0,L_2]$ for some $L_1,L_2>0$.

The functional $E_H$ can be written as
\begin{align*}
	E_H(w)=\frac{1}{2}\int_{\Omega} \left(\frac{\partial^2 w}{\partial x_1^2} \right)^2 + 2\left(\frac{\partial^2 w}{\partial x_1 \partial x_2} \right)^2 + \left(\frac{\partial^2 w}{\partial x_2^2} \right)^2 d\bx.
\end{align*}
The Fr\'{e}chet derivative of $E_H$ with respect to $w$ in the direction $h$ is
\begin{align}
	\left\langle \frac{\partial E_H}{\partial w}, h \right\rangle=& \int_{\Omega} \left(\frac{\partial^4 w}{\partial x_1^4} + 2\frac{\partial^4 w}{\partial x_1^2 \partial x_2^2} + \frac{\partial^4 w}{\partial x_2^4} \right) h \ d\bx \nonumber\\
	&-\left.  \left\langle \frac{\partial^2 w}{\partial x_1^2}, \frac{\partial h}{\partial x_1}\right\rangle\right|_{x_1=0}^{x_1=L_1} +\left.  \left\langle \frac{\partial^3 w}{\partial x_1^3}, h\right\rangle\right|_{x_1=0}^{x_1=L_1} \nonumber\\ 
	&-\left.  \left\langle \frac{\partial^2 w}{\partial x_2^2}, \frac{\partial h}{\partial x_2}\right\rangle\right|_{x_2=0}^{x_2=L_2} +\left.  \left\langle \frac{\partial^3 w}{\partial x_2^3}, h\right\rangle\right|_{x_2=0}^{x_2=L_2} \nonumber\\
	& -\left.  \left\langle \frac{\partial^2 w}{\partial x_1 \partial x_2}, \frac{\partial h}{\partial x_2}\right\rangle\right|_{x_1=0}^{x_1=L_1} +\left.  \left\langle \frac{\partial^3 w}{\partial x_1^2 \partial x_2}, h\right\rangle\right|_{x_2=0}^{x_2=L_2} \nonumber\\
	& -\left.  \left\langle \frac{\partial^2 w}{\partial x_1 \partial x_2}, \frac{\partial h}{\partial x_1}\right\rangle\right|_{x_2=0}^{x_2=L_2} +\left.  \left\langle \frac{\partial^3 w}{\partial x_1 \partial x_2^2}, h\right\rangle\right|_{x_1=0}^{x_1=L_1}.
	\label{eq.variaiton.EH}
\end{align}

The functional $E_L$ can be written as
\begin{align*}
	E_L(w)=\frac{1}{2}\int_{\Omega} \left(\frac{\partial^2 w}{\partial x_1^2} \right)^2 + 2\left(\frac{\partial^2 w}{\partial x_1^2 } \frac{\partial^2 w}{\partial x_2^2 }\right) + \left(\frac{\partial^2 w}{\partial x_2^2} \right)^2 d\bx.
\end{align*}
The Fr\'{e}chet derivative of $E_L$ with respect to $w$ in the direction $h$ is
\begin{align}
	\left\langle \frac{\partial E_L}{\partial w}, h \right\rangle=& \int_{\Omega} \left(\frac{\partial^4 w}{\partial x_1^4} + 2\frac{\partial^4 w}{\partial x_1^2 \partial x_2^2} + \frac{\partial^4 w}{\partial x_2^4} \right) h\ d\bx \nonumber\\
	&-\left.  \left\langle \left(\frac{\partial^2 w}{\partial x_1^2}+\frac{\partial^2 w}{\partial x_2^2}\right), \frac{\partial h}{\partial x_1}\right\rangle\right|_{x_1=0}^{x_1=L_1} +\left.  \left\langle \left(\frac{\partial^3 w}{\partial x_1^3}+\frac{\partial^3 w}{\partial x_1^2\partial x_2}\right), h\right\rangle\right|_{x_1=0}^{x_1=L_1} 
	\nonumber\\
	&-\left.  \left\langle \left(\frac{\partial^2 w}{\partial x_2^2}+\frac{\partial^2 w}{\partial x_2^2}\right), \frac{\partial h}{\partial x_2}\right\rangle\right|_{x_2=0}^{x_2=L_2} +\left.  \left\langle \left(\frac{\partial^3 w}{\partial x_2^3}+\frac{\partial^3 w}{\partial x_1 \partial x_2^2}\right), h\right\rangle\right|_{x_2=0}^{x_2=L_2}.
	\label{eq.variaiton.EL}
\end{align}
By using the boundary conditon
\begin{align*}
	\frac{\partial^2 w}{\partial x_1^2}=\frac{\partial^2 w}{\partial x_2^2}= \frac{\partial^3 w}{\partial x_1^3} = \frac{\partial^3 w}{\partial x_2^3} = \frac{\partial^3 w}{\partial x_1^2\partial x_2}= \frac{\partial^3 w}{\partial x_1\partial x_2^2}=0 \mbox{ on } \partial \Omega,
\end{align*}
all boundary integrals in (\ref{eq.variaiton.EH}) and (\ref{eq.variaiton.EL}) vanish. We get the same Fr\'{e}chet derivative for both $E_H$ and $E_L$
\begin{align*}
	\left\langle \frac{\partial E_H}{\partial w}, h \right\rangle=\left\langle \frac{\partial E_L}{\partial w}, h \right\rangle=\int_{\Omega} \left(\frac{\partial^4 w}{\partial x_1^4} + 2\frac{\partial^4 w}{\partial x_1^2 \partial x_2^2} + \frac{\partial^4 w}{\partial x_2^4} \right) h\ d\bx.
\end{align*}
Therefore, $E_H$ and $E_L$ have the same optimality condition and thus the same set of critical points.

\section{Details on adapting to periodic boundary conditions}
\label{append_periodic}
Without loss of generality,  consider a rectangular domain $\Omega=[0,L_1]\times[0,L_2]$ and define the spaces
\begin{align}
	&\cH^k_P(\Omega)=\{v\in \cH^k(\Omega): v(0,y)=v(L_1,y), \ v(:,0)=v(:,L_2)\}
\end{align}
for $k=1,2$.
We revise the set $\Sigma_f$ to
\begin{align}
	\Sigma_f = \big\{(\bq,w,\bg)\in(&L^2(\Omega))^2\times L^2(\Omega) \times (H^1(\Omega))^2 | \nonumber\\
	&\exists v\in H_P^1(\Omega)~\text{such that}~\bq=\nabla v~\text{and}~\int_\Omega(v+w+\nabla\cdot \bg-f)\,d\bx=0\big\}.
\end{align}

Problem (\ref{eq:reform_model}) and (\ref{eq.vq}) are replaced by
\begin{align}
	&\min_{\substack{\bq\in (\cH_P^1(\Omega))^2,\bmu\in (\cH^1_P(\Omega))^2,\\ w\in \cH^2_P(\Omega),\bg\in (\cH^1_P(\Omega))^2}}\bigg[\alpha_0\int_\Omega \|\bq\|_0\,d\bx+\alpha_{\rm curv}\int_\Omega \left|\nabla\cdot\bmu\right|^2\left|\bq\right|\,d\bx+\alpha_{w}\int_{\Omega}|\Delta w|^2d\bx \nonumber\\
	&\hspace{2cm} +\alpha_{n}\int_{\Omega}|\bg|^2 d\bx +\frac{1}{2}\int_{\Omega}|f-(v_{\bq}+w+\nabla \cdot \bg)|^2d \bx+I_{\Sigma_f}(\bq,w,\bg)+I_S(\bq,\bmu)\bigg]
	\label{eq:reform_model.periodic}
\end{align}
and
\begin{align}
	\begin{cases}
		\Delta v_{\bq,w,\bg}=\nabla\cdot\bq & \mbox{ in } \Omega,\\
		\int_\Omega v_{\bq,w,\bg}\,d\bx=\int_\Omega f-w-\nabla\cdot\bg\,d\bx,\\
		v_{\bq,w,\bg}(0,x_2)=v_{\bq,w,\bg}(L_1,x_2) &\mbox{ for } 0\leq x_2 \leq L_2,\\
		v_{\bq,w,\bg}(x_1,0)=v_{\bq,w,\bg}(x_1,L_2) &\mbox{ for } 0\leq x_1 \leq L_1,\\
		\left(\frac{\partial  v_{\bq,w,\bg}}{\partial x_1}-q_1\right)(0,x_2)=\left(\frac{\partial  v_{\bq,w,\bg}}{\partial x_1}-q_1\right)(L_1,x_2) &\mbox{ for } 0\leq x_2 \leq L_2,\\
		\left(\frac{\partial  v_{\bq,w,\bg}}{\partial x_2}-q_2\right)(x_1,0)=\left(\frac{\partial  v_{\bq,w,\bg}}{\partial x_2}-q_2\right)(x_1,L_2) &\mbox{ for } 0\leq x_1 \leq L_1.
	\end{cases}
	\label{eq.vq.periodic}
\end{align}

For subproblem solvers, we replace (\ref{eq.mu.split1.min}) and (\ref{eq.mu.split1}) by
\begin{align}
	\blambda^{k+2/4}=\argmin_{\bmu\in (\cH_P^1(\Omega))^2} \left[ \frac{\gamma_1}{2} \int_{\Omega} |\bmu-\blambda^{k+1/4}|^2d\bx +\tau\alpha_{w} \int_{\Omega} \left| \nabla \cdot \bmu\right|^2|\bp^{k+2/4}|d\bx \right]
\end{align}
and 
\begin{align}
	\begin{cases}
		\gamma_1 \frac{\blambda^{k+2/4}-\blambda^{k+1/4}}{\tau} -2\alpha_{w}\nabla(|\bp^{k+2/4}|\nabla \cdot \blambda^{k+2/4})=0 &\mbox{ in } \Omega, \\
		\left(|\bp^{k+2/4}|\nabla \cdot \blambda^{k+2/4}\right)(0,x_2)=\left(|\bp^{k+2/4}|\nabla \cdot \blambda^{k+2/4}\right)(L_1,x_2) &\mbox{ for } 0\leq x_2 \leq L_2,\\
		\left(|\bp^{k+2/4}|\nabla \cdot \blambda^{k+2/4}\right)(x_1,0)=\left(|\bp^{k+2/4}|\nabla \cdot \blambda^{k+2/4}\right)(x_1,L_1) & \mbox{ for } 0\leq x_1 \leq L_1,
	\end{cases}
	\label{eq.mu.split1.periodic}
\end{align}
respectively. 

Replace (\ref{eq.step4}), (\ref{eq.step4.v}) and (\ref{eq.step4.opti}) by
\begin{align}
	\min_{\substack{(\bq,w,\bg)\in \Sigma_f, w\in \cH_P^2(\Omega),\\ \bg\in (\cH_P^1(\Omega))^2}} & \bigg[\frac{1}{2} \int_{\Omega} |\bq-\bp^{k+3/4}|^2 d\bx + \frac{\gamma_2}{2} \int_{\Omega} |w-r^{k+3/4}|^2 d\bx + \frac{\gamma_3}{2} \int_{\Omega} |\bg-\bs^{k+3/4}|^2 d\bx \nonumber\\
	&+ \tau\alpha_{w}\int_{\Omega} |\Delta w|^2d\bx +  \tau\alpha_{n}\int_{\Omega}|\bg|^2 d\bx +\frac{\tau}{2} \int_{\Omega} |f-(v_{\bq,w,\bg}+w+\nabla \cdot \bg)|^2d\bx \bigg],
	\label{eq.step4.periodic}
\end{align}
\begin{align}
	\begin{cases}
		(v^{k+1},r^{k+1},\bs^{k+1})= \displaystyle\argmin_{\substack{v\in \cH_P^1(\Omega), w\in \cH_P^2(\Omega),\\ \bg\in (\cH_P^1(\Omega))^2}} \bigg[ \frac{1}{2} \int_{\Omega} |\nabla v-\bp^{k+3/4}|^2 d\bx + \frac{\gamma_2}{2} \int_{\Omega} |w-r^{k+3/4}|^2 d\bx \\
		\hspace{4cm}+ \frac{\gamma_3}{2} \displaystyle\int_{\Omega} |\bg-\bs^{k+3/4}|^2 d\bx
		+ \tau\alpha_{w}\int_{\Omega} |\Delta w|^2d\bx +  \tau\alpha_{n}\int_{\Omega}|\bg|^2 d\bx\\
		\hspace{4cm} +\frac{\tau}{2} \displaystyle\int_{\Omega} |f-(v+w+\nabla \cdot \bg)|^2d\bx \bigg],\\
		\bp^{k+1}=\nabla v^{k+1},
	\end{cases}
	\label{eq.step4.v.periodic}
\end{align}
and
\begin{align}
	\begin{cases}
		\begin{cases}
			\frac{-\Delta v^{k+1}+\nabla \cdot\bp^{k+3/4}}{\tau} + v^{k+1}-(f-r^{k+1}-\nabla\cdot \bs^{k+1})=0,\\
			\left(\frac{\partial  v^{k+1}}{\partial x_1}-p_1^{k+3/4}\right)(0,x_2)=\left(\frac{\partial  v^{k+1}}{\partial x_1}-p_1^{k+3/4}\right)(L_1,x_2)\\
			\left(\frac{\partial  v^{k+1}}{\partial x_2}-p_2^{k+3/4}\right)(x_1,0)=\left(\frac{\partial  v^{k+1}}{\partial x_2}-p_2^{k+3/4}\right)(x_1,L_2) ,
		\end{cases}\\
		\begin{cases}
			\gamma_2\frac{r^{k+1}-r^{k+3/4}}{\tau}+r^{k+1}-(f-v^{k+1}-\nabla\cdot \bs^{k+1}) + 2\alpha_{w}(\nabla^4 r^{k+1})=0,\\
			(\Delta r^{k+1})(0,x_2)=(\Delta r^{k+1})(L_1,x_2), \ \left(\frac{\partial (\Delta r^{k+1})}{\partial x_1}\right)(0,x_2)=\left(\frac{\partial (\Delta r^{k+1})}{\partial x_1}\right)(L_1,x_2),\\
			(\Delta r^{k+1})(x_1,0)=(\Delta r^{k+1})(x_1,L_2), \ \left(\frac{\partial (\Delta r^{k+1})}{\partial x_2}\right)(x_1,0)=\left(\frac{\partial (\Delta r^{k+1})}{\partial x_2}\right)(x_1,L_2),
		\end{cases}\\  
		\begin{cases}
			\gamma_3\frac{\bs^{k+1}-\bs^{k+3/4}}{\tau}-\nabla(\nabla\cdot\bs^{k+1}) +2\alpha_n \bs^{k+1}+\nabla(f-v^{k+1}-r^{k+1})=0 ,\\
			(f-(v^{k+1}+r^{k+1}+\nabla \cdot \bs^{k+1}))(0,x_2)=(f-(v^{k+1}+r^{k+1}+\nabla \cdot \bs^{k+1}))(L_1,x_2),\\
			(f-(v^{k+1}+r^{k+1}+\nabla \cdot \bs^{k+1}))(x_1,0)=(f-(v^{k+1}+r^{k+1}+\nabla \cdot \bs^{k+1}))(x_1,L_2) ,
		\end{cases}
	\end{cases}
	\label{eq.step4.opti.periodic}
\end{align}
respectively, with $\bp=(p_1,p_2)$ and $0\leq x_1\leq L_1,0\leq x_2\leq L_2$.

\section{Further comparisons with HKLM~\cite{huska2021variational}}\label{sec_further_comparison}
\begin{figure}[t!]
	\centering
	\begin{tabular}{c@{\hspace{2pt}}c@{\hspace{2pt}}c@{\hspace{2pt}}c@{\hspace{2pt}}c@{\hspace{2pt}}c}
		
		(a)&(b)&(c)&(d)&(e)\\
		\raisebox{1.2cm}{\multirow{3}{0.18\textwidth}{
				\includegraphics[width=0.18\textwidth]{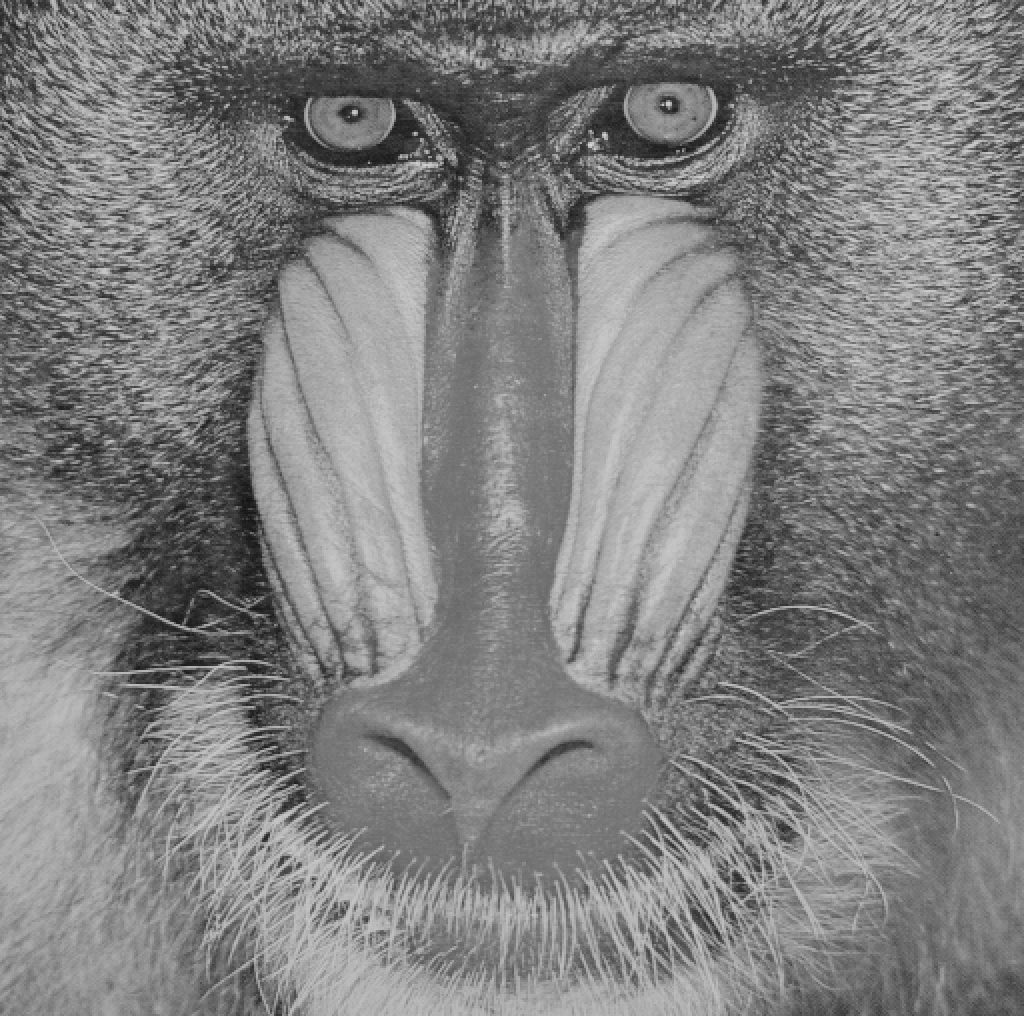}
		}}&\includegraphics[width=0.18\textwidth]{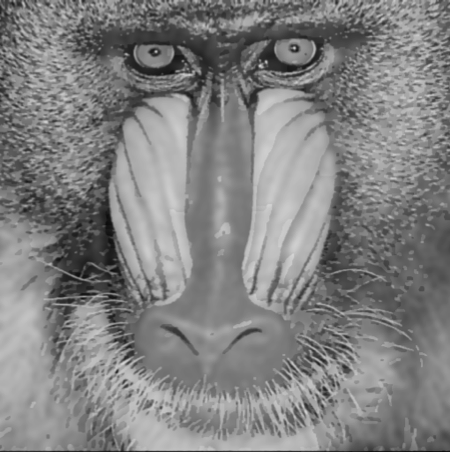}&
		\includegraphics[width=0.18\textwidth]{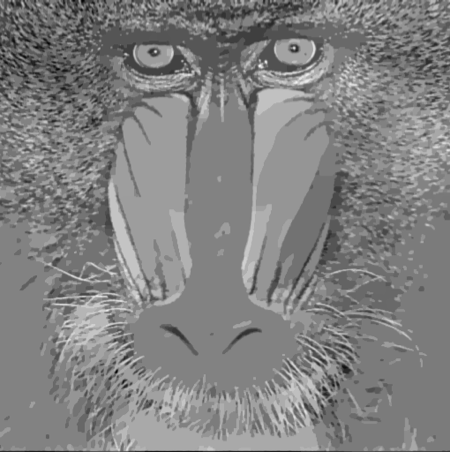}&
		\includegraphics[width=0.18\textwidth]{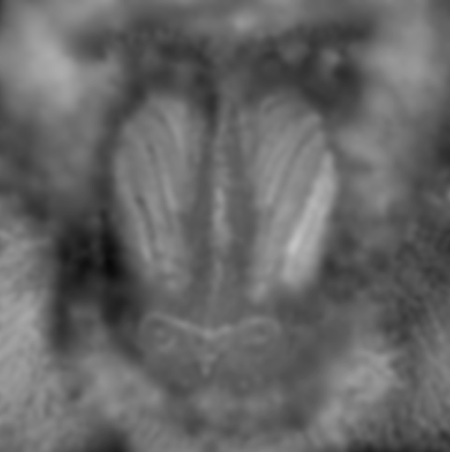}&
		\includegraphics[width=0.18\textwidth]{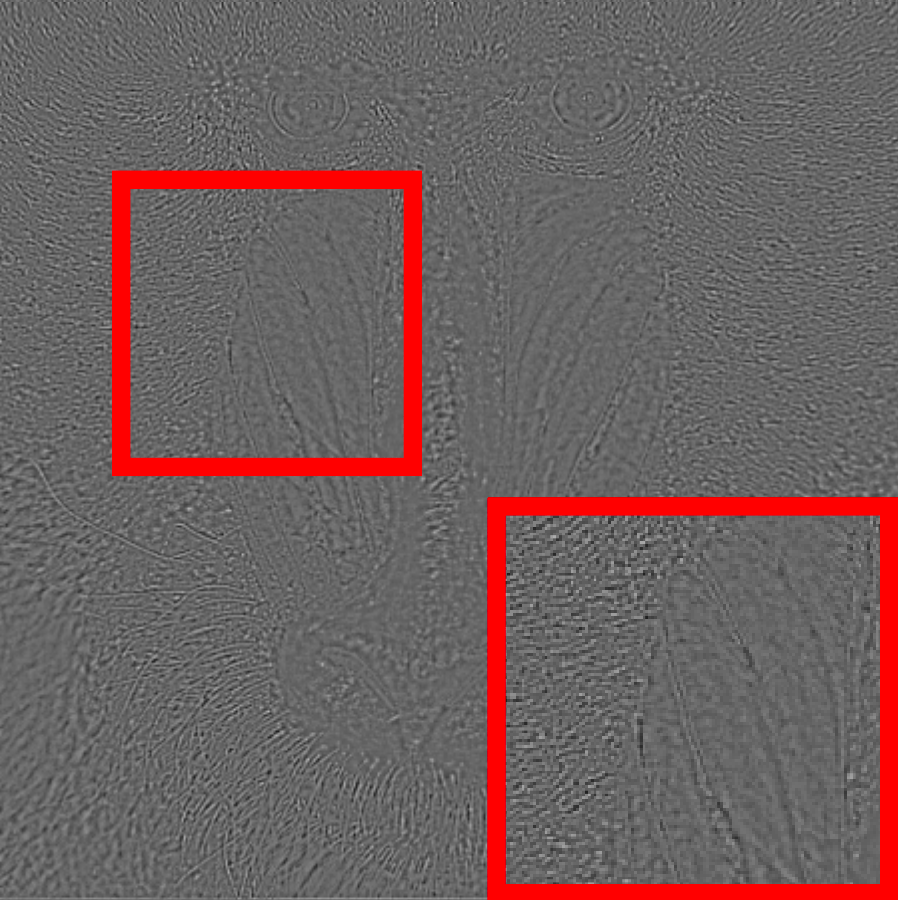}\\
		&(f)&(g)&(h)&(i)\\
		&\includegraphics[width=0.18\textwidth]{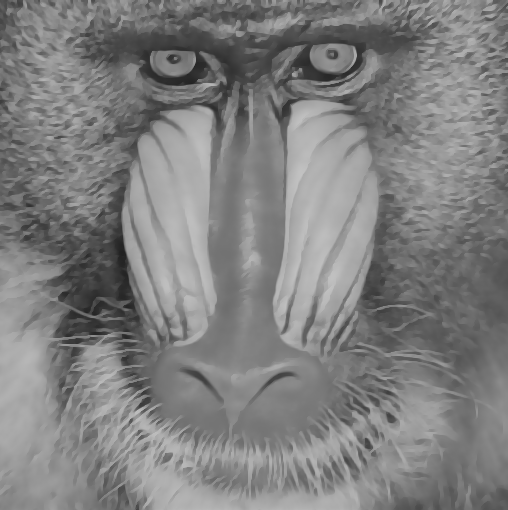}&
		\includegraphics[width=0.18\textwidth]{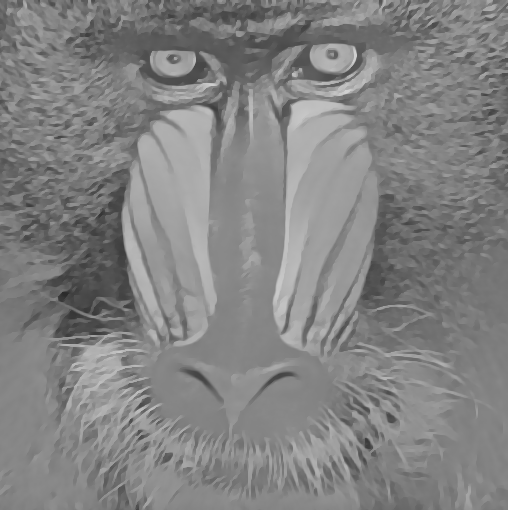}&
		\includegraphics[width=0.18\textwidth]{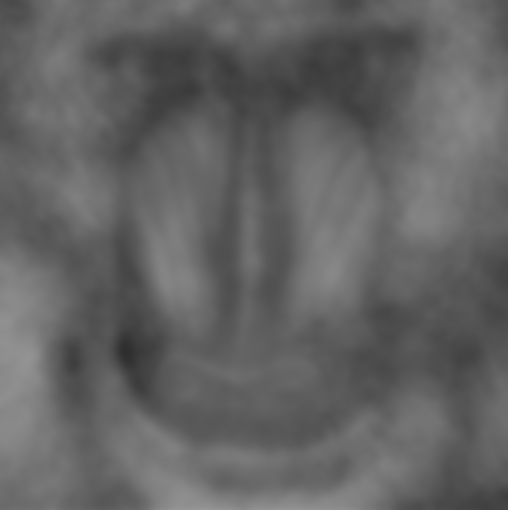}&
		\includegraphics[width=0.18\textwidth]{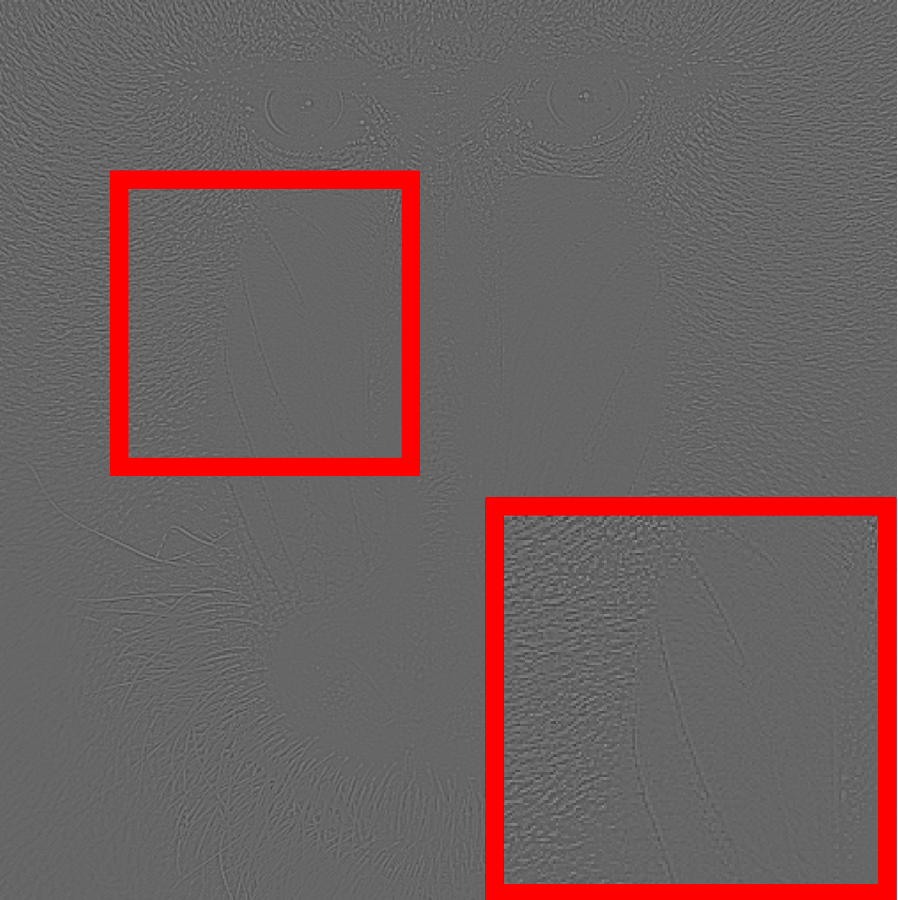}
	\end{tabular}
	\caption{Comparison with results in Figure 6.8. in~\cite{huska2021variational} by HKLM. (a) The input clean image. (b)-(e) show the results from HKLM: (b) $u^*_{\text{HKLM}}$, (c) $v^*_{\text{HKLM}}$, (d) $w^*_{\text{HKLM}}+0.4$, and (e) $2n^*_{\text{HKLM}}+0.4$. (f)-(i) show the results by the proposed method using the default parameters: (f) $u^*_{\text{proposed}}$, (g) $v^*_{\text{proposed}}$, (h) $w^*_{\text{proposed}}+0.4$, (i) $2n^*_{\text{proposed}}+0.4$. The scaling and shifting on the intensity values follow those used in~\cite{huska2021variational}. }\label{fig_original_compare}
\end{figure}

We compare with the \textit{original} results in Figure 6.8 of~\cite{huska2021variational}.  Figure~\ref{fig_original_compare} (a) shows the input image; In the second row, (b) shows the reconstructed oscillation-free part $u^*_{\text{HKLM}}$; (c) shows the structure (or cartoon) part $v^*_{\text{HKLM}}$; (d) shows the shifted smooth part $w^*_{\text{HKLM}}+0.4$; and (e) shows the transformed oscillatory part $2n^*_{\text{HKLM}}+0.4$. Following the same transformations, in the second row, we show the counterparts by our method in (f)-(i), respectively. We use the default model parameters as those in Figure~\ref{fig_real}. We observe that both methods yield similar oscillation-free part, as shown in (b) and (f). For the structure part, our result  in (g) shows less staircase effect than (c). For the smooth part, our result in (h) captures soft transitions of the shadows while avoiding pattern boundaries on the face in (d). For the oscillatory part, our result  in (i) contains overall weaker variations compared to (e); our textures mostly concentrate around the furry region, whereas the pattern boundaries in (e) are retained. See the zoom-in boxes.

\end{document}